# REROUTING CONNECTION: HYBRID COMPUTER VISION ANALYSIS REVEALS VISUAL SIMILARITY BETWEEN INDUS AND TIBETAN-YI CORRIDOR WRITING SYSTEMS

by

Ooha Lakkadi Reddy

Signature Work Product, in partial fulfillment of the
Duke Kunshan University Undergraduate Degree Program

*March 9, 2025*

Signature Work Program
Duke Kunshan University

**APPROVALS**

---

Mentor: Lincoln Rathnam, Division of Arts and Humanities

---

Co-Mentor: Mustafa Misir, Division of Natural and Applied Sciences

---

# CONTENTS

# ABSTRACT

This thesis employs a hybrid CNN-Transformer architecture, in conjunction with a detailed anthropological framework, to investigate potential historical connections between the visual morphology of the Indus Valley script and pictographic systems of the Tibetan-Yi Corridor. Through an ensemble methodology of three target scripts across 15 independently trained models, we demonstrate that Tibetan-Yi Corridor scripts exhibit approximately six-fold higher visual similarity to the Indus script (0.635) than to the Bronze Age Proto-Cuneiform (0.102) or Proto-Elamite (0.078) systems.

Additionally and contrarily to our current understanding of the networks of the Indus Valley Civilization, when measured through direct script-to-script embedding comparisons, the Indus script unexpectedly maps closer to Tibetan-Yi Corridor scripts with a mean cosine similarity of 0.930 (CI: [0.917, 0.942]) than to contemporaneous west Asian signaries mentioned above, which recorded mean cosine similarities of 0.887 (CI: [0.863, 0.911]) and 0.855 (CI: [0.818, 0.891]) despite their close geographic proximity and evident trade relations. Across various dimensionality reduction practices and clustering methodologies, the Indus script consistently clusters closest to Tibetan-Yi Corridor scripts.

Our computational results align with qualitative observations of specific pictorial parallels in numeral systems, gender markers, and key iconographic elements; this is further supported by archaeological evidence of sustained contact networks along the ancient Shu-Shendu road in tandem with the Indus Valley Civilization's decline, providing a plausible transmission pathway. While alternative explanations cannot be ruled out, the specificity and consistency of observed similarities challenge conventional narratives of isolated script development and suggest more complex ancient cultural transmission networks between South and East Asia than previously recognized.

# ACKNOWLEDGEMENTS

This paper has lived in every single thought I've had for almost a year, and I have an endless number of people to be unendingly gratuitous towards. That being said, I will make a hearty attempt to address them sufficiently nonetheless.

I would like to thank my mentors, Professor Lincoln Rathnam and Professor Mustafa Misir, who helped me shape my excited musings into a interdisciplinary work I am incredibly proud of. I could not have accomplished this without their assistance and review.

A special thanks to the Lijiang Dongba Cultural Research Institute and especially Siyu Li for guiding and assisting my anthropological research on the Naxi Dongba script and culture last December.

I also extend a heartfelt thank you to leading Tibetologist John Vincent Bellezza, who reviewed my early evidence and raised questions and considerations that led me towards a much more comprehensive anthropological argument.

Of course, I must thank all the incredible teachers I have had throughout my life; I am forever grateful to my third grade teacher, Doris Barbour, who recognized my early love for learning and has stayed dedicated to cultivating it my entire life onwards. She has supported my academic career at every turn, and I cannot wait to share this work with her now.

My undergraduate years have been tumultuous, to say the least, but I could not have made it through this project without the consistent support of my friends. To those who saw me through late and sleepless nights, edited the early drafts I threw their way, heard me question the merits of my own work, and finally witnessed my joy when the numbers ran: to you, a thank you heard round the world.

And finally, a loving thank you to my mother. The word for mother in Sumerian is *ama*, very similar to the Telugu word *amma* with which I address my own mother. Interestingly, this is also the Tibetan and Dongba word for mother, and—by reconstruction of the name of the indigenous Hindu mother goddess, Amba—is a possible candidate for the Indus word for mother as well. While this can be marked up to the universal sounds we are biologically inclined to make across human evolution and existence, I believe *amma* may very well be the oldest word in the world. My love for her is concrete in such ways; it is universal and unmoving, across oceans and borders. In this work, which took so much of myself to create, I must thank her, as it was her who crafted that self.

This paper has been accepted for presentation at the 52nd International Conference on Computer Applications & Quantitative Methods in Archaeology 2025: ”Digital Horizons” in Athens, Greece.

# LIST OF FIGURES

# LIST OF TABLES

Chapter 1

# INTRODUCTION

## 1.1 Background and Significance

### 1.1.1 An Unexplored Corridor for the Indus Valley

The resilience of language and scripture across millennia has long captivated experts across a range of disciplines, from history to biology, for its fundamental role in memorializing and contextualizing human communication and culture. Yet few problems have captured the intrigue and fascination of the world like the still-undeciphered script used by the Indus Valley Civilization (IVC) of modern-day Pakistan and Northwestern India, dated to circa 2600-1900 BC [75]. Despite holding contemporaneity with other famous Bronze Age writing systems like Sumerian cuneiform and Ancient Egyptian hieroglyphics, the Indus script endures as one of the greatest archaeological mysteries of our time. Several competing hypotheses on the etymological classification and relatives of the script have risen, with proposed connections to reconstructed Proto-Dravidian, the Austroasiatic family [99], or other isolated linguistic groups, as well as a contingent supporting a link to West Asian logosyllabic languages of the era and prior—namely Proto-Elamite and Sumerian proto-cuneiform pictographs, both noted to have visual similarities to the Indus script despite their ongoing classification as language isolates[1] [66]. However, beyond these debates lies an underexplored possibility: a connection to lesser-known ancient pictographic traditions along the Tibetan-Yi corridor (TYC), where minority scripts may offer new insight into the transmission of symbols across early trade and migration networks.

Thus far, the minority languages of the historic corridor have fallen beyond the scope of previous Indologist analysis. Positioned along the eastern edge of the Tibetan Plateau and cradled within the Hengduan Mountains, the TYC has served as a significant axis for trade, migration, and symbolic exchange from prehistory onwards; it remains one of the most linguistically, ethnically, and culturally diverse regions in China [87]. Of the many Tibeto-Burman-speaking ethnic groups that reside in this natural corridor, we take special interest in the Naxi people of Northeastern Yunnan province (located at the southernmost end of the Corridor), whose Dongba script is hailed as the world's last living pictographic script[2]. Although no direct linguistic studies have yet linked the Naxi Dongba script to the paleographic traditions of the West, preliminary visual analyses [14] have noted overlapping iconographic elements, par-ticularly in the representation of natural forces, gender definitives, and numerical notation, that cannot be attributed to the simultaneously developed logosyllabary of Han China to the east.

[1] From this point onward, "West Asia" will refer specifically to Sumer (as well as descendant Mesopotamian civilizations i.e. Babylon, Akkad, Assyria) and Elam within the context of this paper.

[2] It can be argued that the Ba-Shu pictographs share a closer visual affinity to the Indus script than Dongba, and should be the focus of our work; in doing so, this study would be severely limited by the lack of information surrounding the Ba-Shu cultures.

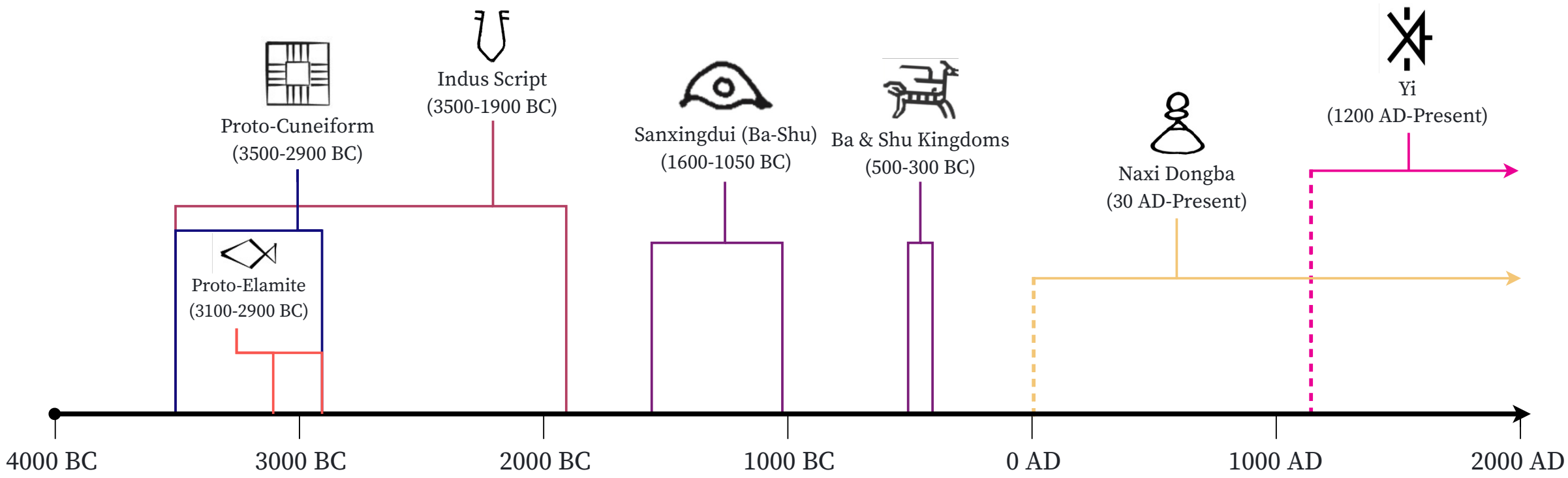

Figure 1.1: A timeline illustrating the approximate periods of relevant script systems.

The specific origin point of the Dongba script remains uncertain, but its symbolic repertoire does share visual similarities with other script relics across the TYC. Specifically, we have chosen to include additional analysis using the earliest pictographic script of the Ba-Shu culture and Classical Yi—case studies in which similarities may suggest a broader continuity of symbolic representation in the region. The non-Sinitic nature of the Ba-Shu script, in particular, has piqued curiosity given the proximity to Han China, signaling it may be more appropriate to contextualize the Sanxingdui and Jinsha cultures within the framework of the TYC rather than as an isolated cultural phenomenon [81].

Despite extensive scholarship on the Indus script and a focused revitalization of the Naxi Dongba script, no comprehensive study has yet attempted to compare these pictographic[3] systems or explore potential linguistic or symbolic transmission between the two regions; this is more than likely due to the geographic and disciplinary divides between Tibetology/Sinology and Indology. This paper employs a hybrid deep learning approach combining convolutional neural network (CNN) and Transformer architecture to analyze visual relationships between the Indus Valley script, Proto-Cuneiform, Proto-Elamite, and ancient Tibetan Yi-Corridor (TYC) scripts. Leveraging similarity analysis through hierarchical clustering and visualization techniques, the study rigorously assesses potential script connections while controlling for biases through statistical validation and multi-model consensus approaches that mitigate individual model artifacts. We seek to potentially suggest, via computer vision praxis, a novel transmission route for the Indus culture and tradition, be it demic or ideologic, that has thus far fallen outside the realm of consideration.

### 1.1.2 Significance

Unlike other languages of this era that are known to us, the Indus script exists in fragments, its meaning locked behind a lack of direct linguistic continuity and a surviving corpus consisting primarily of very short clay inscriptions. Deciphering the scripts goes beyond academic exercise; this work holds the key to contextualizing India's pre-Vedic history—a period which remains largely unknown despite its immense longevity and complexity—in the ways archaeologists have been able to reconstruct the histories of ancient Egypt and Mesopotamia.

[3]The Indus script is understood to likely be logosyllabic and not entirely pictographic; we claim it as pictographic here to encapsulate the nature by which the basic symbols were formed.

At the same time, the TYC has emerged as one of the most promising yet understudied cultural nexuses in early Asia ever since the turn of the 21st century [87] due to its unique positioning as a geographical space challenging existing narratives about the forces that shaped global culture. Despite difficulties recounting the diverse oral traditions of these nomadic people and the loss of many artifacts, the corridor remains a vibrant ethnic and cultural diffusion layer, bridging the two massive tectonic plates that are the Tibeto-Burman people and the Han Chinese. In reconstructing the historical trajectory of migration flows in the TYC, we gain insight not only into the longevity of symbolic and ethnographic artistry but also into the broader transmission of iconography in early Asia.

Meaningful statistical patterns in our computational analysis reveal that the movement of symbols, ideas, and potentially even writing systems followed routes that have long been invisible to traditional scholarship. We aim to introduce a computationally relevant methodological approach to one of the oldest unresolved questions in linguistics and archaeology: how early scripts developed, diffused, and, in some cases, disappeared.

## 1.2 Hypotheses

To mitigate false positives, we introduce dual hypotheses for the origination of the visual similarities we have initially identified.

**Hypothesis 1:** **Indus Influence on the Tibetan-Yi Corridor**
The Dongba script and related scripts in the TYC contain structural or symbolic features in common with the Indus script, suggesting, at the very least, transmission of iconography. If machine learning analysis reveals significant correlations in visual style between the two scripts, closer than the TYC similarities to Proto-Elamite and Proto-Cuneiform, it would indicate that TYC pictographic traditions may have evolved from Indus influence.

**Hypothesis 2:** **Alternative Influence from Proto-Elamite or Sumerian Traditions**
Even if Indus-like elements preliminarily appear in TYC scripts, they may instead originate from an earlier West Asian tradition due to ongoing cultural contact between the civilizations. If TYC scripts show greater structural affinity to Proto-Elamite or Sumerian pictographs, this would suggest that Indus-TYC similarities are secondary to a broader diffusion of West Asian script traditions, likely via the northern edge of the Plateau.

## 1.3 Objectives and Scope

This thesis's objectives are as follows:

1. Establish anthropological plausibility for transmission between the IVC and TYC.
2. Develop and implement a hybrid deep learning methodology to analyze and visualize structural similarity between TYC scripts and Indus/Bronze Age West Asian scripts.
3. Assess potential implications for early Asian anthropology as well as the ongoing decipherment efforts of the Indus script.

The scope of this paper is not an attempt to definitively decipher the Indus script but rather to test for statistically significant visual similarities between ancient Indus and TYC scripts, with a strict focus on morphology rather than phonology. Findings are grounded in pattern recognition and comparative analysis rather than speculative linguistic association.

## Chapter 2

# ANTHROPOLOGICAL FRAMEWORK

This section seeks to historically, archaeologically, and ethnographically establish the framework necessary to contextualize the potential for transmission from the Indus Valley Civilization to the pictographic traditions of the TYC. The interdisciplinary nature of this study requires a range of specialized research providing insights across archaeology, geology, and computational epigraphy, among others, to sufficiently and convincingly span the temporal and spatial gap evident within our hypotheses.

We begin with an assessment of the current state of IVC Indology, particularly in relation to the decipherment of the Indus script and the historical context in which these early inhabitants existed, as well as reviewing current classifications theories surrounding the modern descendants of the Indus script and language. In parallel, we consider the ethnographic and migratory history of the nomadic people of the TYC, namely the Naxi, who, over centuries, maintained pictographic traditions that remain distinct from Chinese logographic evolution and could, therefore, preserve older symbolic continuities. This presents a convincing case for the value of further exploring potential anthropological relations, augmented by a review of the preliminary morphological similarities we have identified between the Dongba and Indus scripts,

This framework will also establish the assumption underlying our study that the pictographic traditions of the TYC have external influence, although potentially indirect. The Dongba script and religion are directly connected to Tibetan Bön, which has a rich history of indigenous rock art glyphs dating to the centuries following the fall and dispersal of the Indus Valley people and culture and predating the arrival of Buddhism in Tibet [4]. The Sanxingdui civilization also rose to prominence during this time, with only seven unearthed pictographs preceding the three distinct epigraphic practices that followed. In parallel, radiocarbon dating proves an uptick in usage along the Shu-Shendu road at this time. In undertaking our work, we attribute the morphological similarities identified to broader global pictographic tradition, although we do not dismiss the plausibility of independent origination altogether.

In establishing this anthropological framework, our goal is to provide historical justification and validity to the pursuit of computational and structural comparisons between the Indus script and the pictographic systems of the TYC. This approach seeks to illuminate understudied dimensions of script diffusion in early Asia, as well as defend the value of expanding multidisciplinary scholarship on Bronze Age South Asia, the Neolithic Qinghai-Tibet Plateau, and their intersections. Our work also establishes possible paths of Indian subcontinental continuity for a collective tradition, given the final results (4) of our computational analysis.

## 2.1 The Decipherment of the Indus Script

The decipherment of the Indus script is an ongoing field of study that has persisted consistently since the first known discovery of an Indus seal in 1875 by Sir Alexander Cunningham [75]. Efforts to interpret the script take a few preset routes, including reconstructions of proto-phonetics, morphological analysis, and classification by comparison. Despite the wealth of contributions from a number of esteemed scholars since 1875, the Indus script is still far from cracked. Scholars have yet to reach any acceptable consensus on the language encoded, the type of signs, or even the overall structure of the script [36]. Additionally, a few key challenges exist that obstruct Indus script decipherment when compared to similar efforts with other lost languages: the small corpus size, short inscription lengths, a lack of accepted descendant(s), and no discovered bilingual artifacts. There has long been hope to discover an equivalent to the Rosetta Stone —heretofore, Indologists have not been similarly fortunate.

The entire discovered script consists of 419 signs across 2,290 artifacts, most of which are clay seals and graffiti (potentially leading to implicit contextual sample bias in signs found vs. those lost to time). In 419 signs, frequency varies massively—219 appear fewer than five times, of which 113 occur only once [104]. This can be partially attributed to the presence of complex graphemes (compound symbols), likely written as one symbol to save space on tablets [26]. Despite there being 2,290 inscribed artifacts, most inscriptions are constrained to about five symbols on average, with some even being only one or two signs. Only eight texts are longer than 15 signs, with the longest continuous text clocking in at 17 signs. The brevity of analyzable Indus datasets has proved a considerable barrier in machine learning decipherment, which has been verifiably effective in deciphering Ugaritic and Linear B (although both had known descendants) [52].

### 2.1.1 Classification

This inability to concretely identify which languages have claim to Indus progenitorship has sparked inflamed debate over the script's etymological classification. Much of this discourse is embroiled within non-academic nationalism and identity politics seeking to tie a specific populace or ideology to the oldest discovered civilization on the subcontinent [37]. Yet, despite the potential for tangentiality, classification within the scope of our paper serves solely to scientifically consider the potential modern-day implications of IVC connection to the TYC and the contextual significance of our results in pertinence to decipherment.

Respected scholars like the late Iravatham Mahadevan [57–65] and Asko Parpola [73–77] ascribed strongly to the theory that the Indus script is proto-Dravidian, a language family now primarily concentrated in the south of India with no known relatives. Both scholars have compiled detailed mappings of reconstructed proto-Dravidian onto Indus pictographic features, corroborated by careful epigraphic evidence. This work saw a major advancement in 2021 by researcher Bahata Ansumali Mukhopadhyay through her well-received reconstruction of the Indus word for ”tooth” using Bronze Age and Dravidian linguistic and archaeogenetic sources [68]. Yet, despite extensive peer-reviewed scholarship supporting the Proto-Dravidian hypothesis, some experts maintain that there is insufficient evidence to definitively classify the Indus script as Dravidian.

Among other prominent theories is a faction that has sought to link the Indus script to the later Brahmi script (first attested in the mid-3rd century BCE Edicts of Ashoka) and thus to spoken Sanskrit or Prakrit. Much of this scholarship is derived from opposition to the Aryan migration theory in a crusade to claim the Indian origin of the Indo-European language family in addition to the aboriginality of the Indo-Aryan people. Nevertheless, this has been firmly disproven—if not solely by archaeogenetic evidence, which definitively confirms the southwards migration of a branch of Indo-Iranians with steppe pastoralist DNA after the fall of the IVC [70, 80]—by the absence of urban cities in early Vedic civilization (c. 1750-500 BCE). The Indus script has been proven to share greater pictorial and stylistic similarities with Proto-Elamite and Sumerian pictographs (Proto-Cuneiform) than Brahmi, which is more closely linked to the Indo-European Greek and Phoenician alphabets [16–18]. Since we know the IVC had direct trade relations with Sumer and Elam, inferable by the number of Indus artifacts discovered in West Asia, it is understandable the script's closest known contemporaries are other pictographic languages of its age (although most scholars agree this is not due to common origin, but rather close contact and diffusion [104].)

## 2.2 The Harappans

### 2.2.1 Discovery of the IVC

While the first Indus seal was discovered in 1875, the IVC was not identified until September 20, 1924: the day a conclave of archaeologists assembled in Shimla for Sir John Marshall's announcement of the discovery of a new urban civilization. This news came off the backs of years of excavation in Harappa and Mohenjodaro by two Indian archaeologists, Daya Ram Sahni and Rakhaldas Banerji [42, 43]. Harappa was the first site unearthed in 1921; henceforth, the IVC has also been colloquially referred to as the Harappan civilization. Thousands of IVC sites along the Ghaggar-Hakra River and the Makran Coast have been identified since the discovery of these two ancient urban centers, recognizable by their meticulous city planning, advanced sanitation networks, and distinctive material works.

### 2.2.2 The IVC, Epitomized

The IVC was vast; undeniably the largest urban civilization of its time, it covered over 1,299,600 square kilometers (more than Ancient Egypt and Mesopotamia combined). The period of time that is attributed to the fully-evolved civilization spanned 700 years (2600 BC-1900 BC)—twice as long as the Ming Dynasty and four times longer than the Akkadian Empire, quantifiable as 28 generations [38]. The first pre-Harappan[1] cities, like Mehrgarh, Amri, and Kotdiji, existed before 3500 BC at the latest, almost a thousand years before the start of what is considered the main timeframe of the IVC. Distinct brick architecture style and urban design in the hills and valleys of modern-day Balochistan stretch the oldest dating of the Indus tradition back even further to around 7000 BC [10, 38]. It is likely the Harappans[2] had no natural enemies, as their geographic boundaries assured them relative safety and security [67]. Instead of conquest, the Harappans engaged in extensive trade networks with other civilizations of their time. Sumerian records mention a land called 'Meluhha', which is often attributed to the IVC, allowing us to reconstruct the commercial contributions of the Indus people. In addition, Indus artifacts have been found across the lands that comprised Sumer and Elam, furthering the case for contact and cultural diffusion. The Harappans were incredibly skilled in naval trade and had developed early technology for long ocean voyages [44], evidenced by relics of Harappan trade on the Arabian peninsula as well as remnants of IVC port of Lothal in modern Gujarat, as well as numerous other smaller ports on Saurashtra. Through Sumerian contact, we are also able to contextualize the potential usage of Indus seals and inscriptions as a mirror of equivalent tablets in Mesopotamia used for identification as well as commercial designation [69].

[1]Time period denotation of the era predating the maturity of Harappa, equivalent to Early Phase (3200-2600 BC).
[2]Used here in reference to the peoples of the entire IVC, rather than just the Harappan site.

### 2.2.3 The Decline of the IVC

We would be amiss to not address the most pertinent yet anticlimactic question surrounding the IVC: the reason and execution of its decline. There is a large temporal gap between the abandonment of IVC cities and the earliest settlements of the Vedic period, leaving much unknown about the cultures of South Asia in the time between. The theory of Indo-Aryan invasion into the subcontinent has not only long been abandoned in favor of a gradual migration [80] but also because it would not explain the lack of continuity in material culture; ergo, it is unlikely the Harappans were defeated in conquest. The most scientifically compelling factor currently attributed to its collapse is a geological change in monsoon patterns and the fluvial recession of the Ghaggar-Hakra River [19, 29], both disastrous for a riverine society. Other potential theories concern the southward migration of the people of the Bactria-Margiana Archaeological Complex (BMAC) and the 3rd millennium fall in sea level disrupting vital Indus trade networks [67], as well as endemic malaria and possibly other diseases supported by biological analysis of human remains at Mohenjodaro [2]. Whatever the reason, it stands that a sizeable demographic migration ensured eastward to the Himalayan foothills and southward to peninsular regions, leading to the abandonment of major IVC urban centers by 1900 BC (Figure 2.1).

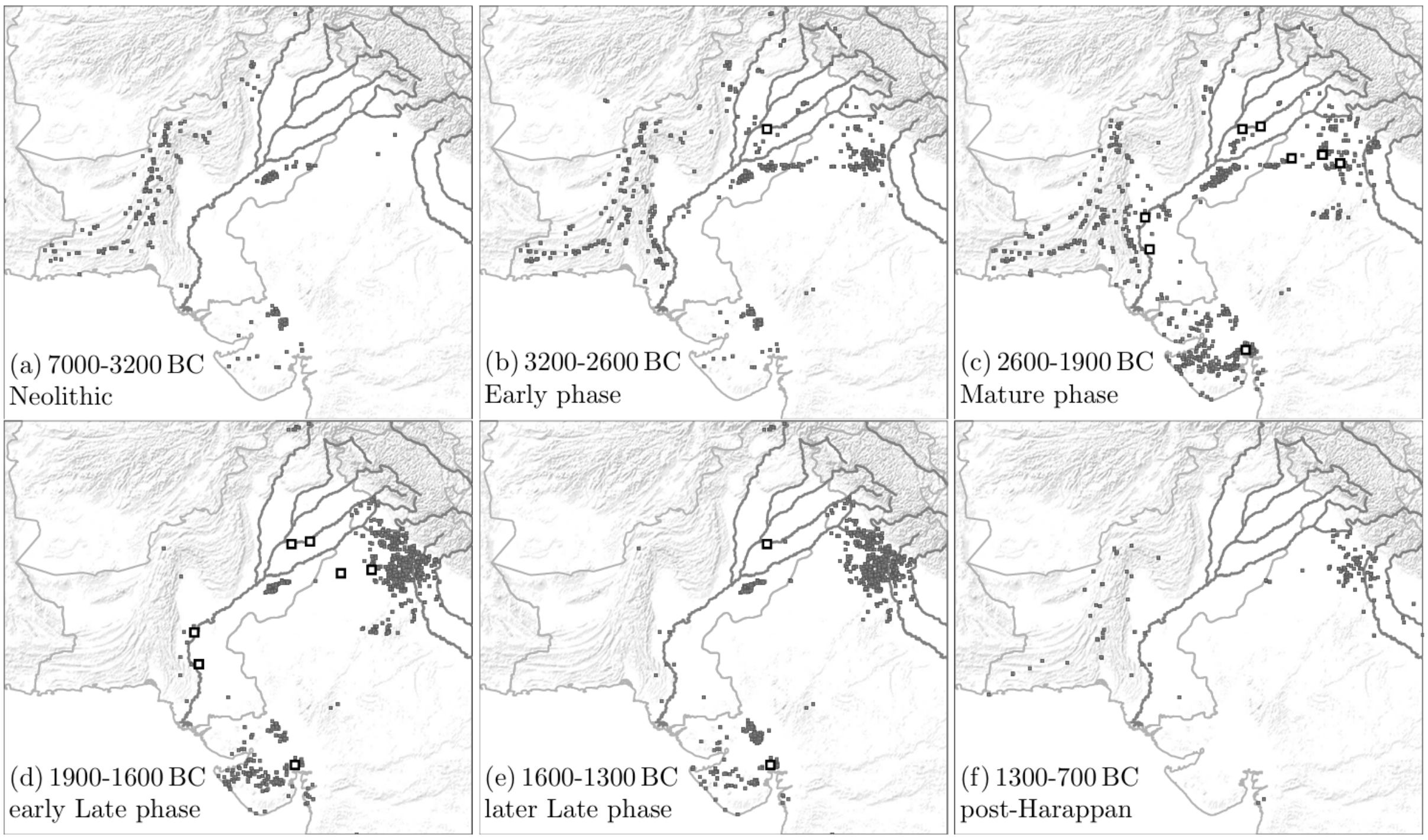

Figure 2.1: Distribution of known IVC sites by time period [38, 46].

## 2.3 Peoples of the Tibetan-Yi Corridor

The TYC is credited as the natural pathway for the ethnic branching and eventual migration of the Tibeto-Burman-speaking peoples onto the Qinghai-Tibet plateau [101]. The unity of varied oral traditions and genetic patterns affirm that these populations originated in the Yellow River basin of the Gansu-Qinghai region in pre-Neolithic times prior to a mass southward migration [87]. From there, the routing and timing of individual groups' movement down the corridor divided these early people into the 16 distinct ethnic minorities currently present within the TYC. The natural geographic isolation of the corridor within the Hengduan Mountains allowed for the preservation of diverse proto-cultures in modernity, making it an ideal environment to study the effects of ancient flows of people, technology, and tradition.

### 2.3.1 Migration Flows of the Naxi

Within this unique cultural petri dish, we encounter the Naxi people of Lijiang, China. Like all TYC peoples, the Naxi can trace their origins in antiquity to the upper reaches of China's Yellow River basin and share close ethnic heritage with the Tibetan people[33, 100]. Not unlike the Ghaggar-Hakra River, paleoclimatological evidence shows that the climate of the basin became significantly colder and drier circa 6000 BC, driving the southward migration and separating Tibeto-Burman from Sinitic within the greater Sino-Tibetan language family (cite: Shuo Shi). This downward distribution trend was dominant in the TYC during the span of time from the beginnings of the pre-Neolithic demographic transmission until the Tang Dynasty, although we cannot pinpoint precisely when the Naxi people became a distinct identity. We do know, however, that they migrated in a fan-like pattern, beginning as one larger people and separating at crucial crossroads (or perhaps 'crossrivers') until they were distributed in a belt among the upper and mid-reaches of the Jinsha, Shuiluo, Wuliang, and Yalong basins ([100]) —geologically divided to the point of diversion.

### 2.3.2 The Dongba Script

The Naxi are famous for their distinct Dongba script, hailed as the world's last living pictographic script [71]. Though the earliest surviving artifact on which the script can be found is dated to 30 AD [39], we cannot say for sure when the Dongba script was created, but analysis of environmental factors regarding the symbols for north and south point to its establishment occurring after the point at which the Naxi reached the Yunnan region [47, 103]. It seems to have disseminated from west to east, with the most rudimentary form of the script beginning in the Western Rerko region (Figure 2.2) and the most "advanced" hailing from the eastern Ludian region (Figure 2.3) [94].

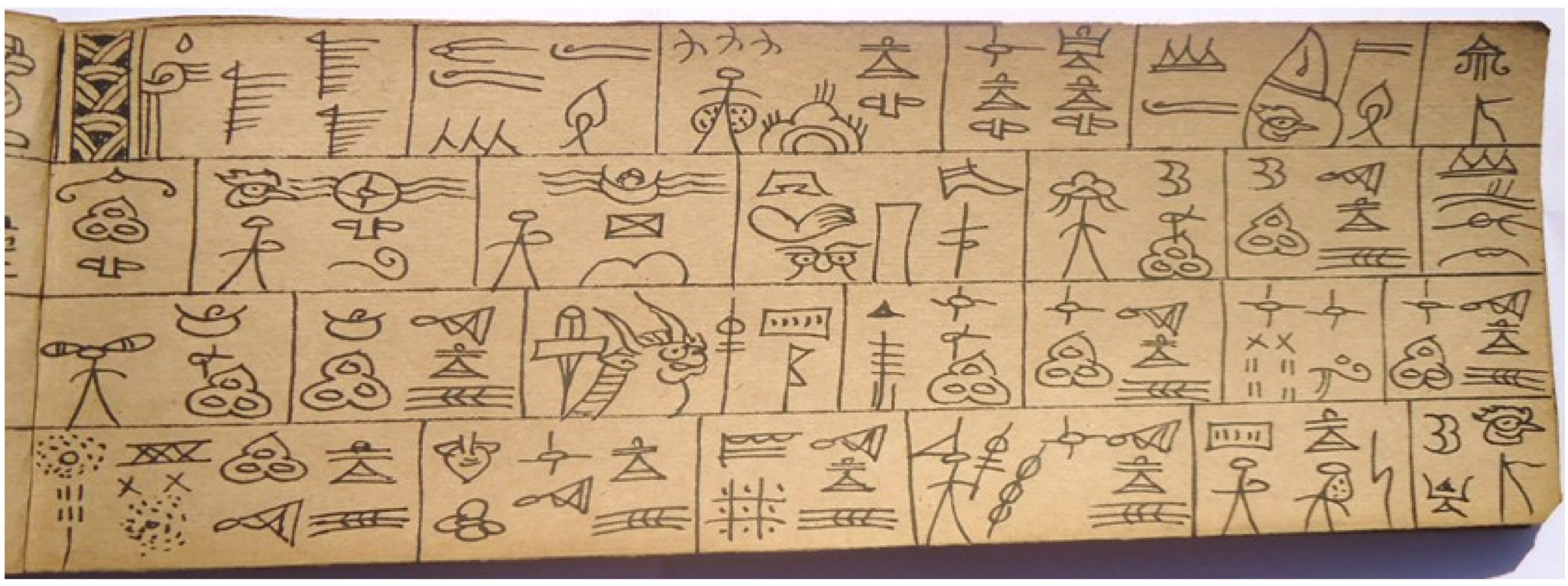

Figure 2.2: "Ancestor-worshiping scripture of Rerko branch" by Dongba Yang Zhashi, Youmi Village, Labo [100].

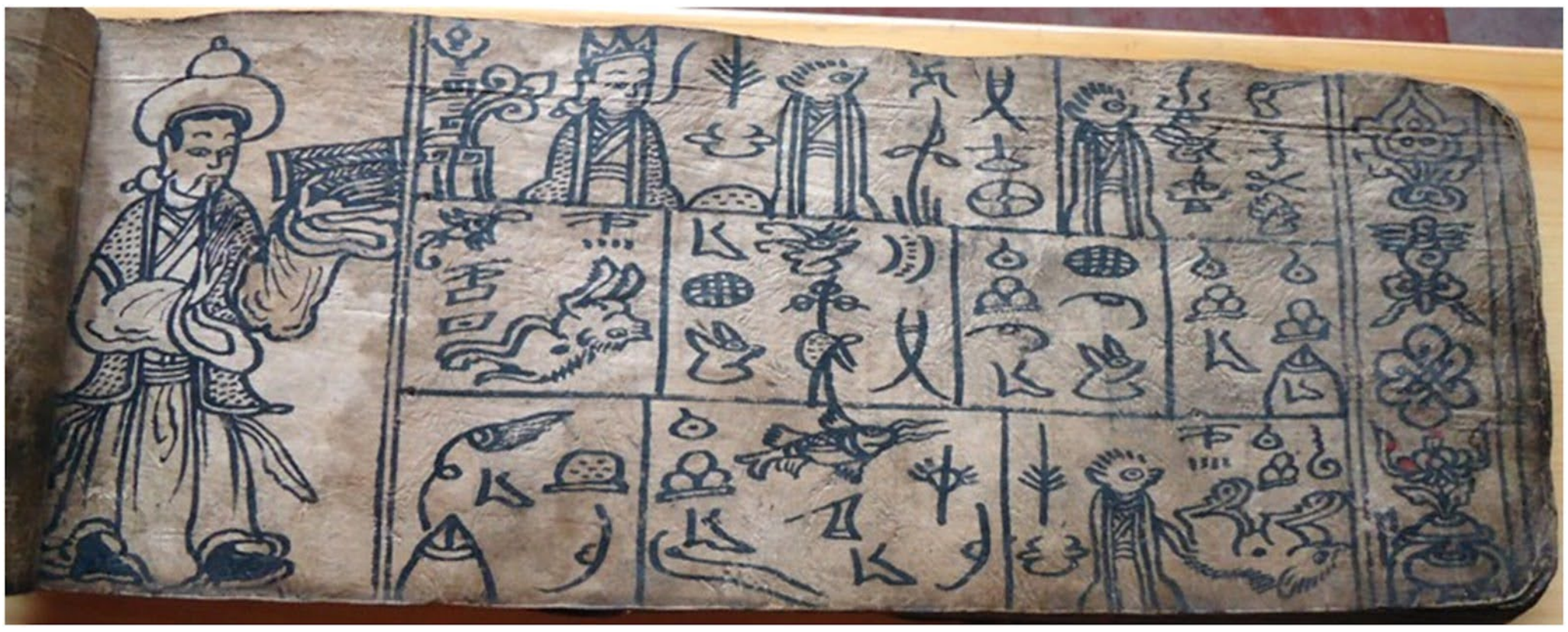

Figure 2.3: "Ancestor-worshiping scripture" by Dongba He Shijun, Xinzhu Village, Ludian [100].

Within the possibility that the Dongba written tradition was developed into a full-fledged script post-southwards-migration, pictographic influence from early civilizations may still have influenced the Naxi people not only at a time prior but also at any position from the Yellow River basin to along the TYC—which, despite having a distinct geographic locale, is first and foremost a heritage route. The name of the Dongba script in the Naxi language is *ser jel lv jel [s⊠⊠ tc ¸⊠⊠ lv ⊠ ⊠ tc¸⊠⊠ ]*, literally meaning "records made on wood and stone" [79]; we can infer the earliest form of these glyphs were not full-fledged manuscripts but rather simple carvings in caves or trees that were later complicated. Arguments for shared influence on the pictorial history of the corridor are found in the first Ba-Shu scripts of early polities in modern Sichuan and isolated works of pre-Buddhist rock art on the Tibetan Plateau (Appendix A.8.5), artistically and semantically similar to the Dongba pictograms.

The Dongba script was initially developed for use by Naxi priests, aptly called Dongbas, in recording divination scripture—originally, these priests were practitioners of the Tibetan Bön religion. Other peoples of the TYC, like the Yi, also have similar priestly traditions as a marker of their shared origin; however, only the Naxi developed and preserved this manner of pictographic script. Its primary use in antiquity was for manuscripts recording stories of Naxi history and mythology, although this has become generalized over time due to post-20th-century revitalization efforts. Despite consisting of relatively complex pictographs, these symbols have always denoted sound—even in times when usage was limited to divination and priestly work.

### 2.3.3 The Yi

The Yi people (self-named Nuosu in the Liangshan region) are one of the largest ethnic minorities of Southwest China, with a population of about 9–10 million spread across Yunnan, Sichuan, Guizhou, and Guangxi provinces [22]. Dwelling in the highlands of the TYC, a hallmark of their distinct culture is their ancient written script, recorded by Yi priests called bimo (comparable to Naxi dongba). Used at least since the 15th century, the original script consisted of thousands of logographic symbols. Over time, many regional variants emerged, as different Yi areas often had their own glyph forms; scholars have documented as many as 8,000–10,000 characters in old manuscripts (with perhaps 90,000 variants). These texts include scriptures, chronicles, and genealogical records that encode the Yi's collective memory. Notably, Yi classical literature does not track Chinese dynastic history but follows its own historical lineage; Yi classical records recount the deeds of ancestors and migrations in a framework independent of Han Chinese historiography [48]. In the 1970s, as part of cultural preservation efforts, the Chinese government worked with Yi scholars to standardize the script. A Modern Yi script was introduced in 1974 (based on the Liangshan dialect) with 756 standardized characters, replacing the multitude of classical forms for official use.

Ethnologists like Gao Zhiying have specifically examined Tibetan and Naxi cultural influences in the western part of the TYC, reflecting how the concept of the corridor has been used to understand inter-ethnic relationships [33]. Naxi legends and Yi legends sometimes reference similar themes (e.g. flood myths, ancestral heroes), hinting at ancient connections or shared narratives in this mountain world. Additionally, genetic studies of various TYC populations (including Yi, Naxi, Hani, and others) have revealed complex admixture and reinforced that the corridor has historically facilitated ethnic flows and diversification rather than social isolation [103]. The Yi people's identity has been shaped not only by their internal traditions but also by this ”corridor context”—a melting pot of Tibeto-Burman cultures.

### 2.3.4 Sanxingdui and the Ba-Shu Pictographs

The Sanxingdui (and Jinsha) civilization refers to a Bronze Age culture that flourished in the Sichuan Basin of Southwest China and is associated with the ancient Shu Kingdom. Centered near modern Guanghan, 40 km northeast of Chengdu, the Sanxingdui site spans roughly 12 square kilometers and dates from around 2700 BCE to 1000 BCE [49]. Despite its long duration, this culture was unknown to recorded history until archaeological discoveries in the 20th century revealed its remains. The breakthrough discovery occurred in 1986, when archaeologists excavated two sacrificial pits at Sanxingdui. These pits yielded an astonishing trove of over 900 artifacts, including large bronzes, gold objects, jade carvings, and elephant tusks. The finds and material culture were unlike anything previously seen in Chinese archaeology, suggesting a complex civilization entirely separate to Shang China. While Sanxingdui revealed no written language, eight inscribed symbols were noted [82].

Colloquially, the Ba-Shu scripts refer to three undeciphered scripts used in the Ba and Shu kingdoms—the presumed descendants of the Sanxingdui and Jinsha civilizations—in the 4th and 5th century BC [81]. The first is a pictographic script used to decorate bronze weapons found in Ba graves, of which nearly 200 individual symbols have been discovered. The later two, interestingly, draw similarities to the later Yi script, and have been understood as potential predecessors.

In the context of the broader TYC, the southward migrations of the surviving Shu people following their defeat by the Qin in 316 AD was the final edition of the mass southwards migration of Tibetan-Burman that defined early demic flow in the TYC [88]. Han, Wei, and Jin dynasty records prove that in the following years, the ”Yi” barbarian names are all connected to the Shu, who integrated into the Yi upon leaving their original locale. Seeing as the TYC was a natural evacuation route, we can assume the Ba-Shu cultures were privy to non-demic transmission along the corridor as well.

## 2.4 Pictorial Parallels

We have noted a few preliminary similarities in qualitative visual analysis of the Indus script and the Naxi Dongba script, augmented by brief notation of connections to the Ba-Shu pictographs (and by extension, the Yi script). Through our study, we specifically examine fundamental iconographic motifs and concepts less likely to be lost or misconstrued across a large spatial and temporal gap, as exists between the IVC and the TYC.

The Indus script is less literal in its representations than the Dongba script. While the Dongba script has discrete pictography and only started being rearranged for phonetic usage later in its lifespan, the Indus script contains more abstract modifiers and signs that cannot be understood intuitively. It is not a primitive system and was in use for at least 700 years, over 1.2 million square kilometers. The Naxi language is also firmly within the Sino-Tibetan language family and bears no similarities to Indo-European or Dravidian languages—we do not argue any phonetic connection. Nor do we claim the Dongba script is a direct descendant of the Indus, as it lacks far too many of the Indus script's modifiers and basic mechanisms to make such a connection. However, the transmission of arts and technologies—like writing—travel in unique ways, and we maintain that the iconography of the IVC in some way, shape, or form made contact with the TYC, as detailed below.

It is important to briefly note that some basic shared symbols also have potential counterparts in the Proto-Cuneiform and Proto-Elamite signaries, positioning them to also be plausible sources of origination and inspiration. This gives our alternate hypothesis salience within the context of our study.

### 2.4.1 Numerals

The original hypothesis inspiring this paper stems from an observation of the similarities between the assumed number system of the IVC and the numerals of the Naxi Dongba script. Standardized weights prove the Harappans had, at least, a concept of enumeration and mathematics long before the first previously known appearance of arithmetic in the ancient world. Measurement tools have been unearthed at IVC sites, revealing that the Indus "inch" was precisely double the Sumerian sushi, possibly to simplify conversion during commerce or mark a common origin [91]. Harappans, distinguishable by their multi-story brick architecture, crafted blocks with remarkable uniform precision using optimal geometrical ratios, proving they had advanced knowledge of angles and proportions prior to the theorems of Pythagoras [28, 38].

It is a common assumption that the stroke signs, as seen in Figure 2.4, in the Indus script represent numerals, as they often do in other systems [40, 73, 97]. As far back as 3500 BC, the pre-Harappans represented numbers in their inscribed works, starting with a system of one to 13 vertical strokes [28].

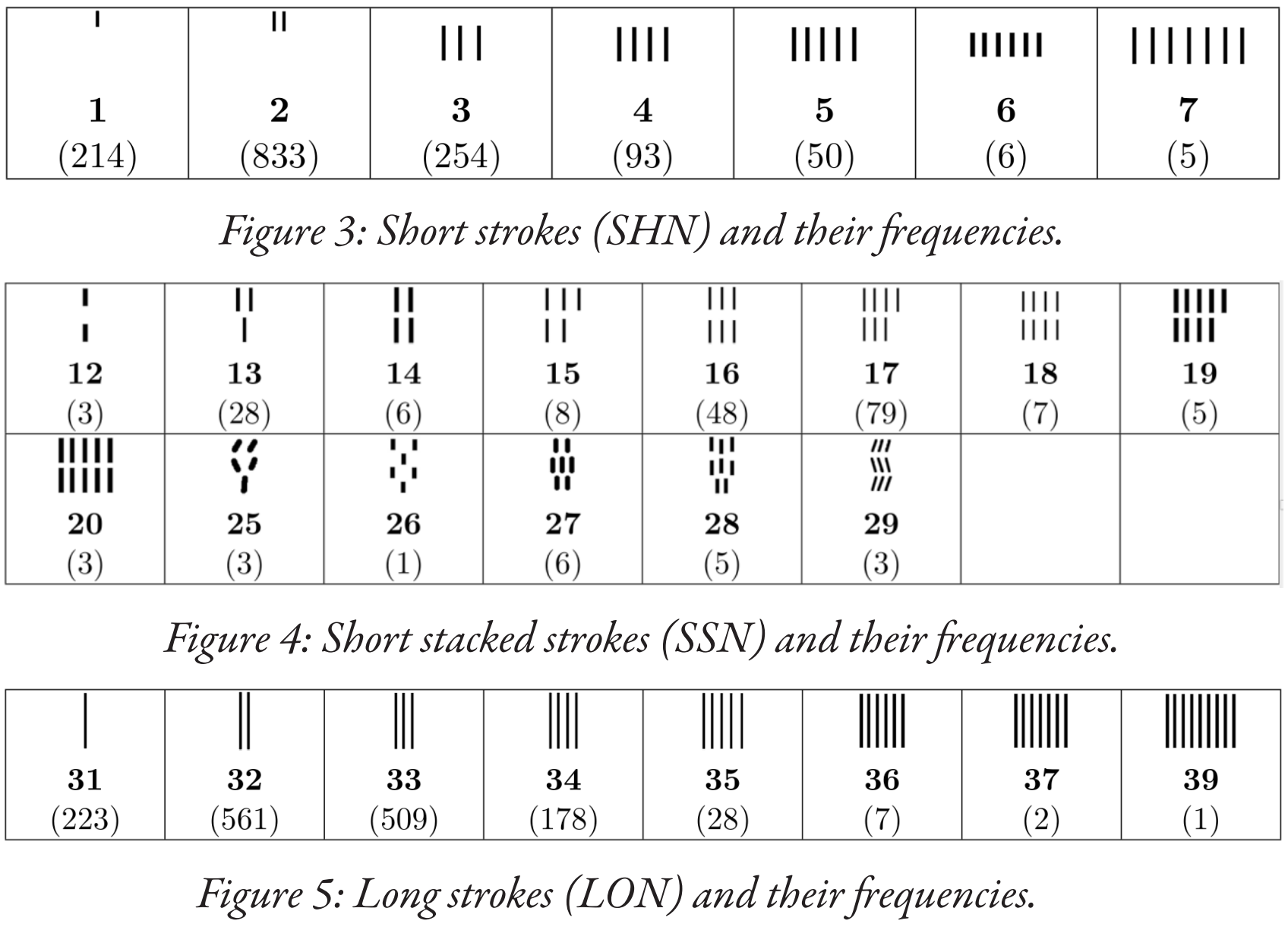

Figure 2.4: Indus stroke signs and frequencies, as documented in the Interactive Corpus of Indus Text (ICIT) [26, 96].

By 2500 BC, the Indus script had switched to an abridged system, though we do not know for sure the base or even whether they used more than one. Some suggest an octal system, based on a semantically related reconstructed Dravidian root connecting ”8” with ”to count/number,” as well as there being a maximum of 7 short strokes at any one point [20, 26, 66]. Others suggest a base-10 decimal system as the maximum stroke count expands to 9 when we consider long strokes as well [97]; a substantial piece of evidence for the decimal system is a fragmented length measure from Mohenjodaro, subdivided into 10 units (Appendix A.10) [30, 56]. Indus numeral notation appears to be relatively complex, as evidenced by recurring sign pairings that suggest non-numerical meaning, variations in stroke length, and positional relationships that indicate value significance [26].

Pictorially, the strokes are reminiscent of the counting systems of Sumer and Elam (Figure 2.5). However, the Mesopotamian and Elamite counting systems are sexagesimal (base-60, or perhaps base-10 mod 60) [23, 24], and operate differently from a traditional base-10 counting system. If the Harappans had used base-10, cooperation with Sumer and Elam would have been notably streamlined, whereas an octal system would have required regular conversion. It remains possible the IVC used both base-8 and base-10 systems, similar to the dual usage of base-60 and base-10 in Mesopotamia and Elam.

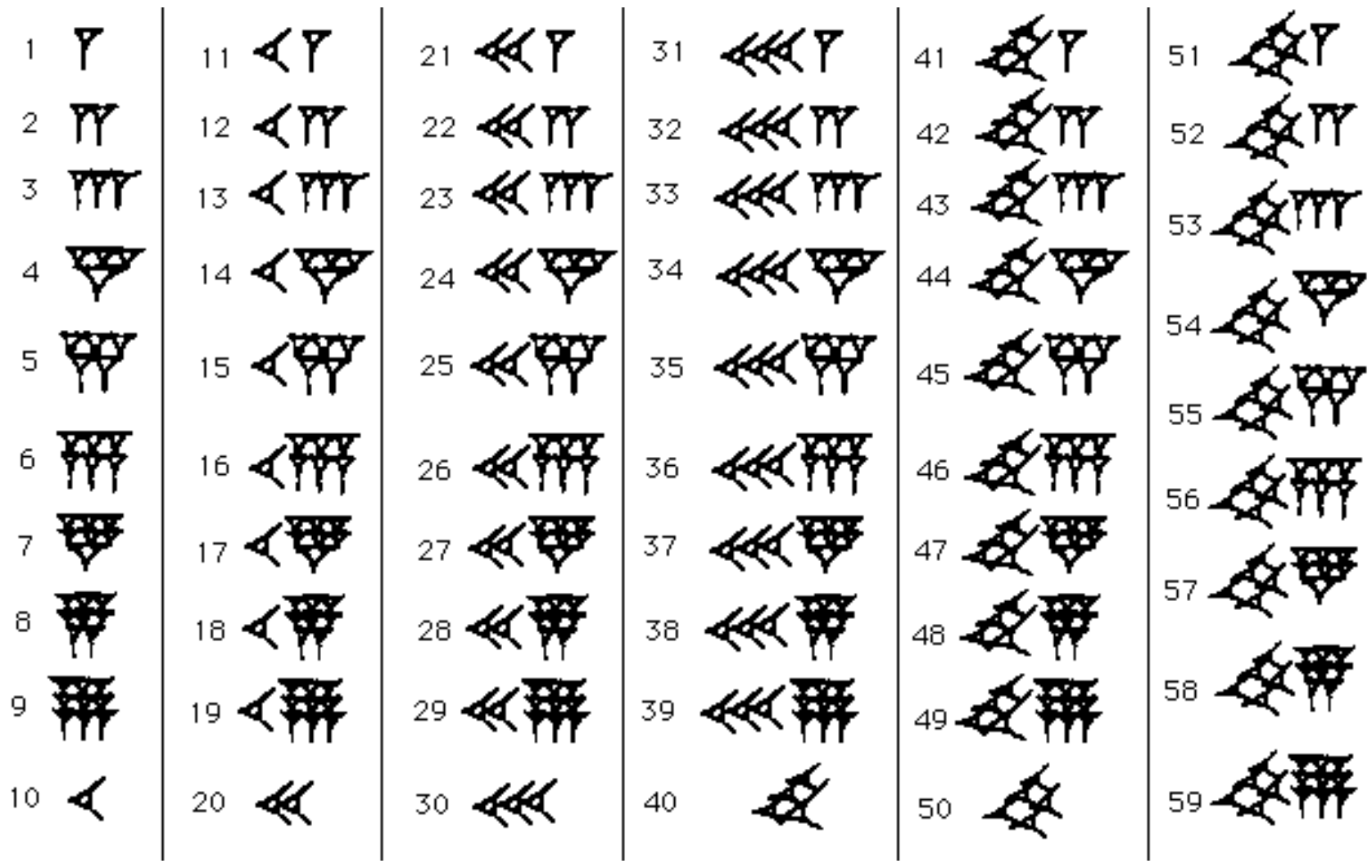

Figure 2.5: Cuneiform base-60 numerals up to 59. Each triangle represents a clay wedge impression [13].

Dongba numerals are represented in small strokes in the same style as the civilizations to the West and are explicitly base-10. Despite this tally system being arguable as an intuitive way to record numbers, the Bronze Age civilization(s) to the east of the TYC adhered to entirely separate conventions in their numeral systems. The earliest Chinese numerals were tallied horizontally for the first four numerals and then had unique forms to represent values from five onwards (Figure 2.6). Other groups in the TYC that developed a script, like the Yi, did not develop independent numerals and borrowed the Chinese system when necessary.

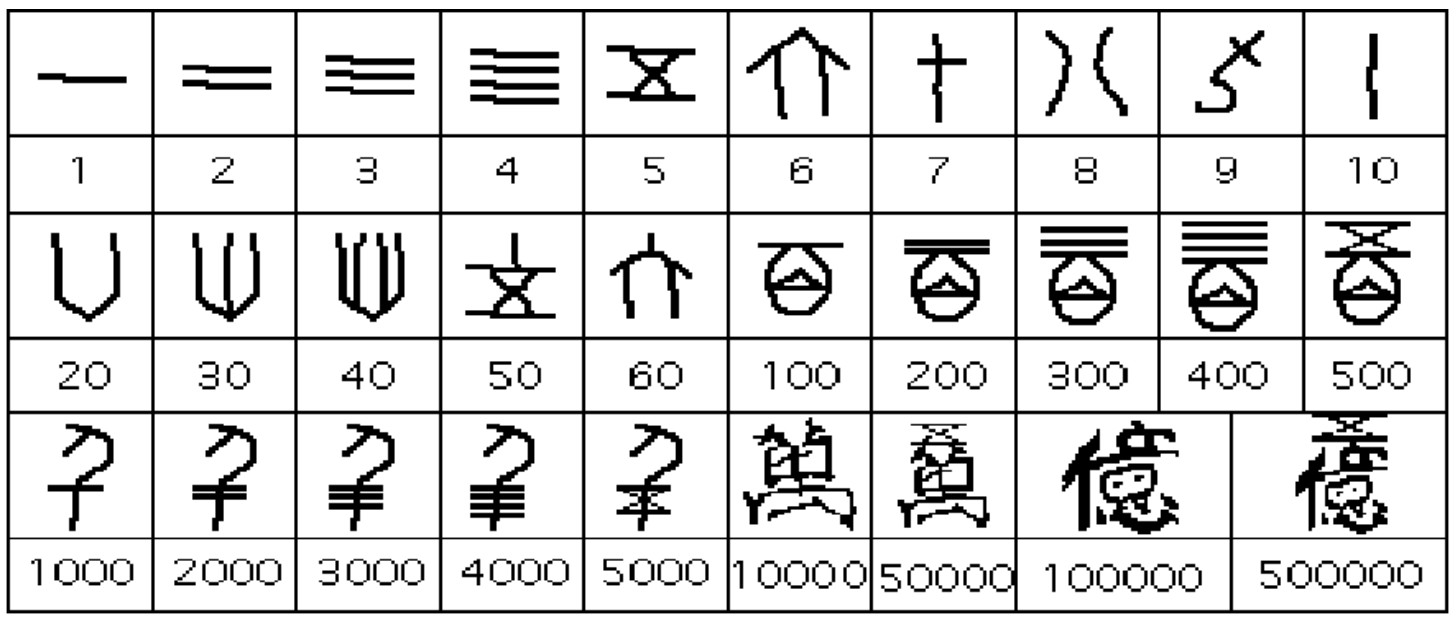

Figure 2.6: Archaic Chinese numerals during the Shang Dynasty, beginning with horizontal counters [72].

Stylistically, Dongba strokes are not wedges (like cuneiform) or dots (like Mayan glyphs), but straight strokes such as seen in the IVC—this mostly has to do with the materials on which these scripts were traditionally written, as West Asia often carved and impressed into softer clay while the IVC and the Naxi carved in harder stone and wood that would inhibit wedges. In later manuscripts and the standardized font used by the Chinese government in the city of Lijiang, the strokes have a Chinese *gōu* (钩) appended at the end of each stroke, perhaps due to an intentional Sinicization of the script. This feature is not present in older-style manuscripts (Figure 2.2, 2.7) in which the strokes resemble those of the Indus script. There remains a possibility of independent origination, but the uncanny structural and visual similarities of Dongba numerals to older writing systems support our hypothesis of the IVC or West Asia being a stylistic source.

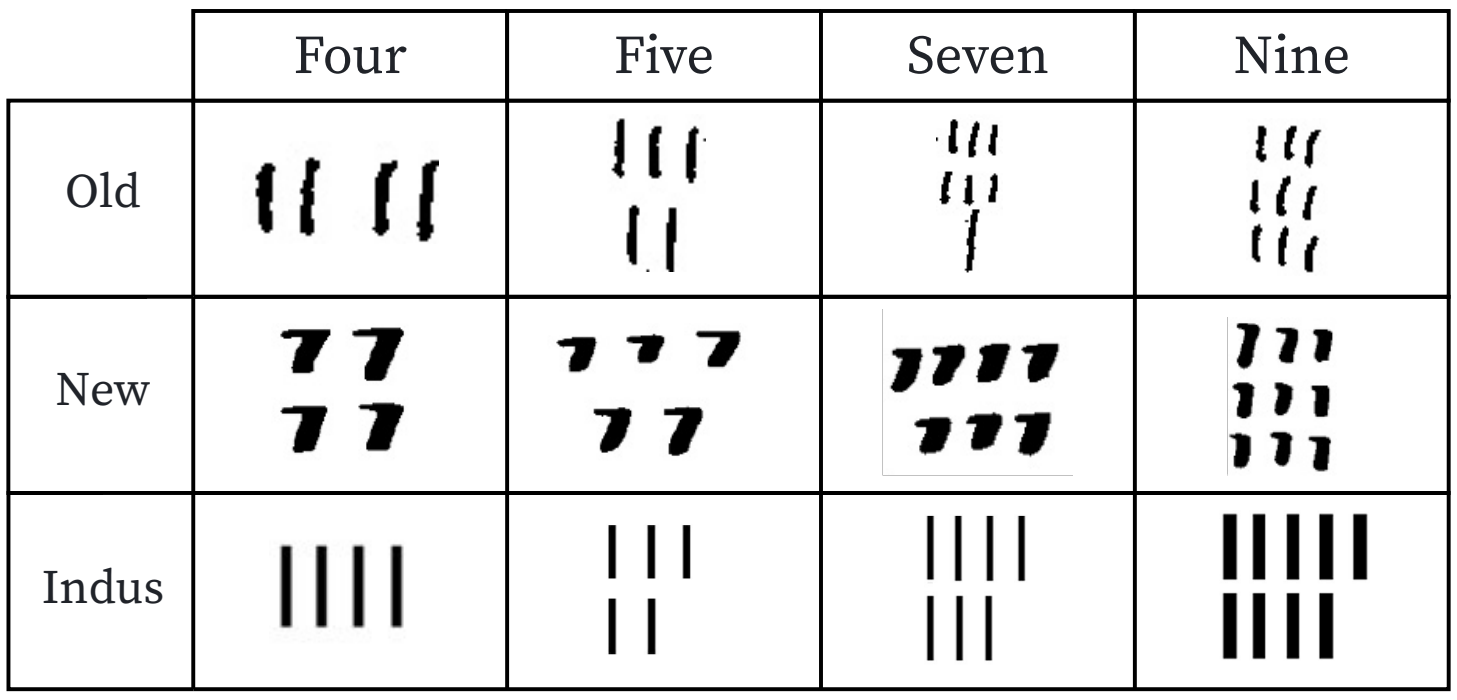

Figure 2.7: A comparative illustration of stroke patterns in the Indus script alongside early and contemporary Dongba numeral systems.

### 2.4.2 Gender Markers

The second preliminary finding guiding my hypothesis is similarities in theorized gender symbols across the Indus and Dongba scripts, based on Mahadevan's final interpretation of the Indus jar and arrow signs [58, 59, 61], two of the most frequently appearing signs in the Indus signary.

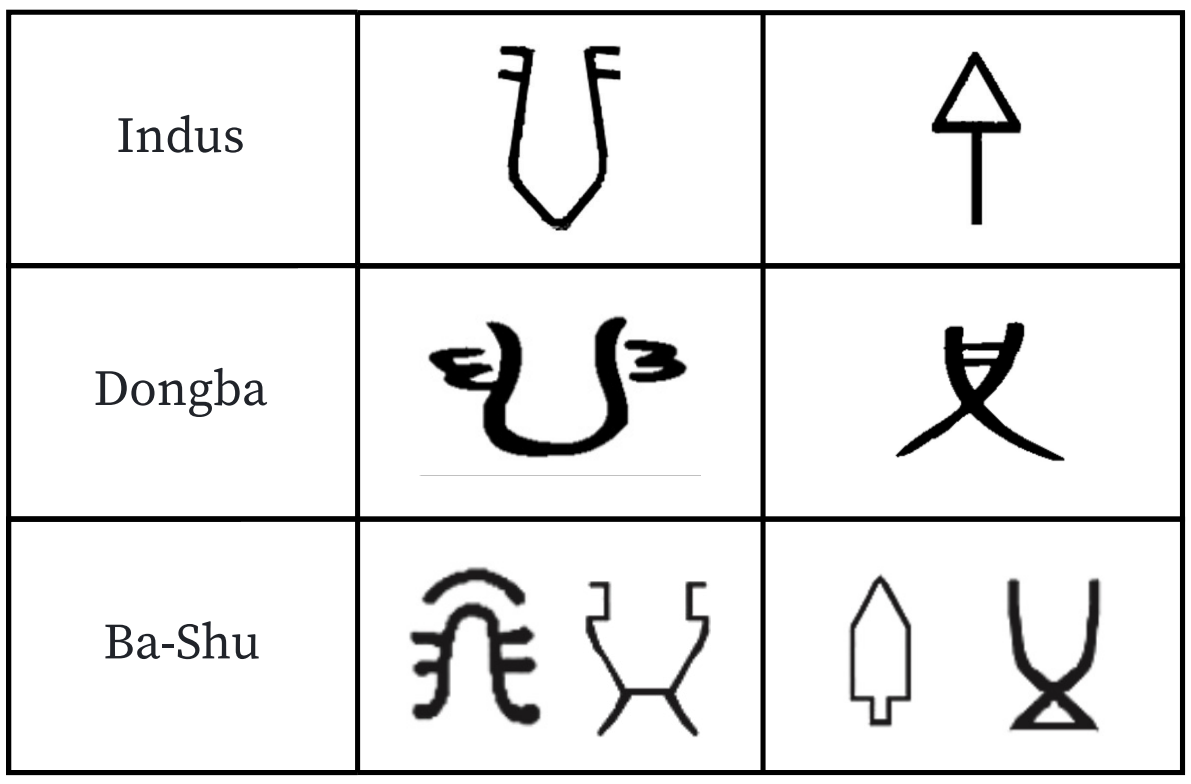

Figure 2.8: Dongba gender modal particles, Indus jar and arrow symbols, and additional similar Sanxingdui/Ba-Shu pictographs.

Across all 2,290 artifacts, these signs have never appeared as a pair despite appearing relatively often in inscriptions; in fact, the jar is the most frequent sign by far. In 2011, Mahadevan, who wavered on his interpretation of these signs often, deemed this mutual exclusivity as indicative of gender markership [61]. In adherence to the Proto-Dravidian hypothesis, he connected via rebus the arrow sign (*ampu*) and the Old Telugu suffix *-(a)mbu* [83], deeming the Indus arrow the non-masculine singular suffix. Naturally, its counterpart, the jar, falls as the masculine suffix. The phallic shape of the jar and, more distantly, the triangular shape of the arrow mirror equivalent iconography in the Dongba script (Figure 2.8, in which gender signs are typically used as non-verbalized modal particles. It is worth noting similarly shaped symbols also appear in Ba-Shu pictography.

### 2.4.3 Other Similarities

There are a few other iconographic similarities worth looking into. One is the Naxi sky character (see Figure 2.9), not traditionally used alone but rather as a modifier on top of another symbol. The Indus roof character functions quite similarly, often placed above the com-mon "fish" and other symbols. The Ba-Shu corpus seems to have similar characters as well, although we cannot confidently attest to their function.

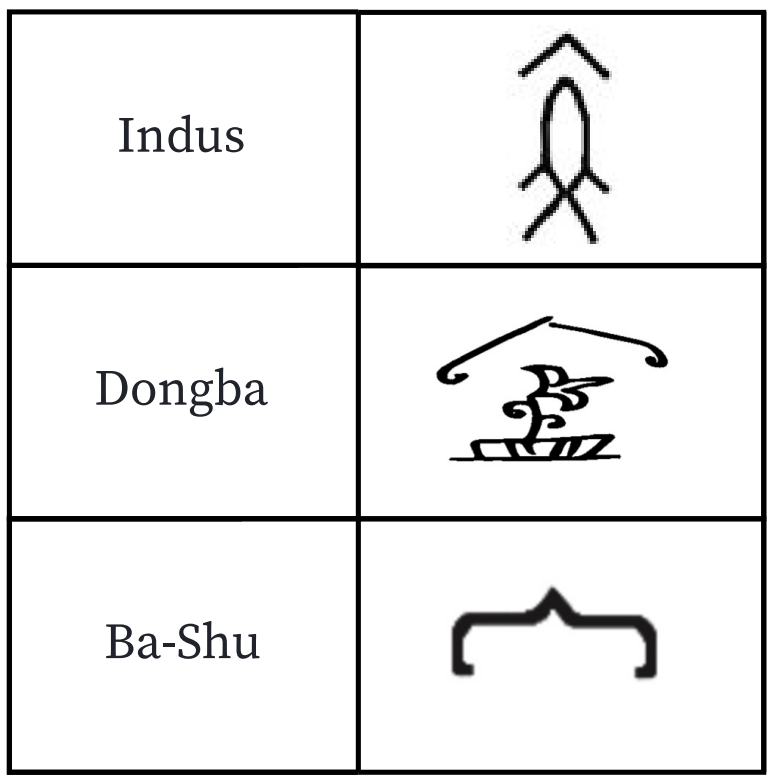

Figure 2.9: The Dongba sky modifier, the Indus roof symbol, and a related Ba-Shu pictograph.

We will also consider the Dongba symbol for fire, identical to an Indus and a Ba-Shu sign (Figure 2.10). This is not proof in and of itself, as many ancient civilizations retained the same rudimentary sign. Both the oracle bone script (甲骨文) of the Han Chinese and the scripts of civilizations to the West connected this sign to hills or mountains, and scholars have assumed the same for the Indus script thus far [73], [57]. The Dongba script uniquely considers the triple triangle sign to mean fire.

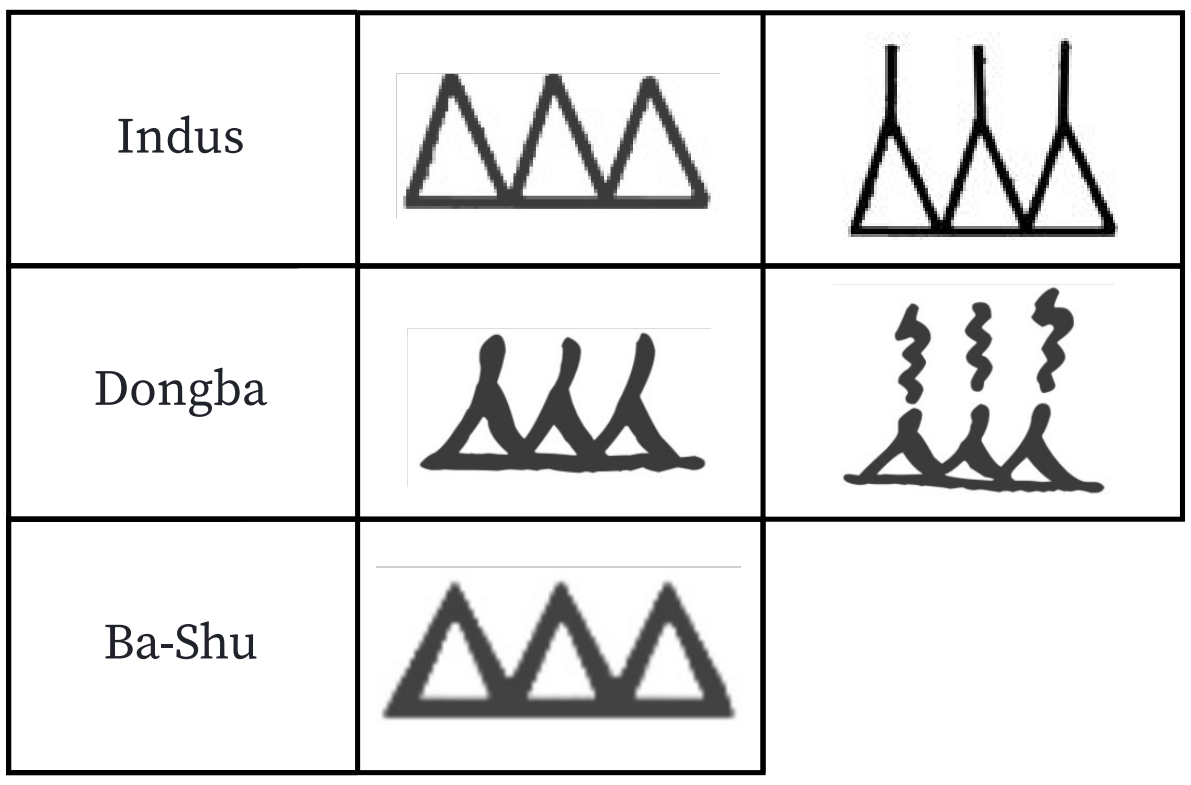

Figure 2.10: The Dongba fire symbol alongside the three-triangle motifs present in both Indus and Ba-Shu scripts.

Our argument lies in the modified Dongba symbol for "smoke." It depicts a line extended from each of the triple peaks, zigzagged in modern writing but often a straight stroke in older manuscripts. This same variation exists in the Indus corpus—and would not make as much sense if referring to a mountain. This is not to say the mountain interpretation is wholly incorrect, as the Indus script is not ideographic and individual symbols often refer to separate, abstract meanings. Yet, given the pictorial closeness, this connection is worth considering.

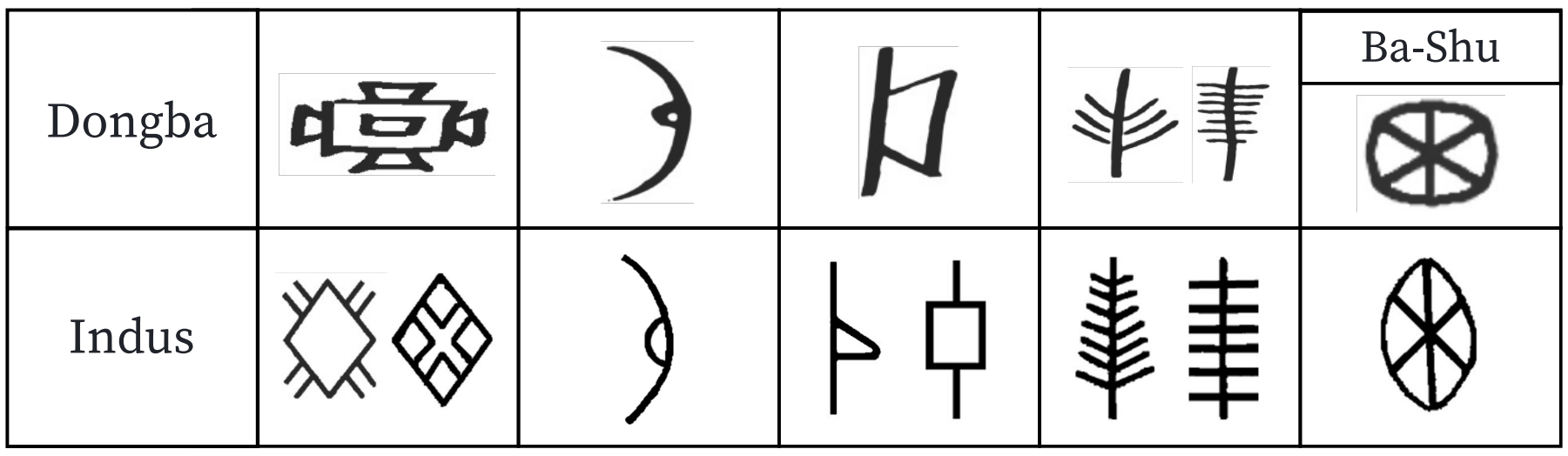

Figure 2.11: Further comparisons highlighting similarities between TYC and Indus script signs.

As shown in Figure 2.11, there are other signs with visual similarity at first glance, though this list is not exhaustive.

## 2.5 Routing Transmission

There is sound evidence for intraconnection within the corridor [87, 101]; yet, in order to convincingly argue transmission from the IVC to the TYC, we must establish plausibility in spanning such a gap in space and time. Evidence via radiocarbon dating leads us to consider the ancient Shu-Shendu road (Figure 2.12, connecting Northeastern India to the Hengduan Mountains of the TYC. Future expansions of this route during the Tang and Yuan dynasties extend it to the base of the Himalayas, where we know the Indus Valley people migrated towards following the civilization's fall [38, 46].

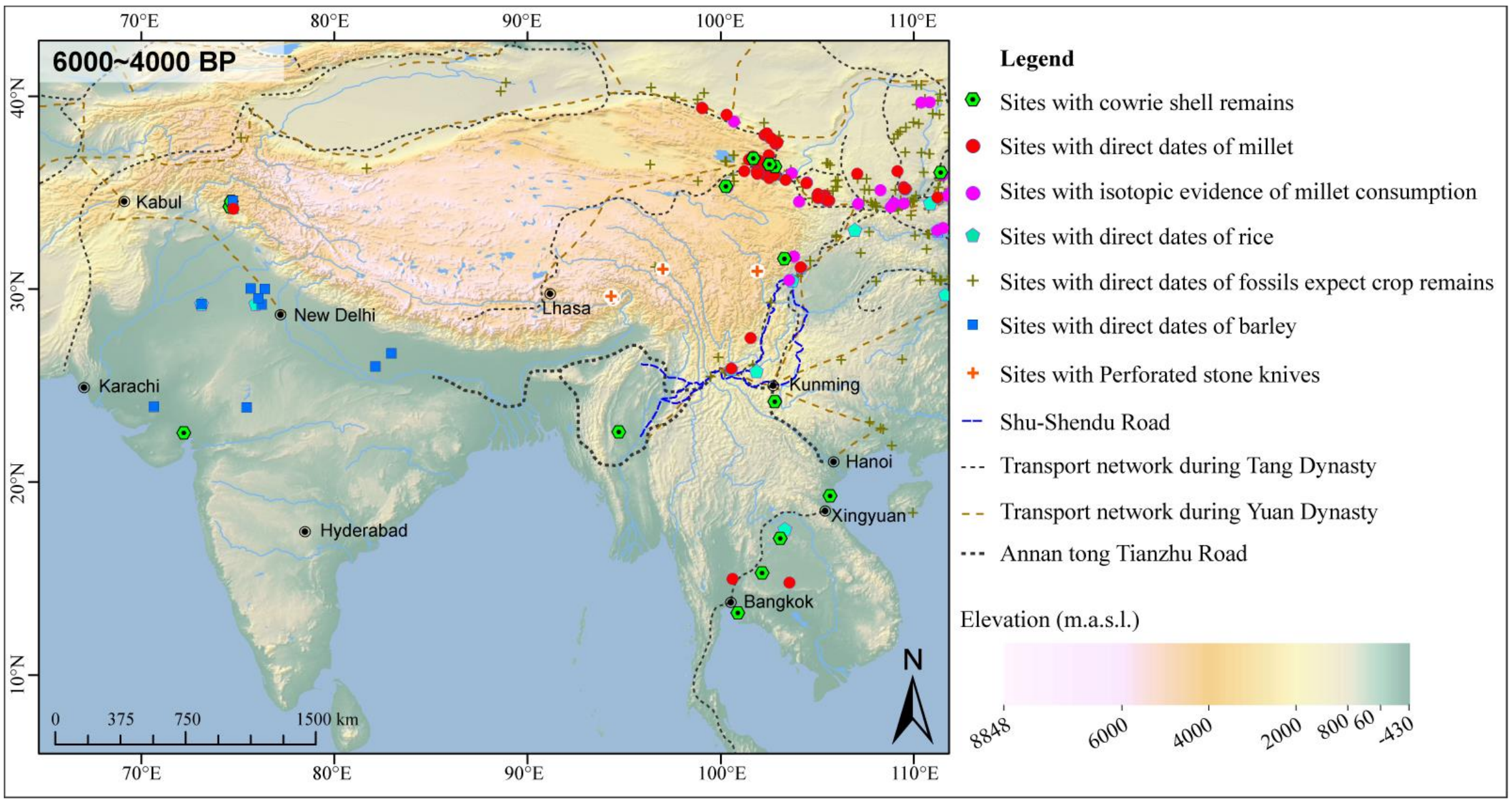

Figure 2.12: Map depicting material culture exchange and transmission routes between the TYC region and South Asia during the period 4050–2050 BC (6000–4000 BP) [53].

It is possible transmission may have occurred during the second millennium BC, when the IVC cities were abandoned and cultural exchange between China and South Asia significantly intensified along the trans-Hengduan mountain routes [53]. Studies have proven that this route, also sometimes called the Southwestern or Southern Silk Road, was a major transmission route for millet cultivation from the Indus Valley into China beginning as early as the third millennium BCE [35]. The discovery of cowrie shells from the Indian Ocean at sites in Southwestern China, along with perforated stone knives of Northern Chinese origin in the Swat Valley and Kashmir[3], demonstrates robust bi-directional cultural transmission [27]. Other valued commodities, such as ivory (found at Sanxingdui and Jinsha sites), traveled via merchant intermediaries through the Lancang, Jinsha, and Nujiang valleys, forming natural corridors through the challenging Hengduan terrain and facilitating intellectual movement. Additionally, hybridization between japonica and indica rice varieties in South Asia after 2050 BC provides further evidence of sustained contact networks through which culture and iconography could have disseminated, influencing indigenous symbolic patterns along these highland routes [25]. All matters considered, the point stands that, albeit somewhat unexplored, multidisciplinary evidence supports plausible linkage between the period and locale of the Indus Valley and the early heritage of the Tibeto-Burman people.

[3] We note that at the time of the IVC, Kashmir and the Swat Valley remained Neolithic despite being within close geographic proximity to the urban Harappan society [6].

Chapter 3

# METHODOLOGY

## 3.1 Dataset Composition

The dataset for this study includes three target ancient scripts and three comparison datasets, with a focus on visual structure (for complete documentation of our data acquisition, see Appendix A.1). The target scripts are the Indus Valley script, Proto-Cuneiform, and Proto-Elamite, representing possible originations for influence on the TYC. For comparison, we collected datasets of the Naxi Dongba script as well as from the broader TYC region—specifically, one set of older Naxi Dongba pictographs, another of a more modern Naxi Dongba standardization[1], and a set consisting of ancient TYC scripts. The TYC comparison set was constructed with equal proportions of its constituent scripts, combining historical Naxi, Ba-Shu/Sanxingdui symbols, and handwritten Classical Yi[2]) via augmentation to ensure a balanced representation. This enables a controlled comparison to see if TYC-region scripts share more visual commonalities with Indus script than the West Asian tradition. We did not filter for frequency—hapax legomena or rare occurring glyphs are frequent in low-resource ancient data and do not stray from cumulative stylistic tradition [26, 73].

## 3.2 Data Preprocessing

All script images underwent uniform preprocessing to facilitate joint training. We standardized the image resolution across datasets to a fixed size of 224×224 pixels—the default input size for both the Swin Transformer and ResNet models we used—padding or scaling images as needed while preserving aspect ratio. This was implemented using a custom `SquarePad` transform followed by a standard resize operation. Resolution normalization ensured that differences in image size or scale do not bias the model.

Images were converted to consistent grayscale intensity since color information was not relevant to the structural qualities of each glyph. We applied intensity normalization through PyTorch's standard normalization transform (mean=[0.485, 0.456, 0.406], std=[0.229, 0.224, 0.225]) to account for varying lighting or contrast in source photographs.

Only the historical Dongba data was denoised as it was retrieved directly from manuscript scans, while the other datasets came preprocessed. The denoising process involved converting color images to grayscale, applying adaptive thresholding (`ADAPTIVE_THRESH_GAUSSIAN_C`) for binarization, performing morphological closing operations with a 2×2 kernel to reduce noise, and applying Non-Local Means denoising (`fastNlMeansDenoising`) with parameters optimized for manuscript clarity. This preprocessing approach follows established practices in computer-vision-assisted paleography [16].

[1] Modern standardized Dongba is Sinicized and formed of Chinese-like strokes. It was easier to find a complete modern Dongba dataset, but to properly contextualize our study, we also test a less comprehensive secondary Dongba dataset sourced from old-style Naxi manuscripts.

[2] It is an interesting note that the standardized Modern Yi, at first glance, shares more visual similarities to the Indus script in its angularity; stylized handwriting is slightly more stroke-like and rounded. Nonetheless, to maintain data integrity, we proceeded with Classical Yi in its older form.

**Augmentation techniques** Given the limited scope of our corpus, we employed extensive data augmentation to expand the effective dataset size and, ergo, increase variability and model robustness in accordance with published literature employing similar applications. Our `AugmentationPipeli` class applied multiple transformations while preserving essential glyph characteristics:

1. **Geometric transformations:** Random rotations (±45°) [41], scaling (0.85-1.15x) [90], translations (±10% of dimensions) [34], slight affine transformations [78], and elastic deformations that simulate material warping while maintaining glyph identity [90].
2. **Visual quality variations:** Brightness and contrast adjustments (0.7-1.5x range) [90], sharpness modifications (0.7-1.6x) [90], Gaussian blur (radius 0.5-1.2) [90], and multiple noise types (random, speckle) to simulate aging effects and digitization artifacts [34].
3. **Background modifications:** Custom texture additions and color jittering to mimic tablet surface variations and preservation conditions [90] [41].

The pipeline combined these transformations into strategic combinations (e.g., rotation+brightness+noise, elastic transform+contrast) to maximize diversity while ensuring the essential structural features of each script remained intact. For the three target scripts (Indus, Proto-Cuneiform, and Proto-Elamite), we generated four augmented variations per original image, expanding our effective dataset by 5×. Each augmentation was applied through an `AugmentedScriptDataset` class that generated variations on-the-fly during training, ensuring optimization of both storage and computational resources.

## 3.3 Model Architecture

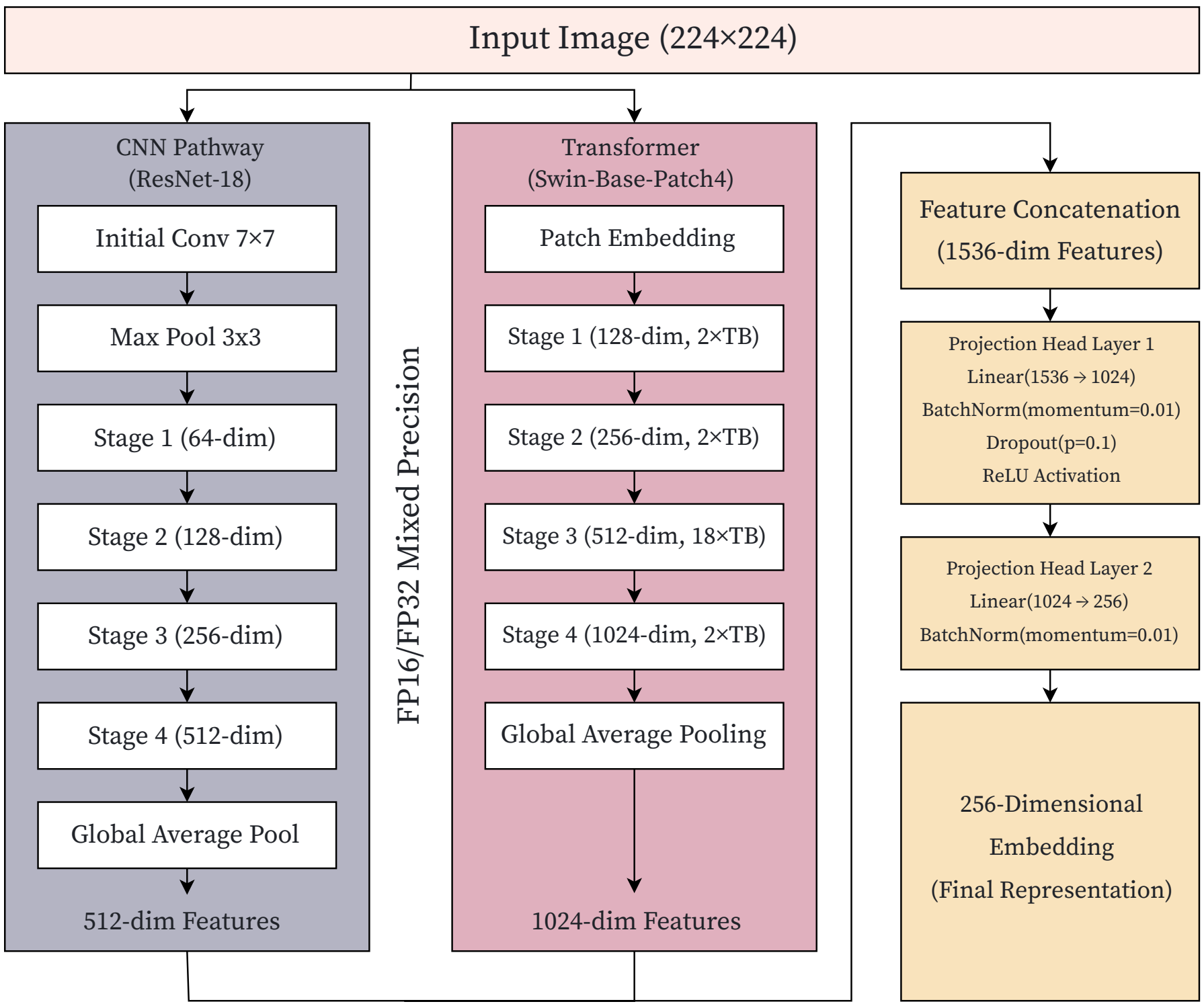

Figure 3.1: Overview of the ResNet-18-Swin Transformer hybrid model architecture.

After preliminary experiments with Swin Transformer-only architectures (following past successful applications in Dongba glyph retrieval [54]), we observed limitations in capturing fine-grained details across multiple signaries critical for ancient script comparison that resulted in feature collapse. This led us to develop our own novel hybrid CNN-Transformer architecture, optimized via self-supervised contrastive learning with custom weighting parameters, combining complementary strengths: ResNet-18 for local feature extraction and Swin Transformer [51] for capturing long-range dependencies.

### 3.3.1 CNN (ResNet-18) component

Neural networks are computational systems inspired by biological brains, consisting of interconnected artificial neurons that process input data through multiple layers to produce desired outputs. These systems learn patterns from data through iterative weight adjustments, enabling them to perform complex tasks without explicit programming. Convolutional Neural Networks (CNNs), in specific, are specialized neural network architectures designed to process grid-like data, particularly images, and map them to numerical embeddings. The fundamental operation in CNNs is the convolution, which applies the following equation filter upon the input to detect spatial patterns:

$$(f * g)(x, y) = \sum_{i=-a}^{a} \sum_{j=-b}^{b} f(i, j) \cdot g(x - i, y - j) \tag{3.1}$$

Where $f$ represents the input feature map, $g$ is the kernel (or filter), and $(x, y)$ denotes spatial coordinates. This enables CNNs to capture local patterns while still maintaining translation invariance, critical in image recognition.

The CNN pathway in our hybrid model employs a ResNet-18 backbone [32] to extract hierarchical visual features from input images. ResNet architectures revolutionized deep learning by introducing residual connections to help mitigate the vanishing gradient problem in deep networks. Beginning with a 7×7 convolutional layer (64 filters, stride 2) followed by max-pooling, the architecture continues with four sequential stages containing residual blocks with increasing filter dimensions (64, 128, 256, 512). Each stage comprises two residual blocks with dual 3×3 convolutional layers, batch normalization, and ReLU activation. The residual connections follow the formulation:

$$y = \mathscr{F}(x, \{W_i\}) + x \tag{3.2}$$

where $\mathscr{F}$ represents the residual mapping to be learned and $x$ is the identity input. This shortcut connection allows gradients to flow directly through the network and enables the training of deeper architectures. Each convolutional layer in a residual block can be expressed as:

$$z^{(l+1)} = \sigma\left(\text{BN}\left(W^{(l)} * z^{(l)} + b^{(l)}\right)\right) \tag{3.3}$$

where $*$ denotes the convolution operation, $\sigma$ is the ReLU activation function defined as $\sigma(x) = \max(0, x)$, and BN represents batch normalization. The CNN pipeline concludes with global average pooling, producing a 512-dimensional feature vector representing the input glyph.

### 3.3.2 Transformer (Swin) Component

Transformers are another neural network architecture that revolutionized sequence modeling by simplifying transduction down to only attention, rather than recurrence or convolution [93]. Following the publication of this 2017 landmark paper, this methodology enabled models to directly relate different positions in a sequence. The following attention function is the core mechanism, which computes weighted sums of input features:

$$\text{Attention}(Q, K, V) = \text{SoftMax}\left(\frac{QK^T}{\sqrt{d_k}}\right)V \tag{3.4}$$

Where $Q$ (query), $K$ (key), and $V$ (value) are learned linear transformations of the input, and $d_k$ is the dimensionality of the key vectors, serving as a scaling factor to prevent excessive magnitudes in the dot product.

Operating in parallel to our CNN, the Swin Transformer backbone [51] extends this concept through hierarchical vision processing with localized shifting self-attention windows.

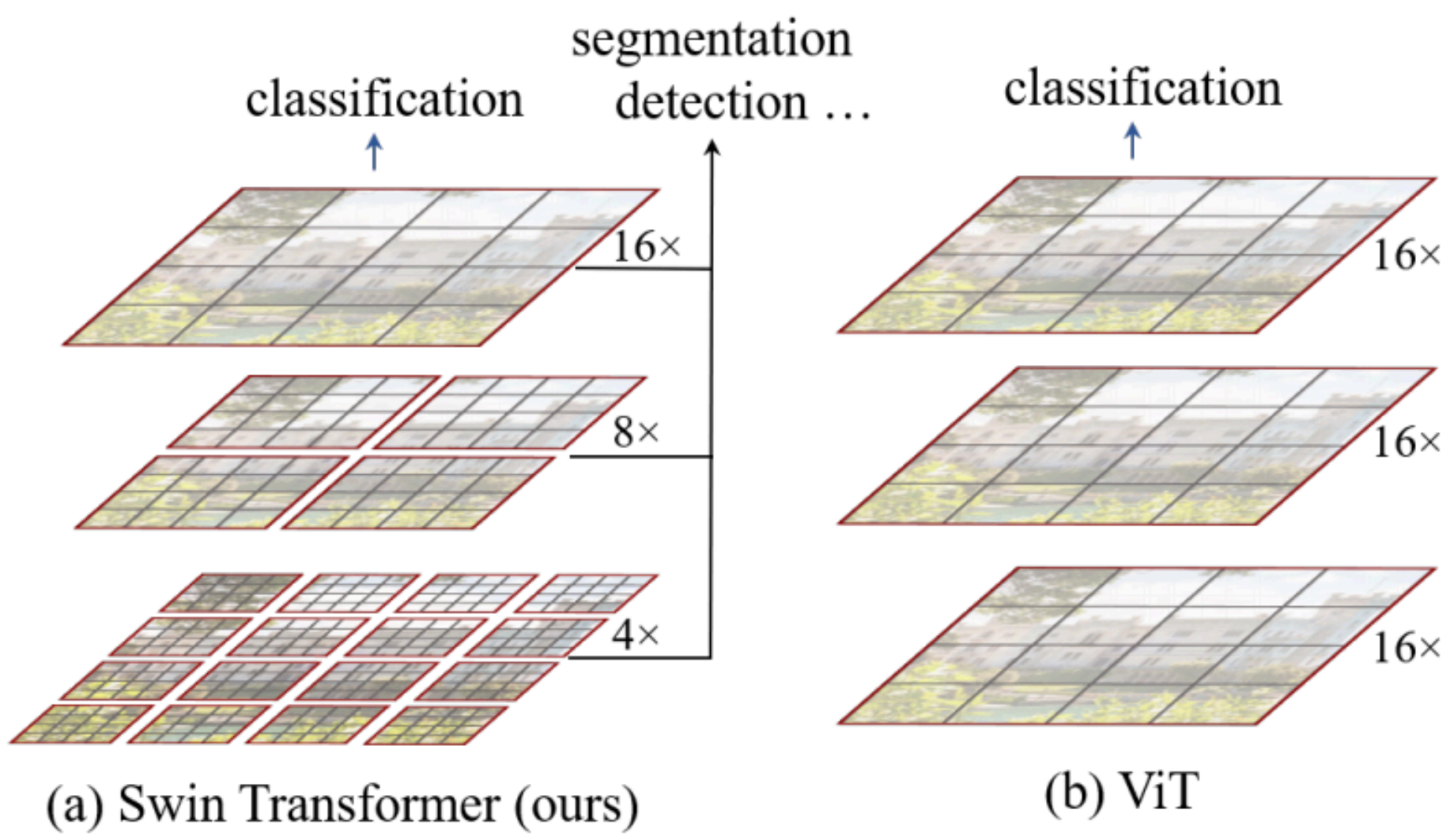

Figure 3.2: Swin Transformer creates hierarchical features by merging patches, enabling both classification and dense prediction tasks while achieving linear complexity $O(N)$ through windowed self-attention, unlike traditional Transformers' quadratic $O(N^2)$ global attention [51].

Unlike the original Transformer which computes attention globally, Swin Transformer computes attention within local windows, making it more efficient for visual data. The multi-head self-attention mechanism in each Swin Transformer block is formulated as:

$$\text{MHSA}(Q, K, V) = \text{Concat}(\text{head}_1, \text{head}_2, \dots, \text{head}_h)W^O \tag{3.5}$$

where each attention head is computed as:

$$\text{head}_i = \text{Attention}(QW_i^Q, KW_i^K, VW_i^V) \tag{3.6}$$

$$\text{Attention}(Q, K, V) = \text{SoftMax}\left(\frac{QK^T}{\sqrt{d_k}} + B\right)V \tag{3.7}$$

$Q$, $K$, and $V$ are query, key and value matrices, $d_k$ is the feature dimension, and $B$ is the relative position bias. The position bias term is parameterized as:

$$B_{i,j} = B'_{\hat{p}_x, \hat{p}_y} \tag{3.8}$$

where $\hat{p}_x = p_x + M - 1$ and $\hat{p}_y = p_y + M - 1$ are the normalized relative coordinates, and $B'$ is a learnable bias matrix.

The Swin-Base architecture comprises four stages with increasing dimensionality: Stage 1 (128-dimension, 2 blocks, 4 attention heads), Stage 2 (256-dimension, 2 blocks, 8 heads), Stage 3 (512-dimension, 18 blocks, 16 heads), and Stage 4 (1024-dimension, 2 blocks, 32 heads). Between stages, patch merging layers reduce spatial resolution while increasing feature dimensions. The model employs shifted window self-attention ($7 \times 7$ window size), restricting attention within non-overlapping windows in one layer and shifting window configuration in the next, enabling efficient computation with global connectivity. This pathway outputs a 1024-dimensional feature representation complementary to the CNN features.

### 3.3.3 Feature Fusion

After the CNN and Transformer components process an image in parallel, we obtain two feature vectors: a 512-d CNN-based feature and a 1024-d Transformer-based feature. These two representations are concatenated to form a combined 1536-dimensional embedding, which is then processed through a projection head:

$$h = \text{Dropout}_{0.1}(\text{ReLU}(\text{BN}(W_1 z + b_1))) \tag{3.9}$$

$$g = \text{BN}(W_2 h + b_2) \tag{3.10}$$

Where $z$ is the concatenated feature vector, $W_1 \in \mathbb{R}^{1024\times1536}$, $W_2 \in \mathbb{R}^{256\times1024}$, $b_1 \in \mathbb{R}^{1024}$, $b_2 \in \mathbb{R}^{256}$ are learnable parameters, BN represents batch normalization with a momentum of 0.01, and $\text{Dropout}_{0.1}$ applies dropout regularization with probability 0.1.

We refer to ablation studies conducted by Guo et al. [31] in assessing the value of a hybrid model rather than one or the other; frankly, it seems counterintuitive to reintroduce convolution into a technology novel for doing away with it. Yet, by leveraging both types of features, the model can recognize subtle shape details and broader structural relationships, which is critical for comparing scripts that might share high-level stylistic traits despite different low-level details. This hybrid approach provides advantages over either independent architecture by capturing both detailed stroke patterns (CNN) and global structural relationships (Transformer) critical for ancient script comparison [1, 5, 31].

## 3.4 Ensemble Modeling

To mitigate false positives and boost the reliability of our findings, we implemented ensemble modeling based on principles from Lakshminarayanan et al. [45], while adapting their approach to our specific context of ancient script analysis. Ensemble variance reduction—the statistical principle of combining multiple models to decrease prediction variability—forms the theoretical foundation of our methodology. Our specific implementation involved training five independent instances of our hybrid CNN-Transformer model for each target script analysis (Indus, Proto-Cuneiform, Proto-Elamite), totaling 15 models organized as three ensembles (one per target script). Each model in an ensemble was initialized with different random weights and trained from scratch using identical training protocols and data, a setup we designed specifically for our research questions about ancient script relationships between systems with established visual closeness[17].

Table 3.1: Ensemble Training Data Summary

| model_idx | seed | val_loss | epoch | model_path |
|---|---|---|---|---|
| 1 | 42 | 8.801242896488730 | 11 | Proto-Cuneiform_ensemble |
| 2 | 43 | 8.71736410685948 | 11 | Proto-Cuneiform_ensemble |
| 3 | 44 | 8.819251537323000 | 7 | Proto-Cuneiform_ensemble |
| 4 | 45 | 8.773778438568120 | 9 | Proto-Cuneiform_ensemble |
| 5 | 46 | 9.27264860698155 | 1 | Proto-Cuneiform_ensemble |
| 1 | 42 | 9.65621223449707 | 3 | Indus_ensemble |
| 2 | 43 | 9.49264907836914 | 1 | Indus_ensemble |
| 3 | 44 | 9.3032808303833 | 11 | Indus_ensemble |
| 4 | 45 | 9.921699905395510 | 1 | Indus_ensemble |
| 5 | 46 | 10.057604217529300 | 1 | Indus_ensemble |
| 1 | 42 | 9.465515518188480 | 13 | Proto-Elamite_ensemble |
| 2 | 43 | 10.229036331176800 | 5 | Proto-Elamite_ensemble |
| 3 | 44 | 9.9594895362854 | 1 | Proto-Elamite_ensemble |
| 4 | 45 | 9.304905700683590 | 12 | Proto-Elamite_ensemble |
| 5 | 46 | 9.408125591278080 | 11 | Proto-Elamite_ensemble |

All models were trained for up to 20 epochs with an early stopping criterion to prevent overfitting. Early stopping monitored the model's performance on a validation split, and training was stopped if no improvement was seen for five epochs. Because each ensemble member had a random initialization and mini-batch order, they each learned somewhat different feature nuances, even though they saw the same overall training data.

During inference, embeddings from all five models in an ensemble were averaged to produce a consensus representation for each glyph. The diversity among ensemble members allows the ensemble to act like a decision-making committee, offering individual perspectives and mitigating biases, leading to a more robust overall representation [45]. We found that ensemble embeddings had lower variance and improved generalization in downstream evaluations compared to any single model, consistent with ensemble theory. Additionally, using multiple models reduces the chance of missing a true pattern: if one model converges to a poor local optimum (where meaningful similarities are not well captured), the others likely will not, and the averaging will down-weight such failures. This was crucial for reliably detecting subtle script relationships.

## 3.5 Training Objective and Implementation

We formulated the training objective as a self-supervised contrastive learning problem to learn an embedding space where visually similar glyphs are close together. This was accomplished via an enhanced version of the NT-Xent (Normalized Temperature-Scaled Cross-Entropy) contrastive loss [12] with a temperature parameter of 0.1:

$$\mathscr{L} = -\log \frac{\exp(\mathrm{sim}(z_i, z_j)/\tau)}{\sum_{k=1}^{2N} \mathbb{1}_{[k\neq i]} \exp(\mathrm{sim}(z_i, z_k)/\tau)} + \mathscr{L}_{\mathrm{var}} + 0.1 \cdot \mathscr{L}_{\mathrm{unif}} \tag{3.11}$$

Where $\mathrm{sim}(z_i, z_j) = \frac{z_i^T z_j}{\|z_i\| \cdot \|z_j\|}$ is the cosine similarity between normalized embeddings, $\tau$ is the temperature parameter, $\mathbb{1}_{[k\neq i]}$ is an indicator function evaluating to 1 when $k \neq i$, and $N$ is the batch size.

Beyond the standard contrastive term, we incorporated two regularization terms: a variance term that prevents feature collapse and a uniformity term that promotes even distribution of embeddings on the hypersphere:

$$\mathcal{L}_{\text{var}} = \frac{\lambda_{\text{reg}}}{\text{Var}[\mathbf{z}] + \epsilon} \tag{3.12}$$

$$\mathcal{L}_{\text{unif}} = \log \mathbb{E}_{x,y \sim p_{\text{data}}} \left[ e^{-2\|f(x)-f(y)\|^2} \right] \tag{3.13}$$

where $\text{Var}[\mathbf{z}]$ represents the feature variance calculated as the mean squared distance of representations from their mean, $\lambda_{\text{reg}}$ is a regularization coefficient, $\epsilon$ is a small constant ($10^{-4}$) to prevent division by zero, and $f(x)$ and $f(y)$ are normalized embeddings of randomly sampled data points.

In each training batch, augmented versions of the same glyph served as positive pairs, and other glyphs from the same script within the batch functioned as negative pairs, allowing the model to learn meaningful representations of each script's internal visual structure. We used the AdamW optimizer with decoupled weight decay and a learning rate of $3 \times 10^{-5}$ with a cosine decay schedule that included a 10% warmup period. This relatively conservative learning rate was chosen after careful experimentation, as higher values could cause training instability when jointly optimizing the CNN and transformer components.

### 3.5.1 Computational Implementation

All 15 of our models were trained on an NVIDIA A100 GPU accelerator (40 GB VRAM) accessed through Google Colab's premium subscription services and implemented in Python using PyTorch (Appendix A.2). This process presented several engineering challenges due to the complexity of our hybrid architecture; the Swin Transformer component in particular was highly demanding, as it consists of approximately 327 distinct layers/parameters and tens of millions of parameters total—making training both time-consuming and memory-intensive.

To address these computational constraints, we employed several optimization techniques:

1. **Mixed precision training:** To optimize computational efficiency, we implemented PyTorch's `amp` (Automatic Mixed Precision) API (Appendix A.2). This technique is incorporated in our training loop through `torch.amp.autocast()` and `GradScaler` in our `SCRIBE` class, and can be mathematically modeled with:

$$\text{scale}_{t+1} = \begin{cases} \text{scale}_t \times \alpha_{\text{inc}}, & \text{if no overflow for } n_{\text{stable}} \text{ consecutive updates} \\ \text{scale}_t \times \alpha_{\text{dec}}, & \text{if overflow detected} \end{cases} \tag{3.14}$$

   where typical values are $\alpha_{\text{inc}} = 2$, $\alpha_{\text{dec}} = 0.5$, and $n_{\text{stable}} = 2000$.

2. **Training stability techniques:** To ensure reliable model convergence, we implemented:

   - Gradient clipping with a maximum norm of 1.0 using `torch.nn.utils.clip_grad_norm_`, which prevents "exploding gradients"—a common issue in deep networks where accumulated gradients become excessively large, causing training instability. Our implementation explicitly applies this to both model and projection head parameters.

   - Cosine learning rate schedule with 10% linear warmup period, implemented via our custom `warmup_cosine_schedule` function:

$$\eta_t = \eta_{\min} + \frac{1}{2}(\eta_{\max} - \eta_{\min})(1 + \cos(\pi \cdot \frac{t - t_{\text{warmup}}}{t_{\max} - t_{\text{warmup}}})) \tag{3.15}$$

for $t > t_{warmup}$ and linear ramp-up for $t \leq t_{warmup}$. This schedule adjusts the learning rate throughout training, helping our model avoid local optima.

- **Adaptive weight decay** ($1 \times 10^{-3}$) applied through our `AdamW` optimizer configuration, which helps prevent overfitting on our limited archaeological dataset by penalizing large weights.

These implementation strategies collectively enabled efficient training of our hybrid architecture, with full ensemble training completed in approximately 4 hours. The combination of architectural innovations and computational optimizations allowed us to effectively learn meaningful script representations despite the challenging nature of the dataset and the multi-component architecture of our model.

## 3.6 Evaluation and Validation

To verify that the model learned meaningful representations and to validate our hypothesis about script similarities, comprehensive evaluation ensued; this included qualitative visualization of model attention, quantitative clustering metrics, embedding similarity analysis, dimensionality reduction for visualization, and rigorous statistical testing.

We utilized Gradient-weighted Class Activation Mapping (Grad-CAM) to visualize which image regions our model found important when analyzing ancient scripts. While the original Grad-CAM technique [85] was designed for classification tasks, our custom implementation adapts it for representation learning by tracking which image parts contribute most to the model's overall understanding. Our code uniquely handles both CNN and Transformer pathways—something not present in the original approach—allowing us to compare how different network architectures process the same glyph. This approach backpropagates, or traces backward, through our model to identify influential regions in the original image, creating visual heatmaps of attention.

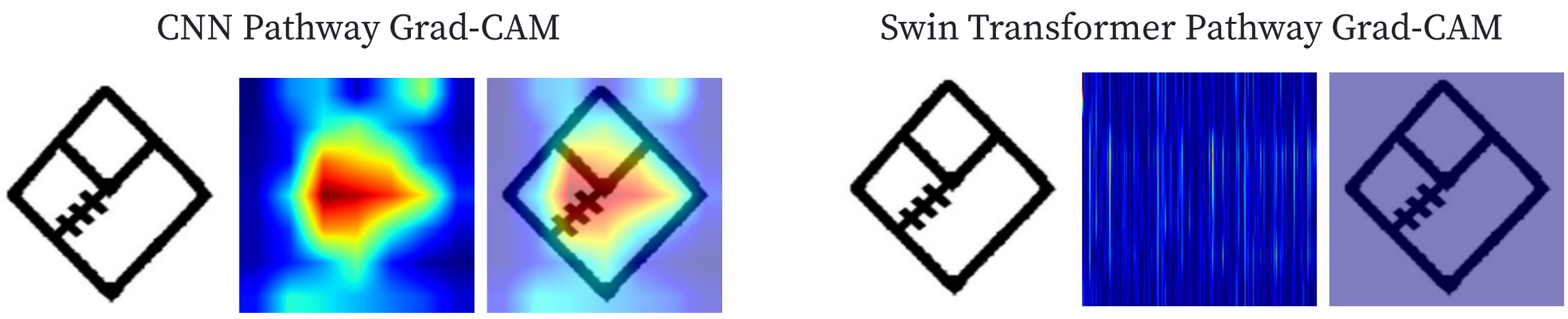

Figure 3.3: Dual-model Grad-CAM visualization. *Left:* CNN pathway highlights local details, focusing on intersection. *Right:* Swin Transformer attention forms grid-like patterns with more globally distributed focus.

Heatmaps of "important" regions in an input image are generated by using the gradients flowing into the last convolutional layers or transformer blocks. We observed distinct and complementary patterns in our dual-pathway approach (Figure 3.3). The CNN Grad-CAM visualizations highlighted specific local features of glyphs, such as curved segments, junctions, or distinct patterns (as seen in 3.3). In contrast, the Transformer Grad-CAM visualizations revealed a more holistic pattern recognition approach, highlighting broader regions that defined overall structure and proportions. Due to its shifted window attention mechanism, the map, when projected back into image space, reads as stratified grid lines.

Key glyph components were demarcated as high-importance regions in the CNN path-way, indicating these features drove the model's embedding. We performed Grad-CAM on examples from each script in both the target and comparison sets, consistently finding that models focused on meaningful glyph features rather than artifacts (Appendix A.8). The complementary nature of our CNN and Transformer attention mechanisms proved particularly valuable for comparing scripts with different levels of visual complexity, allowing our model to identify similar structural patterns despite differing complexity levels across script traditions.

### 3.6.1 Similarity Analysis

Using learned embeddings, we conducted a pairwise similarity analysis to test our hypotheses directly. For each pair of script sets (for example, Indus vs. TYC), we computed the cosine similarity between all embeddings of images from the first script and all embeddings of images from the second script. Mathematically, for scripts $S_1$ and $S_2$ with embedding sets, the mean similarity was calculated as:

$$\text{sim}(S_1, S_2) = \frac{1}{|S_1||S_2|} \sum_{x \in S_1} \sum_{y \in S_2} \frac{z(x)^T z(y)}{\|z(x)\| \cdot \|z(y)\|} \tag{3.16}$$

where $z(x)$ and $z(y)$ represent the $d$-dimensional learned embedding vectors of script images $x$ and $y$, respectively, extracted from the trained model, and $z(x)^T z(y)$ denotes their dot product, which computes the numerator of the cosine similarity between the two embeddings. This produced a distribution of similarity scores for each script pair. The cosine similarity values (ranging from -1 to 1) indicate how closely related the feature representations of two scripts are—higher values mean the model sees those script images as more similar. After compiling these similarity computations across the ensemble, each of the five models produced its own set of similarities, and so we took the mean similarity and standard deviation across the ensemble. This allowed us to quantitatively evaluate our hypothesis: if Indus is visually closer to TYC scripts, we expect the Indus–TYC similarity scores to be higher on average than Indus–Proto-Cuneiform or Indus–Proto-Elamite scores, and this should hold consistently across ensemble members (which it did, as our results (chapter 4) will show). The consistency of results across the five models (all showing the same ordering of similarities) was noted as an important validation that no single model was aberrant.

#### Cross-Script Embedding Comparison

In addition to the glyph-level similarity analysis described above, we performed a direct script-to-script embedding comparison to evaluate the higher-level relationships between script systems. While the first similarity metric (around 0.63) measured the average similarity between individual glyphs across different scripts, this second metric directly compared the aggregate representations of each script.

For script-to-script comparison, we first compute comprehensive representations of each script from our ensemble of trained models, then measure the cosine similarity between these script-level embeddings:

$$\text{ScriptSim}(S_1, S_2) = \frac{\bar{z}(S_1)^T \bar{z}(S_2)}{\|\bar{z}(S_1)\| \cdot \|\bar{z}(S_2)\|} \tag{3.17}$$

where $\bar{z}(S)$ represents the aggregate embedding vector for script $S$. This approach revealed stronger connections between scripts (with values typically in the 0.85-0.93 range) than observed in the glyph-level comparison (0.61-0.68 range), highlighting deeper structural relationships between writing systems that transcend individual character variations.

### 3.6.2 Clustering

Beyond internal metrics, we applied unsupervised clustering algorithms on the embeddings to see if the model's features would naturally group the scripts correctly, and to explore the structure of the embedding space. We experimented with bottom-up hierarchical clustering for each model individually, producing dendrograms of script relationships using various linkage methods (single, complete, average, and Ward). By comparing results across different linkage criteria and across models, we ensured the observed hierarchical relationships were consistent and not artifacts of a particular model or linkage method, following best practices established by Saxena et al. [84].

### 3.6.3 Dimensionality Reduction

We also implemented dimensionality reduction to visualize our embeddings in 2D/3D space.

- **t-SNE (t-Distributed Stochastic Neighbor Embedding):** We applied t-SNE [55] to produce a 2D plot; this technique minimizes the Kullback-Leibler divergence between probability distributions in high and low dimensional spaces:

$$\mathrm{KL}(P\|Q) = \sum_i \sum_j p_{ij} \log \frac{p_{ij}}{q_{ij}} \tag{3.18}$$

  While the t-SNE plots by individual model were more abstract and showed more spread by each language, the ensemble plots revealed clear clusters of points corresponding to each script (Appendix A.3), which provided a visual confirmation of the separation (as well as a validation of the ensemble technique).

- **PCA (Principal Component Analysis):** We used PCA as a linear projection method. By examining the first two principal components of the embeddings, we checked if script groupings were evident in a linear subspace. PCA explained a reasonable portion of variance and again showed clear patterns between script clusters (though not as cleanly as t-SNE, given PCA is linear).

These clustering and visualization methods served as an important validation step, demonstrating that the model's embeddings could recover script categories. The fact that the spatial arrangement of clusters in visualization seemed to reflect known or hypothesized relationships provides confidence in the model's effectiveness; such tendencies were indeed observed in the embedding space plots (detailed in chapter 4).

### 3.6.4 Statistical Analysis

Finally, we implemented a rigorous statistical framework to test our primary hypothesis: *that the Tibetan Yi-Corridor scripts (Naxi, Ba-Shu, Yi) are more visually similar to the Indus Valley script than the other ancient scripts (Proto-Cuneiform, Proto-Elamite) are*. Rather than relying on qualitative observations, we quantified this and applied hypothesis tests. We focused on comparing the cosine similarity distributions between script pairs of interest. The key comparisons were:

- **TYC to Indus vs. TYC to Proto-Cuneiform/Proto-Elamite:** We tested whether TYC-Indus similarities are significantly higher than Indus–Proto-Cuneiform/Proto-Elamite similarities.

- **Indus to TYC vs. Indus to Proto-Cuneiform/Proto-Elamite:** Similarly, we tested whether Indus–TYC similarity exceeds Indus–Proto-Elamite/Proto-Cuneiform similarity significantly.

For each comparison, we conducted Welch's t-test [95] to determine whether the observed differences in similarity distributions were significant. For similarity distributions from scripts $S_1$, $S_2$, and $S_3$ (e.g., Indus, TYC, and Proto-Cuneiform), we tested whether the mean similarity between $S_1$ and $S_2$ significantly differs from the mean similarity between $S_1$ and $S_3$. The t-statistic for this comparison is calculated as:

$$t = \frac{\overline{\text{sim}(S_1, S_2)} - \overline{\text{sim}(S_1, S_3)}}{\sqrt{\frac{s_1^2}{n_1} + \frac{s_2^2}{n_2}}} \tag{3.19}$$

where $\overline{\text{sim}(S_1, S_2)}$ and $\overline{\text{sim}(S_1, S_3)}$ are the mean similarities, $s_1^2$ and $s_2^2$ are the variances of the similarity distributions, and $n_1$ and $n_2$ are the sample sizes. For pairs with large numbers of similarity comparisons (>1000), we employed a strategic subsampling approach, selecting approximately $\sqrt{1000}$ samples from each distribution to maintain computational efficiency while preserving statistical validity.

By aggregating at the model level, we were able to simplify our analysis: for each of the five ensemble models, we computed the mean similarity for each pair (Indus–TYC, Indus–Cuneiform, Indus–Elamite). This gave five values for each type of comparison. We then performed a paired t-test on those values (paired by model) to see if, e.g., Indus–TYC mean similarity is consistently higher than Indus–Cuneiform across models. For this model-level paired t-test, the test statistic is:

$$t = \frac{\bar{d}}{s_d/\sqrt{N}} \tag{3.20}$$

where $\bar{d}$ is the mean difference between the similarity scores across models, $s_d$ is the standard deviation of these differences, and $N$ is the number of models (5 in our case).

Despite consistently significant t-test results ($p < 0.001$), we encountered a common challenge in large-sample comparisons: p-values approached zero across all tests, potentially overstating the importance of subtle differences [92]. To address this limitation and provide meaningful interpretation within our compressed high-similarity space (>0.85 for related scripts), we incorporated Cohen's d effect size analysis:

$$d = \frac{\overline{\text{sim}(S_1, S_2)} - \overline{\text{sim}(S_1, S_3)}}{\sqrt{\frac{(n_1-1)s_1^2+(n_2-1)s_2^2}{n_1+n_2-2}}} \tag{3.21}$$

This standardized measure quantifies the practical significance of differences, distinguishing between statistically significant results that represent negligible effects ($d < 0.2$) versus those indicating substantial relationships ($d \geq 0.8$) [15].

Additionally, a Bonferroni correction was applied to our significance threshold [7, 8]. This means we required each test to be significant at $\alpha/n$ (where n is the total number of tests performed):

$$\alpha_{\text{corrected}} = \frac{\alpha}{n} \quad (3.22)$$

This maintains an overall family-wise error rate of $\alpha=0.05$, controlling for multiple comparisons. This is a conservative approach to ensure that any finding of greater similarity is not a false positive.

The results of these statistical tests (reported in chapter 4) did indicate a significant difference: the Indus–TYC similarity was higher, with p-values well below the adjusted threshold, supporting our hypothesis. We also confirmed consistency across the ensemble models as mentioned: all 5 models showed the same trend of Indus being closest to TYC with no outlier model contradicting this. This was further quantified by making sure that removing any single model from the ensemble did not change the overall conclusion and the result was not driven by one peculiar model—otherwise known as leave-out testing—following ensemble stability evaluation practices suggested by [21]. This consistency analysis adds robustness: it's unlikely that the observed similarity is due to chance or overfitting in one model, given all independent models converged on a similar solution.

## 3.7 Limitations and Controls

We acknowledge several limitations in our methodology and data, and implemented specific controls to mitigate their impact:

- **Limited sample size:** The scarcity of data (Indus, Old Naxi, Proto-Elamite) limits generalization and risks overfitting. We mitigated this with data augmentation, ensemble learning, early stopping, and contrastive learning with variance/uniformity regularization. Cross-validation ensured robustness across random splits.
- **Digitization bias:** Image sources varied—from standardized corpora to manuscript photos—posing risk of bias. Standardized preprocessing (grayscale, contrast normalization, glyph isolation) and Grad-CAM visualizations confirmed focus on glyph structure, not artifacts. Some bias may remain, and so we interpret results cautiously.
- **Script complexity differences:** Our CNN-Transformer model extracts local and global features, reducing bias from stroke complexity. Grad-CAM showed attention to subcomponents, and clustering patterns were not driven by stroke count alone.
- **Generalization safeguards:** We used a 70/20/10 split, early stopping, dropout (0.1), weight decay (1e-3), and gradient clipping. Warmup-cosine learning rates and validation-based model selection ensured stability and generalization despite data limitations.

In summary, our approach combines deep learning with ensemble techniques and multifaceted validation to ensure discovered patterns are genuine and interpretable, establishing a solid foundation for investigating historical script connections through visual analysis.

Chapter 4

# RESULTS

This section presents our comparative analysis of TYC scripts (newer Dongba, older Dongba, and ancient TYC) against ancient writing systems (Indus, Proto-Cuneiform, and Proto-Elamite) using feature representations generated by our hybrid computer vision model[1].

## 4.1 Cross-Script Similarity Metrics

Table 4.1: Mean Cosine Similarity Between Comparison and Target Scripts

| Comparison Script | Indus | Proto-Cuneiform | Proto-Elamite |
|---|---|---|---|
| TYC | 0.635 | 0.102 | 0.078 |
| Naxi-Dongba | 0.634 | 0.109 | 0.087 |
| Old-Naxi | 0.617 | 0.106 | 0.076 |

Table 4.1 presents the mean cosine similarity scores across our ensemble of five independently trained models. These results reveal a striking pattern: all three comparison scripts show substantially higher similarity to the Indus script (60.7%-67.8%) than to either Proto-Cuneiform (10.2%-10.9%) or Proto-Elamite (7.6%-8.7%). The magnitude of this difference—approximately six-fold higher similarity to the Indus script—suggests a specific relationship that supports our initial hypothesis of a potential historical connection between these writing systems. Notably, the Indus script maps closer to TYC scripts than to Proto-Elamite or Proto-Cuneiform, despite the latter's proximal positions in time, space, and documented interactions (Figure 4.1). As expected, West Asian scripts identify each other as their closest match.

[1]For clarity in our data, Naxi-Dongba refers to the newer Dongba dataset, and Old-Naxi refers to the older. TYC refers to the combined dataset of Ba-Shu pictographs, Classical Yi, and the aforementioned older Dongba dataset in equal parts.

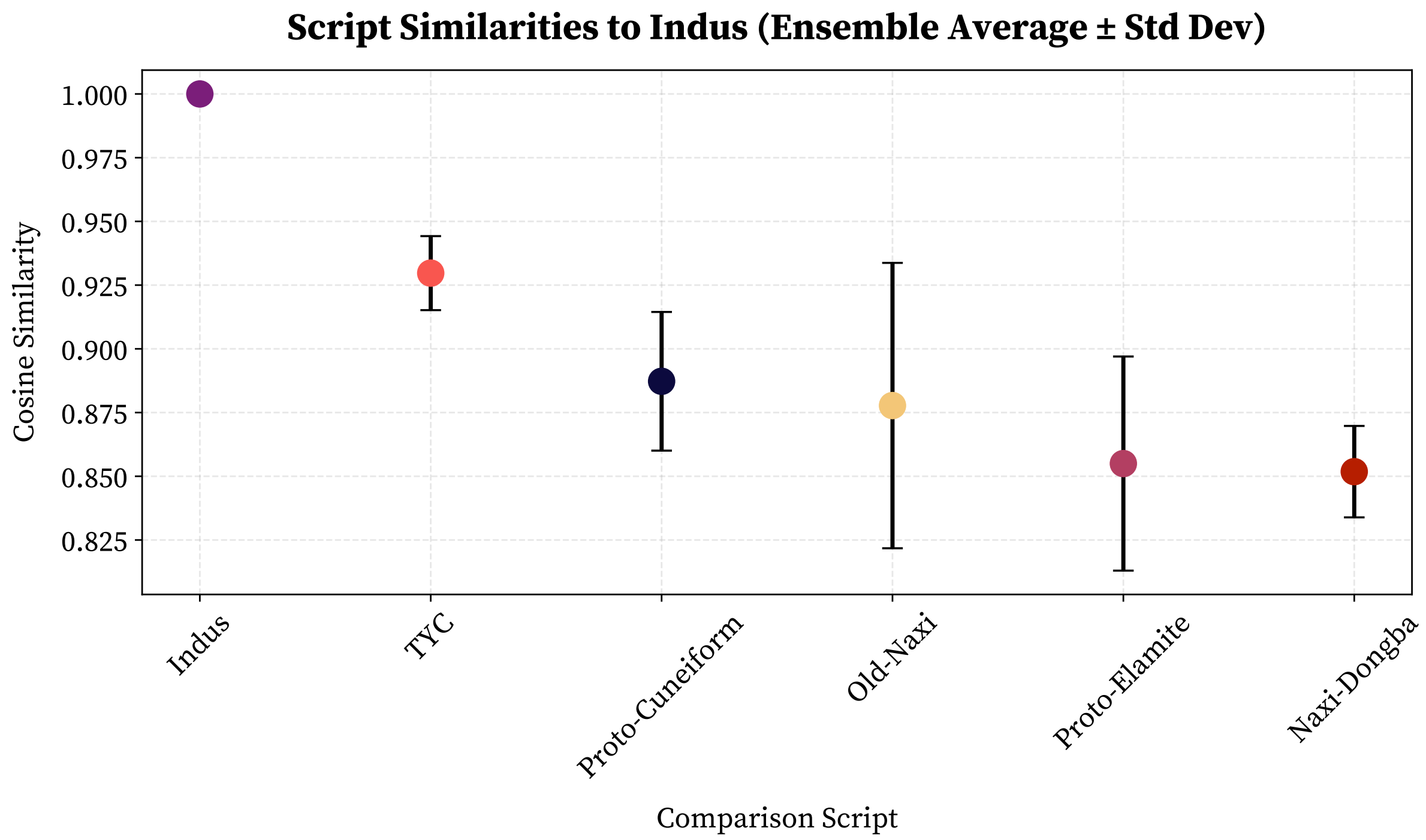

Figure 4.1: Indus Ensemble Cosine Similarity Plot

### 4.1.1 Direct Script-to-Script Embedding Comparison

As described in our methodology, we also performed a direct script-to-script embedding comparison to evaluate higher-level relationships between writing systems. Table 4.2 presents the results of this comparison with Indus as the anchor script, which reveal even stronger connections between scripts than observed in the glyph-level analysis.

Table 4.2: Direct Script-to-Script Embedding Similarity

| Source Script | Target Script | Mean Similarity | 95% CI |
|---|---|---|---|
| Indus | TYC | 0.930 | [0.917, 0.942] |
| Indus | Proto-Cuneiform | 0.887 | [0.863, 0.911] |
| Indus | Proto-Elamite | 0.855 | [0.818, 0.892] |
| Indus | Old-Naxi | 0.878 | [0.829, 0.927] |
| Indus | Naxi-Dongba | 0.852 | [0.836, 0.868] |

These direct comparisons reinforce our glyph-level findings but with notably higher overall similarity values, as they capture higher-order structural relationships between writing systems. The Indus script shows the strongest connection to TYC scripts (0.930), significantly higher than to Proto-Cuneiform (0.887) or Proto-Elamite (0.855). This pattern is particularly striking given the Indus Valley Civilization's documented trade relations and geographic proximity to West Asian cultures. Interestingly, Old-Naxi (0.878) shows a stronger relationship to Indus than does the newer Naxi-Dongba script (0.852), suggesting that older TYC scripts may preserve features more similar to the Indus writing system.

The elevated similarity values in direct script comparisons (0.852-0.930) versus glyph-level comparisons (0.078-0.635) highlight how aggregate script representations capture deeper structural relationships that may not be immediately apparent when comparing individual characters. These findings provide additional and independent support for our hypothesis regarding the connection between Indus Valley and Tibetan-Yi Corridor writing systems.

### 4.1.2 Effect Size Analysis

While the statistical significance ($p < 0.001$) of our findings established that the observed similarity patterns were unlikely due to chance, the effect size analysis provided crucial context about the magnitude of these differences. Table 4.3 presents Cohen's d values for key script comparisons across our ensemble models.

Table 4.3: Mean Effect Size (Cohen's d) Between Script Comparisons

| Comparison Script | Indus vs. Proto-C | Indus vs. Proto-E | Proto-C vs. Proto-E |
|---|---|---|---|
| TYC | 7.13 (large) | 7.45 (large) | 0.46 (small) |
| Naxi-Dongba | 8.15 (large) | 8.36 (large) | 0.41 (small) |
| Old-Naxi | 7.16 (large) | 7.57 (large) | 0.61 (medium) |

The effect sizes for the comparisons between Indus and West Asian scripts were consistently enormous ($d > 7.13$), far exceeding Cohen's threshold for "large" effects ($d \geq 0.8$). The comparison between Proto-Cuneiform and Proto-Elamite showed small to medium effect sizes (0.41-0.61), aligning with the established historical understanding that these West Asian scripts shared certain visual conventions and potentially influenced each other. Understandably, this effect is dwarfed by the much stronger Indus-TYC relationship our analysis revealed.

## 4.2 Ensemble Model Consistency

Table 4.4 presents the complete similarity matrix demonstrating consistent patterns across all five models in our ensemble.

Table 4.4: Complete Similarity Matrix Across All Models

| Comparison Script | Model | Indus | Proto-Cuneiform | Proto-Elamite |
|---|---|---|---|---|
| TYC | model_0 | 0.607199 | 0.064351 | 0.098311 |
| TYC | model_1 | 0.613448 | 0.067255 | 0.065887 |
| TYC | model_2 | 0.626782 | 0.072332 | 0.081921 |
| TYC | model_3 | 0.650874 | 0.073373 | 0.070953 |
| TYC | model_4 | 0.678075 | 0.231436 | 0.074483 |
| Naxi-Dongba | model_0 | 0.577379 | 0.070971 | 0.082362 |
| Naxi-Dongba | model_1 | 0.655400 | 0.054296 | 0.098983 |
| Naxi-Dongba | model_2 | 0.578332 | 0.051874 | 0.093169 |
| Naxi-Dongba | model_3 | 0.684051 | 0.082636 | 0.077289 |
| Naxi-Dongba | model_4 | 0.675574 | 0.282917 | 0.083985 |
| Old-Naxi | model_0 | 0.620153 | 0.072526 | 0.101854 |
| Old-Naxi | model_1 | 0.570650 | 0.076422 | 0.052467 |
| Old-Naxi | model_2 | 0.653981 | 0.075445 | 0.086738 |
| Old-Naxi | model_3 | 0.596160 | 0.086085 | 0.071727 |
| Old-Naxi | model_4 | 0.645789 | 0.220979 | 0.065582 |

The model-by-model analysis confirms the robust pattern of Indus script dominance across independently trained models—a mean similarity of 0.635 compared to 0.102 for Proto-Cuneiform and 0.078 for Proto-Elamite. While model_4 shows elevated Proto-Cuneiform similarity scores (22.1%-28.3%), these remain significantly below the corresponding Indus similar-ities, likely reflecting model-specific feature capture rather than fundamental script relationships.

## 4.3 Script-Specific Patterns

Analysis of individual script comparisons reveals:

- **TYC Script:** Consistent relationship with Indus script (similarity 60.7%-67.8%)
- **Naxi-Dongba Script:** Highest absolute Indus similarity (68.4%) but greatest variance (57.7%-68.4%), potentially reflecting stylistic evolution
- **Old-Naxi Script: Old-Naxi Script:** Most consistent Indus relationship with lowest variance (57.1%-65.4%), possibly preserving archaic features aligning with Indus script

The consistency across these historically and geographically varied writing systems suggests genuine structural and visual affinities rather than coincidental similarities.

## 4.4 Dimensionality Reduction Analysis

Our t-SNE visualization reveals three distinct clusters: (1) Proto-Cuneiform and Proto-Elamite, (2) Indus, TYC, and older Dongba, and (3) newer Dongba script. This clustering pattern supports our theoretical framework by distinguishing West Asian scripts, ancient TYC-IVC scripts, and Sinicized TYC scripts. Additional t-SNE plots can be found in Appendix A.3.

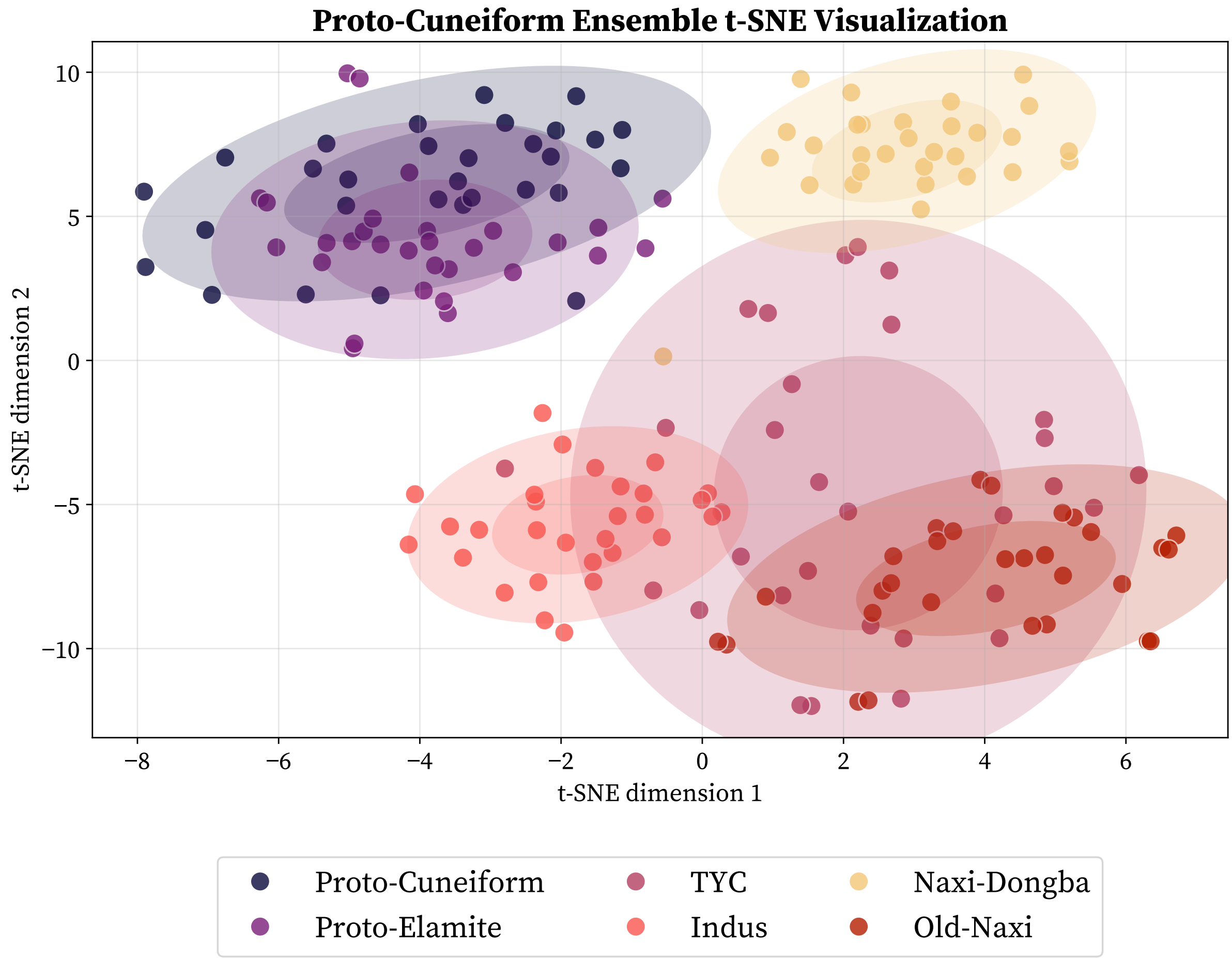

Figure 4.2: Proto-Cuneiform Ensemble Embeddings t-SNE Visualization

PCA mapping further corroborates these findings. In 3D plots, older TYC scripts (including older Dongba) cluster with Indus and West Asian languages, while newer Dongba appears as a consistent outlier due to Sinitic influences. This aligns with the findings of Daggumati and Revesz [16–18], who discovered through similar computer vision analysis that the Indus script visually clusters with Proto-Cuneiform and Proto-Elamite rather than other ancient writing systems, such as Brahmi or early European scripts. Additional PCA plots can be found in Appendix A.4.

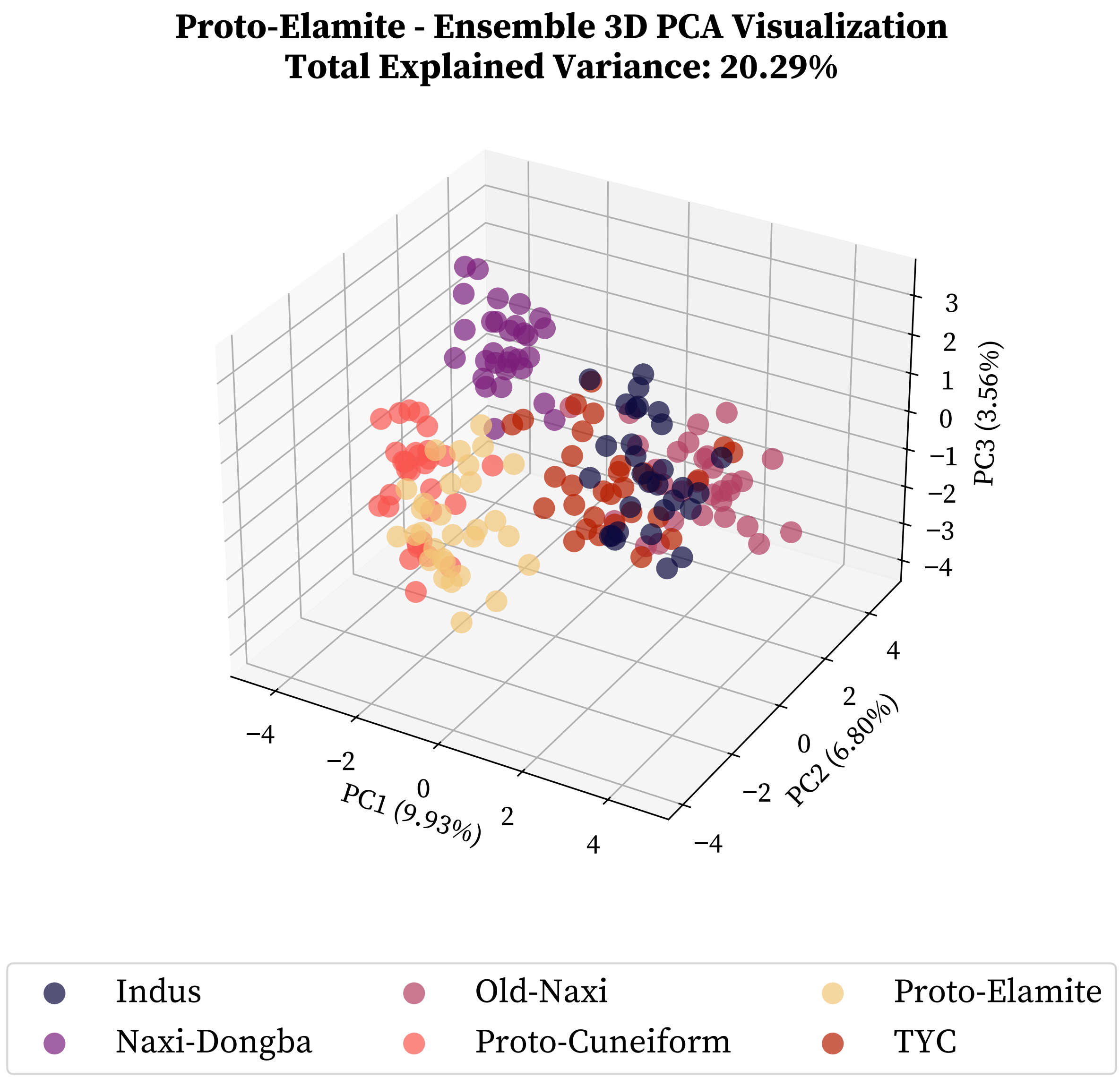

Figure 4.3: 3D Proto-Elamite Ensemble Embeddings PCA Visualization

## 4.5 Hierarchical Clustering

We applied hierarchical agglomerative clustering to script embeddings using cosine similarity as the distance metric. Multiple linkage methods (single, complete, average, and Ward's) were tested to assess the robustness of the structural relationships between scripts. All ensemble dendrograms can be found in Appendix A.5.

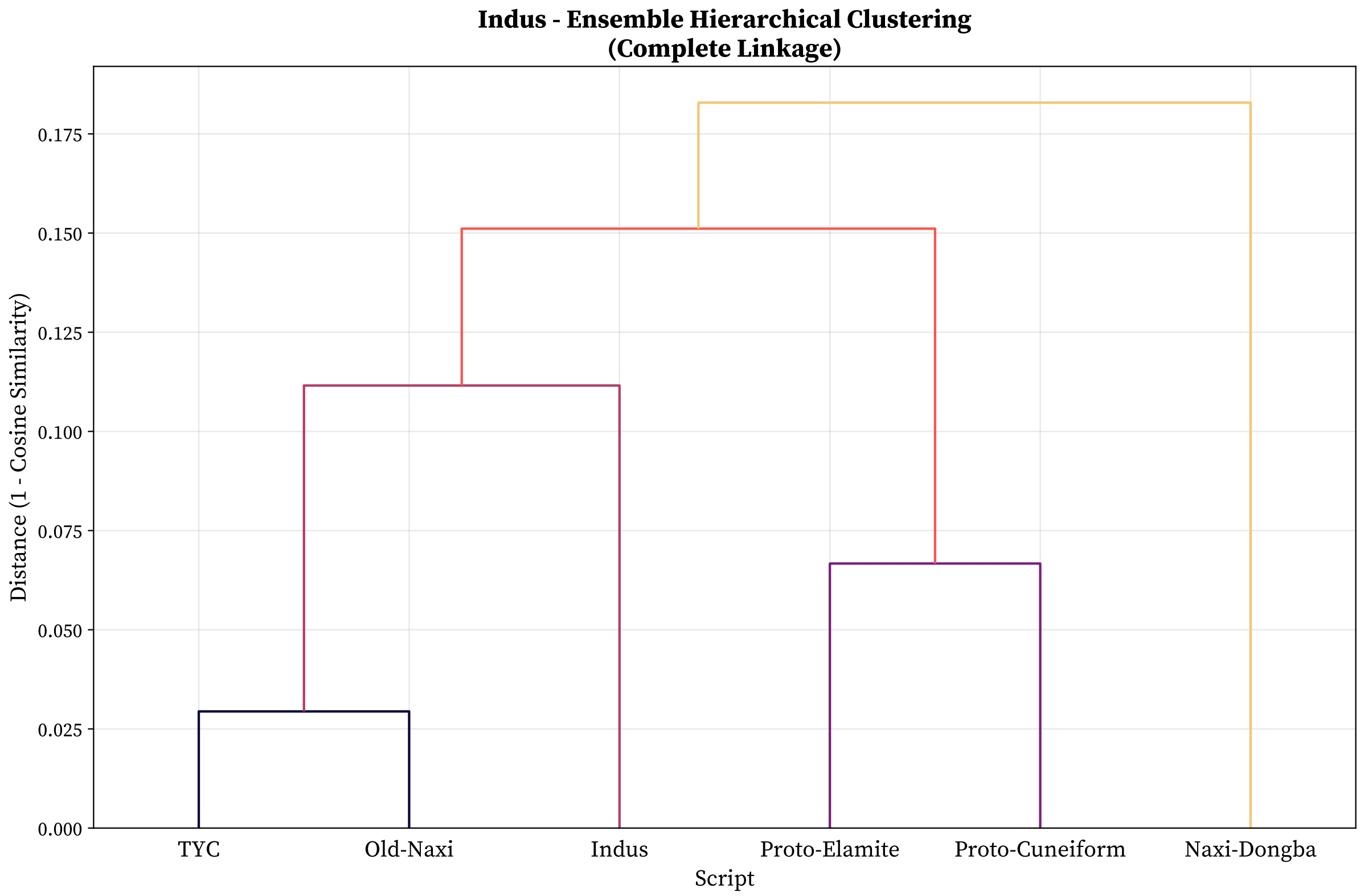

Figure 4.4: Indus Ensemble Embeddings Complete Linkage Dendrogram

The dendrograms across our ensemble models (Figure 4.4 and Appendix A.5) consistently reveal two primary clusters: (1) Indus, old Dongba, and TYC forming one tight group, and (2) Proto-Cuneiform and Proto-Elamite forming another. Interestingly, while newer Dongba typically appears as an outlier in previous analyses, hierarchical clustering shows it tends to merge with the West Asian scripts at higher levels rather than with the Indus/TYC cluster.

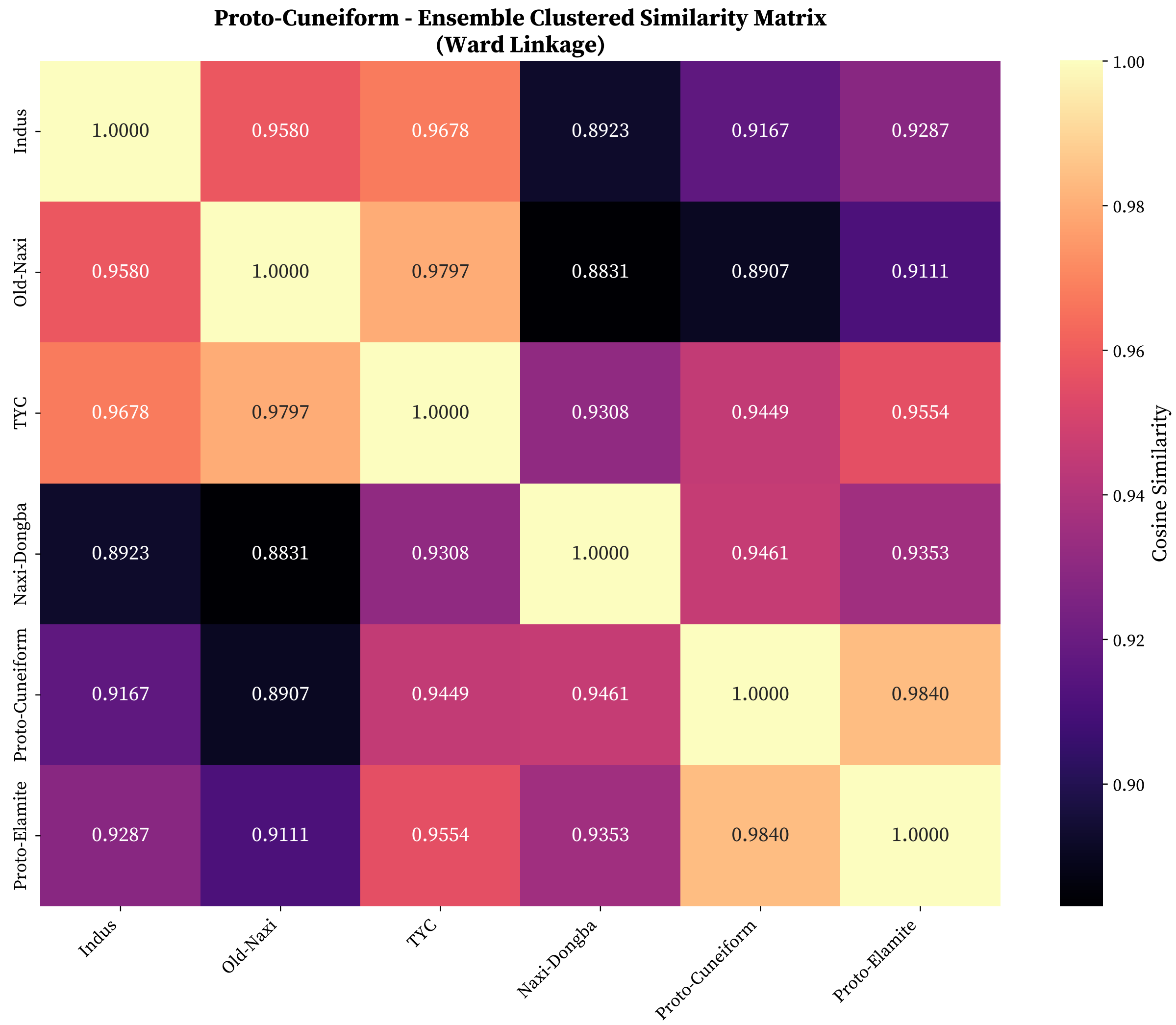

Figure 4.5: Proto-Cuneiform Ensemble Ward Linkage Heatmap

As seen in Figure 4.5, our heatmaps display two distinct high-similarity regions (dark patches) at the intersections of West Asian scripts and at the TYC/Indus script intersection. This consistent pattern across multiple visualization techniques and mathematical approaches provides substantial evidence for the relationships identified in our feature embeddings. All ensemble heatmaps for all metrics can be found in Appendix A.6.

Chapter 5

# DISCUSSION

## 5.1 Interpretation of Script Similarity Patterns

Our computational analysis reveals a striking consistent pattern: TYC scripts (including both older and newer Dongba variants) demonstrate approximately six-fold higher similarity to the Indus Valley script compared to West Asian writing systems. This robust finding, consistent across 15 independently trained models and multiple analytical approaches, suggests a potential historical relationship between the Indus civilization and early writing systems of the TYC that warrants serious consideration.

The significantly higher cosine similarities between TYC scripts and the Indus script (0.635) compared to Proto-Cuneiform (0.102) or Proto-Elamite (0.078) suggest that the visual and structural characteristics shared between TYC and Indus scripts transcend coincidental resemblance. This relationship is further supported by our direct script-to-script embedding comparison, which shows an even stronger connection between Indus and TYC scripts (0.930) compared to Proto-Cuneiform (0.887) and Proto-Elamite (0.855). This pattern is particularly noteworthy given that Proto-Cuneiform and Proto-Elamite, despite their documented historical interactions with the Indus civilization, show markedly lower similarity scores; we take particular consideration of the previously mentioned work of Daggumati and Revesz [16–18]. Within this context, we do not discount the measured similarities between the Indus and West Asian scripts altogether; rather, the visual closeness of the TYC scripts inside the scope of this relationship should be valued even more so in conjunction.

In our visualizations, namely our t-SNE plots (Figure 4.2 and Appendix A.3) and dendrograms (Figure 4.4 and Appendix A.5, the Indus script clusters with the TYC and old Dongba datasets. Proto-Cuneiform and Proto-Elamite form their own clusters, with modern Dongba either forming its own subcategory or merging eventually with the West Asian scripts (at very high levels). Even more interesting is the treatment of modern Dongba within our PCA plots—with more variance, all other datasets tend to cluster together in the 3D space, leaving modern Dongba as the true outlier (Figure 4.3 and Appendix A.4. This signifies a larger ancient script tradition rooted in antiquity which modern Dongba may have outgrown over centuries, and that while we are nitpicking small differences, these scripts may all be more similar than we have yet to consider.

### 5.1.1 Interpreting Cohen's d with Low Variance

The extreme Cohen's $d$ values (up to $d = 8.36$) observed in our Indus-TYC comparisons arise from both substantial differences in group means and exceptionally low within-group variance in similarity scores. While Cohen's $d$ quantifies the standardized difference between two means, its magnitude is inversely proportional to the pooled standard deviation ($SD_p$), as defined:

$$d = \frac{M_1 - M_2}{SD_p} \tag{5.1}$$

where $M_1$ and $M_2$ are the group means, and $SD_p$ is the pooled standard deviation calculated by:

$$SD_p = \sqrt{\frac{(SD_1^2 + SD_2^2)}{2}} \tag{5.2}$$

In our similarity distributions, the standard deviations are consistently small ($SD_1, SD_2 \approx 0.02 - 0.05$), while the differences in means are substantial ($M_1 - M_2 \approx 0.50 - 0.60$). This results in inflated $d$ values, as even moderate mean differences divided by minimal dispersion yield extreme standardized effects.

Such effect sizes vastly exceed conventional benchmarks—namely, the $d = 0.8$ classification for "large" effects [15]). While these magnitudes are mathematically valid, they do warrant careful interpretation. The bounded nature of similarity scores ($0 \leq x \leq 1$) and the tight clustering within groups inherently suppress variance, inflating standardized measures.

Nonetheless, the consistency of these patterns across independent models and across different script datasets suggests that the relative relationships—rather than absolute magnitudes—are robust. Specifically, the persistent gap between Indus-TYC similarities and those involving Proto-Cuneiform or Proto-Elamite, reflected in large effect sizes (7.13-8.36) for these comparisons compared to small-to-medium effect sizes (0.41-0.61) between Proto-Cuneiform and Proto-Elamite, substantiates the hypothesis of a closer structural alignment between Indus and TYC script forms.

## 5.2 Historical Implications

These findings align with the aims of our experiment in identifying the potential for Indus influence on the TYC scribal traditions, with both our glyph-level and script-level analyses pointing to the same conclusion. The particularly high script-level similarity (0.930) between TYC and Indus scripts, compared to the similarities with Proto-Cuneiform (0.887) and Proto-Elamite (0.855), further solidifies our main anthropological hypothesis of an existent relationship of some sort between the two cultures and peoples. The consistent clustering of Indus, old Dongba, and TYC scripts across multiple visualization techniques (t-SNE, PCA, hierarchical linkage) provides numerical and statistical evidence that may work to support unexplored historical transmission theories and lend statistical support to the early usage of the Shu-Shendu trans-Hengduan road.

The divergence of the newer Dongba script—consistently appearing as an outlier or clustering with West Asian scripts at higher levels—reflects its documented Sinicization and modern distance from the original ancient TYC tradition. This evolution provides a valuable control within our analysis, demonstrating that our model successfully distinguishes between indigenous and imported scribal traditions in the TYC region.

This newfound connection aims to spark further study and exploration, especially into potential connection of the IVC to Ba-Shu cultures and the preceding civilization of Sanxingdui—both of which remain a large unknown compared to the documented Han dynasties of China. We also hope to have opened a new avenue within the existing study of Indology and provide experts with a novel approach to deciphering the Indus script (and consequently, the Ba-Shu scripts), which remains a pressing question of our time [37]. The IVC also serves as a doorway to surviving cultures; determining the script and language's proper classification and, therefore, descendancy creates a pathway of study linking the TYC to modern Indian subcontinental demographics.

## 5.3 Methodological Contributions

Our hybrid CNN-Transformer approach combines the complementary strengths of convolutional layers (for local feature extraction) and self-attention mechanisms (for capturing long-range dependencies), effectively capturing both fine-grained stroke patterns and global structural relationships in ancient scripts. Combined with our contrastive loss functionality, ensemble learning strategy, and rigorous statistical testing, we establish a robust quantitative foundation for comparing ancient writing systems that moves beyond subjective visual assessment.

Our dual-metric approach—comparing both individual glyphs and aggregate script representations—provides a more comprehensive understanding of script relationships than either method alone. The glyph-level analysis reveals specific visual similarities between characters, while the script-level comparison captures broader structural patterns that might not be apparent from character-by-character comparisons. By applying multiple dimensionality reduction techniques (t-SNE, PCA) and hierarchical clustering with various linkage methods, we ensure that the observed script relationships are genuine and not artifacts of any particular analytical approach, drawing meaningful patterns.

## 5.4 Alternative Interpretations

While our findings strongly suggest a potential historical connection between the Indus script and TYC writing systems, several alternative explanations warrant consideration:

1. **Independent origination:** The similarity between TYC and Indus scripts could potentially arise from independent development of similar graphical solutions for representing comparable concepts. Pictographic systems often begin with similar intuitive representations before evolving into more complex forms.
2. **Common ancestry:** Both script traditions might derive from an even older, as-yet-undiscovered symbolic system that predates both the Indus Valley Civilization and the early symbolic traditions of the TYC region.
3. **Convergent evolution:** Similar environmental or cultural constraints could lead to parallel symbolic development. The functional requirements of recording certain types of information (counting systems, agricultural cycles, or religious practices) might naturally produce similar pictographic solutions across cultures.

The highly specific nature of some similarities, particularly in numerical systems and gender markers, makes these alternative explanations less likely than some form of cultural transmission—direct or indirect—but they cannot be conclusively ruled out without additional archaeological evidence.

## Chapter 6

# CONCLUSIONS

This study provides substantial evidence supporting our primary hypothesis of a significant visual relationship between the Indus Valley script and writing systems of the TYC. Through our hybrid CNN-Transformer computational approach and ensemble modeling, we consistently observed approximately six-fold higher similarity between TYC scripts and the Indus script (0.635) compared to Proto-Cuneiform (0.102) or Proto-Elamite (0.078) in our glyph-level analysis. Our direct script-to-script embedding comparison further strengthened these findings, showing an even stronger connection between Indus and TYC scripts (0.930) compared to Indus and Proto-Cuneiform (0.887) or Proto-Elamite (0.855). These findings remained robust across five independently trained models and multiple analytical approaches, including cosine similarity analysis, dimensionality reduction techniques, and hierarchical clustering.

These computational results, combined with our qualitative observations of specific pictorial parallels, particularly in numerical systems, gender markers, and key iconographic elements like the sky/roof character and city representations, are remarkably consistent across analytical methods and suggest these similarities are not coincidental but rather may reflect genuine historical connections.

Our findings in conjunction with the presented anthropological evidence give credit to the theory of possible transmission along the ancient Shu-Shendu road, active during the critical period following the abandonment of IVC cities in the second millennium BCE. Archaeological evidence of cross-regional exchanges—including cowrie shells from the Indian Ocean at sites in Southwestern China and evidence of crop transmission—substantiates the existence of sustained contact networks through which symbolic systems could have diffused.

While alternative explanations remain plausible (independent invention, common ancestry, or convergent evolution), the specific nature of the similarities identified through both computational and anthropological approaches is difficult to attribute to chance alone. The consistent clustering of Indus, old Dongba, and TYC scripts, coupled with specific iconographic parallels, suggests a more direct relationship than previously recognized.

These findings challenge conventional narratives about the isolated development of writing systems in South and East Asia and invite reconsideration of complex cultural transmission networks that may have connected the Indus Valley and the TYC following the decline of the Indus Valley Civilization. By bridging computational paleography with anthropological evidence, this research offers new perspectives on the interactions of the IVC people and culture that may aid in uncovering the unknowns of their civilizations and especially the encoding of their elusive inscriptions.

## 6.1 Limitations

Our study faces several methodological constraints that should be acknowledged. The available corpus of images from each script tradition is limited—only 5% of the IVC has been excavated, and the symbols found are constrained by their contextual usage primarily on seals. Similarly, our old Dongba database is not comprehensive—many older Rerko manuscripts have been destroyed over time—and biased towards frequency. The visual features captured by our model may not fully represent the functional or phonological characteristics of these writing systems, as our approach focuses primarily on visual similarity rather than linguistic or semantic content.

Additionally, differences in image preservation and digitization across the various script sources introduce potential biases. While our preprocessing steps and Grad-CAM validations helped mitigate these concerns, subtle artifacts may remain that could influence similarity assessments.

## 6.2 Future Research

Building on our findings, several promising research directions emerge. First, we propose applying our computational methodology to analyze the extensive corpus of rock art on the Tibetan Plateau [4], which may reveal additional connections to TYC morphological systems and, ergo, Indus. Second, we recommend intensifying archaeological investigations along the Shu-Shendu Road corridor, particularly focusing on Northeast India and the eastern edges of the Tibetan Plateau—regions that remain critically underexplored despite their potential significance in understanding ancient routes of transmission across less understood civilizations like the IVC and Sanxingdui/Jinsha [53]. Third, our visual embedding framework could be extended to incorporate linguistic constraints and semantic annotations where available, potentially advancing decipherment efforts for both the Indus script and Ba-Shu pictography. Finally, expanding our comparative analysis to include other ancient writing systems from Central and East Asia could further elucidate the complex network of cultural and symbolic exchange that characterized these historically significant regions, as well as introduce control variables to contextualize the strength of similarity.

## 6.3 Code Availability

All code, data, models, and experimental results pertaining to this thesis are publicly available on GitHub at https://github.com/oohalakkadi/ivc2tyc.

# Appendix A

# ADDITIONAL MATERIAL

## A.1 Data Acquisition

All data processing code used in this project can be located in `data/datasets.ipynb` (except Old-Naxi, which can be found in `data/old_naxi.ipynb`) within the project GitHub repository. All datasets were saved to Google Drive for streamlined access from Google Colab environments.

### Indus Dataset

The Indus script dataset was retrieved from the *Interactive Corpus of Indus Texts* (ICIT) [96] hosted at https://www.indus.epigraphica.de/. The complete HTML corpus was downloaded via authenticated requests and parsed using `BeautifulSoup` (Appendix A.2). Sign images were extracted from the ICIT site and downloaded programmatically using Python's `requests` library (Appendix A.2), authenticated via HTTP Basic Authentication . The dataset includes the full corpus of Indus signs, saved as `.jpg` images for use in visual similarity analysis.

### Proto-Cuneiform Dataset

The Proto-Cuneiform dataset was sourced from the `proto-cuneiform_signs` GitHub repository maintained by the Cuneiform Digital Library Initiative (CDLI) [11]. The dataset includes `.jpg` images of Proto-Cuneiform signs located in the `archsigns` directory. All images are licensed under CC BY, in accordance with Robert K. Englund's permissions. Files were cloned and extracted for analysis in this project.

### Proto-Elamite Dataset

The Proto-Elamite dataset was obtained from the `pe-decipher-toolkit` GitHub repository [86], released in connection with Born et al. [9]. The dataset includes categorized `.png` images of Proto-Elamite signs from the `PE_mainforms` and `PE_num` directories. This resource was developed to support computational decipherment research and accompanies the paper on sign clustering and topic extraction in Proto-Elamite.

### New Naxi Dongba Dataset

The New Naxi Dongba dataset was generated from the BabelStone Naxi LLC font [3], containing 2,162 glyphs derived from Lǐ Líncàn's *Naxi Pictographic Symbols Dictionary* [47]. Glyph images corresponding to Unicode points `U+E000` to `U+E849` were rendered as `.png` files at 224×224 resolution using the Python `Pillow` library (Appendix A.2). This dataset was created specifically for this project to support pictographic analysis.

### Old Naxi Dongba Dataset

The Old Naxi Dongba dataset was extracted from the VGTS (Versatile Glyph Text Spotting) GitHub repository [98]. This dataset includes Dongba Buddhist Hieroglyphic manuscripts (DBH), with 3,633 annotated bounding boxes across 253 categories. Glyphs were cropped from annotated manuscripts and additional support images were processed for one-shot learning tasks. Data includes hand-written character samples for training and evaluation.

### Classical Yi Dataset

The Classical Yi dataset was sourced from the *Ancient Yi Script Handwriting Sample Repository*, published in *Scientific Data* [50]. The dataset includes 427,939 handwritten samples across 2,922 classes, collected from 310 participants. Images were downloaded from the Science Data Bank repository and organized by category. The dataset covers a wide variety of handwriting styles and is used as a benchmark for handwritten character recognition.

### Ba-Shu Dataset

The Ba-Shu dataset was compiled from images sourced from the Sichuan Museum [89] and the Sanxingdui Museum [82], viewable in *Presentation and Translation of the Linguistic Landscape in the Sichuan Museum and the Spread of the Ba-Shu Culture from the Perspective of Cultural Communication* [102].

## A.2 Packages Used

The complete implementation is available in the GitHub repository.

### System and File Management

- `os`, `shutil`, `glob`, `zipfile`, `pathlib`, `pickle`
  For file operations, directory management, and handling archives.
- `logging`, `warnings`, `traceback`, `datetime`
  For system logging, error handling, and timestamping processes.

### Data Science and Numerical Computing

- `numpy`, `pandas`
  For numerical operations, data manipulation, and tabular data processing.
- `seaborn`, `matplotlib`
  For data visualization and plotting.

### Image Processing

- `Pillow (PIL)`, `opencv-python (cv2)`
  For image loading, preprocessing, and augmentation.

- `skimage`
  For image filtering and additional processing.
- `fontTools`
  For working with font files and glyph extraction.

### Machine Learning and Deep Learning

- `torch`, `torchvision`
  For building and training neural networks on GPU.
- `timm`, `transformers`
  Pretrained models and advanced architectures for vision and NLP tasks.

### Dimensionality Reduction and Clustering

- `sklearn.manifold.TSNE`, `sklearn.decomposition.PCA`
  For visualizing high-dimensional data.
- `KMeans`, `cosine_similarity`, `StandardScaler`
  For unsupervised clustering and similarity analysis.

### Evaluation Metrics and Analysis

- `confusion_matrix`, `classification_report`
  For evaluating model classification performance.

### Hierarchical Clustering and Statistical Analysis

- `scipy.cluster.hierarchy`, `scipy.spatial.distance`
  For hierarchical clustering and distance computation.
- `scipy.stats.ttest_ind`
  For statistical significance testing.

### Graph Theory and Network Analysis

- `networkx`
  For network-based graph analysis and visualization.

### Web Scraping and Requests

- `requests`, `BeautifulSoup (bs4)`
  For accessing web resources and parsing HTML/XML data.

### Progress Tracking and Utilities

- `tqdm`
  For tracking progress of loops and operations.

### Google Colab Specific

- `google.colab.drive`, `google.colab.files`
  For mounting Google Drive and managing file uploads/downloads within Colab.

### Hardware Acceleration

- `torch.device("cuda")`
  For leveraging GPU acceleration when available.

All packages were installed using `pip` within the Colab environment. CUDA acceleration was used for deep learning tasks when available.

## A.3 Ensemble t-SNE Visualizations

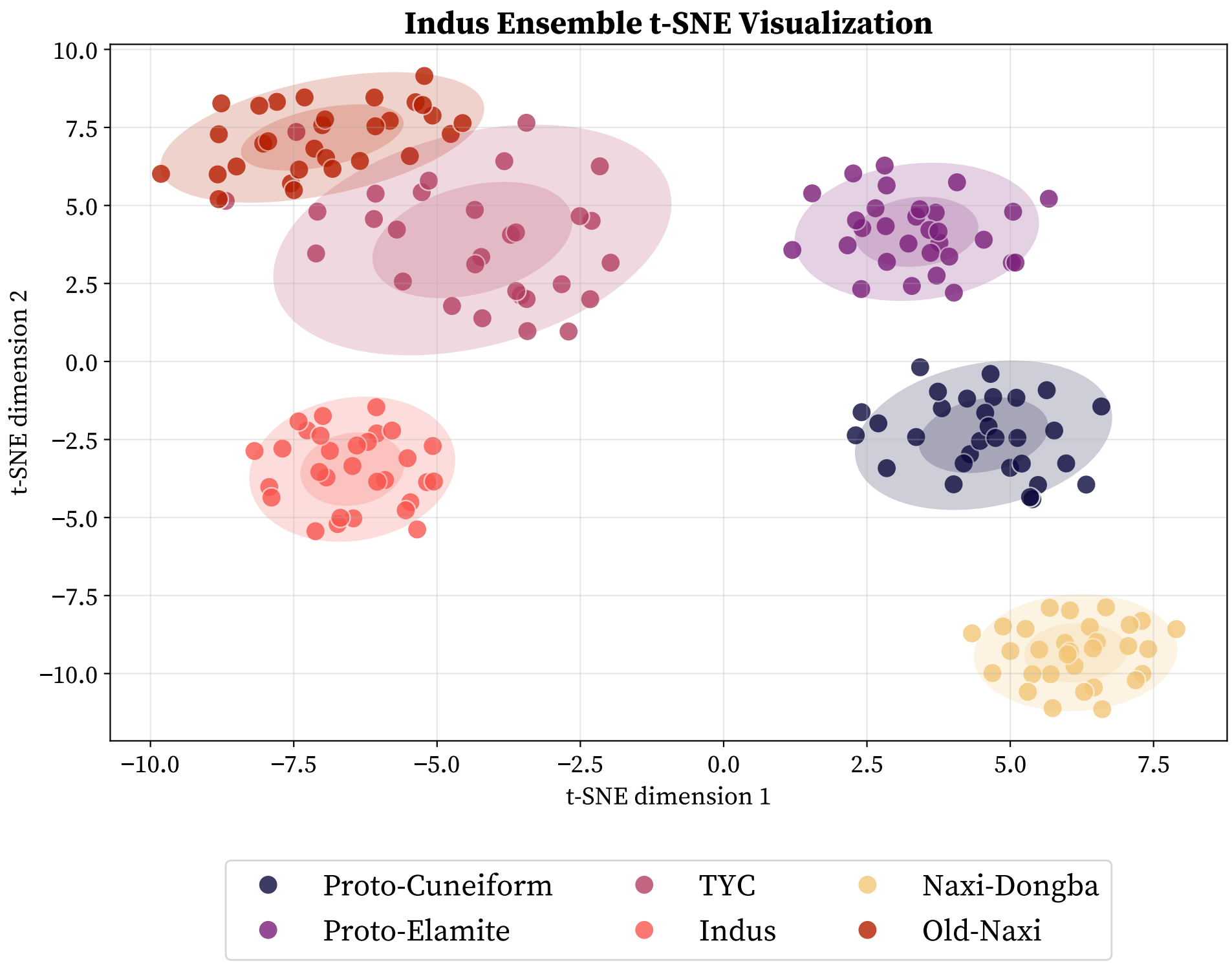

Indus Ensemble t-SNE

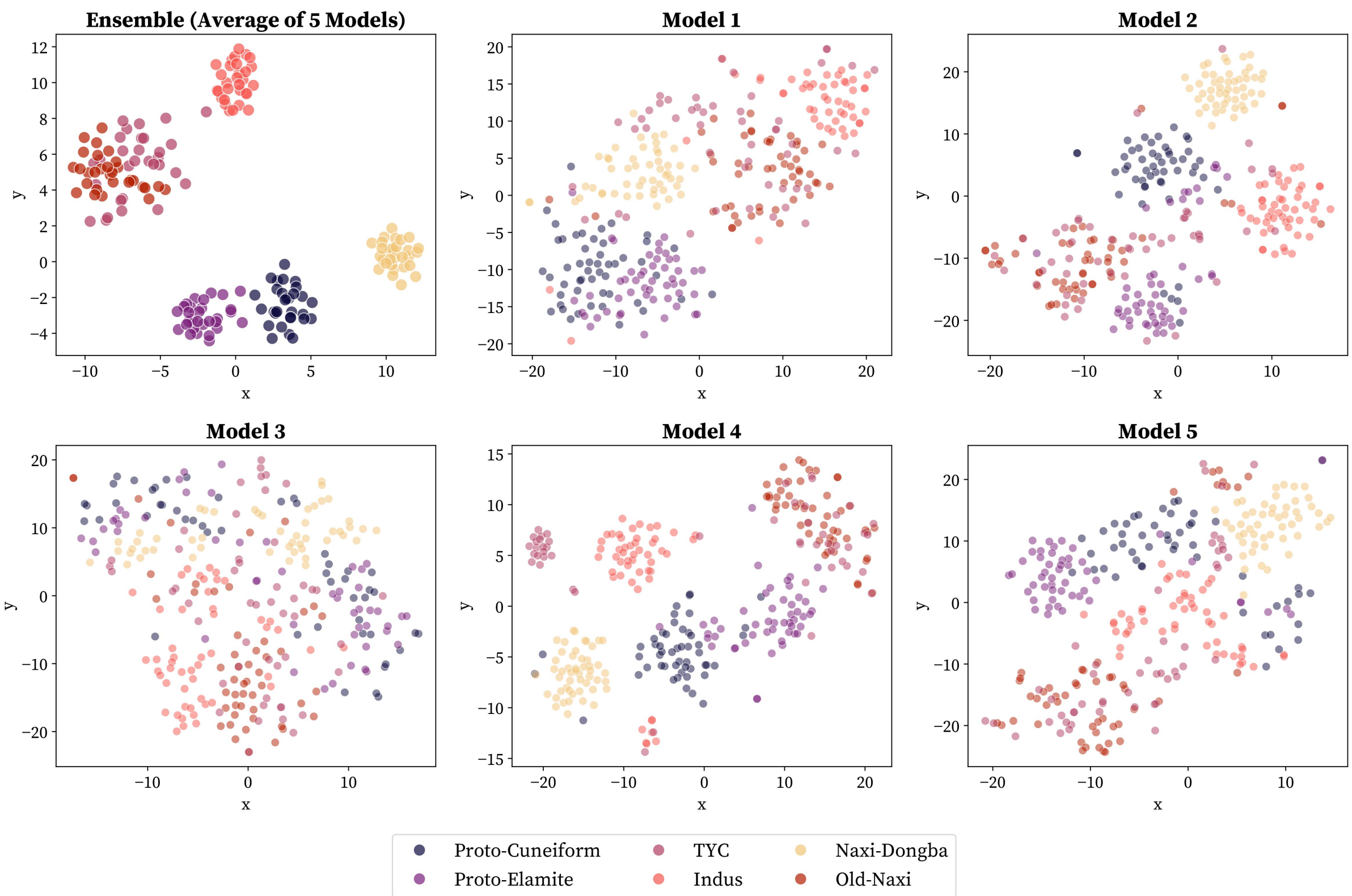

Indus Ensemble Model t-SNE Comparison Grid

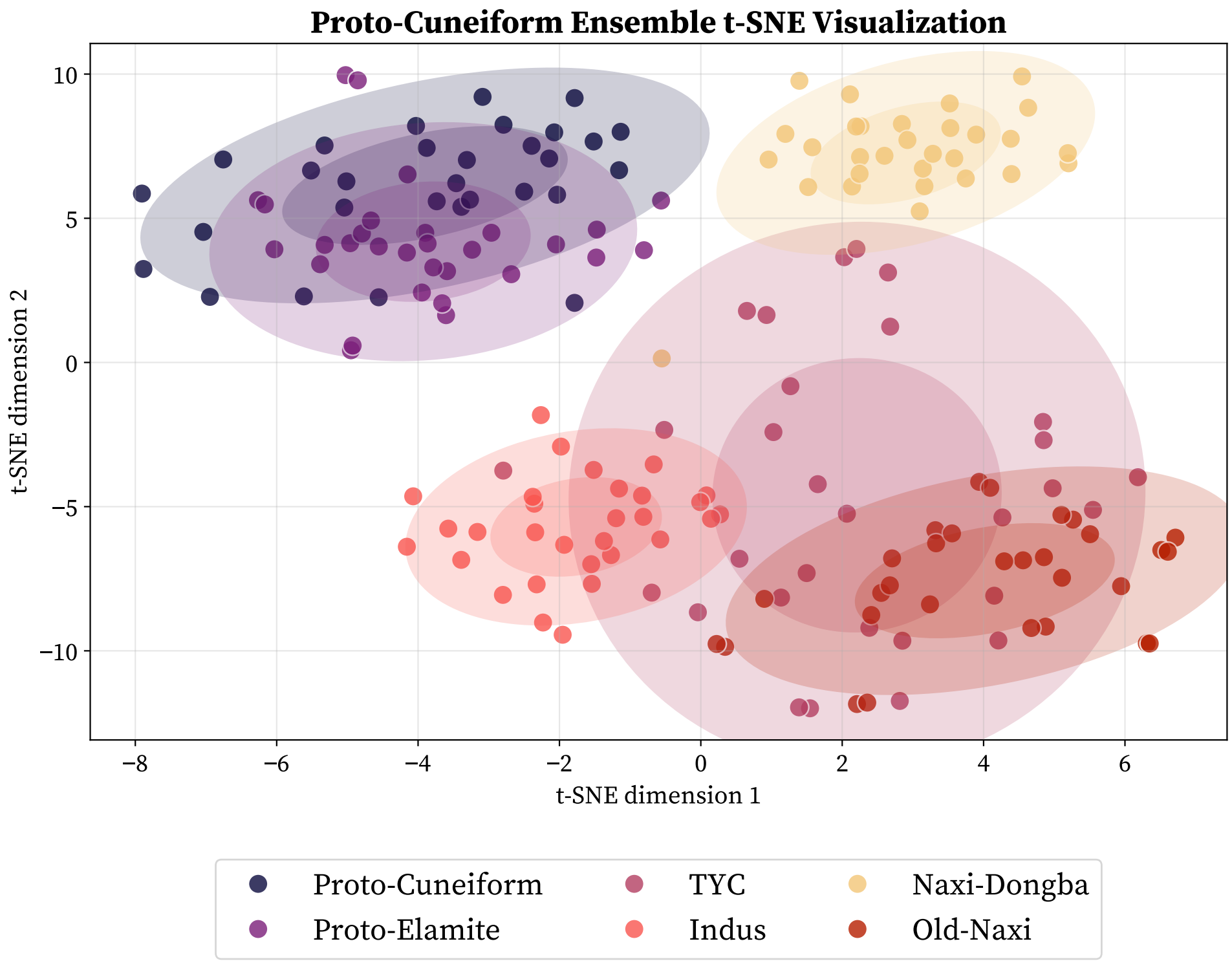

Proto-Cuneiform Ensemble t-SNE

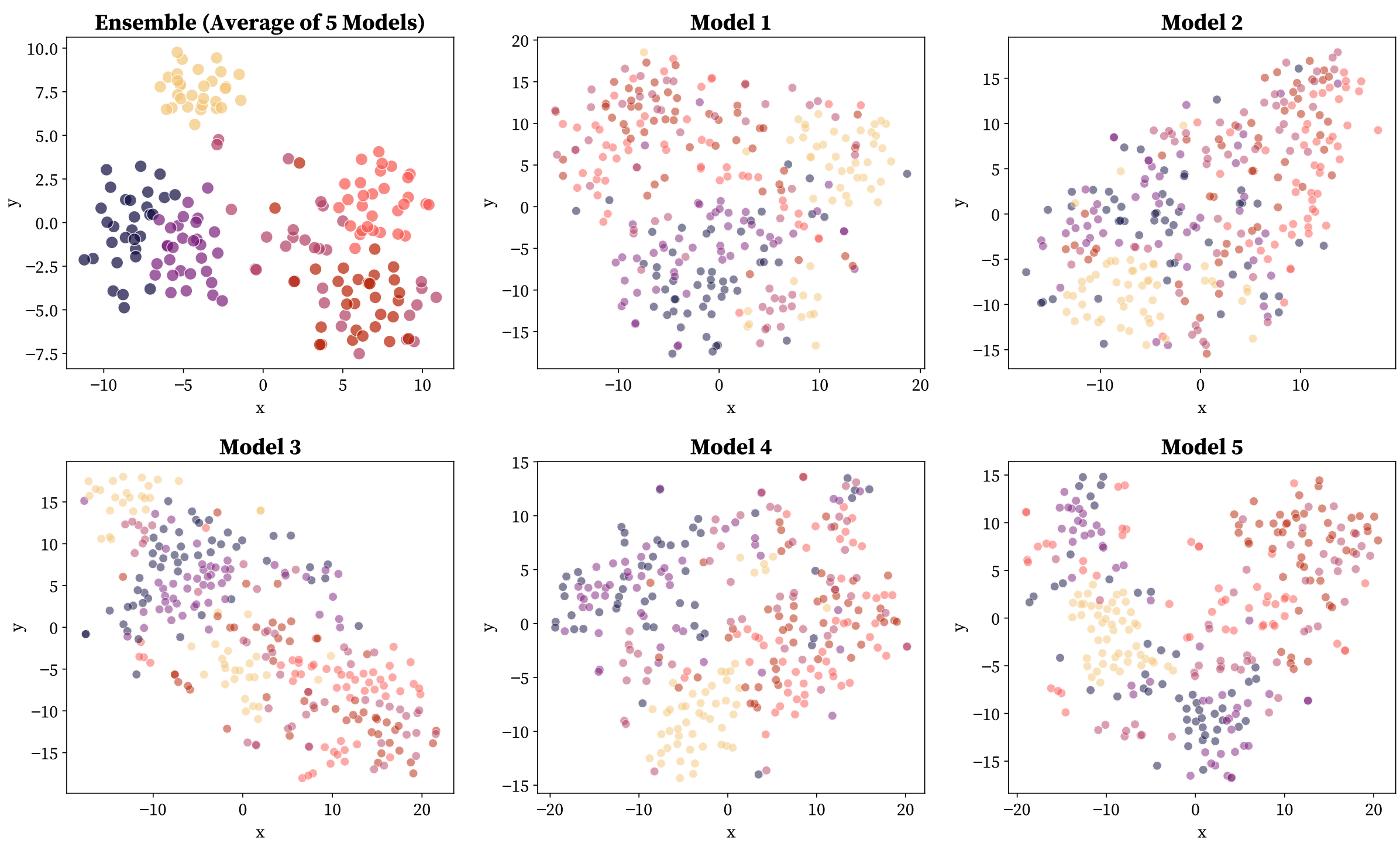

Proto-Cuneiform Ensemble Model t-SNE Comparison Grid

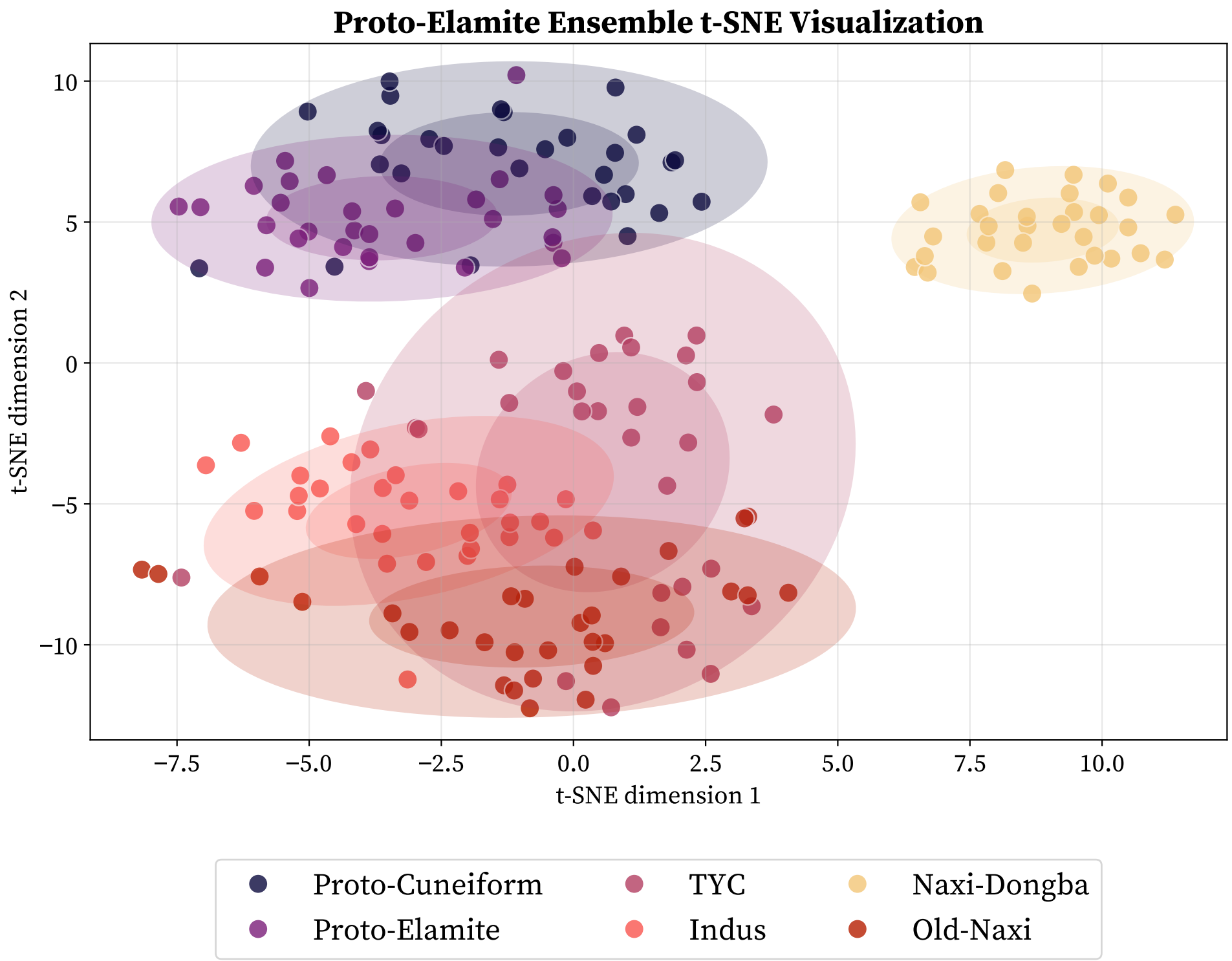

Proto-Elamite Ensemble t-SNE

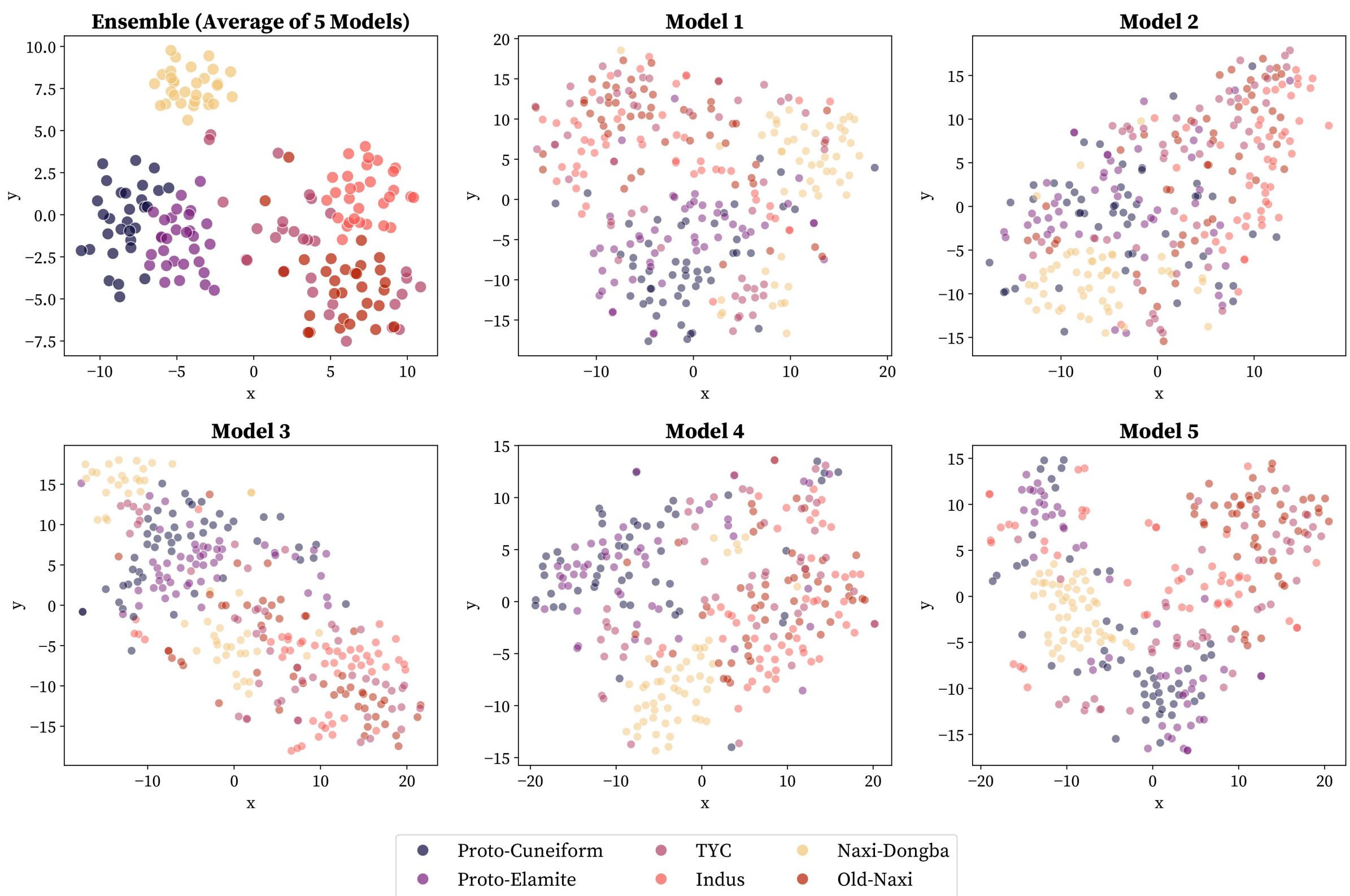

Proto-Elamite Ensemble Model t-SNE Comparison Grid

## A.4 Ensemble PCA Visualization

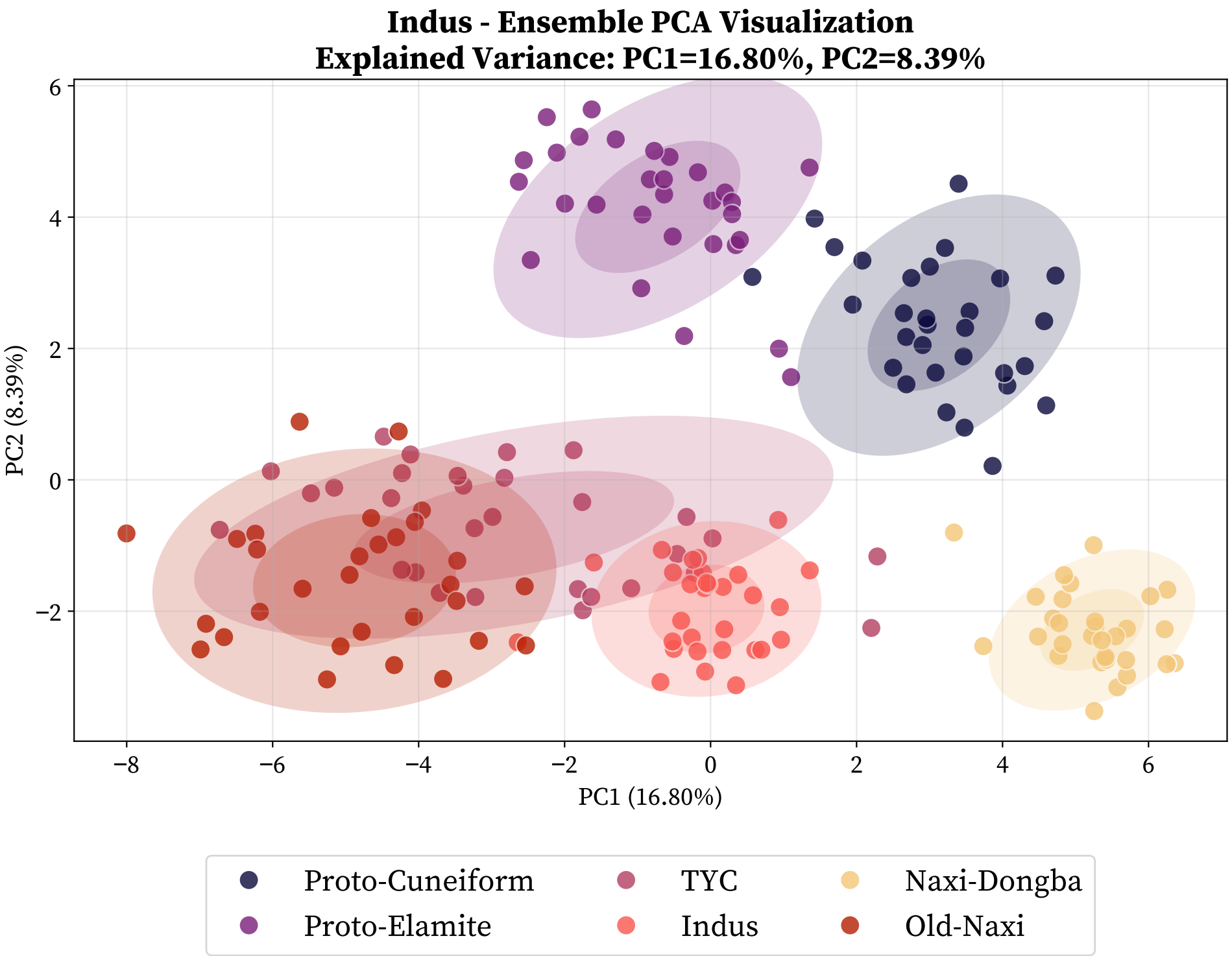

Indus Ensemble PCA

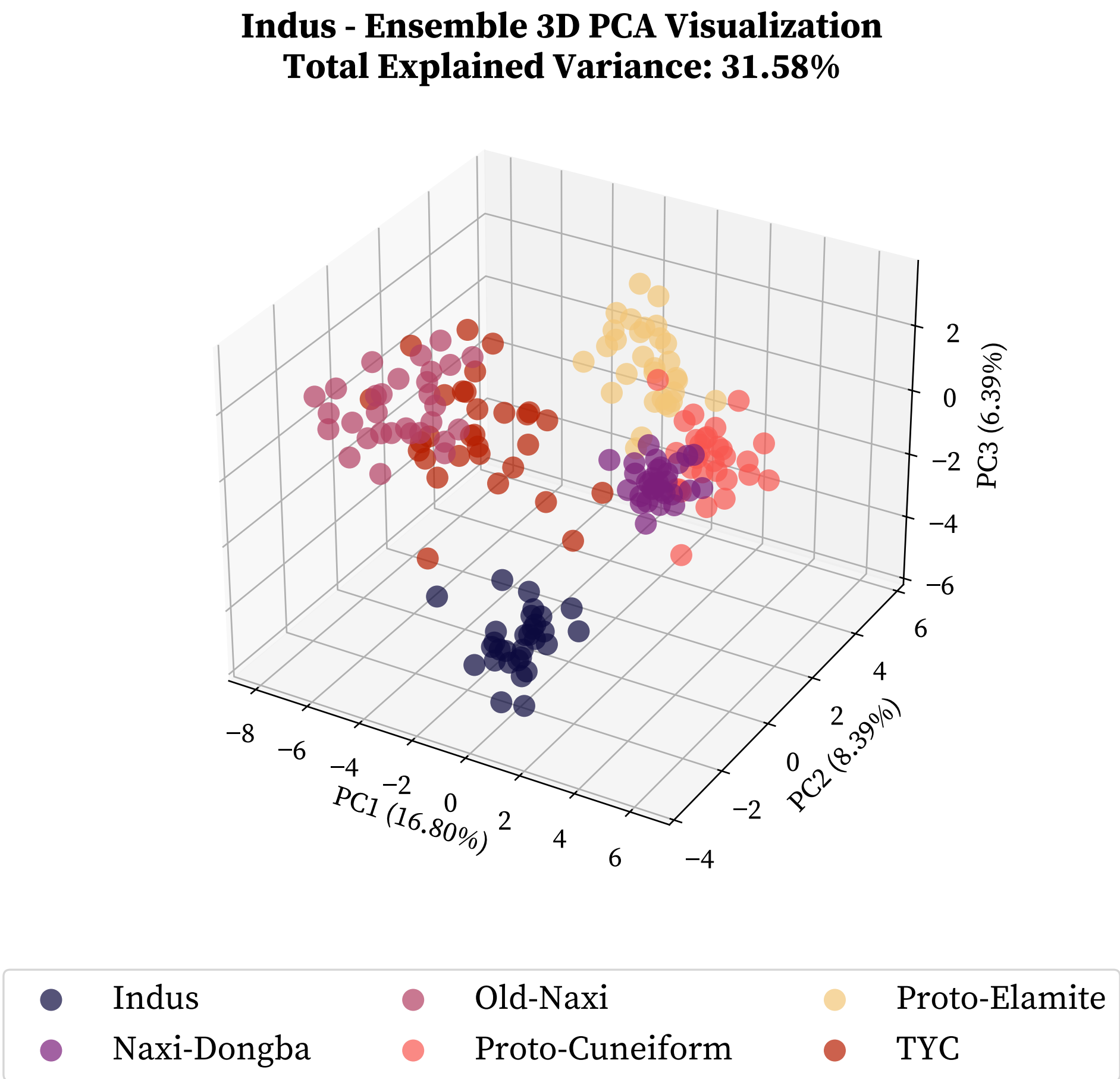

Indus Ensemble 3D PCA

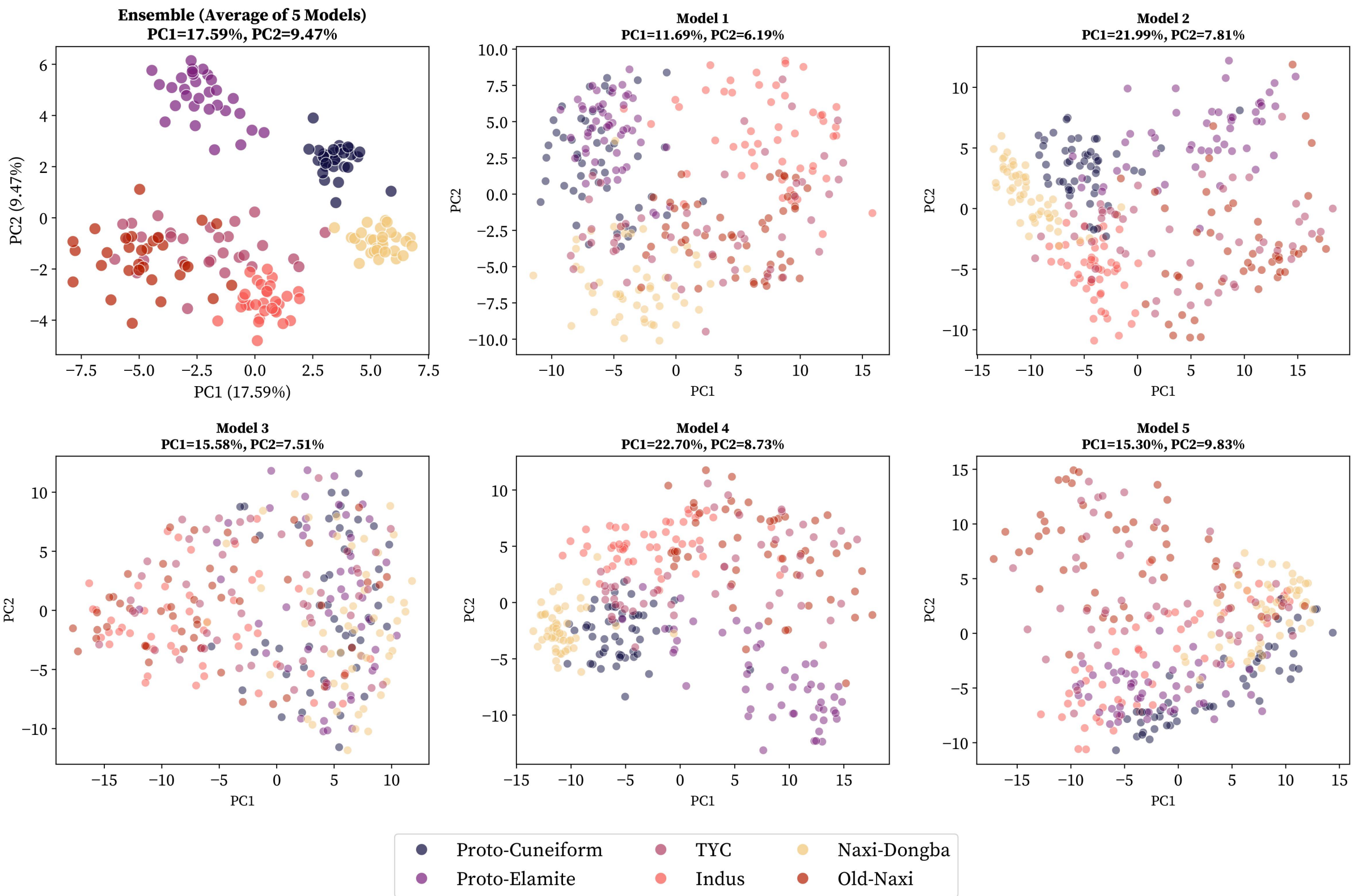

Indus Ensemble Model PCA Comparison Grid

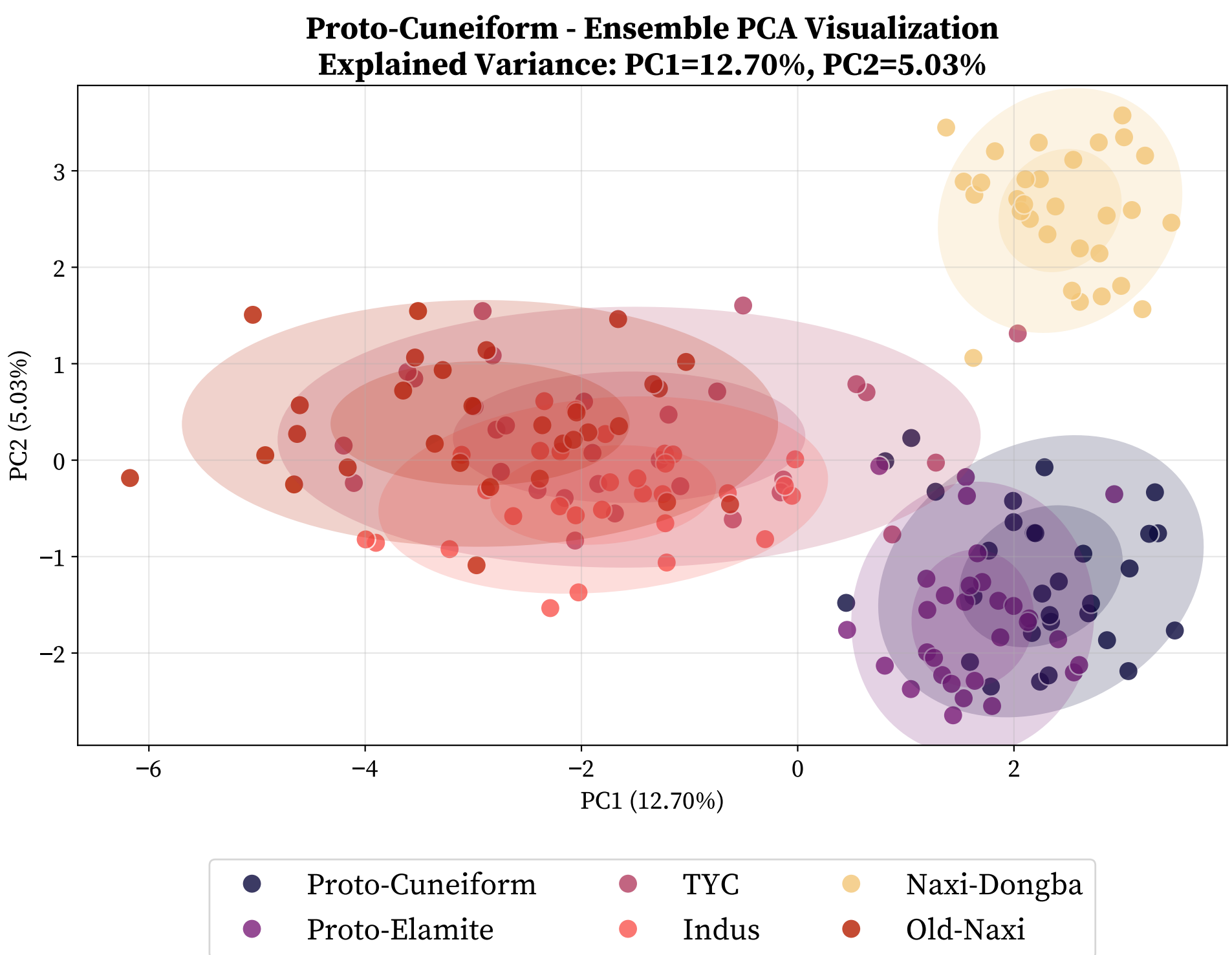

Proto-Cuneiform Ensemble PCA

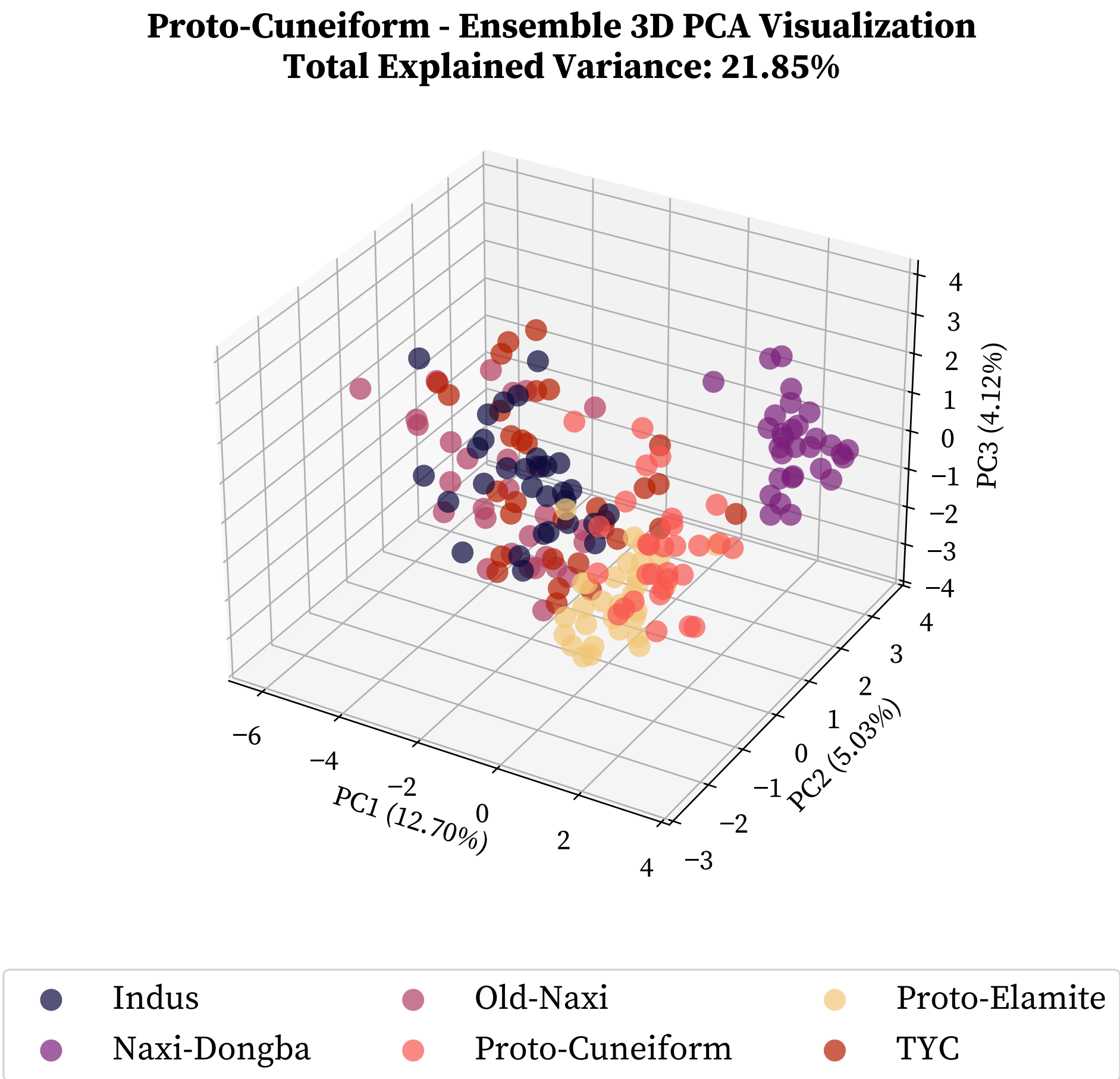

Proto-Cuneiform Ensemble 3D PCA

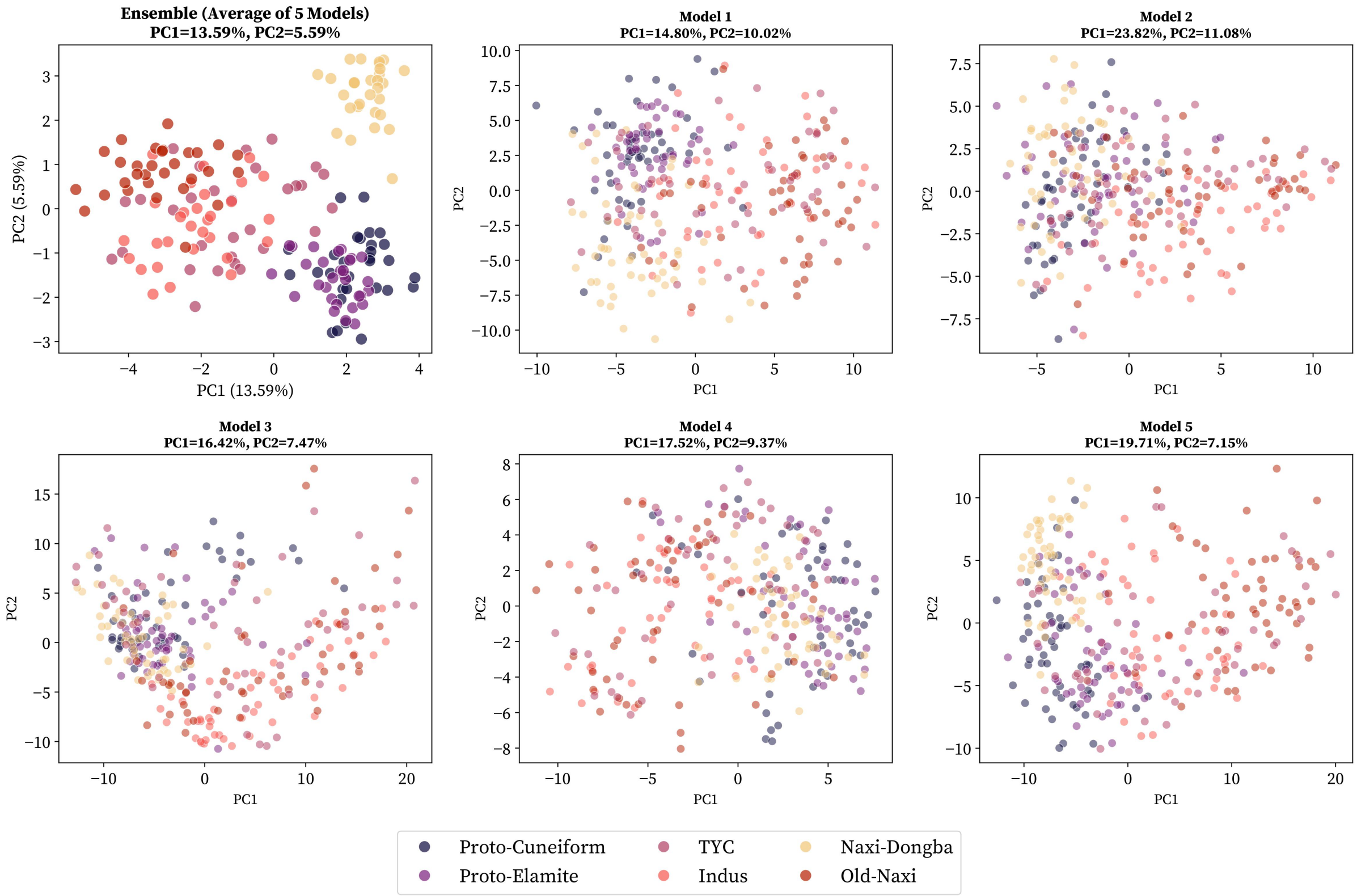

Proto-Cuneiform Ensemble Model PCA Comparison Grid

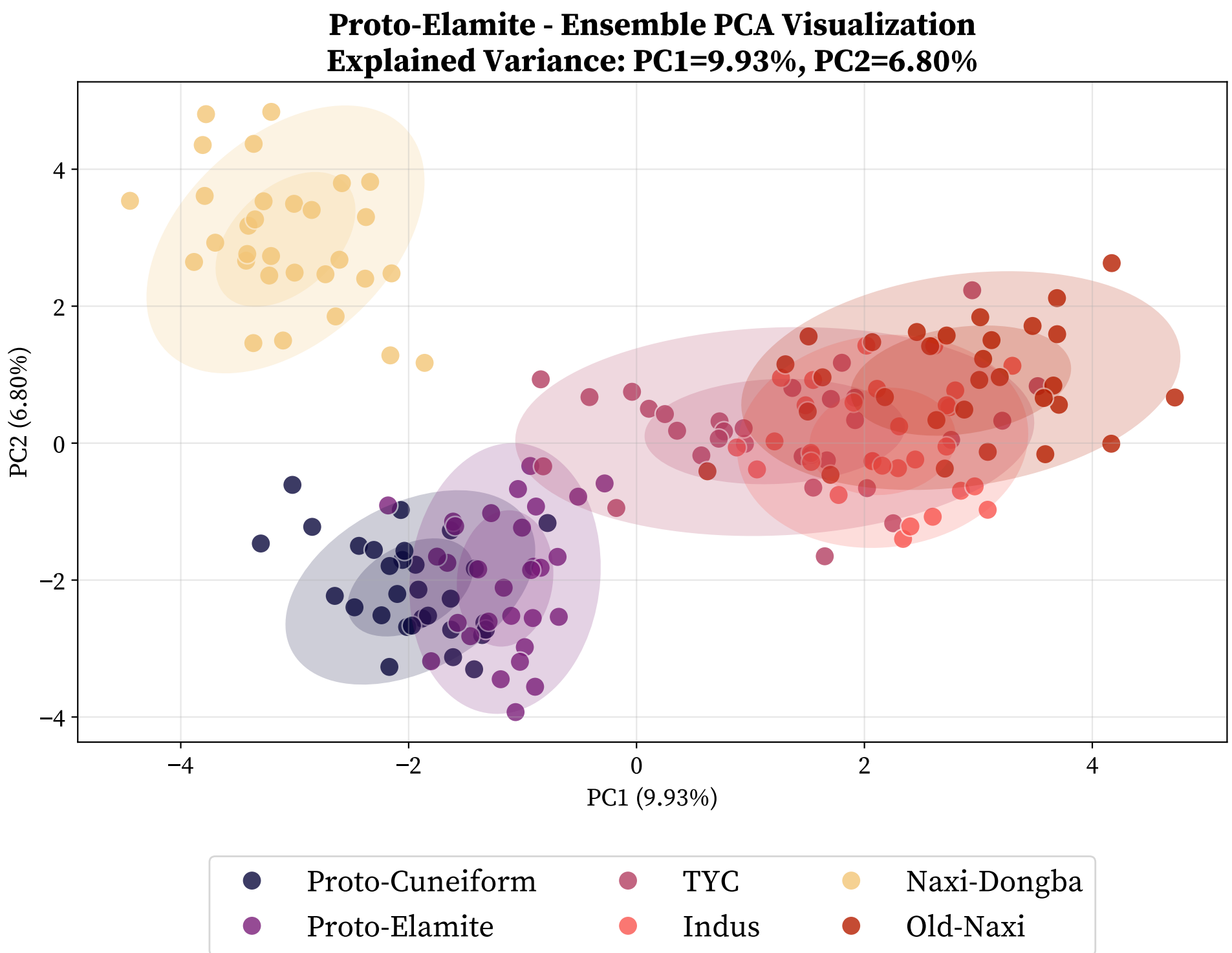

Proto-Elamite Ensemble PCA

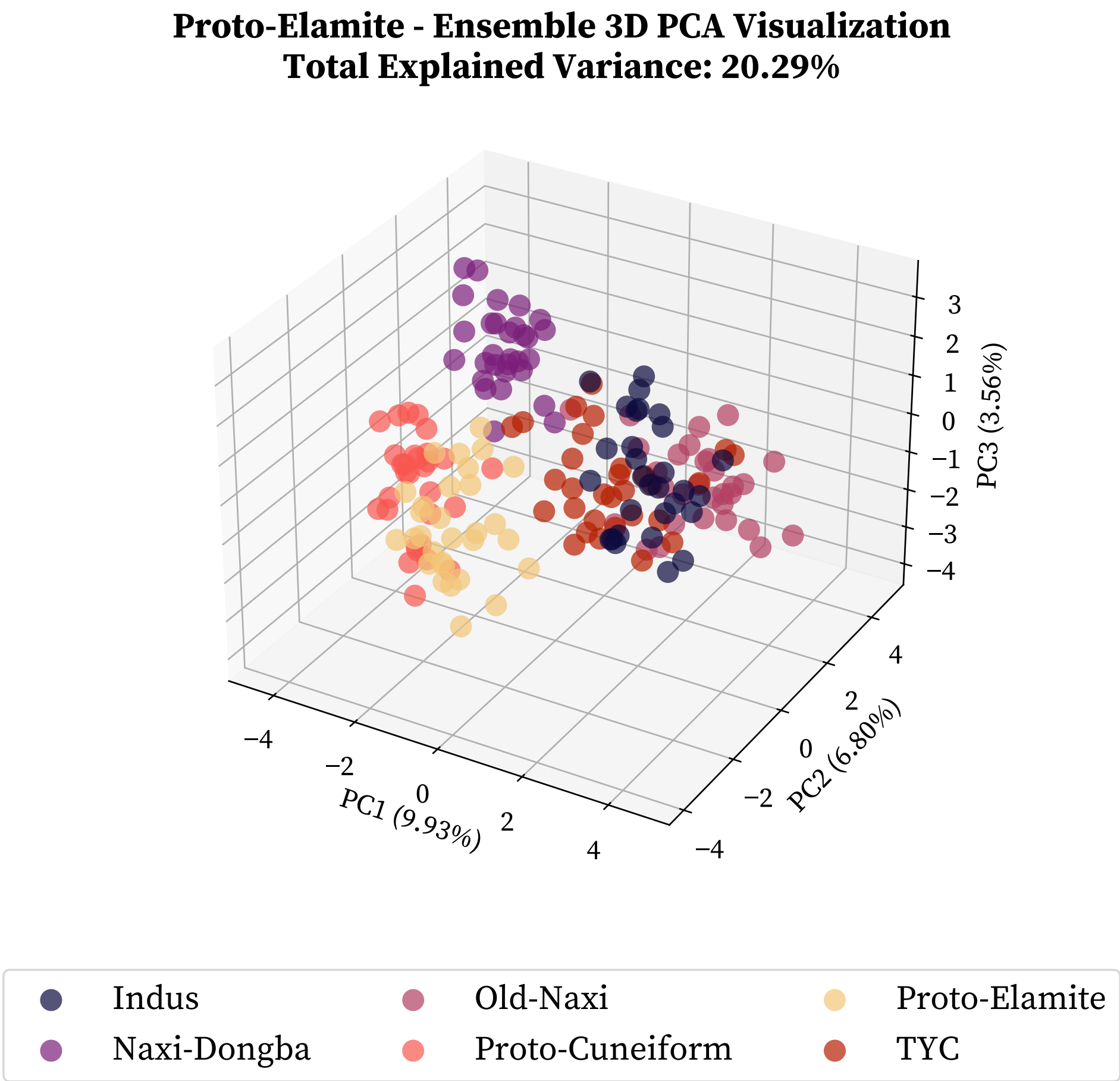

Proto-Elamite Ensemble 3D PCA

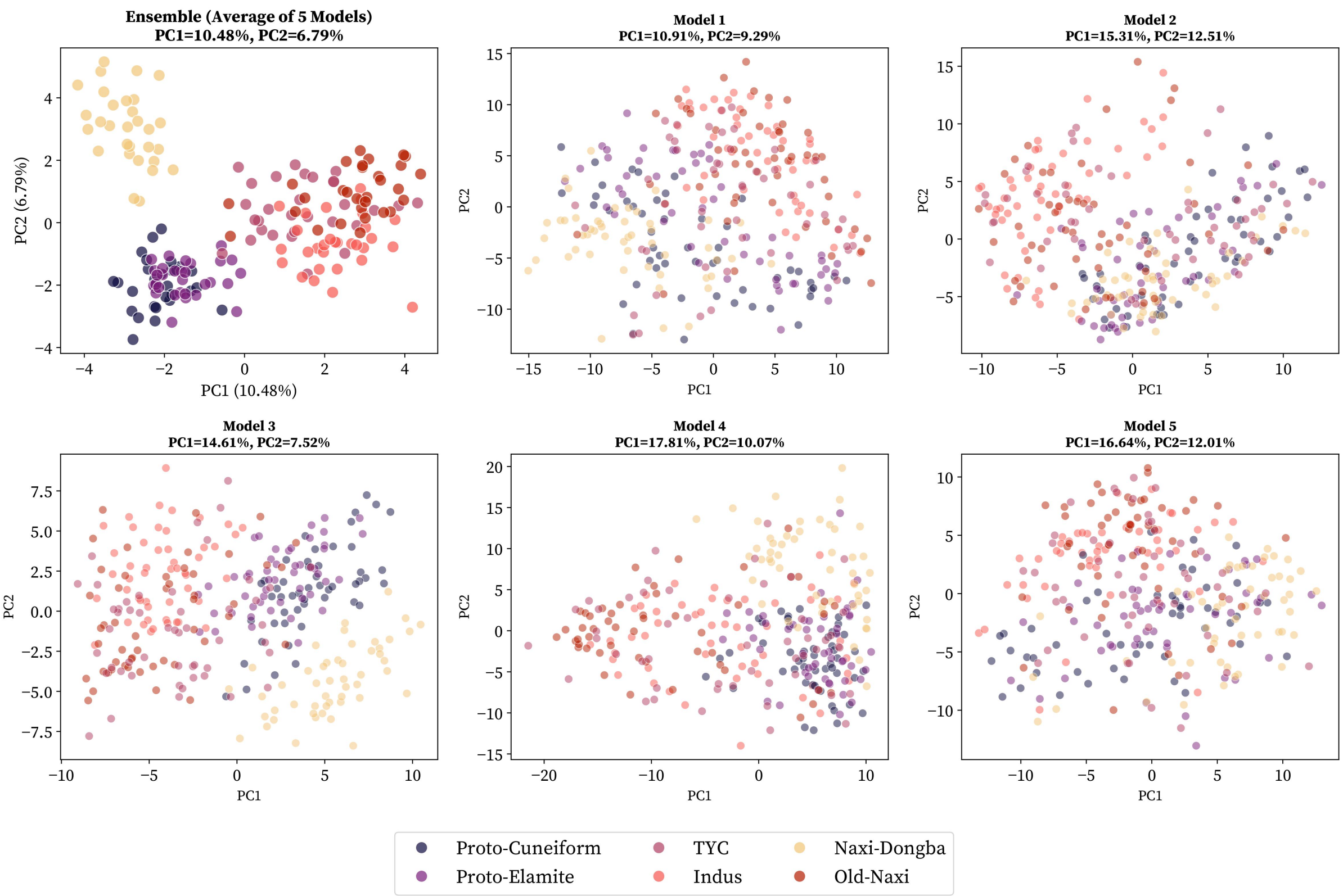

Proto-Elamite Ensemble Model PCA Comparison Grid

## A.5 Ensemble Hierarchical Clustering

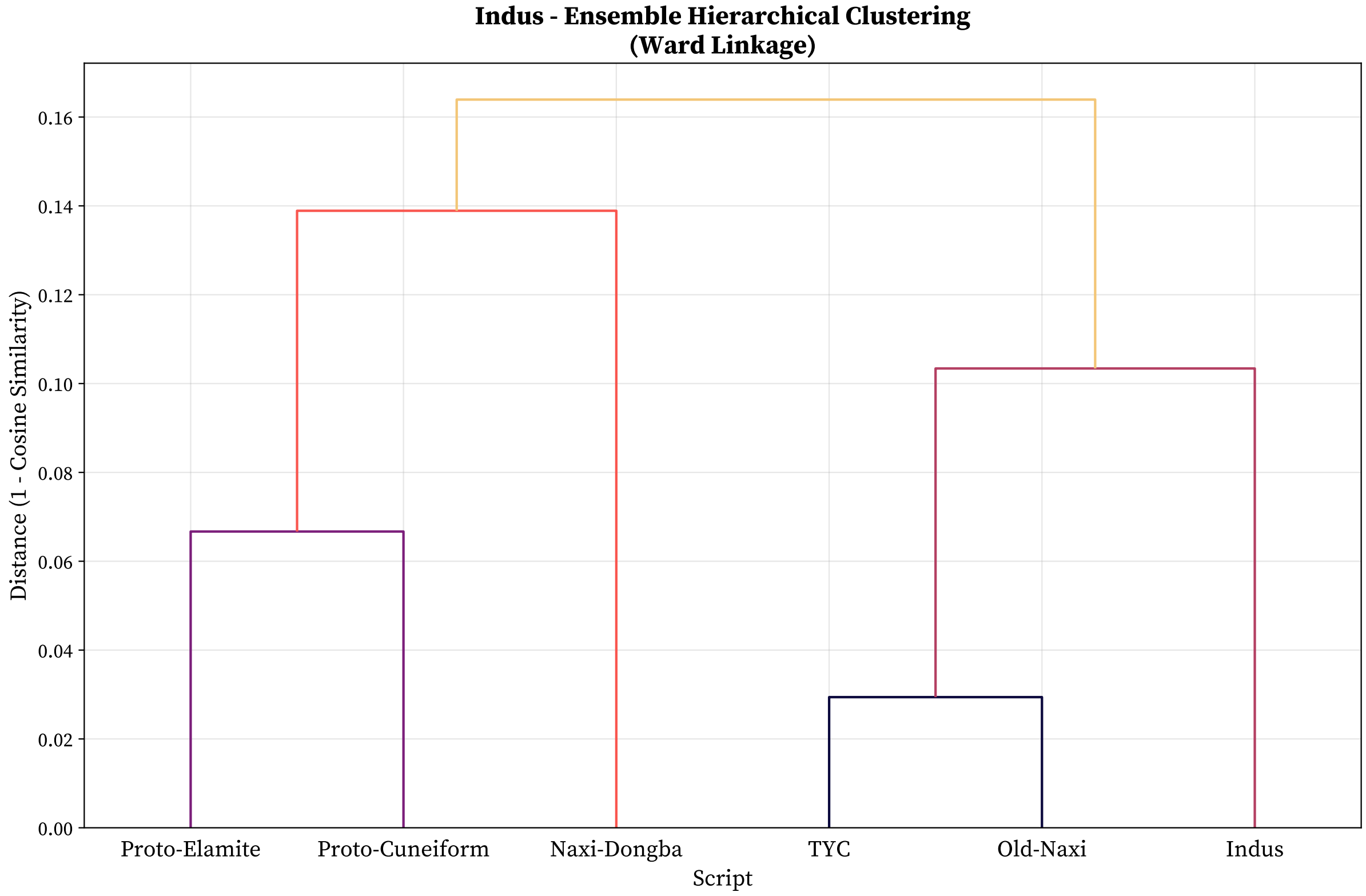

Indus Ensemble Ward Linkage Dendrogram

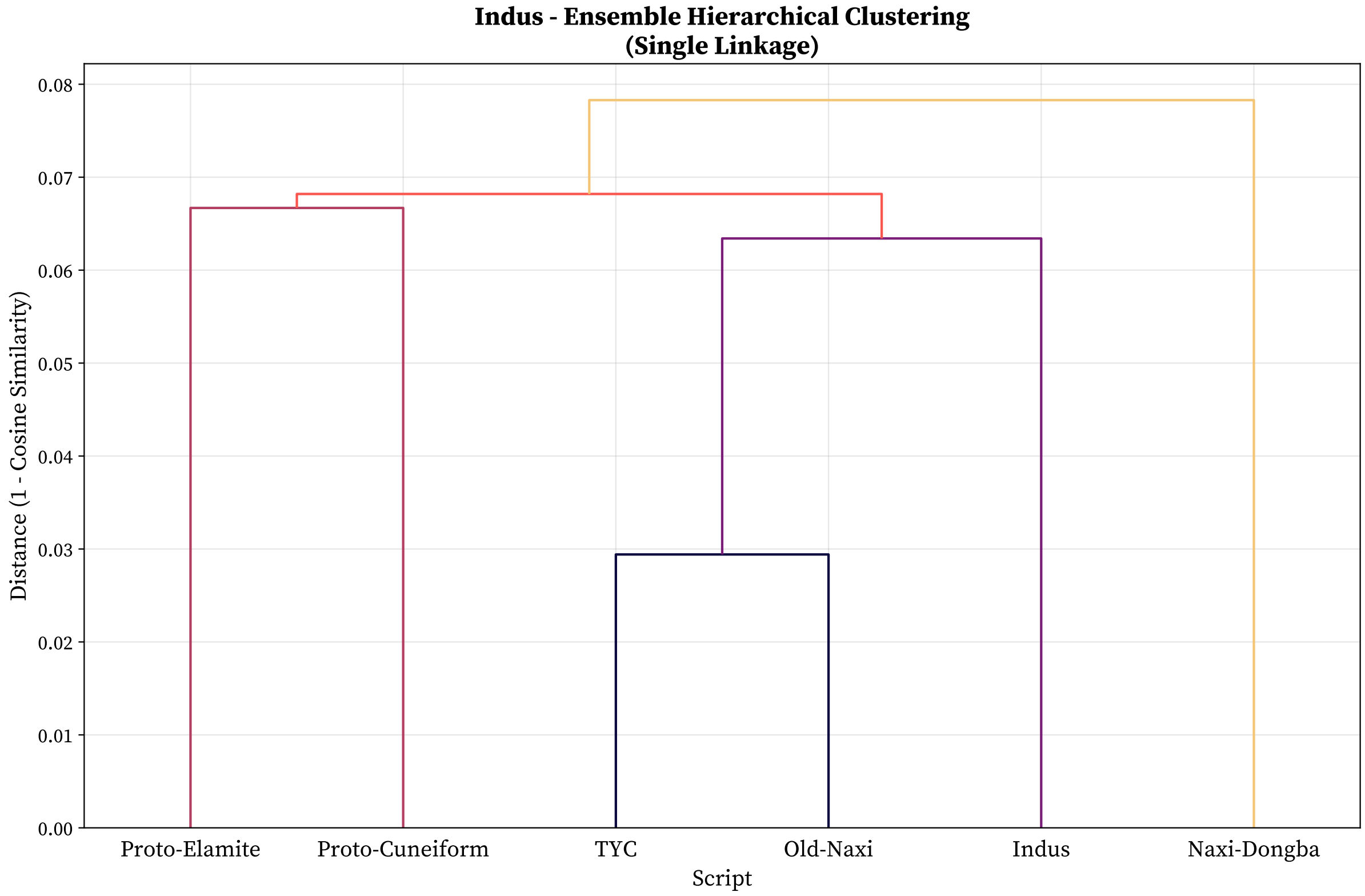

Indus Ensemble Single Linkage Dendrogram

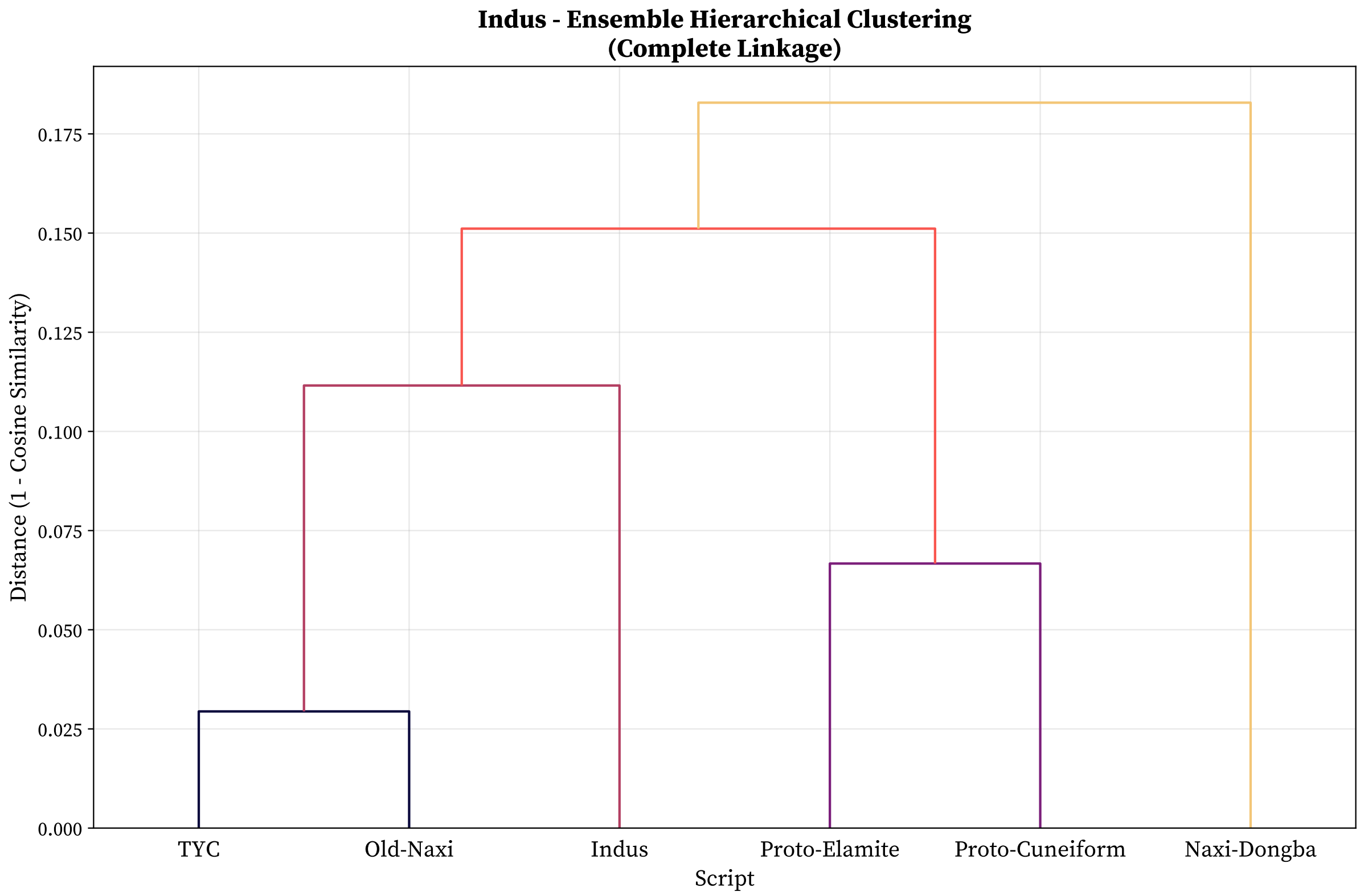

Indus Ensemble Complete Linkage Dendrogram

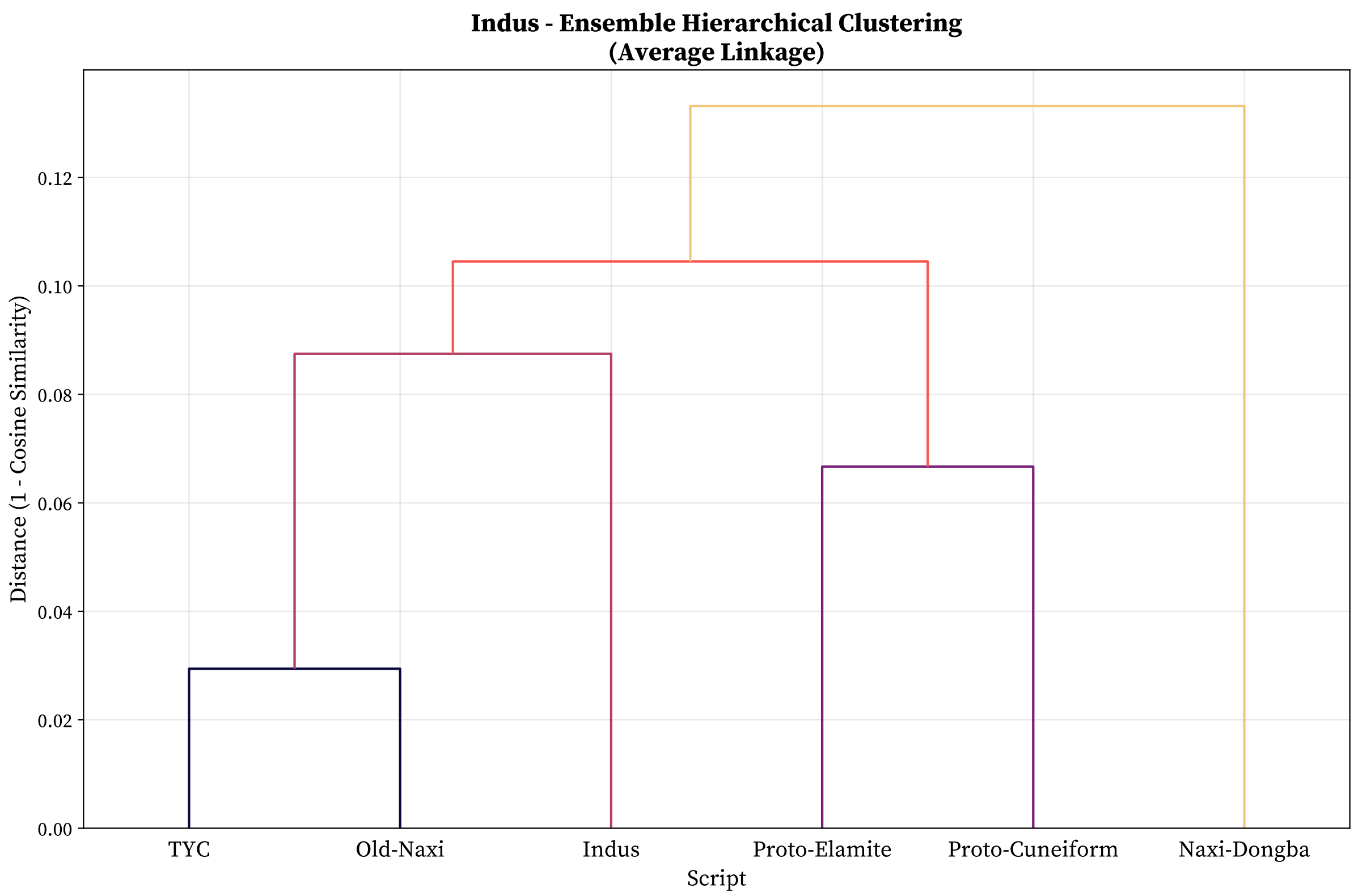

Indus Ensemble Average Linkage Dendrogram

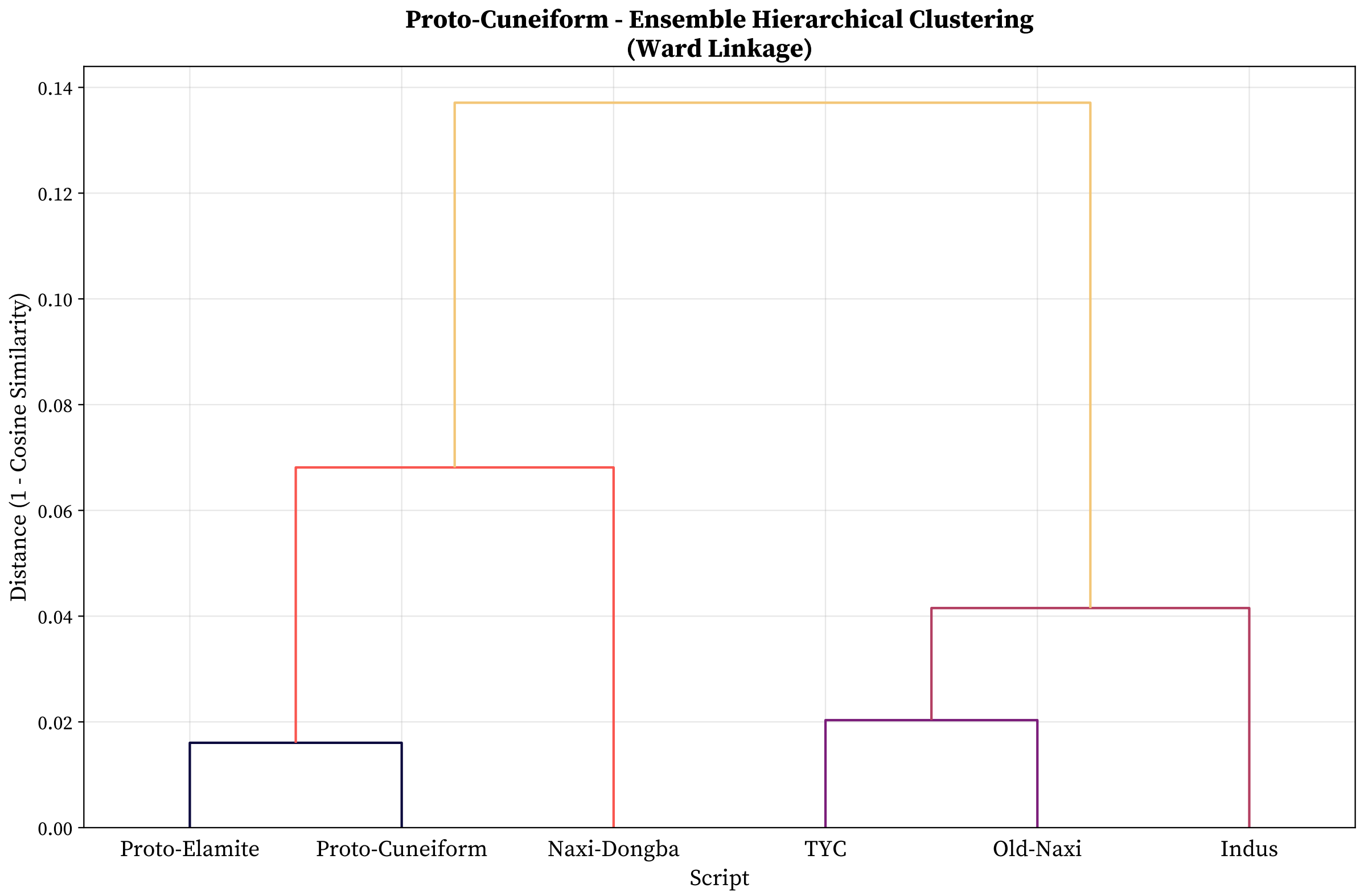

Proto-Cuneiform Ensemble Ward Linkage Dendrogram

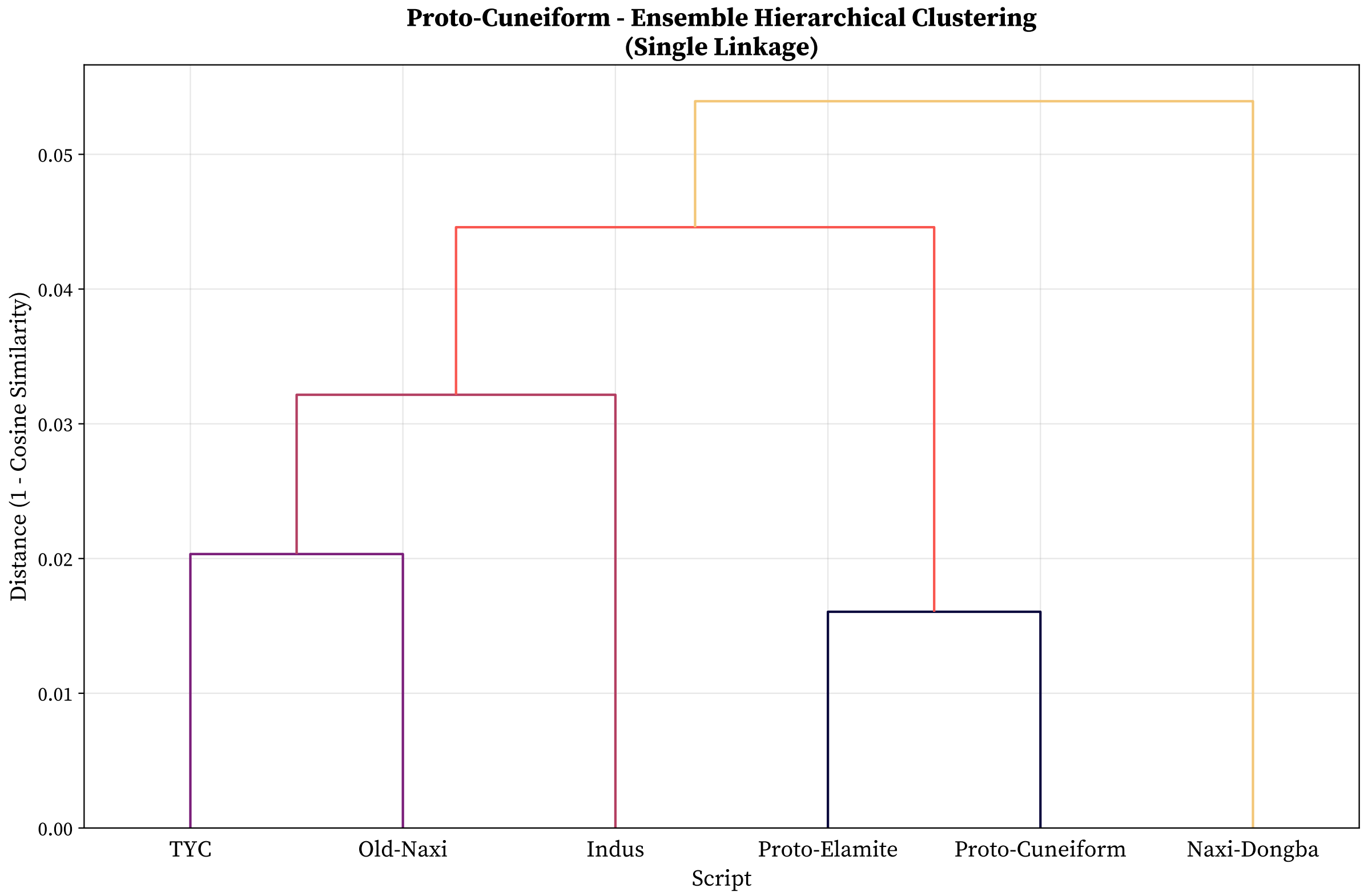

Proto-Cuneiform Ensemble Single Linkage Dendrogram

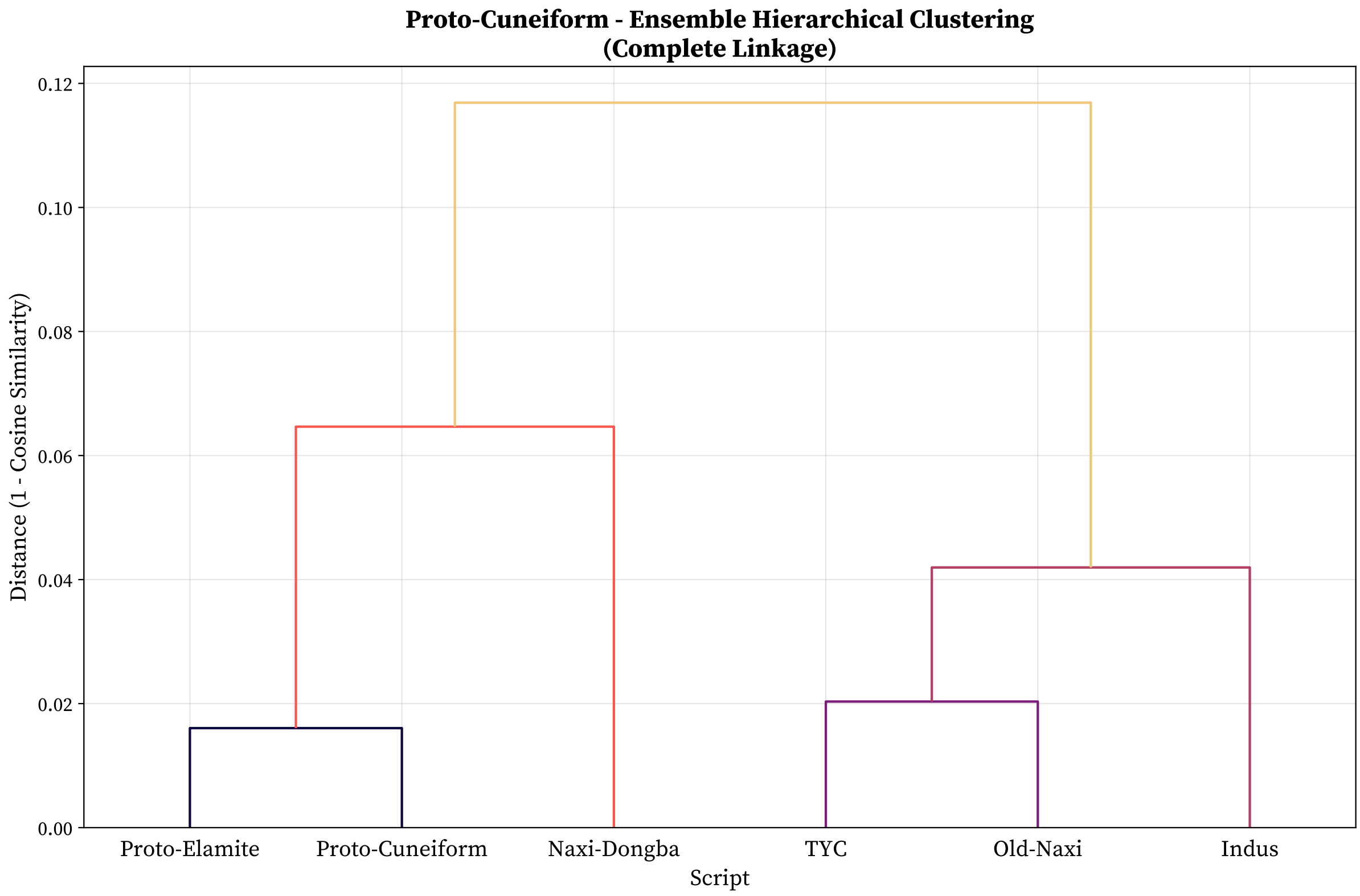

Proto-Cuneiform Ensemble Complete Linkage Dendrogram

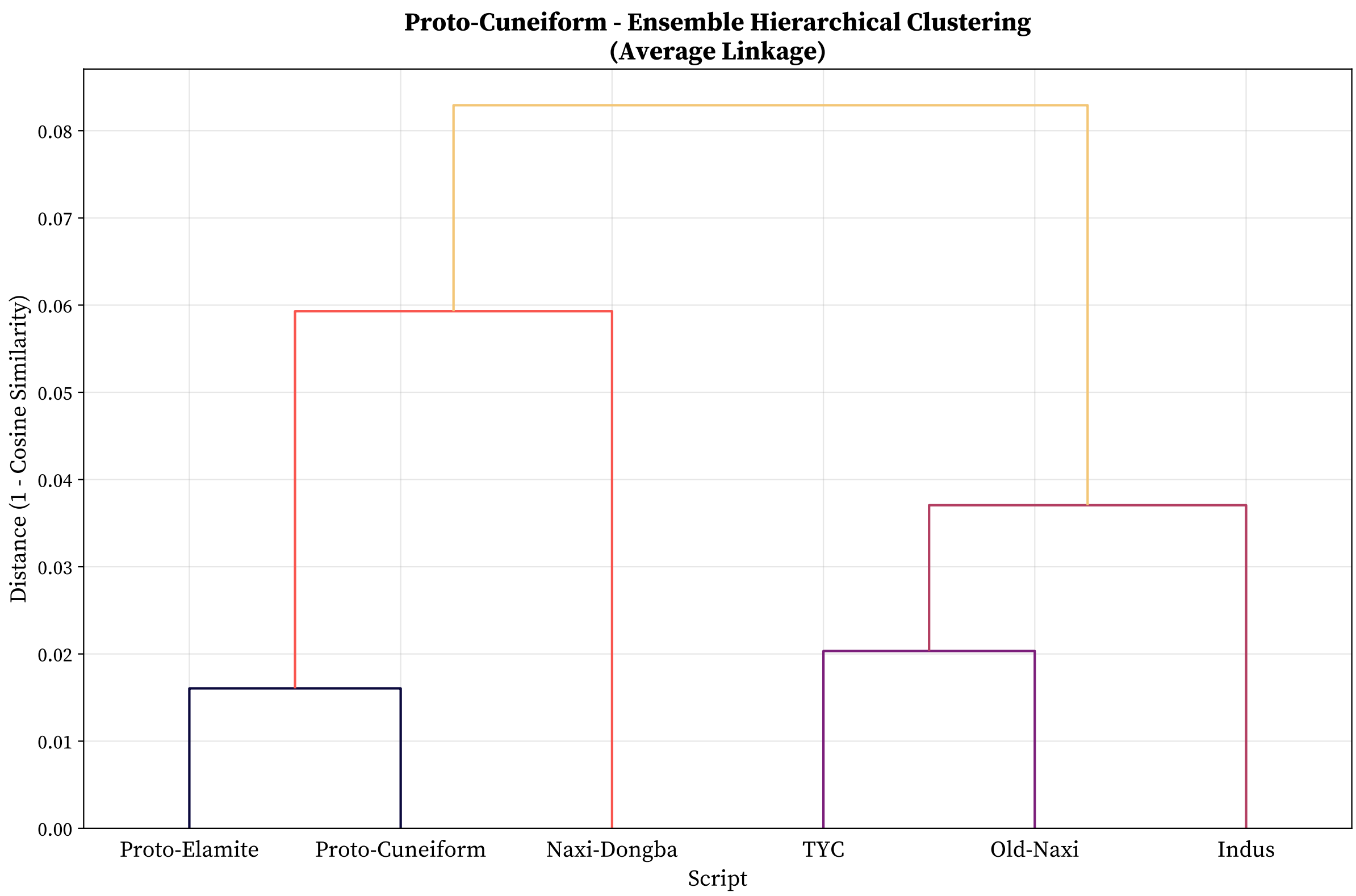

Proto-Cuneiform Ensemble Average Linkage Dendrogram

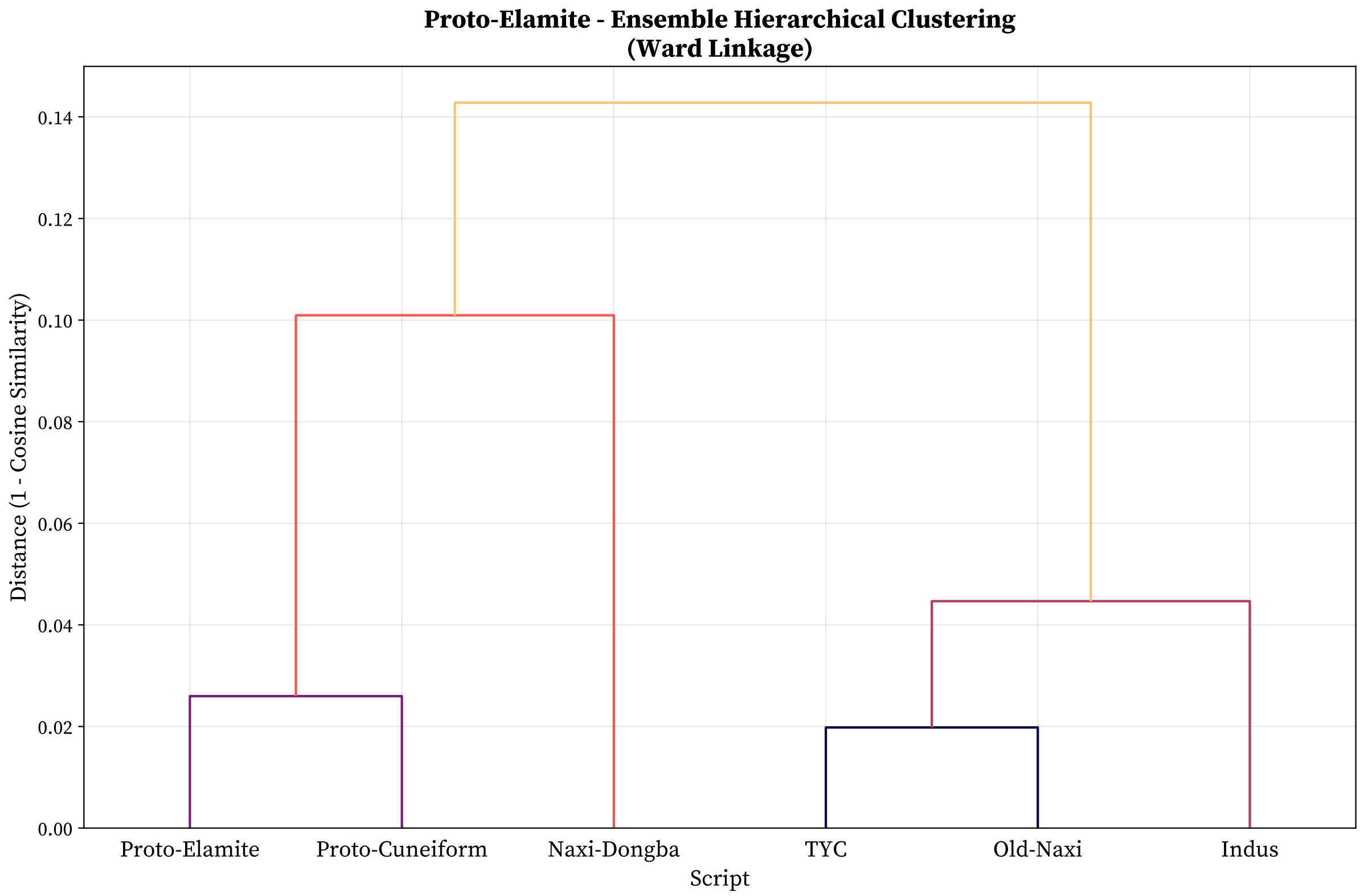

Proto-Elamite Ensemble Ward Linkage Dendrogram

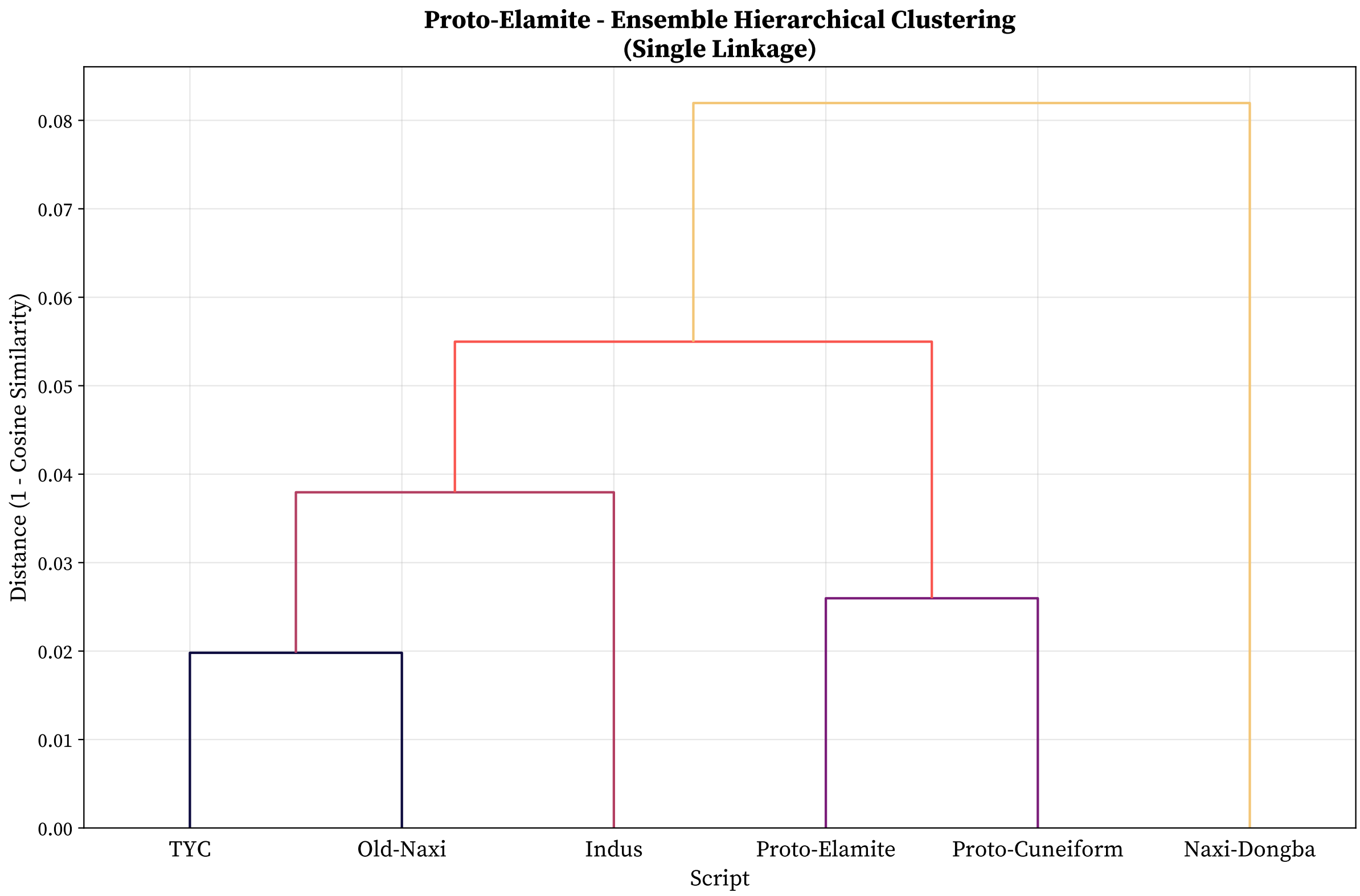

Proto-Elamite Ensemble Single Linkage Dendrogram

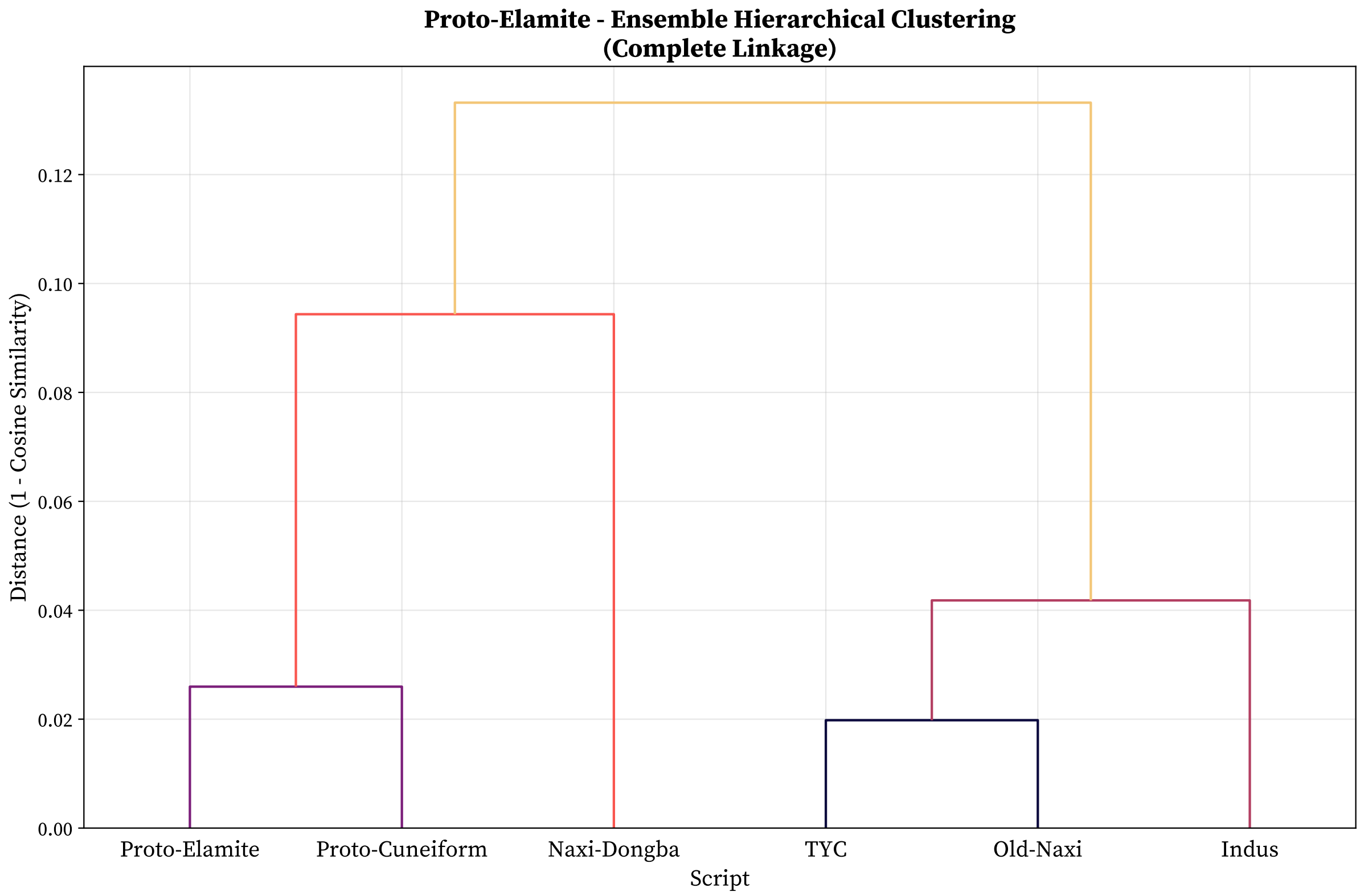

Proto-Elamite Ensemble Complete Linkage Dendrogram

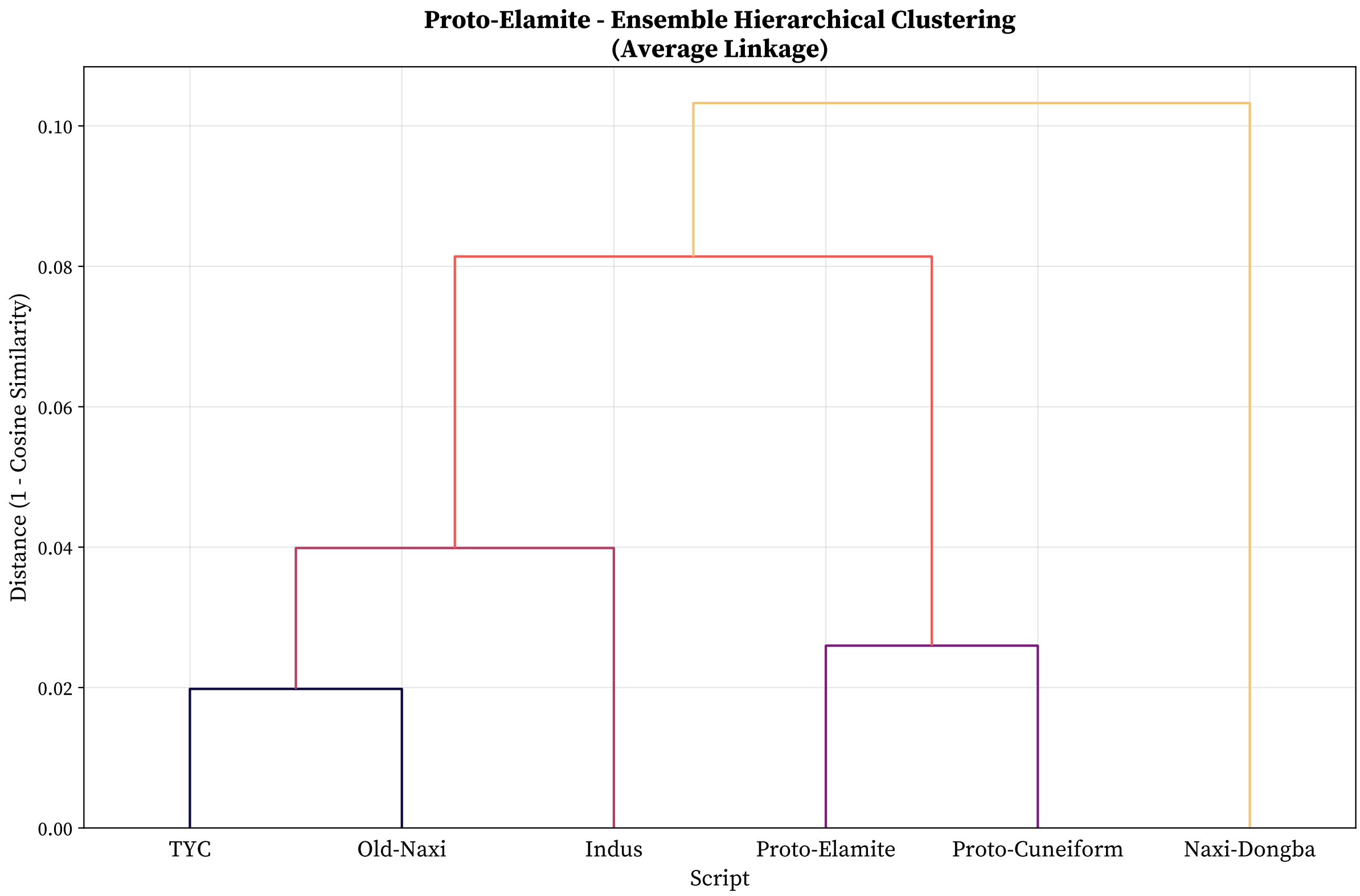

Proto-Elamite Ensemble Average Linkage Dendrogram

## A.6 Clustering Heatmaps

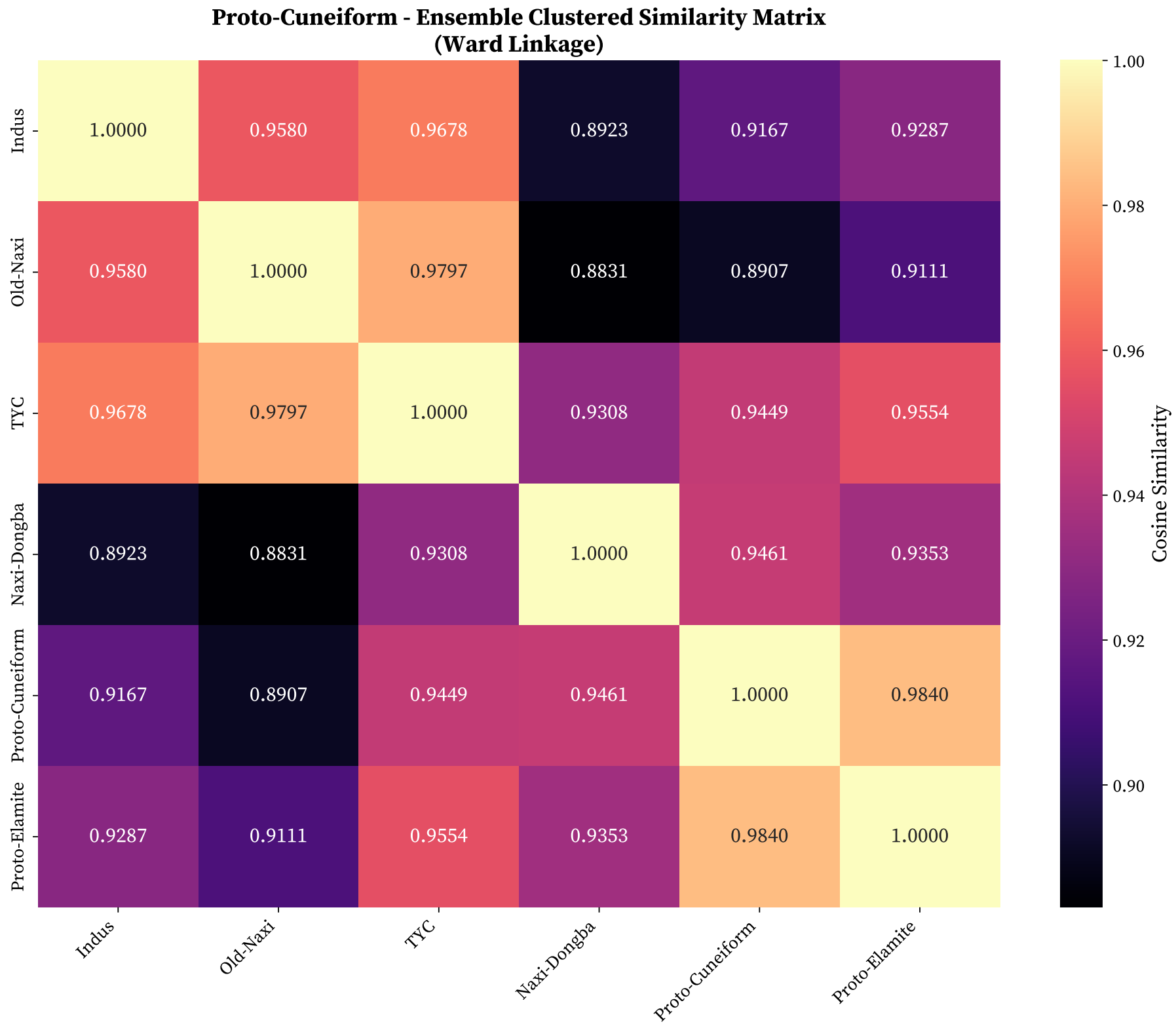

Proto-Cuneiform Ward Similarity Heatmap

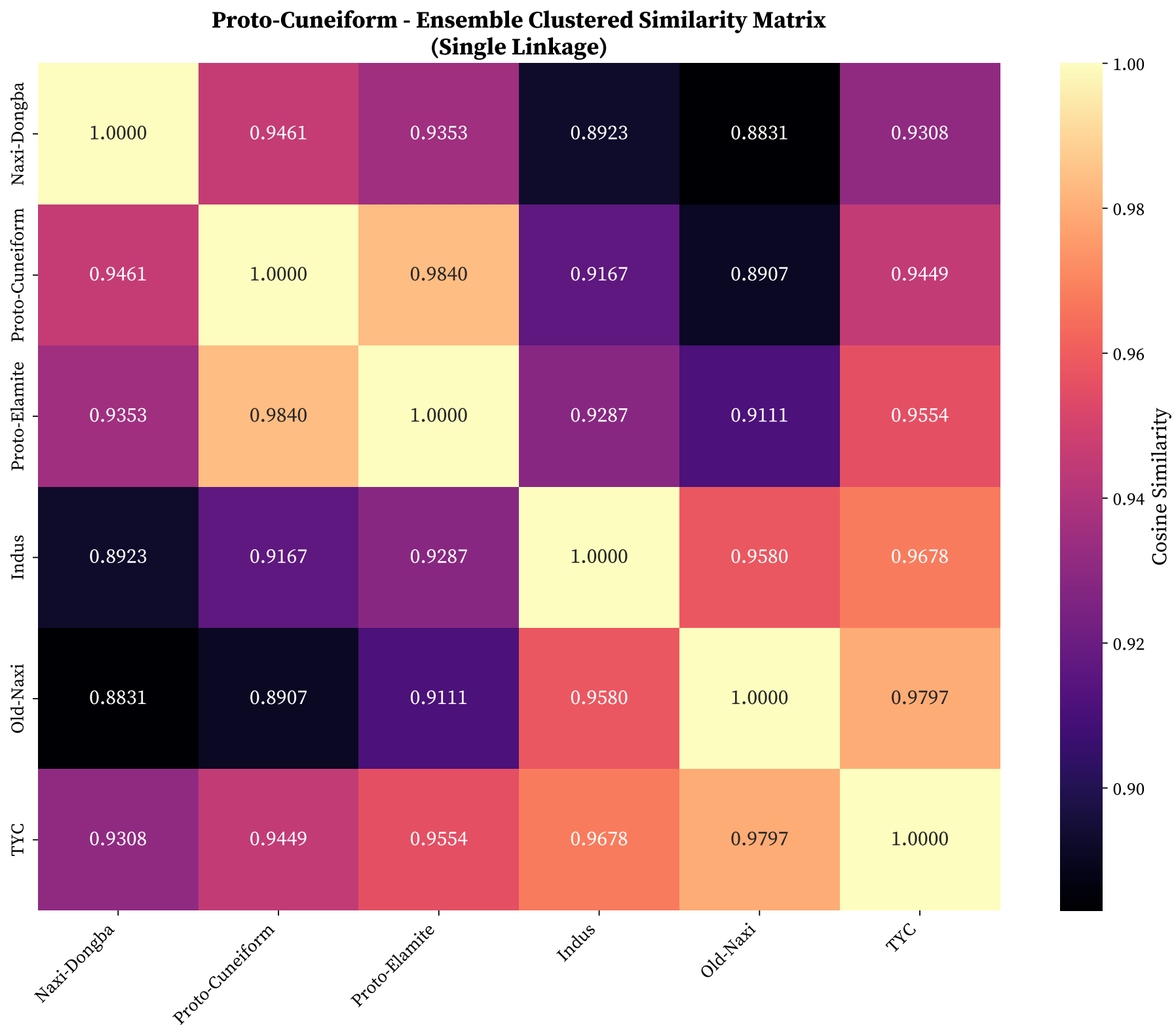

Proto-Cuneiform Single Similarity Heatmap

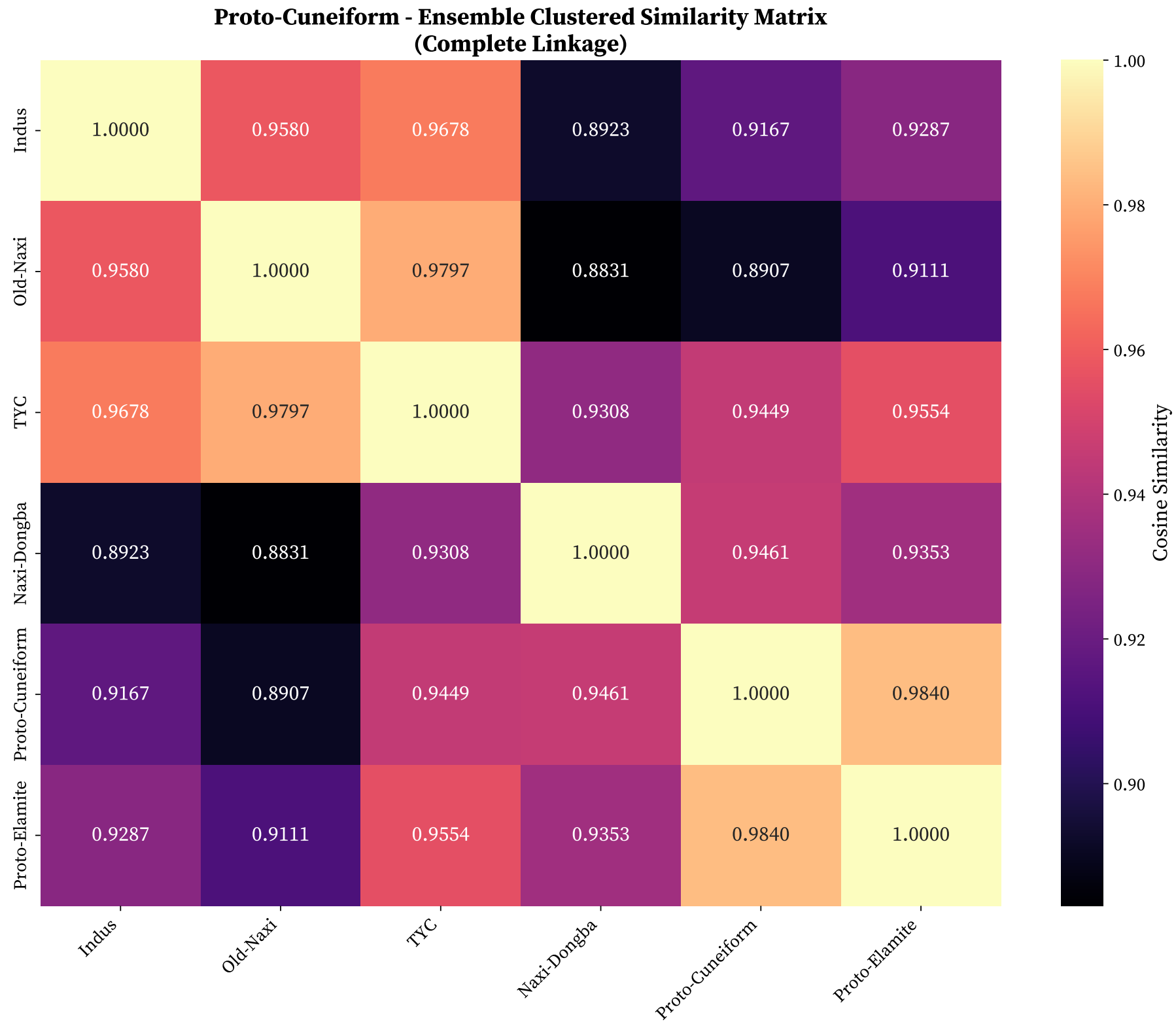

Proto-Cuneiform Complete Similarity Heatmap

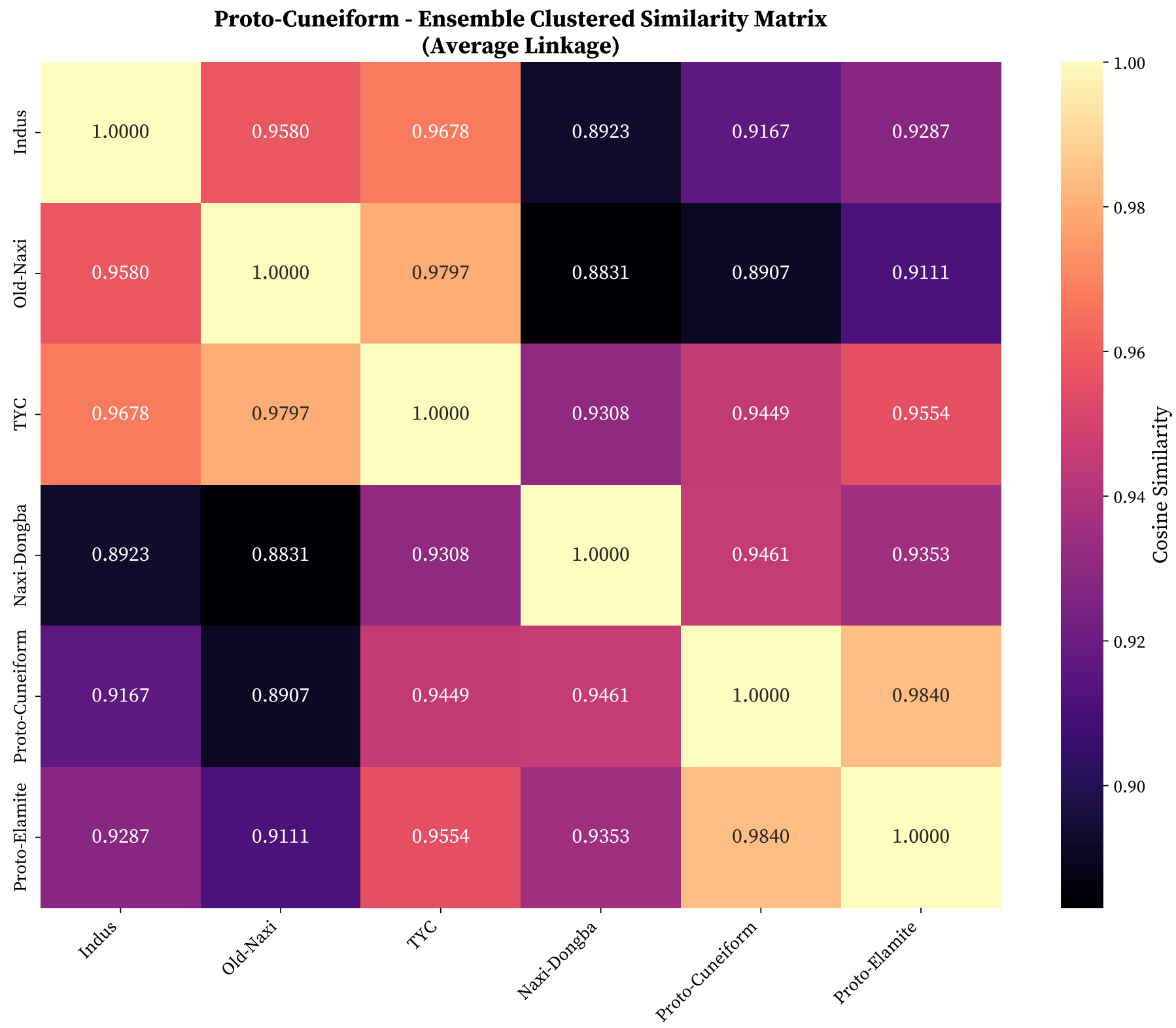

Proto-Cuneiform Average Similarity Heatmap

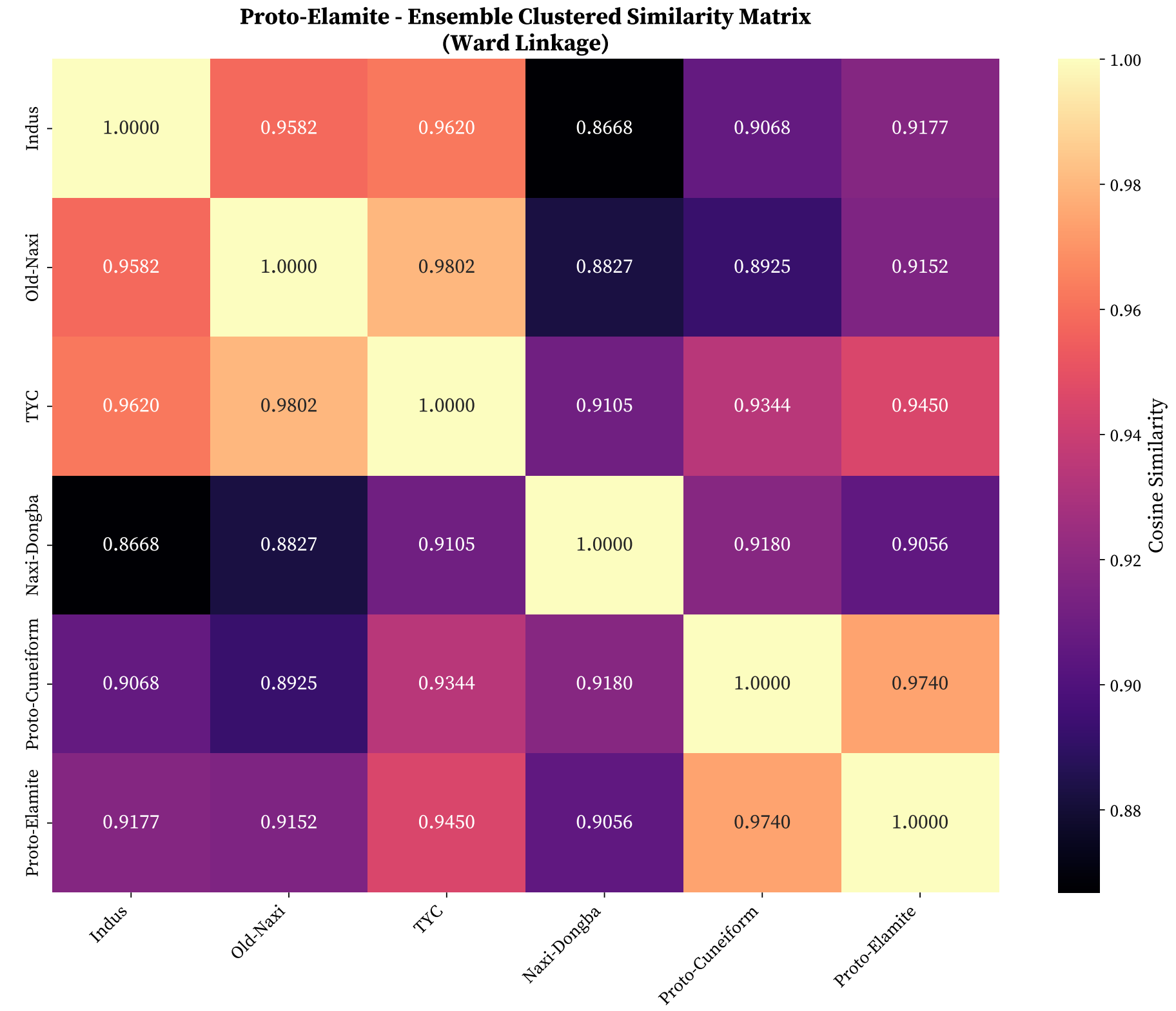

Proto-Elamite Ward Similarity Heatmap

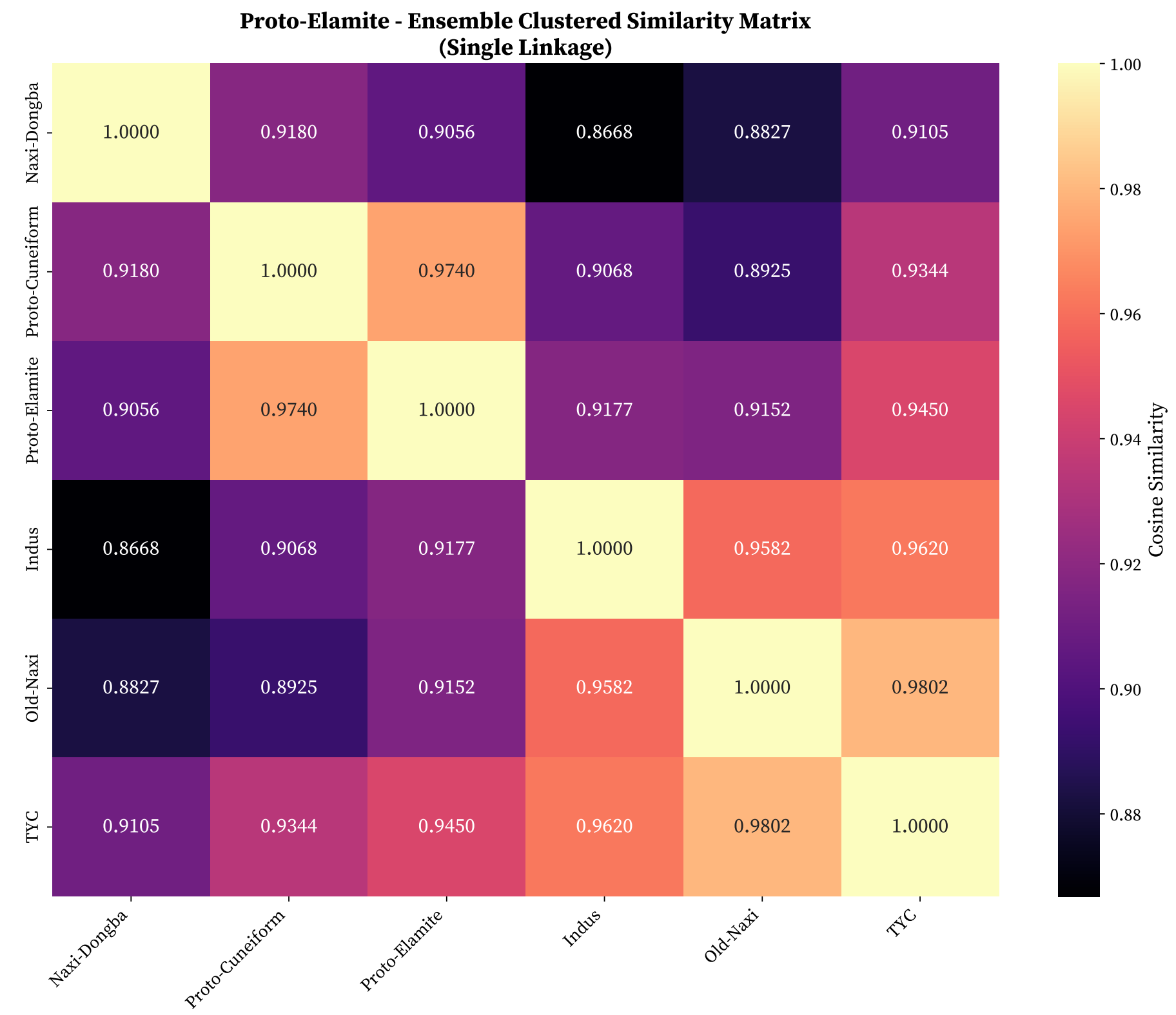

Proto-Elamite Single Similarity Heatmap

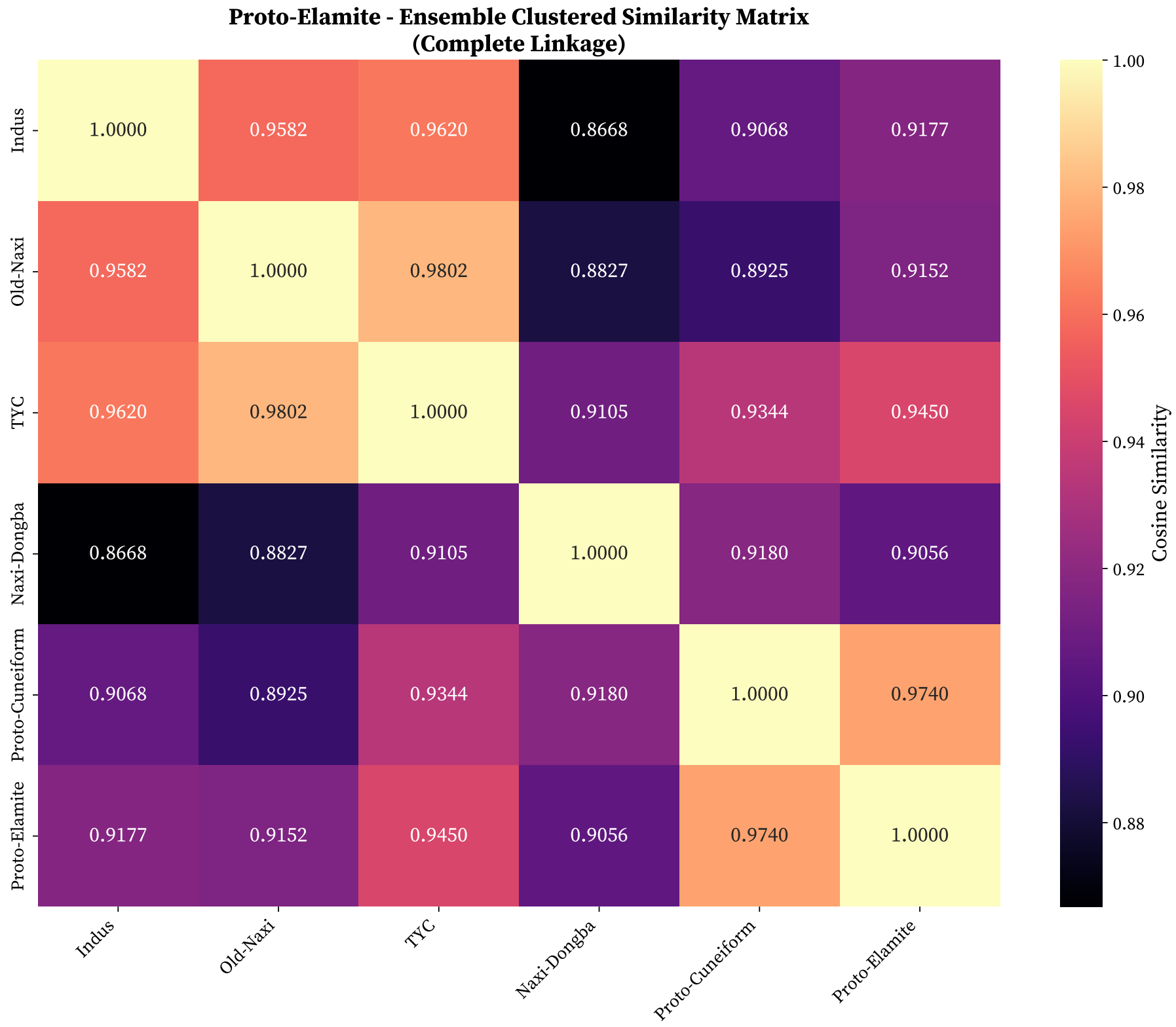

Proto-Elamite Complete Similarity Heatmap

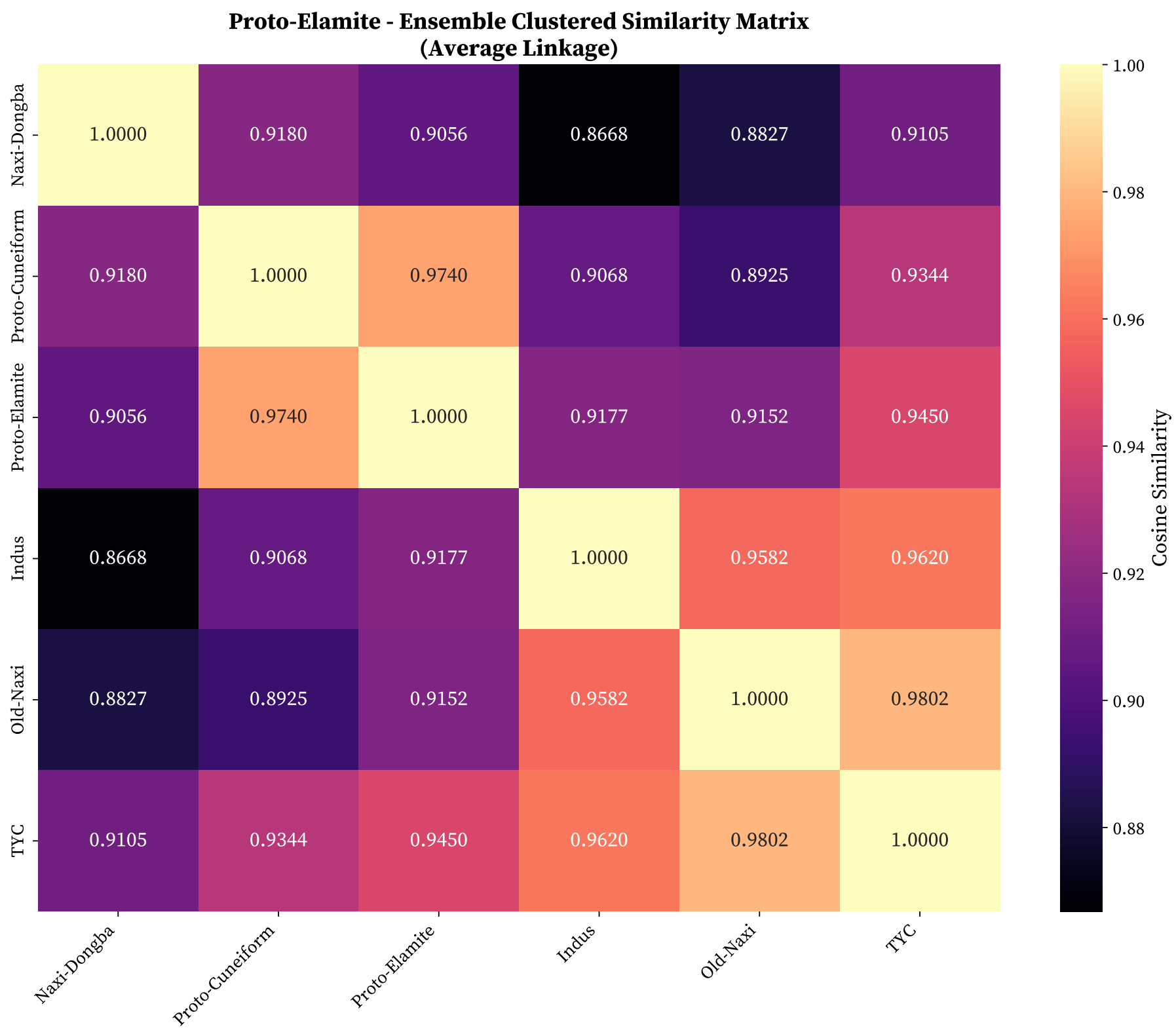

Proto-Elamite Average Similarity Heatmap

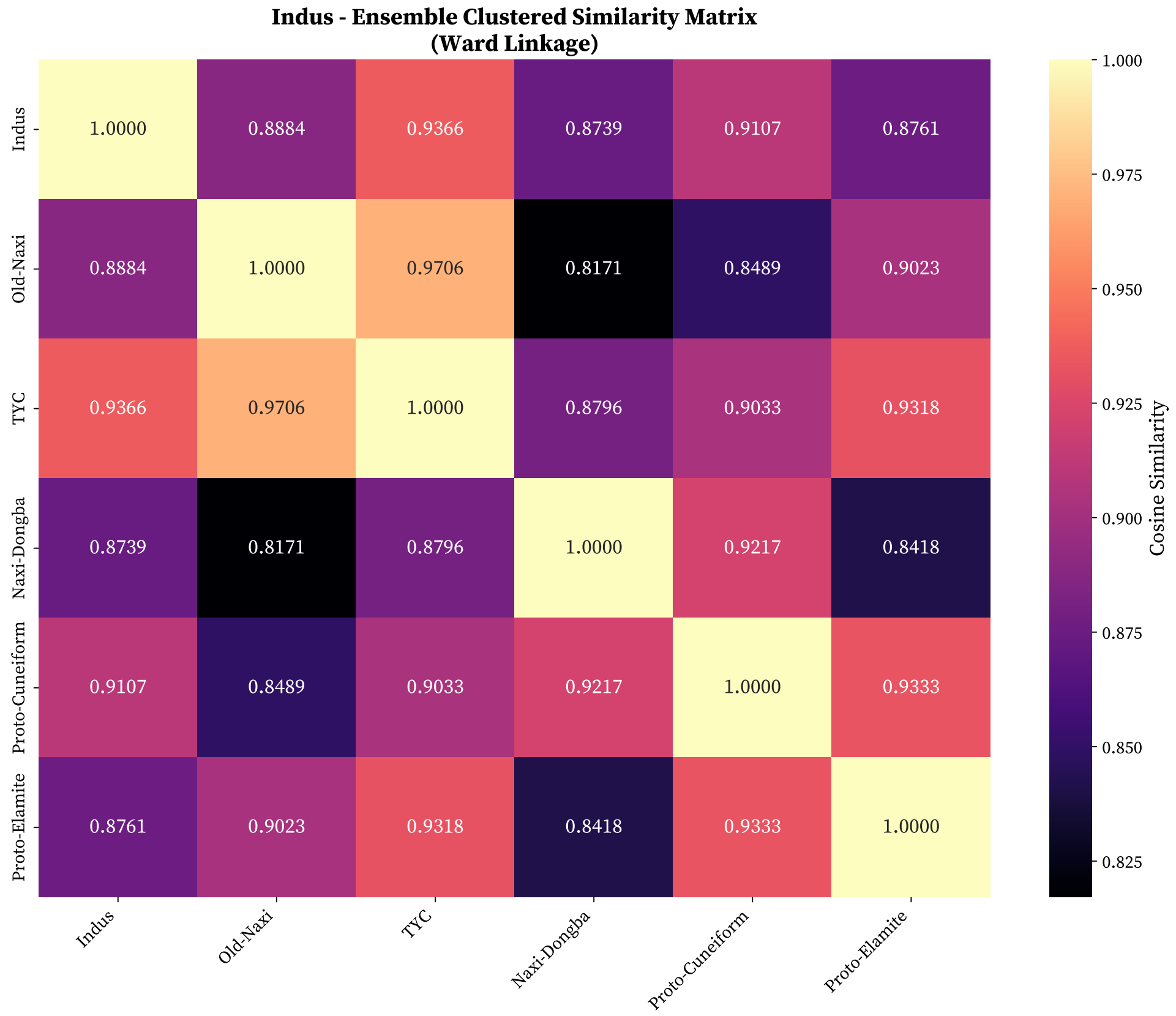

Indus Ward Similarity Heatmap

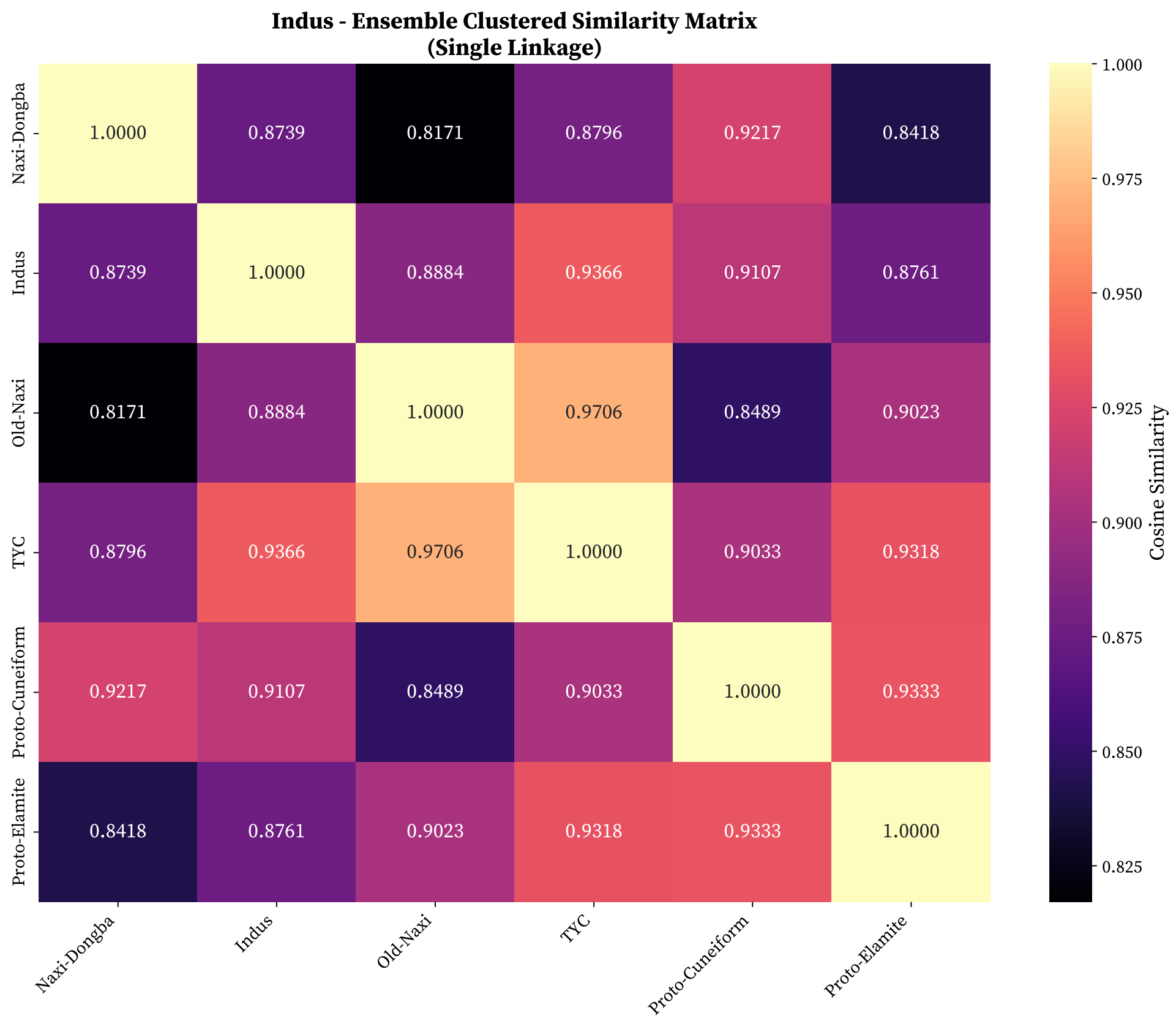

Indus Single Similarity Heatmap

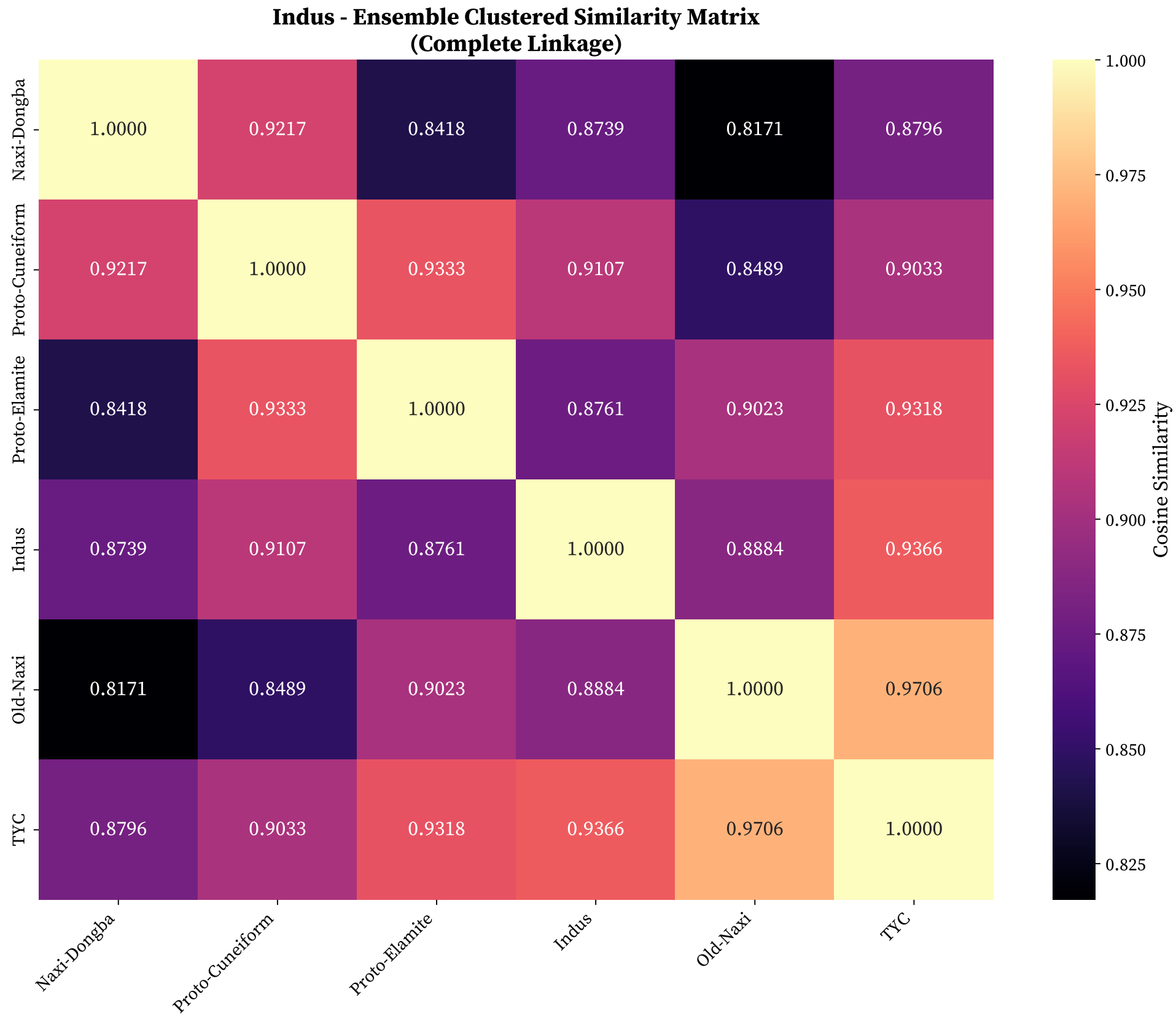

Indus Complete Similarity Heatmap

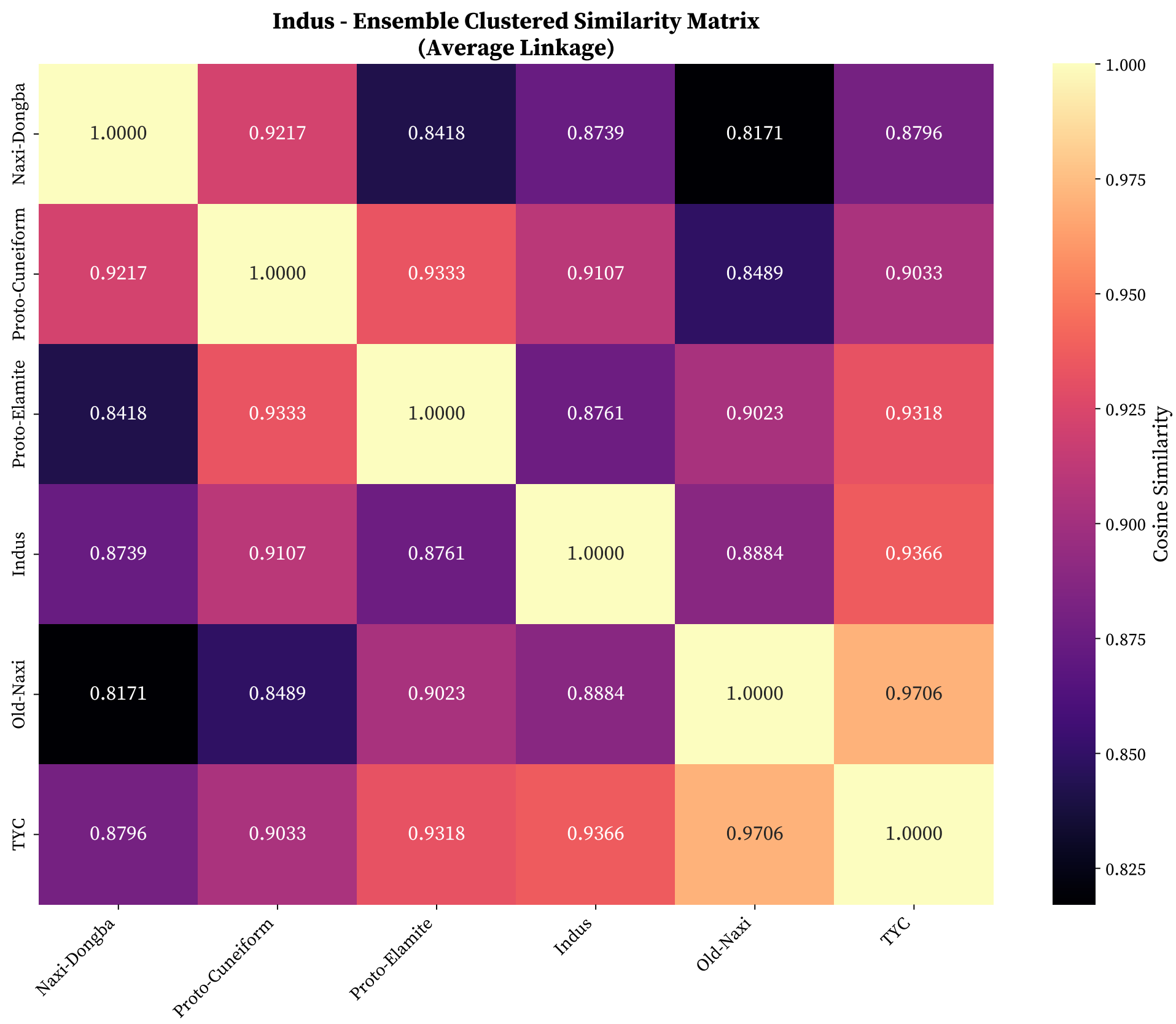

Indus Average Similarity Heatmap

## A.7 Cosine Similarity Plots

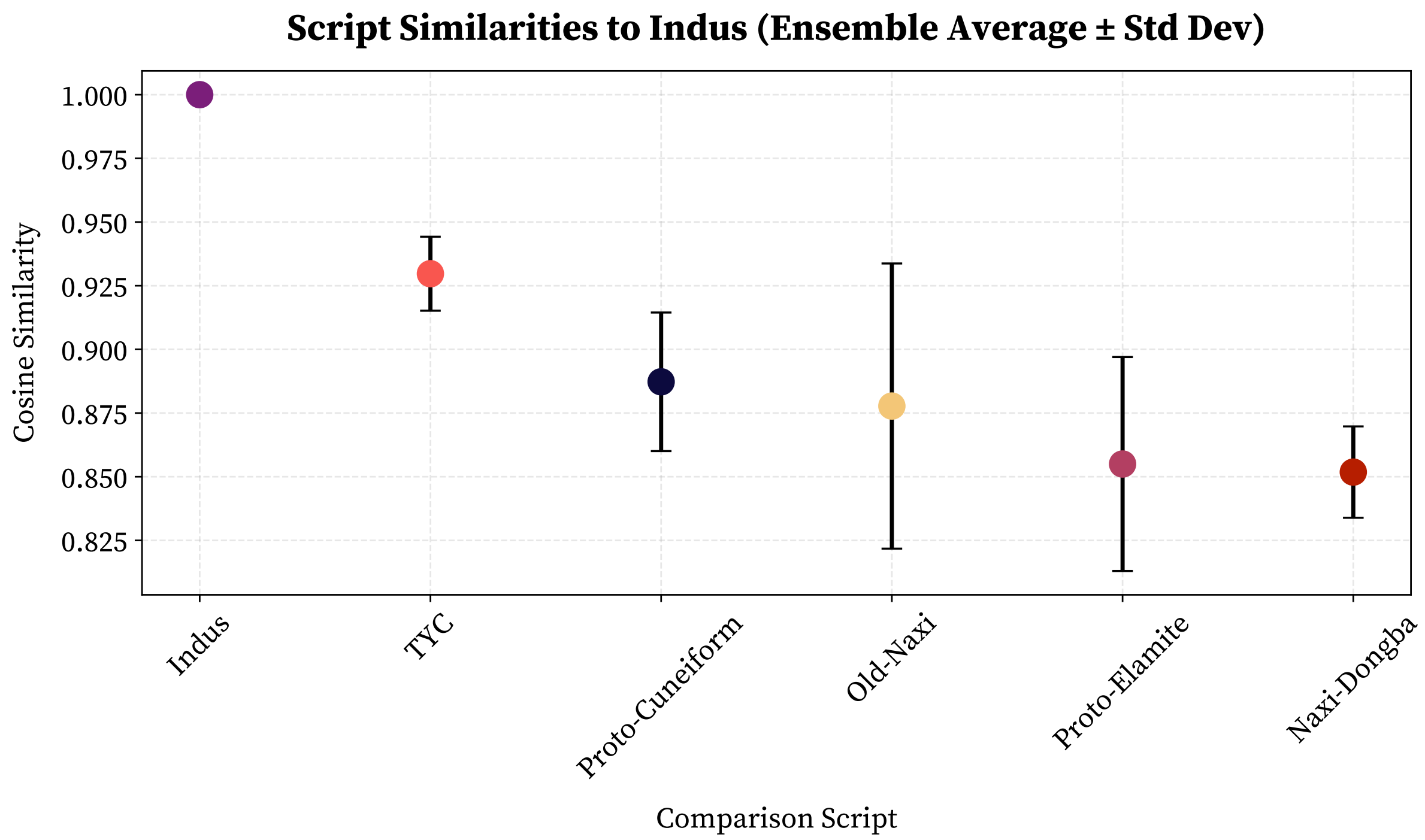

Figure A.1: Indus Ensemble Cosine Similarity Plot

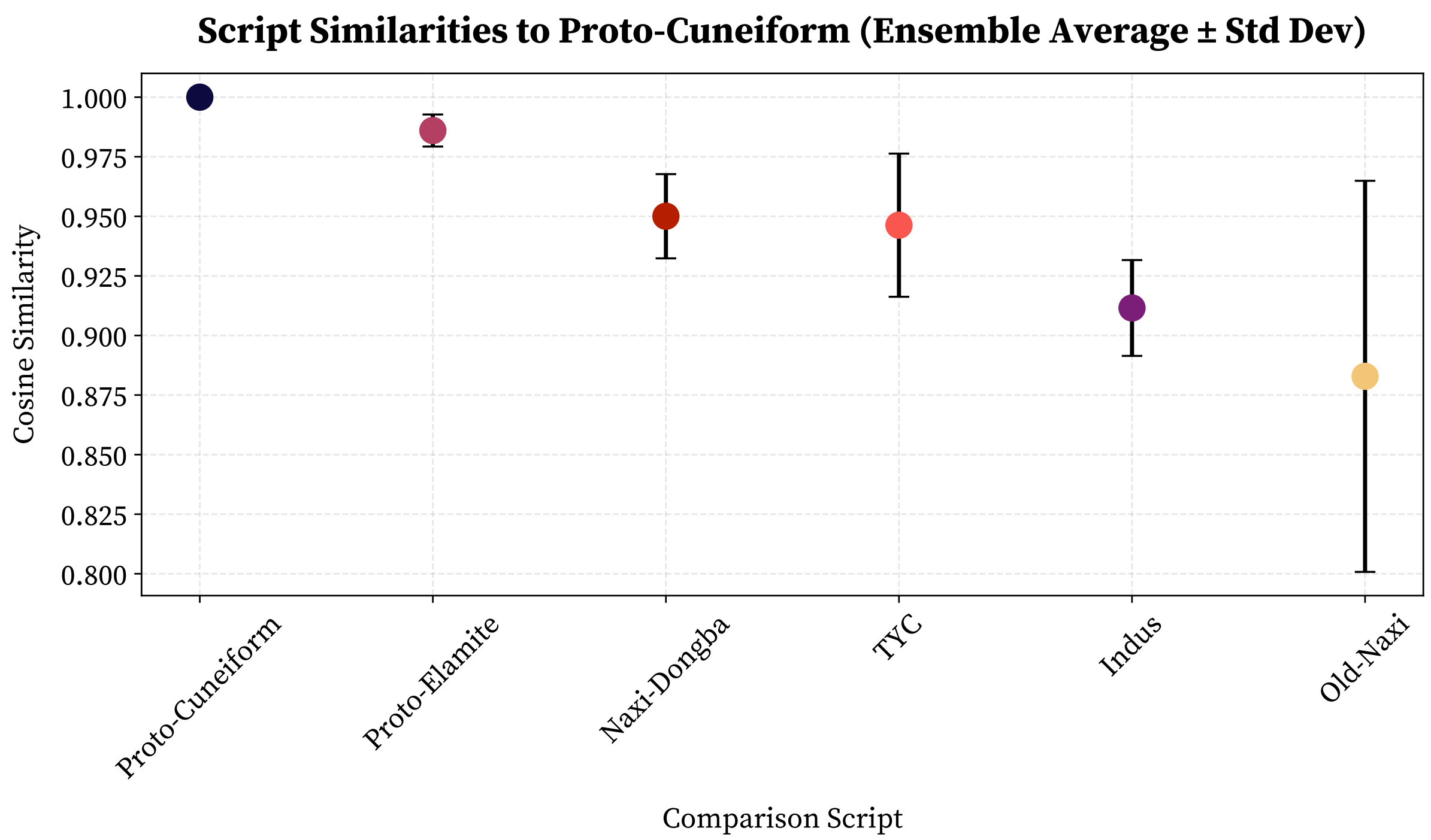

Figure A.2: Proto-Cuneiform Ensemble Cosine Similarity Plot

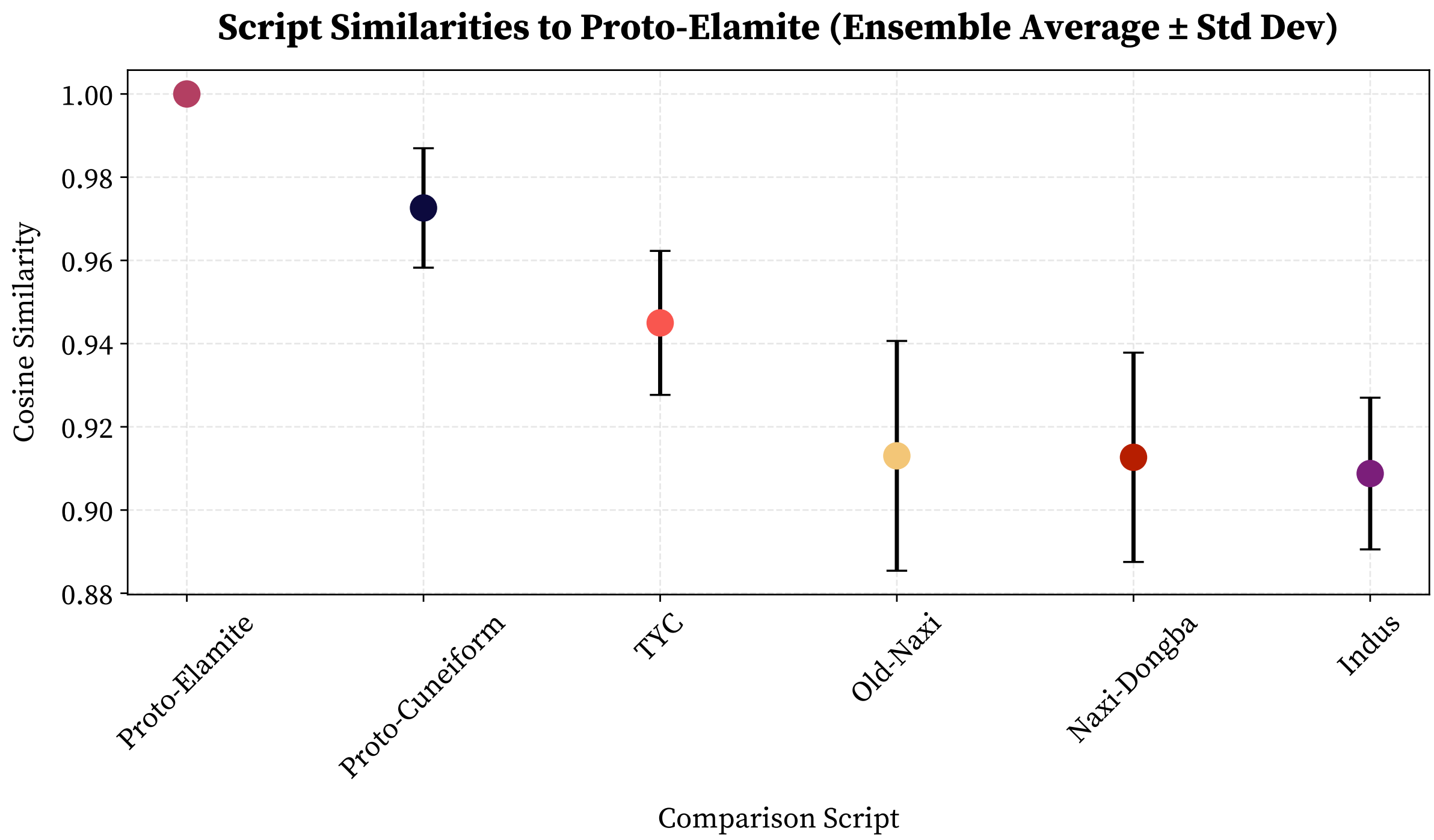

Figure A.3: Proto-Elamite Ensemble Cosine Similarity Plot

## A.8 Grad-CAM Heatmaps

### A.8.1 Indus Script

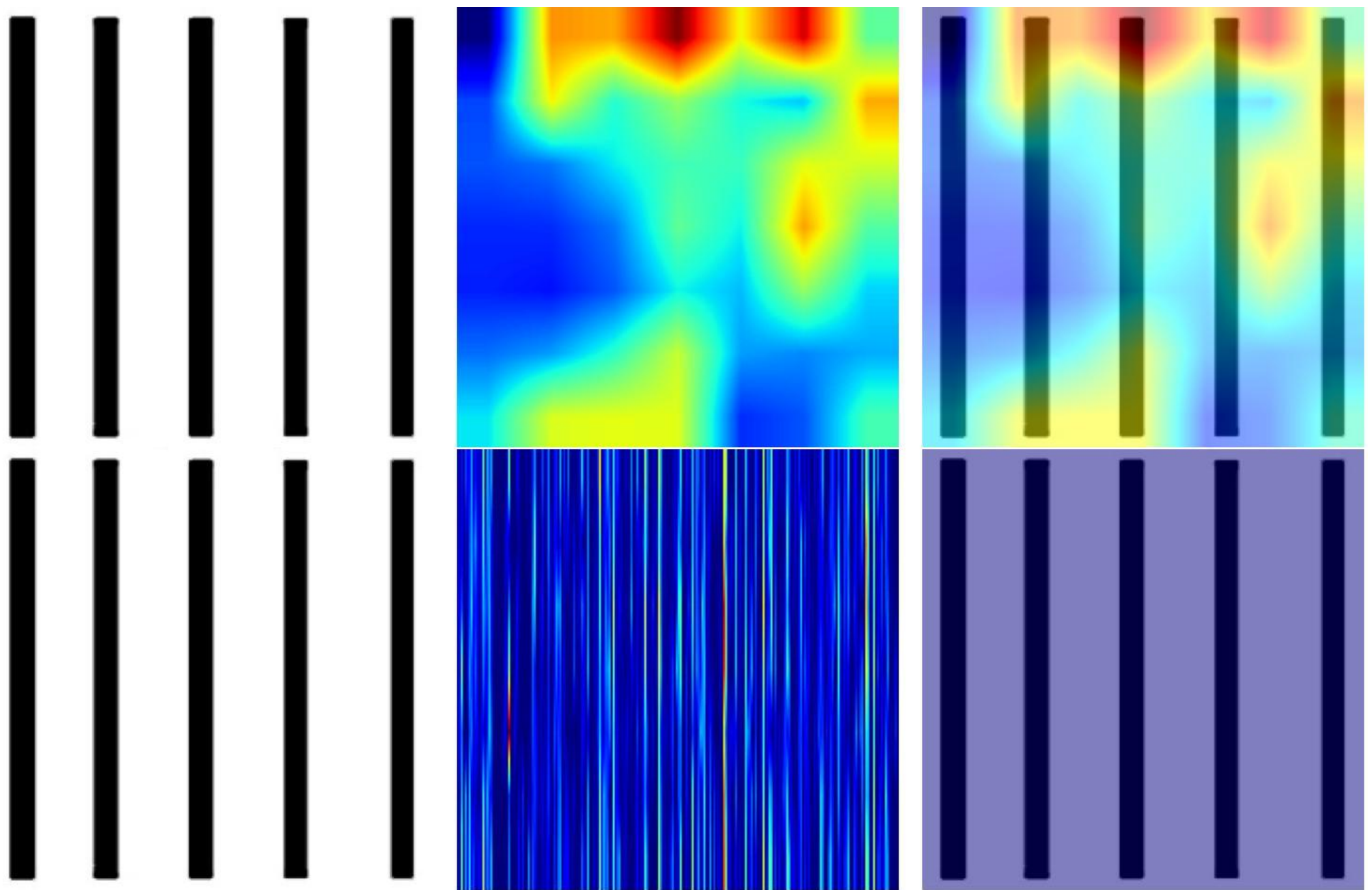

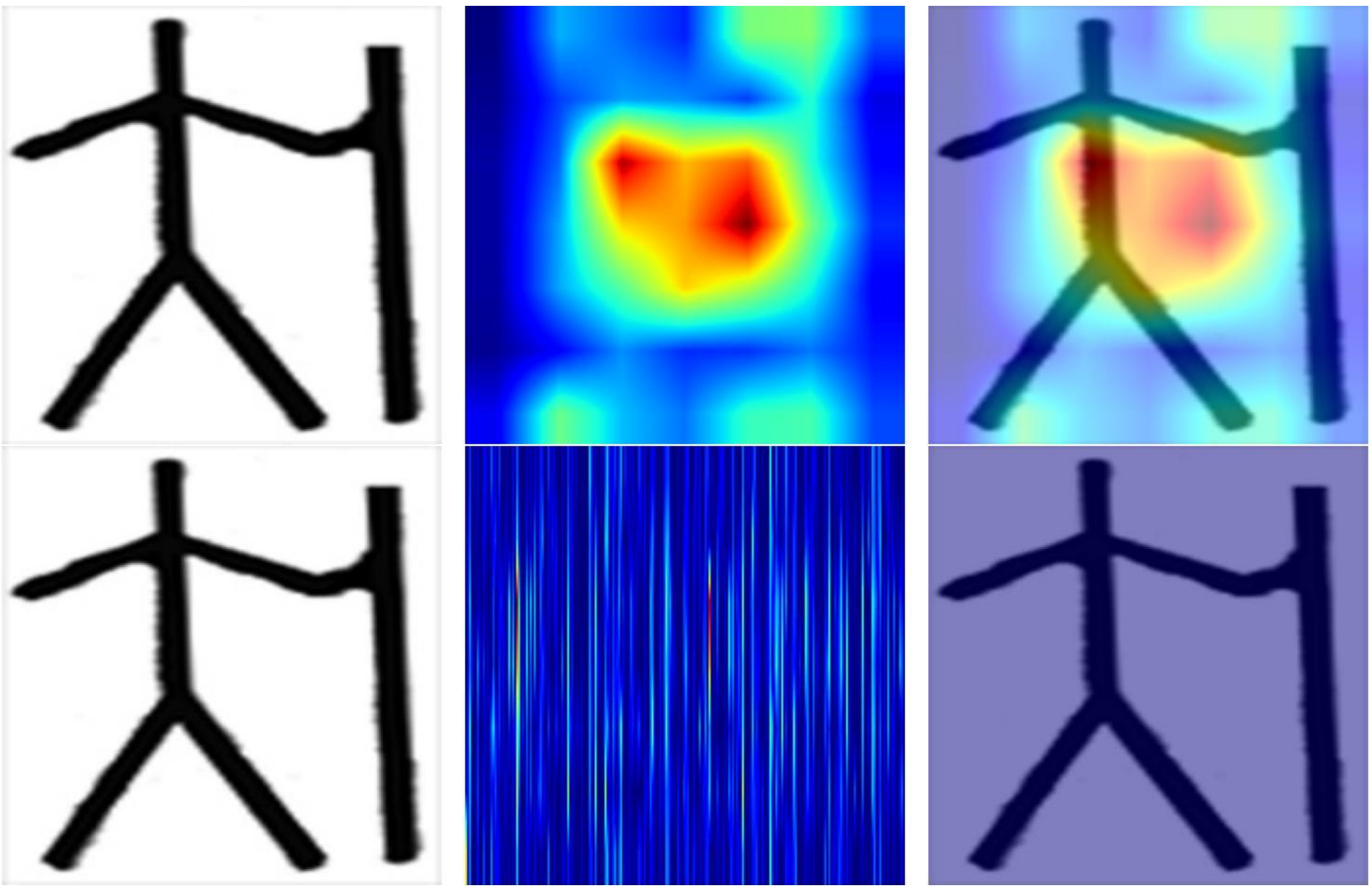

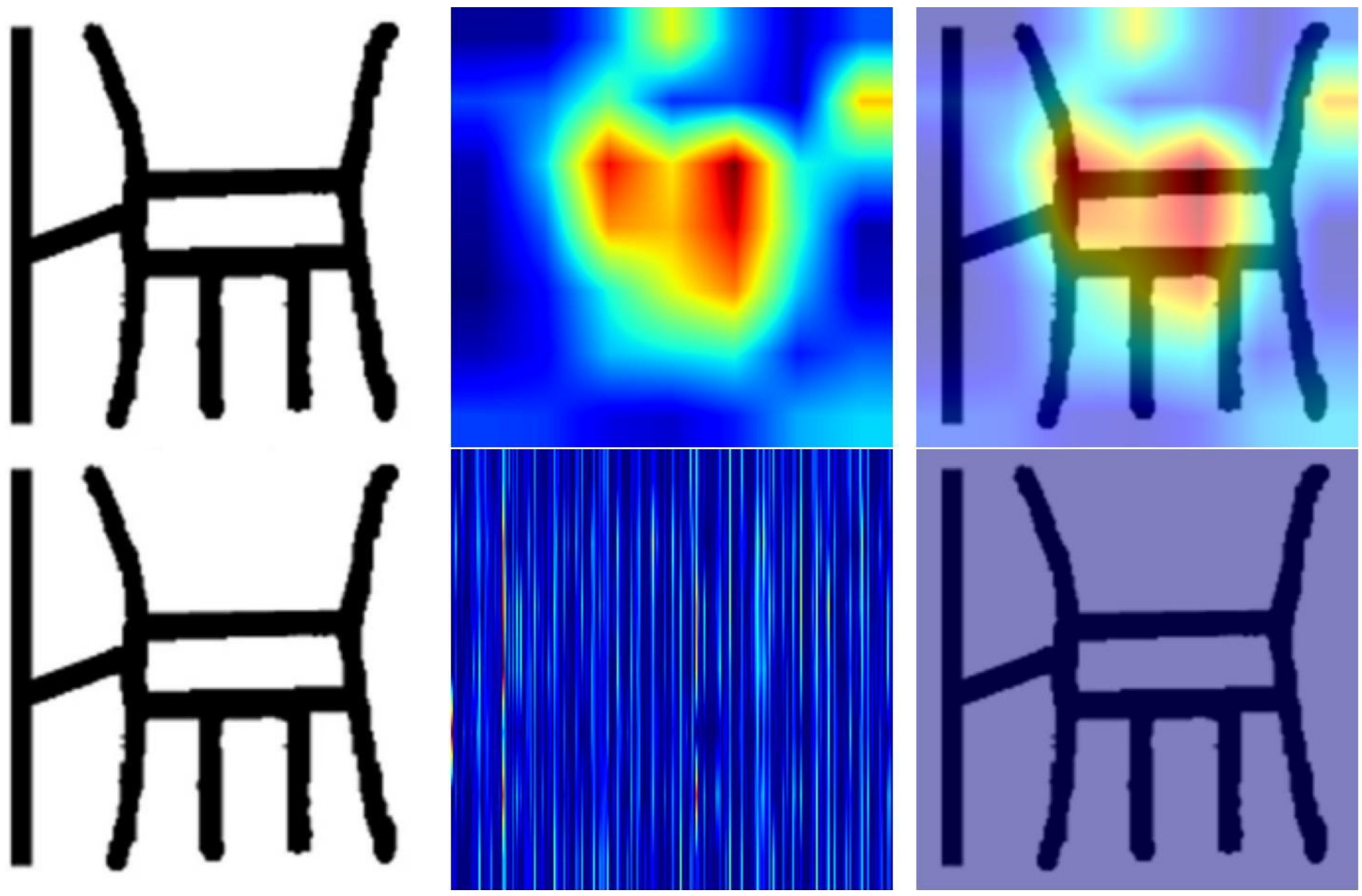

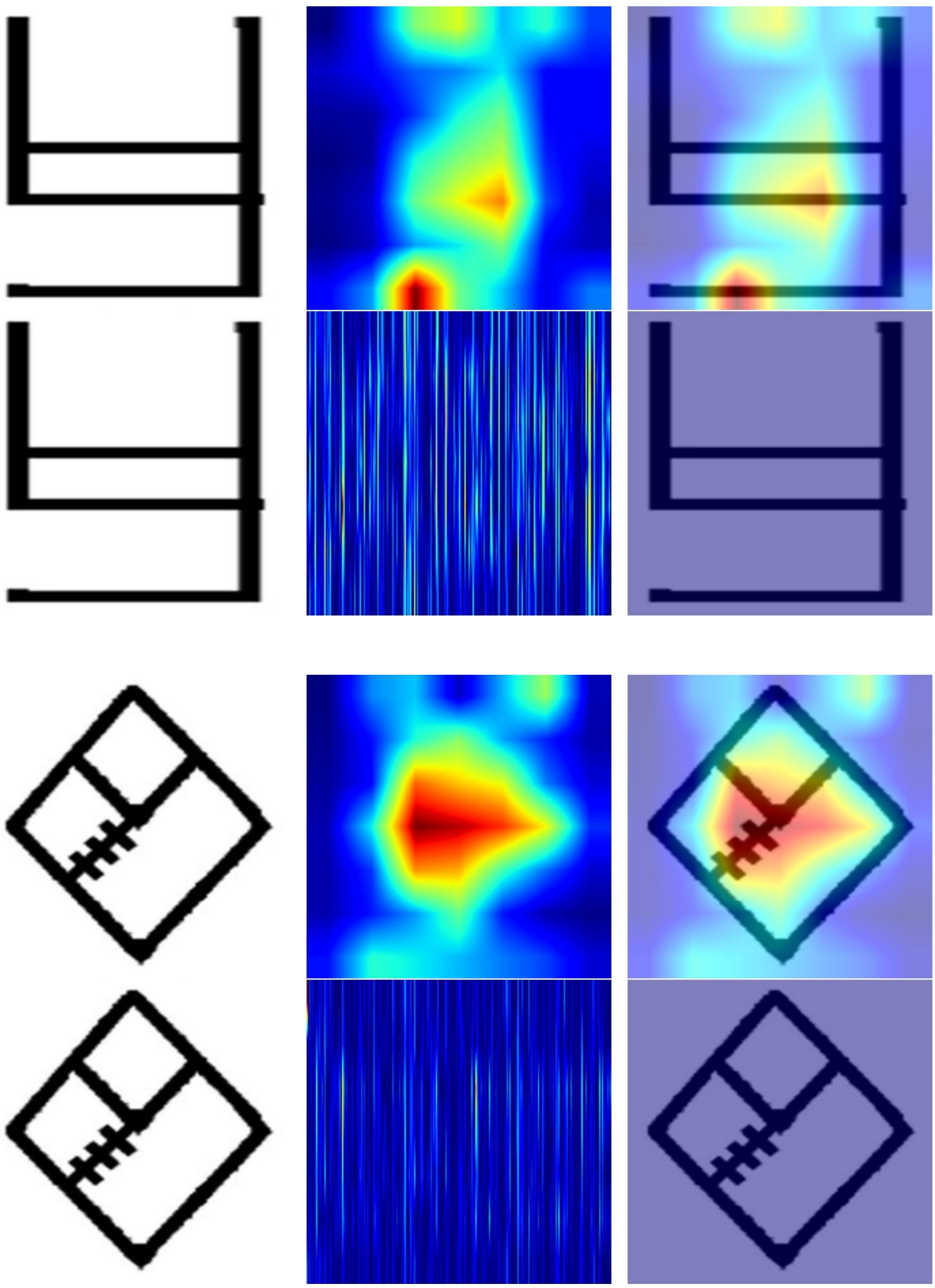

### A.8.2 Proto-Cuneiform

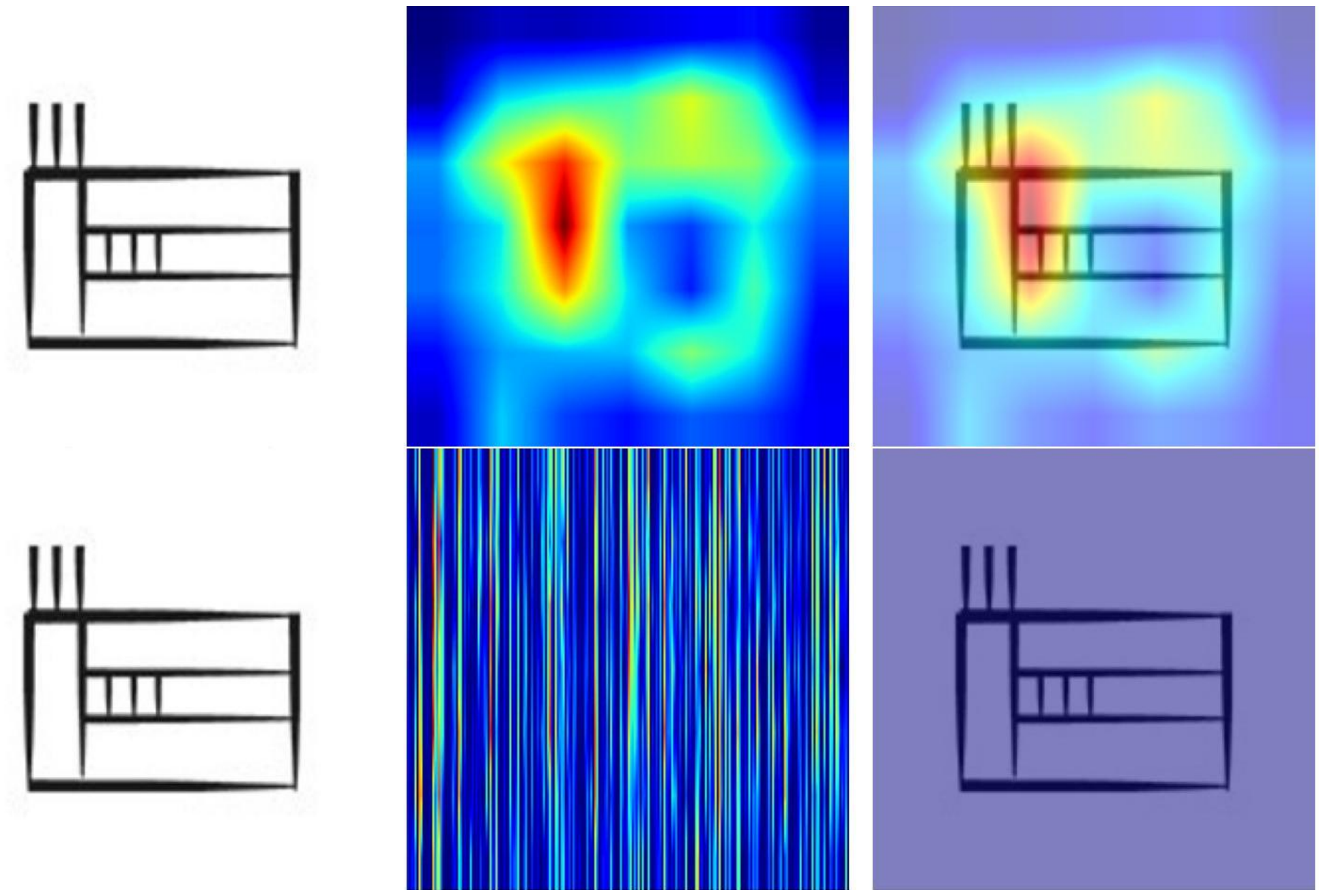

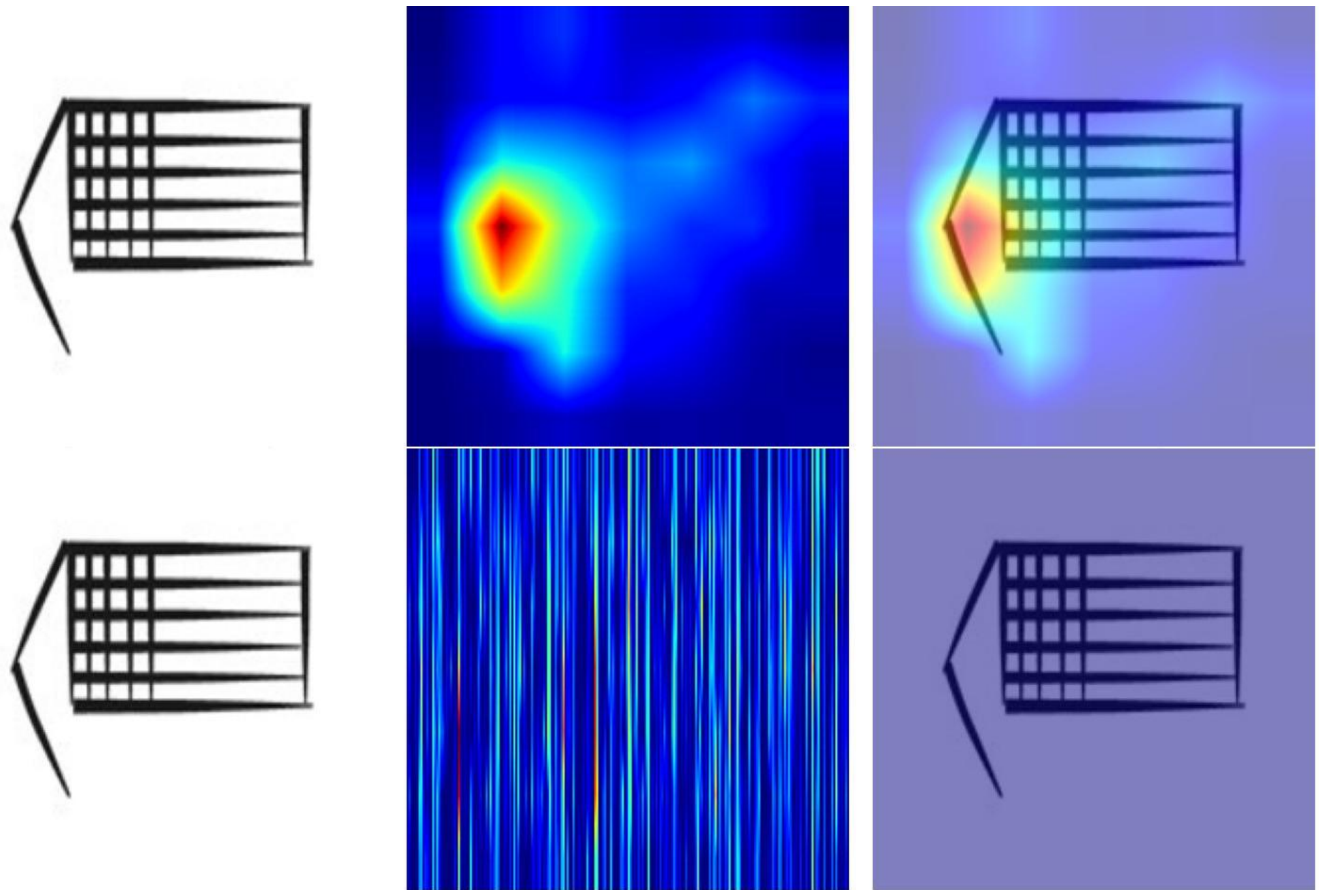

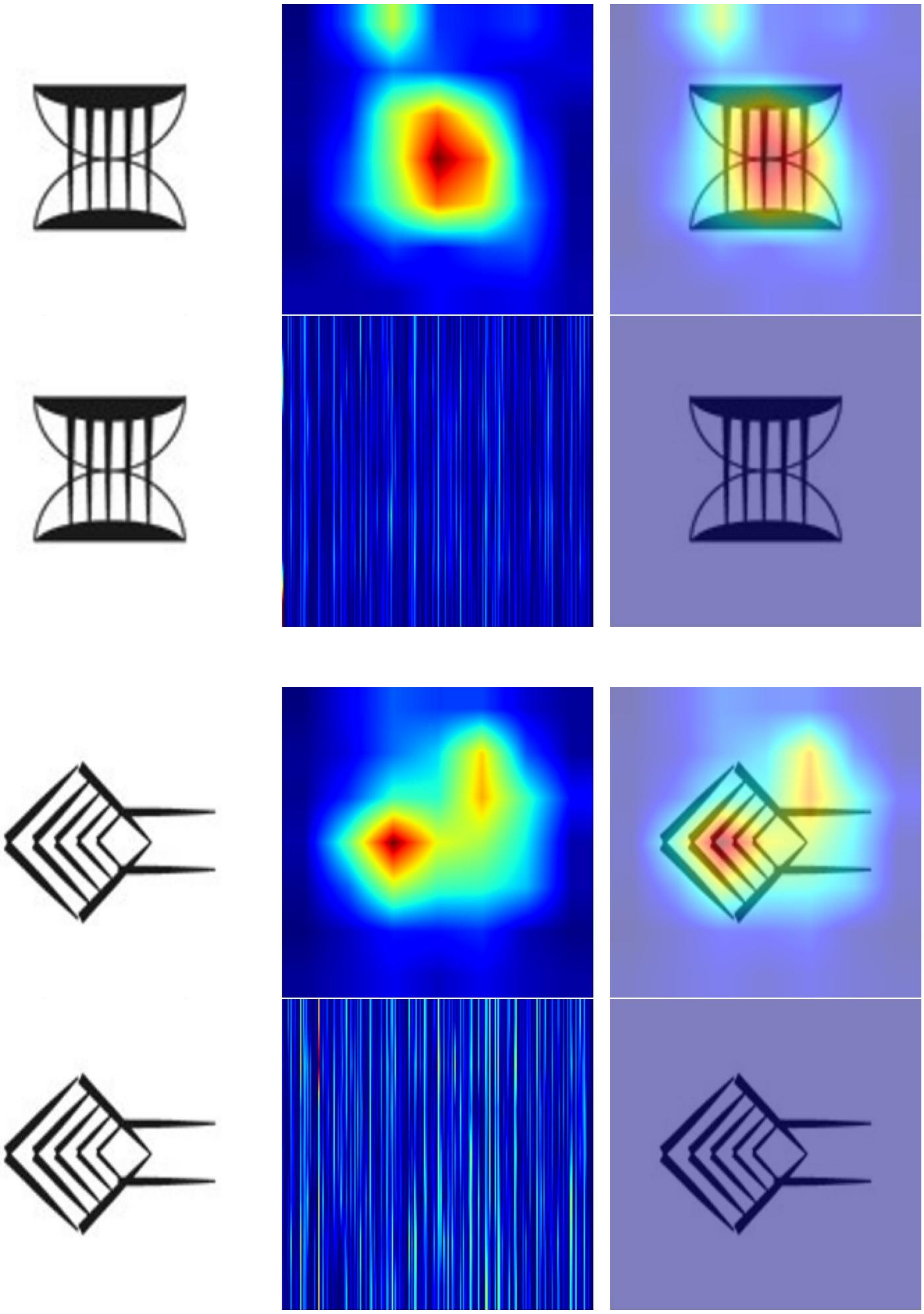

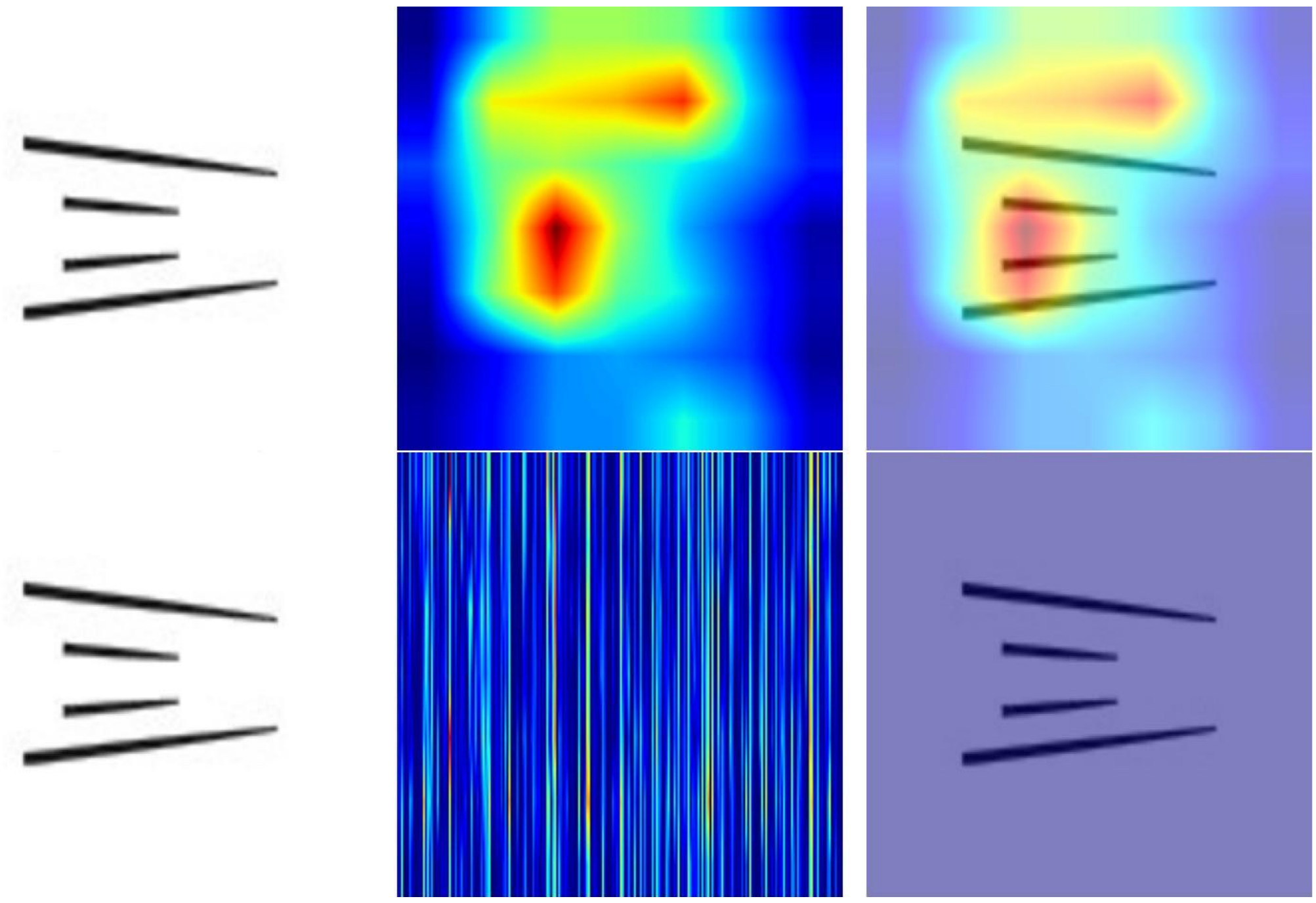

### A.8.3 Proto-Elamite

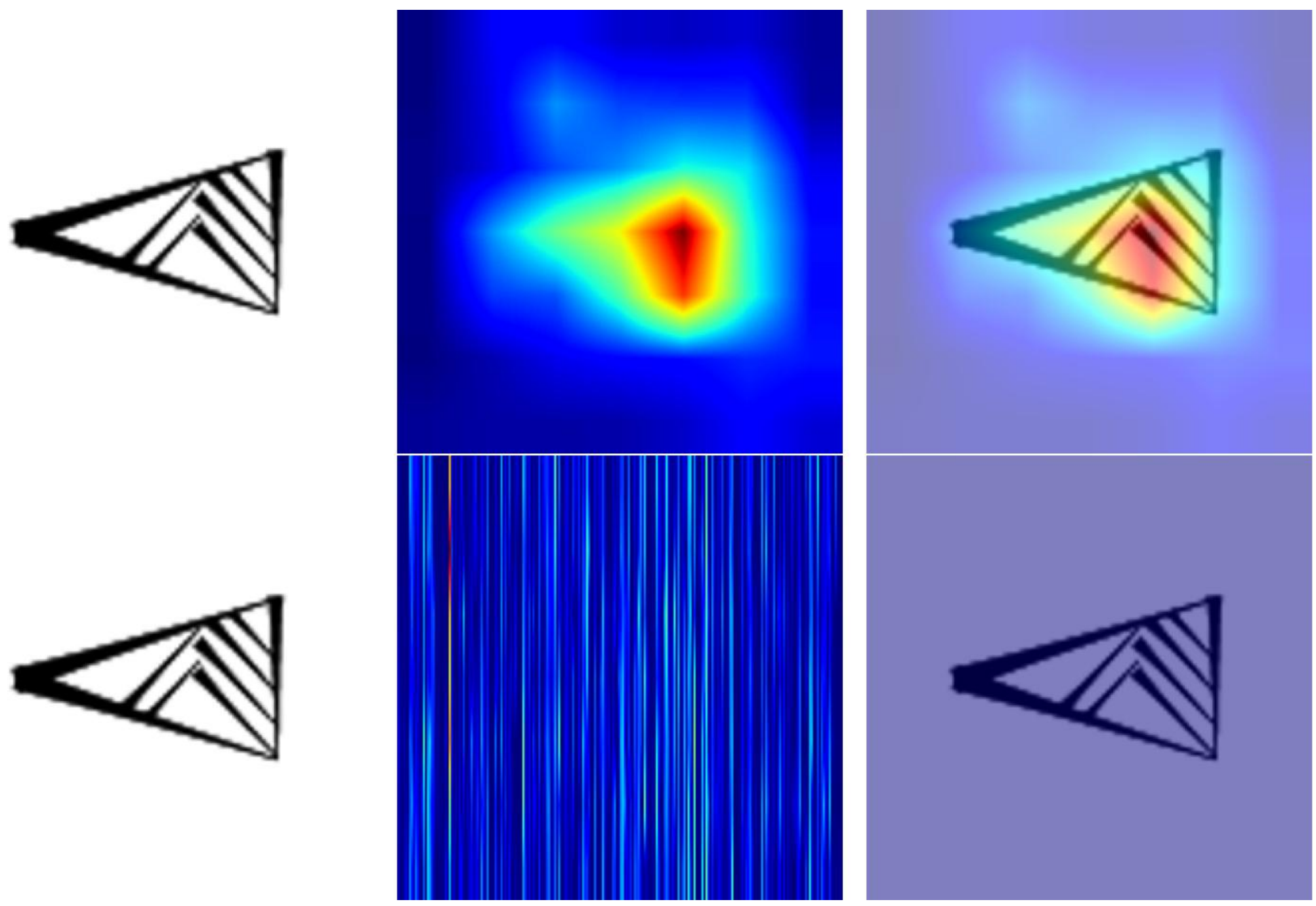

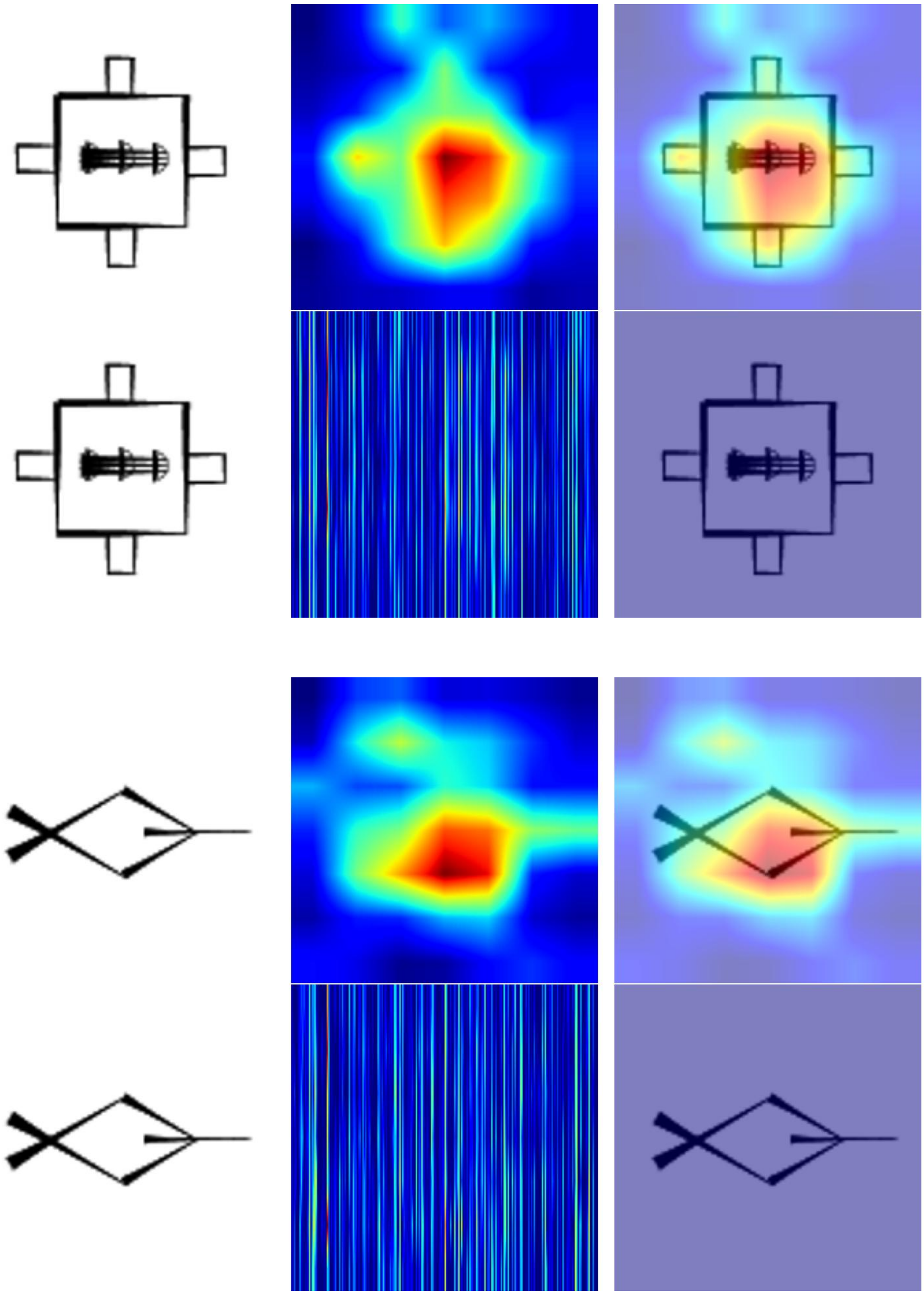

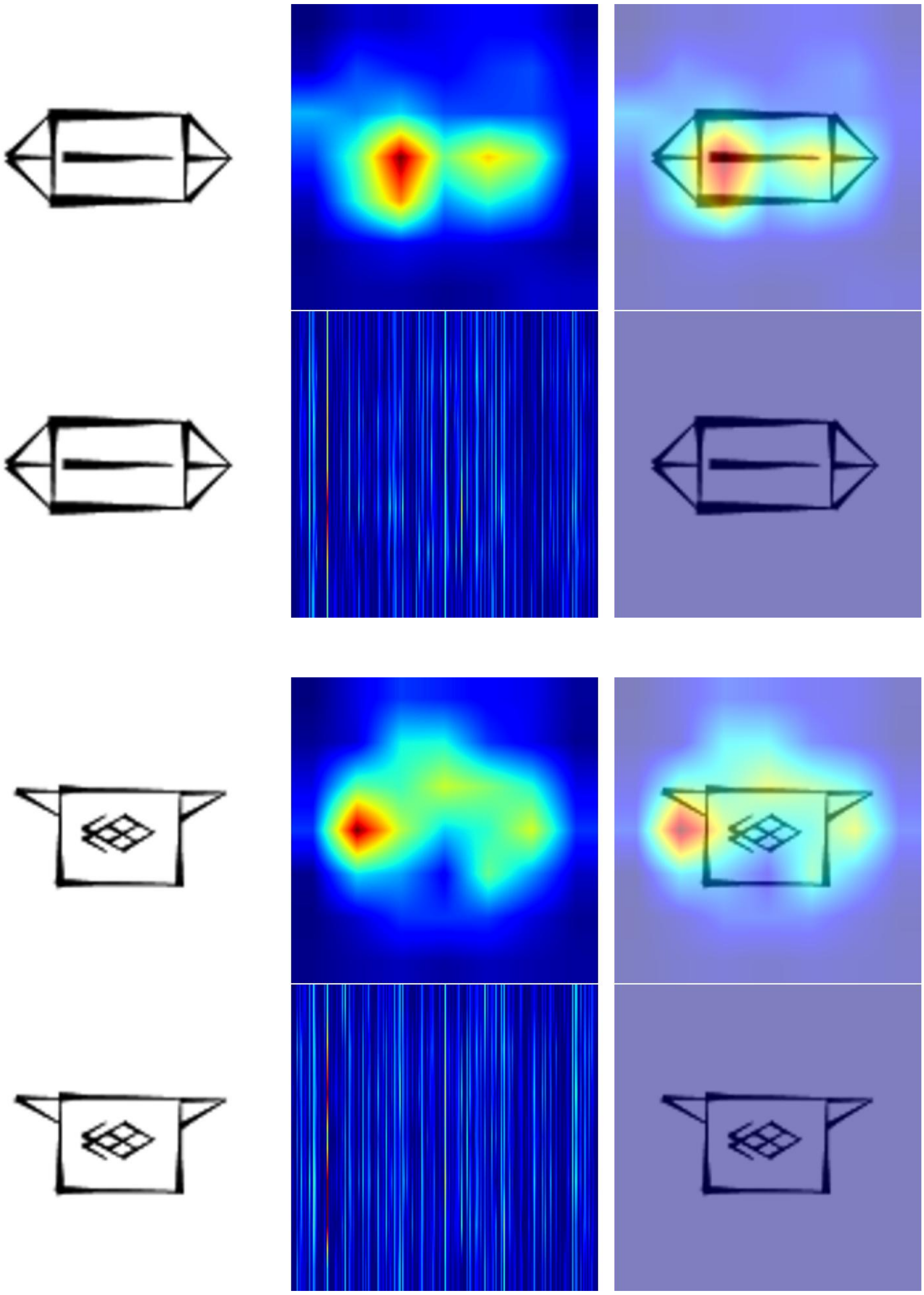

### A.8.4 New Dongba

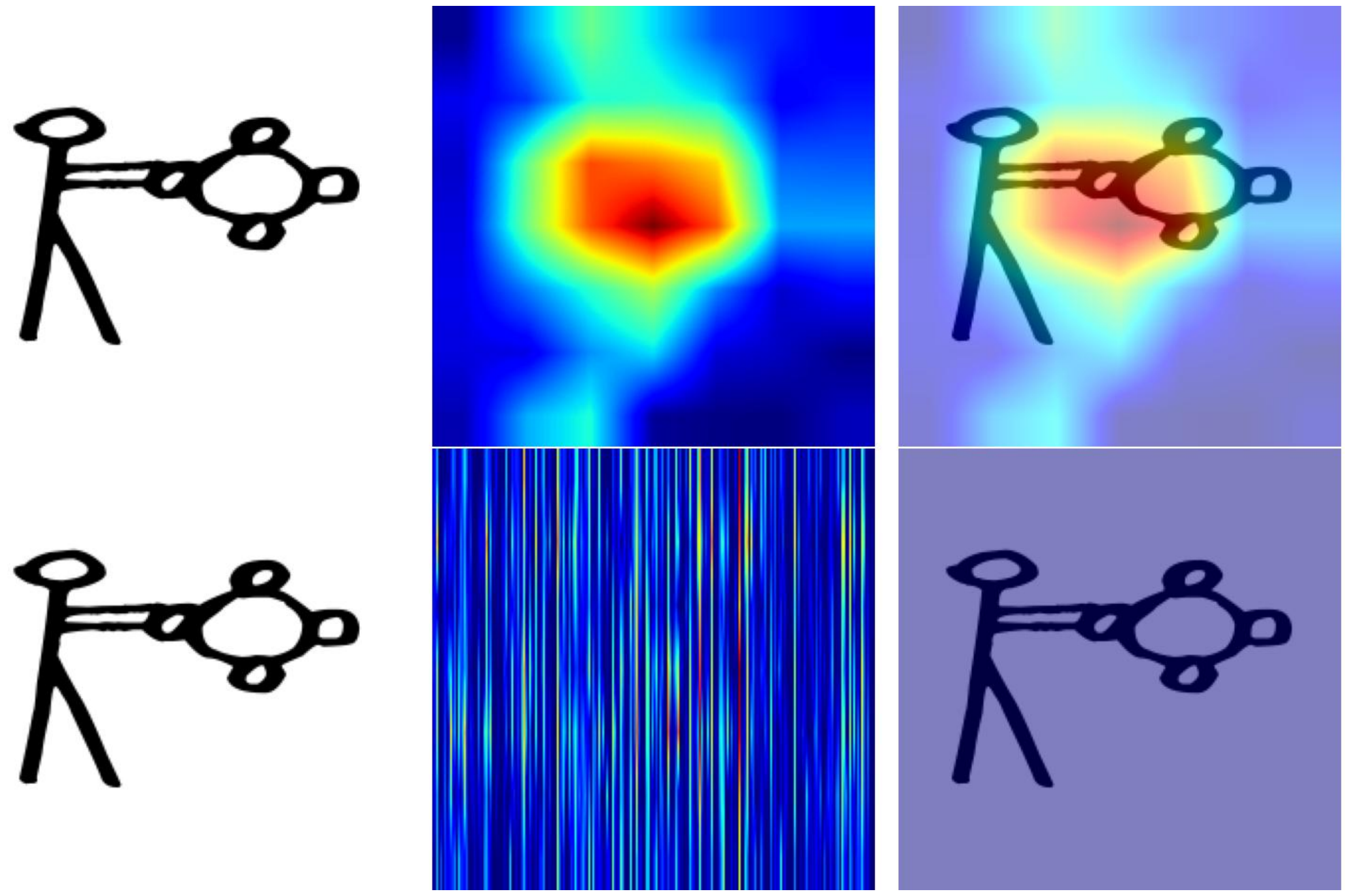

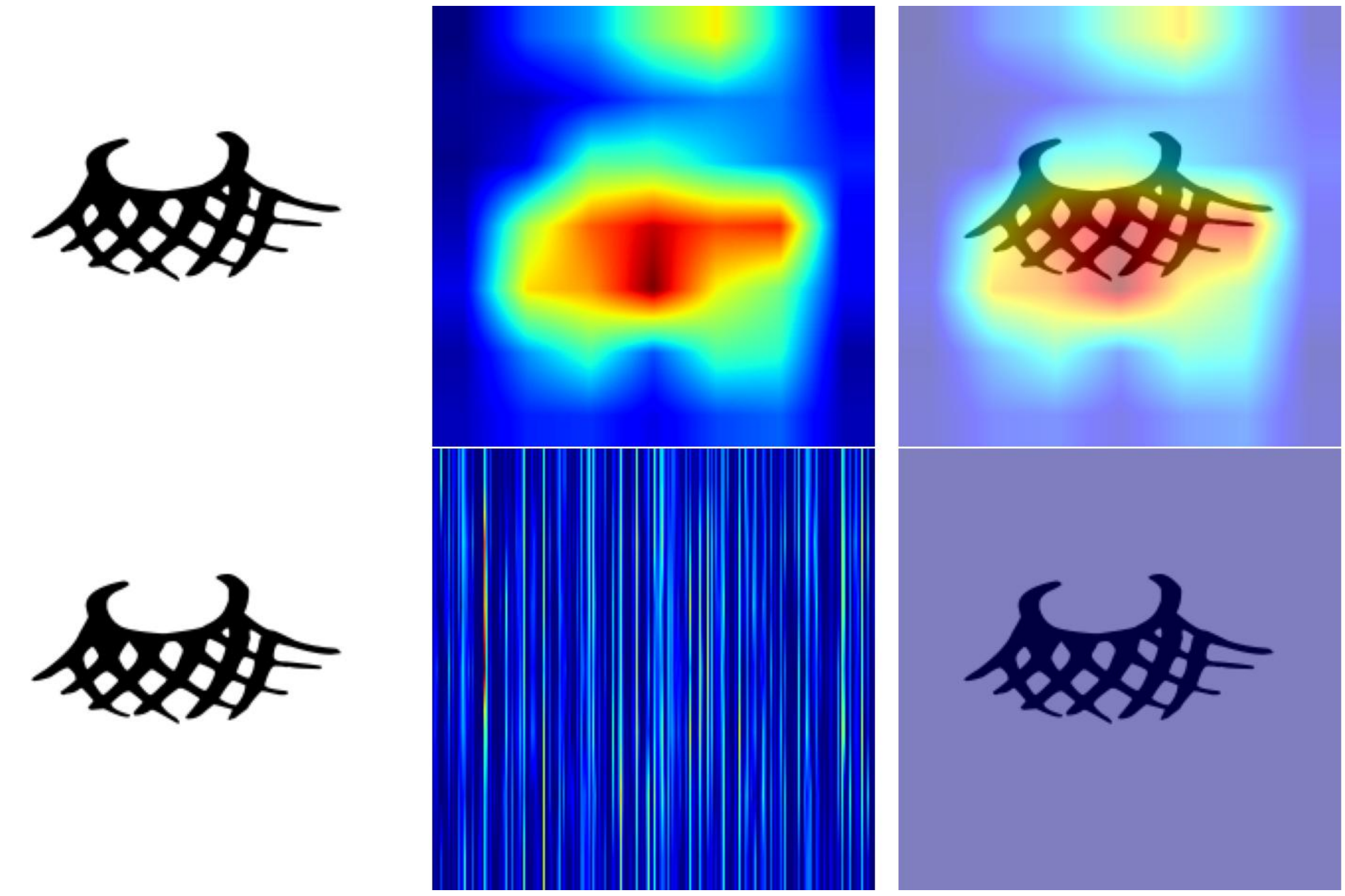

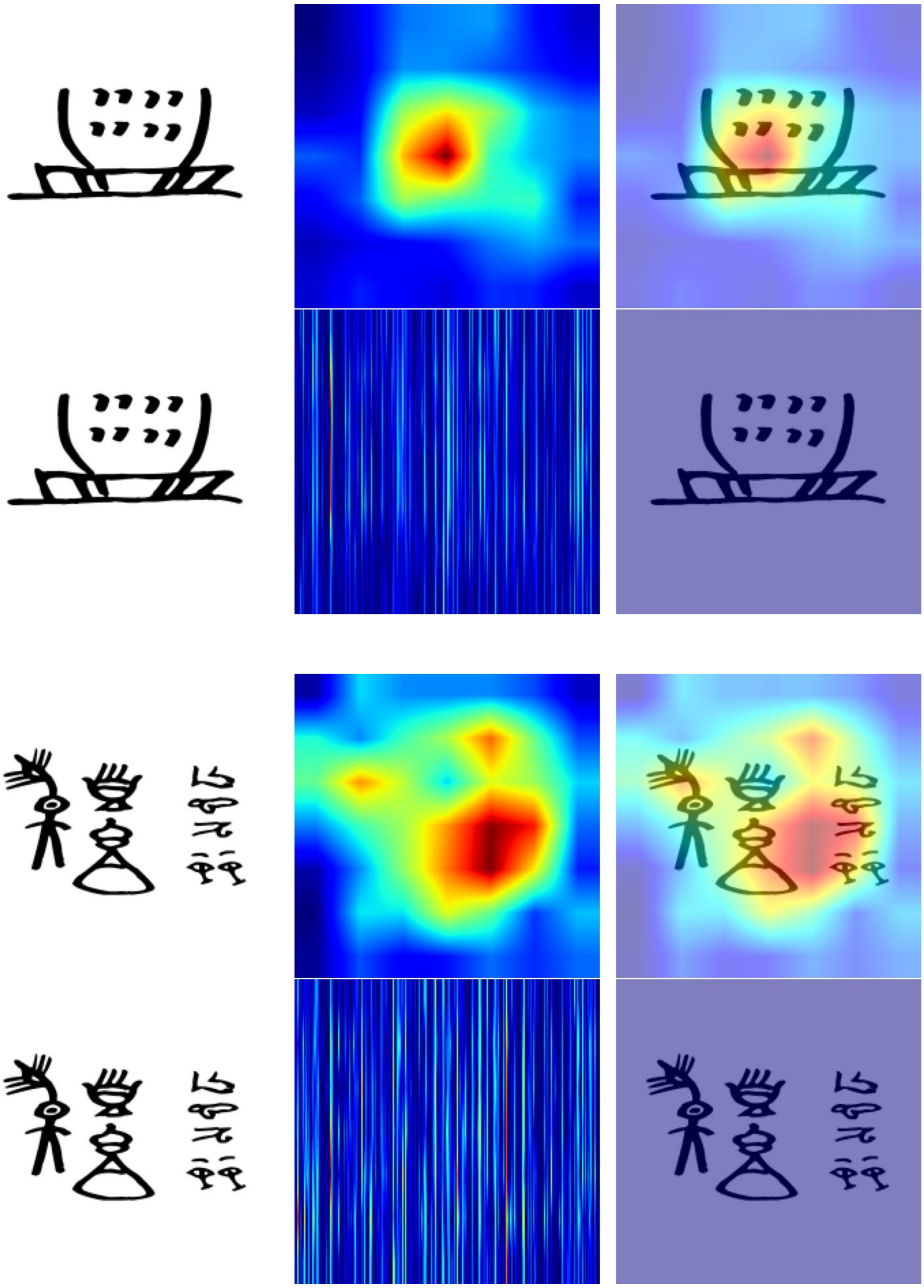

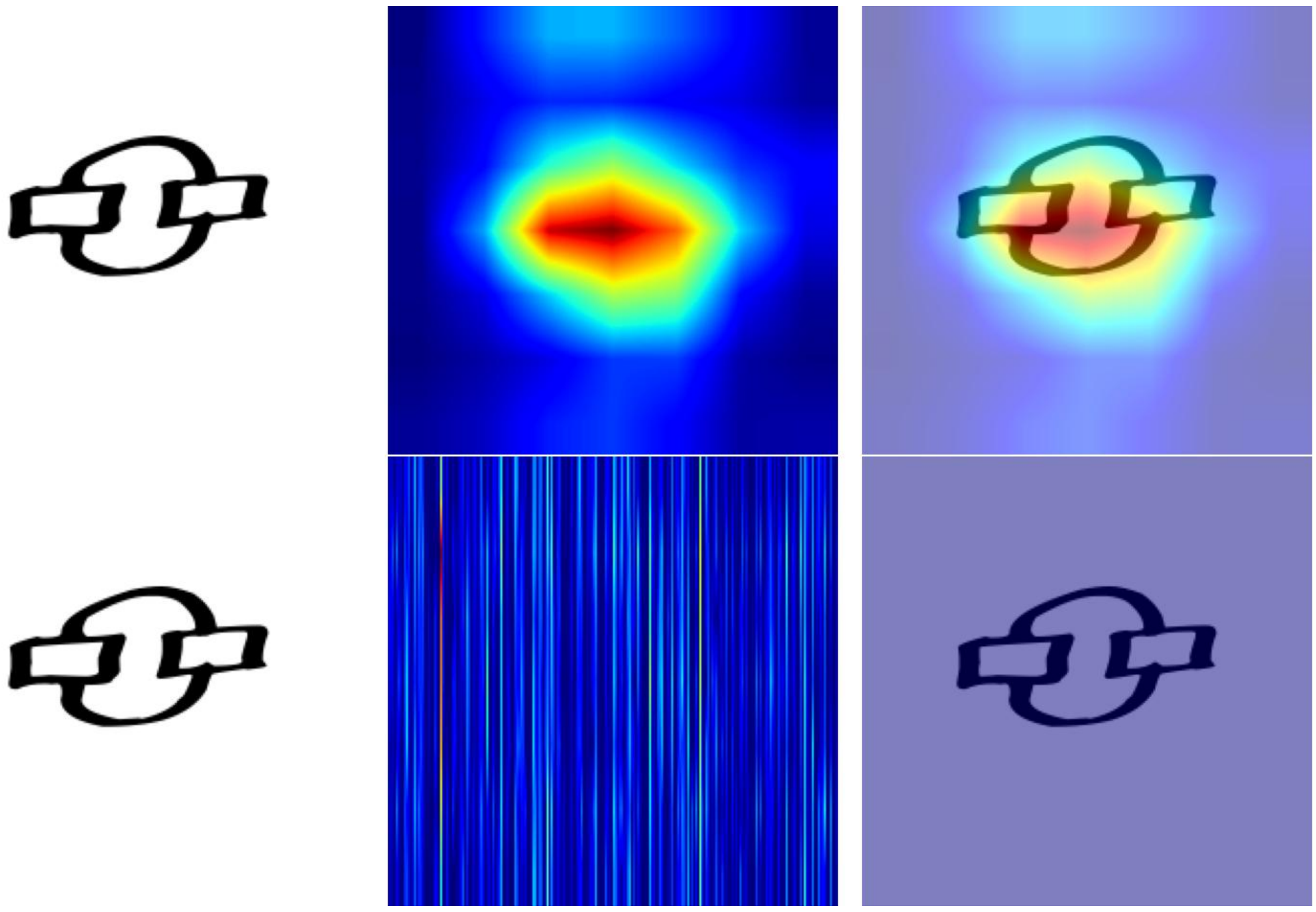

### A.8.5 Old Dongba

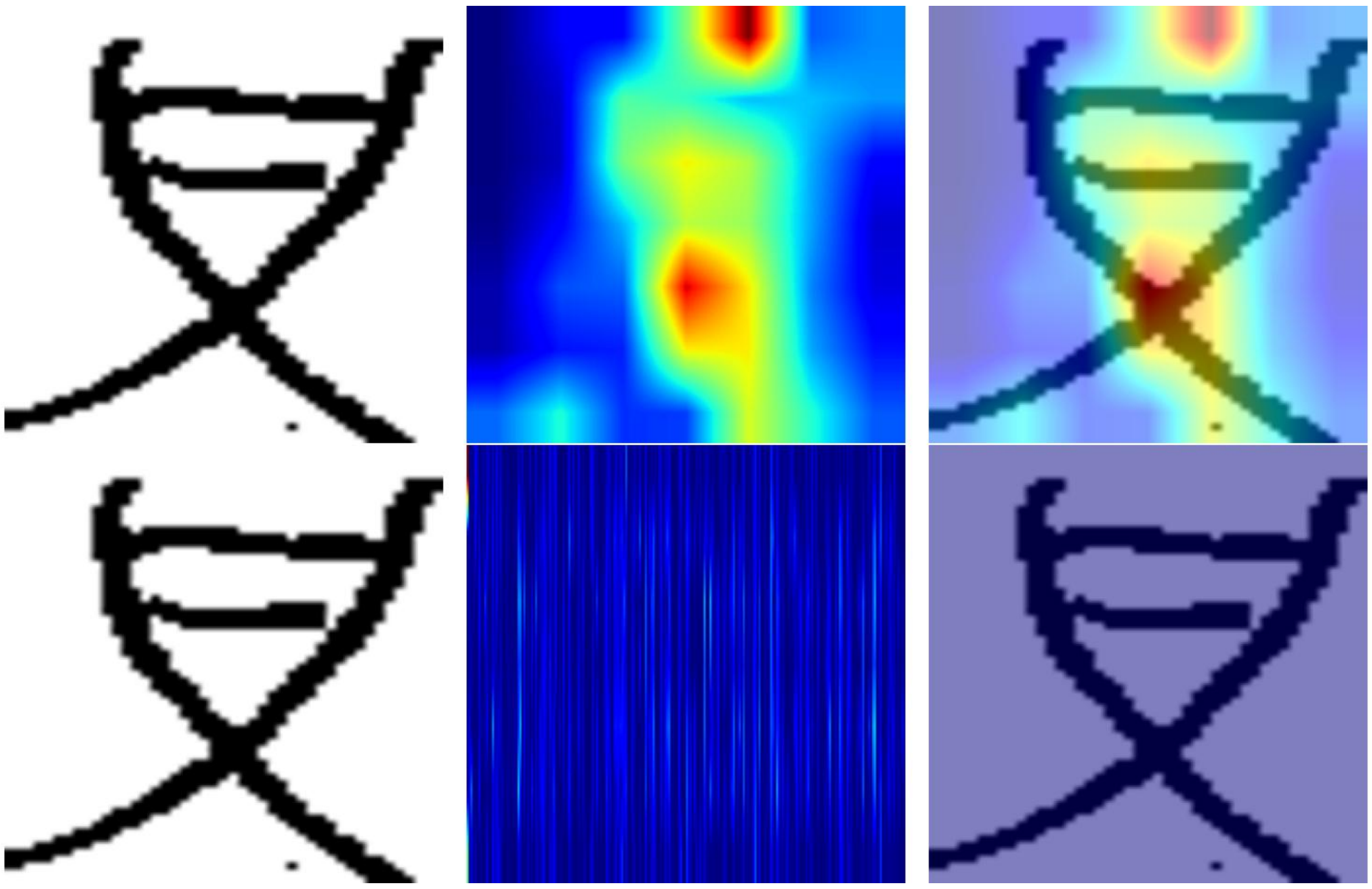

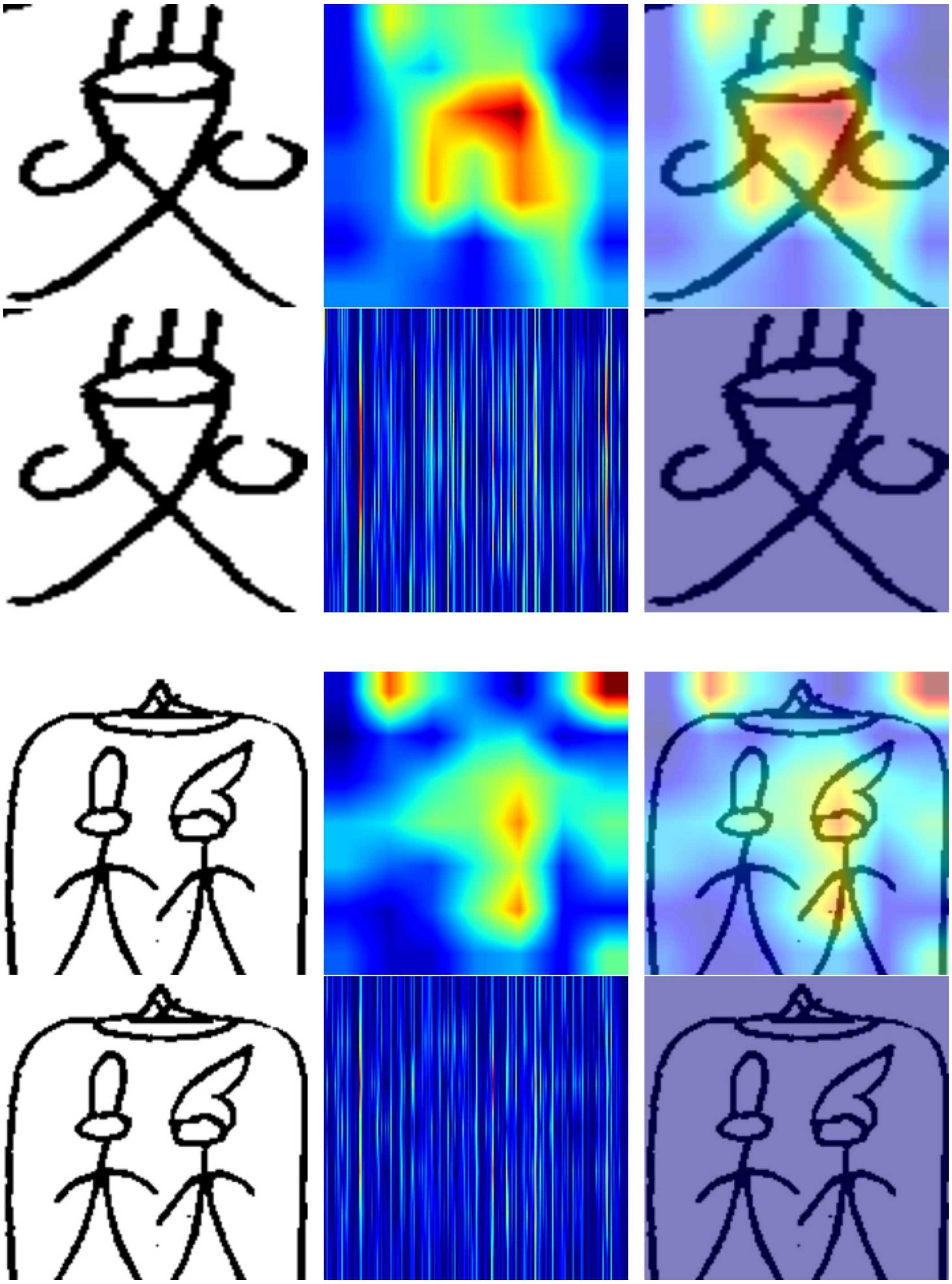

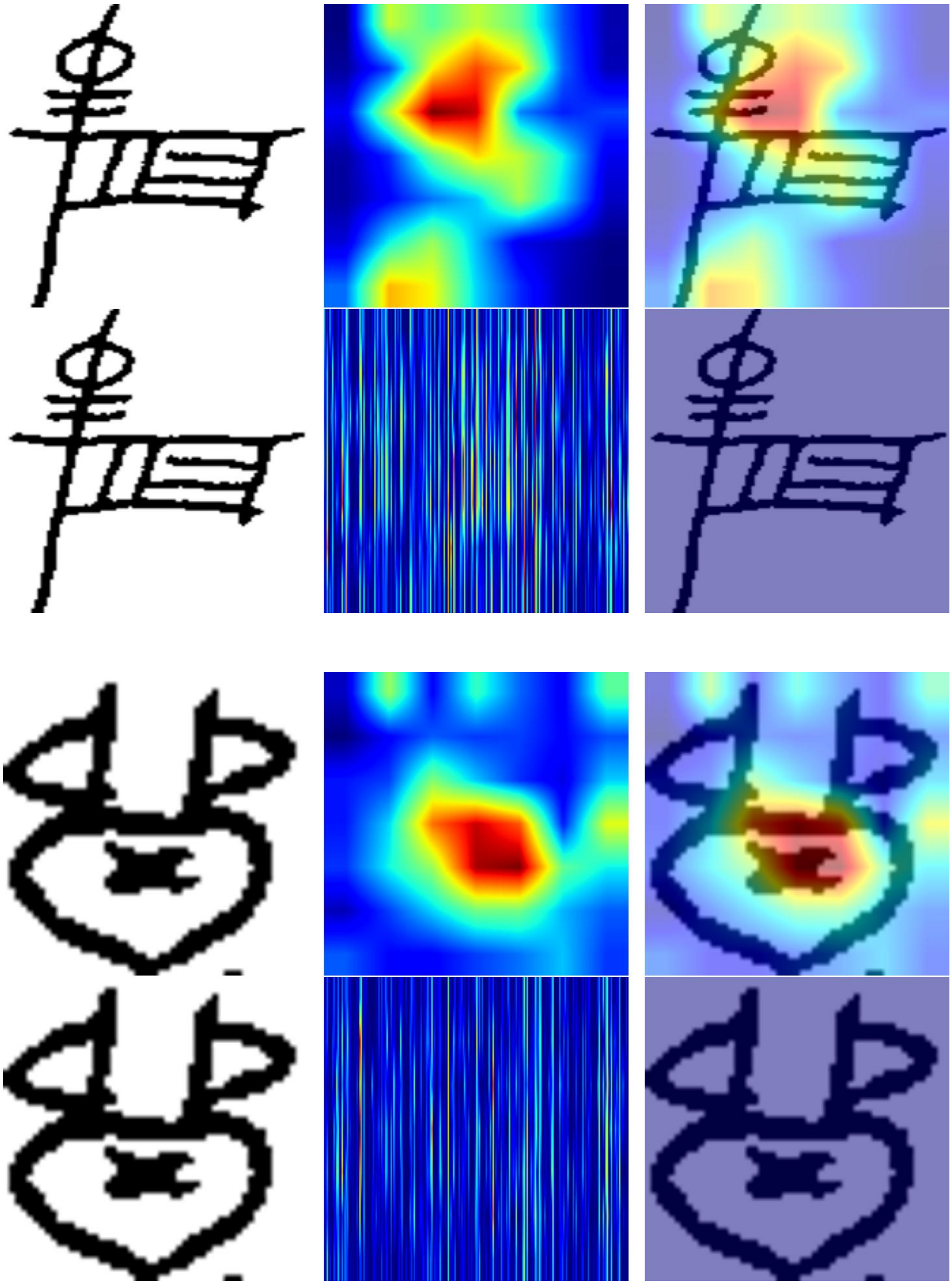

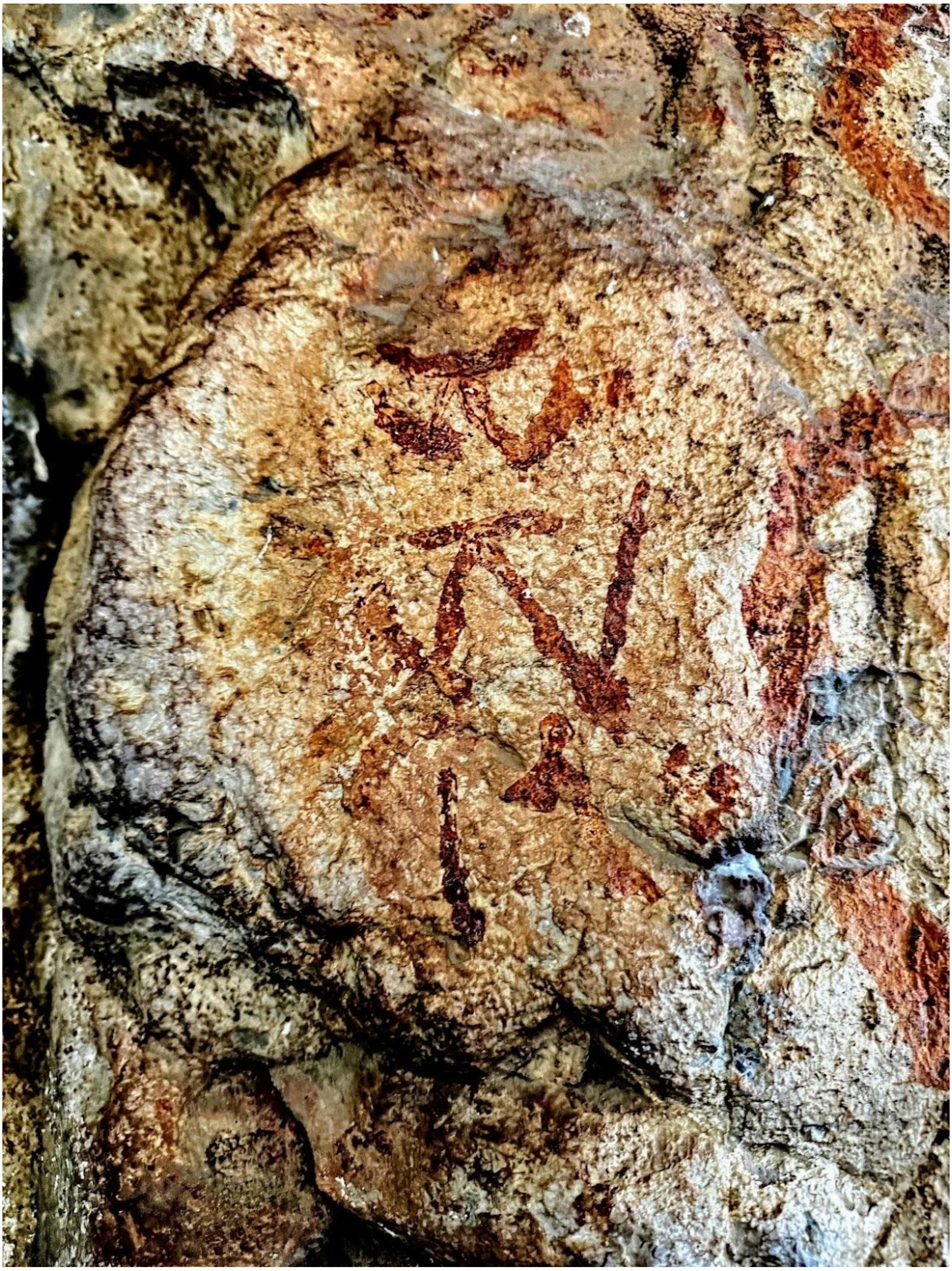

Figure A.4: Early characters depicted in a cave near Namtso Lake, Tibet. Photo taken by author on December 31, 2024.

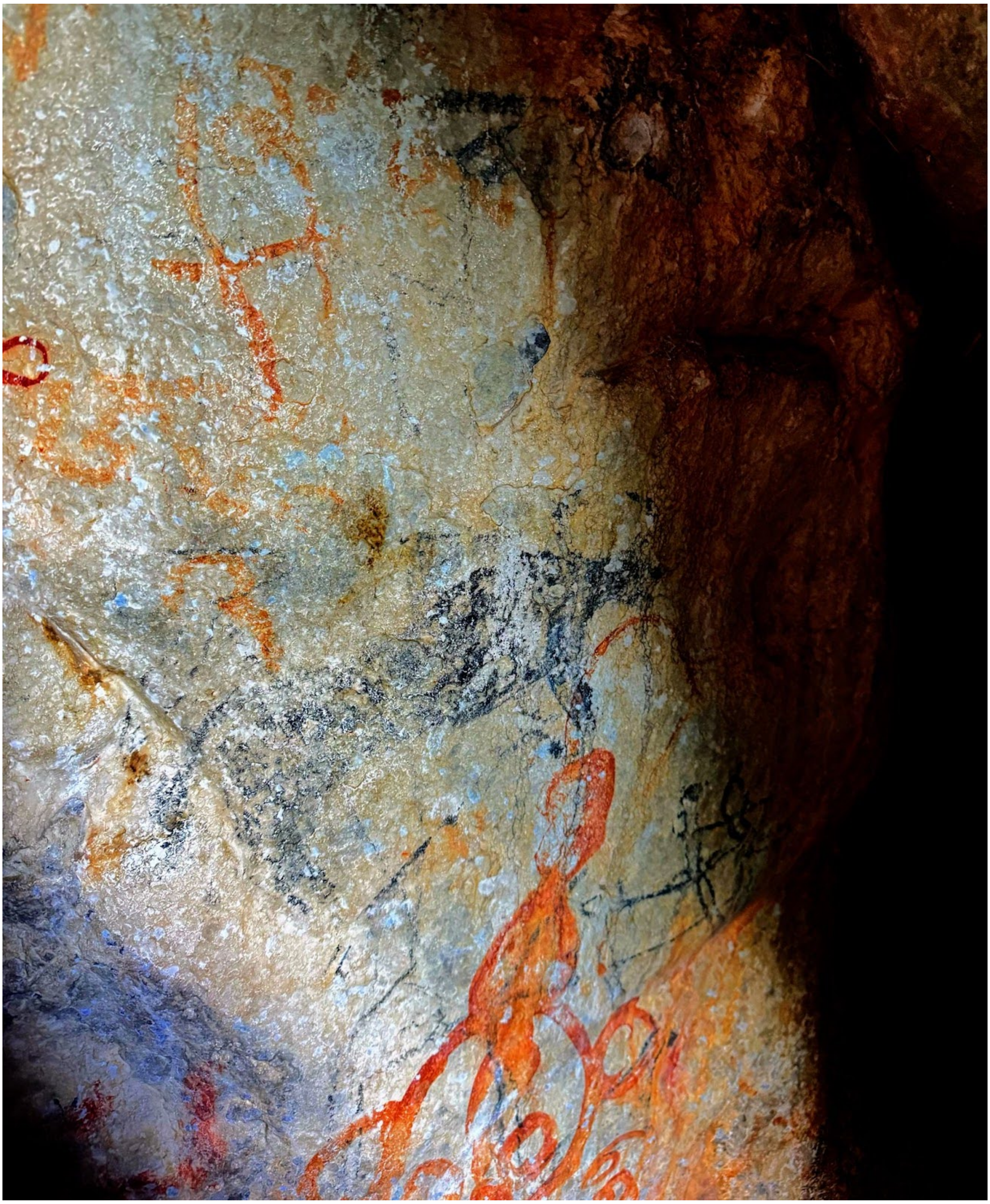

Figure A.5: ]
Black yak on the walls of a cave near Namtso Lake, Tibet. Photo taken by author on December 31, 2024.

## A.9 Indus Scale

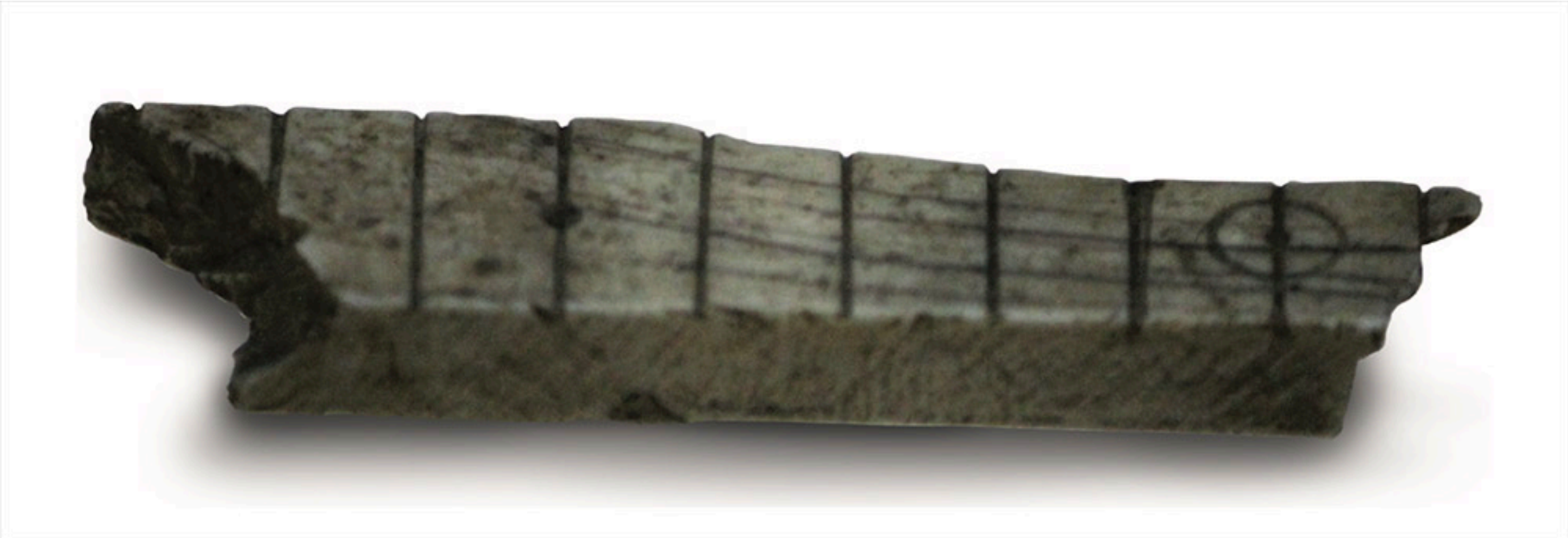

Figure A.6: Fragmented scale measure from Mohenjodaro [30, 56].

## A.10 Statistical Tests

The following pages contain the results of our statistical relationship analysis, including our t-tests and Cohen's d evaluations.

Table A.1: Statistical Tests Summary Output

| Comparison Script | Model | Test | Mean 1 | Mean 2 | Difference | T-statistic | P-value | Cohen's d | Effect Size | Significant | Better Match |
|---|---|---|---|---|---|---|---|---|---|---|---|
| TYC | model_0 | Indus_vs_Proto-Cuneiform | 0.61 | 0.07 | 0.54 | 3430.39 | 0.00 | 9.70 | large | Yes | Indus |
| TYC | model_0 | Indus_vs_Proto-Elamite | 0.61 | 0.10 | 0.51 | 3143.15 | 0.00 | 8.89 | large | Yes | Indus |
| TYC | model_0 | Proto-Cuneiform_vs_Proto-Elamite | 0.07 | 0.10 | -0.03 | -412.76 | 0.00 | -1.17 | large | Yes | Proto-Elamite |
| TYC | model_1 | Indus_vs_Proto-Cuneiform | 0.61 | 0.07 | 0.54 | 2362.56 | 0.00 | 6.68 | large | Yes | Indus |
| TYC | model_1 | Indus_vs_Proto-Elamite | 0.61 | 0.07 | 0.54 | 2328.76 | 0.00 | 6.59 | large | Yes | Indus |
| TYC | model_1 | Proto-Cuneiform_vs_Proto-Elamite | 0.07 | 0.07 | 0.00 | 1.77 | 0.08 | 0.01 | negligible | No | Proto-Cuneiform |
| TYC | model_2 | Indus_vs_Proto-Cuneiform | 0.63 | 0.07 | 0.56 | 2720.97 | 0.00 | 7.70 | large | Yes | Indus |
| TYC | model_2 | Indus_vs_Proto-Elamite | 0.63 | 0.08 | 0.55 | 2696.50 | 0.00 | 7.63 | large | Yes | Indus |
| TYC | model_2 | Proto-Cuneiform_vs_Proto-Elamite | 0.07 | 0.08 | -0.01 | -117.34 | 0.00 | -0.33 | small | Yes | Proto-Elamite |
| TYC | model_3 | Indus_vs_Proto-Cuneiform | 0.65 | 0.07 | 0.58 | 2331.90 | 0.00 | 6.60 | large | Yes | Indus |
| TYC | model_3 | Indus_vs_Proto-Elamite | 0.65 | 0.07 | 0.58 | 2385.25 | 0.00 | 6.75 | large | Yes | Indus |
| TYC | model_3 | Proto-Cuneiform_vs_Proto-Elamite | 0.07 | 0.07 | 0.00 | 4.93 | 0.00 | 0.01 | negligible | Yes | Proto-Cuneiform |
| TYC | model_4 | Indus_vs_Proto-Cuneiform | 0.67 | 0.23 | 0.44 | 1868.95 | 0.00 | 5.29 | large | Yes | Indus |
| TYC | model_4 | Indus_vs_Proto-Elamite | 0.66 | 0.07 | 0.59 | 2670.99 | 0.00 | 7.55 | large | Yes | Indus |
| TYC | model_4 | Proto-Cuneiform_vs_Proto-Elamite | 0.23 | 0.07 | 0.16 | 1391.71 | 0.00 | 3.94 | large | Yes | Proto-Cuneiform |
| Naxi-Dongba | model_0 | Indus_vs_Proto-Cuneiform | 0.58 | 0.07 | 0.51 | 3428.45 | 0.00 | 9.70 | large | Yes | Indus |
| Naxi-Dongba | model_0 | Indus_vs_Proto-Elamite | 0.58 | 0.08 | 0.50 | 3406.41 | 0.00 | 9.63 | large | Yes | Indus |
| Naxi-Dongba | model_0 | Proto-Cuneiform_vs_Proto-Elamite | 0.07 | 0.08 | -0.01 | -145.12 | 0.00 | -0.41 | small | Yes | Proto-Elamite |
| Naxi-Dongba | model_1 | Indus_vs_Proto-Cuneiform | 0.65 | 0.05 | 0.60 | 3414.21 | 0.00 | 9.66 | large | Yes | Indus |
| Naxi-Dongba | model_1 | Indus_vs_Proto-Elamite | 0.66 | 0.10 | 0.56 | 3329.02 | 0.00 | 9.42 | large | Yes | Indus |
| Naxi-Dongba | model_1 | Proto-Cuneiform_vs_Proto-Elamite | 0.06 | 0.10 | -0.04 | -523.73 | 0.00 | -1.48 | large | Yes | Proto-Elamite |
| Naxi-Dongba | model_2 | Indus_vs_Proto-Cuneiform | 0.58 | 0.05 | 0.53 | 2773.65 | 0.00 | 7.85 | large | Yes | Indus |
| Naxi-Dongba | model_2 | Indus_vs_Proto-Elamite | 0.58 | 0.10 | 0.48 | 2461.76 | 0.00 | 6.96 | large | Yes | Indus |
| Naxi-Dongba | model_2 | Proto-Cuneiform_vs_Proto-Elamite | 0.05 | 0.09 | -0.04 | -483.31 | 0.00 | -1.37 | large | Yes | Proto-Elamite |
| Naxi-Dongba | model_3 | Indus_vs_Proto-Cuneiform | 0.68 | 0.08 | 0.60 | 2883.93 | 0.00 | 8.16 | large | Yes | Indus |
| Naxi-Dongba | model_3 | Indus_vs_Proto-Elamite | 0.68 | 0.08 | 0.60 | 3069.56 | 0.00 | 8.68 | large | Yes | Indus |
| Naxi-Dongba | model_3 | Proto-Cuneiform_vs_Proto-Elamite | 0.08 | 0.08 | 0.00 | 59.78 | 0.00 | 0.17 | negligible | Yes | Proto-Cuneiform |
| Naxi-Dongba | model_4 | Indus_vs_Proto-Cuneiform | 0.68 | 0.28 | 0.40 | 1672.96 | 0.00 | 4.73 | large | Yes | Indus |
| Naxi-Dongba | model_4 | Indus_vs_Proto-Elamite | 0.68 | 0.08 | 0.59 | 2651.29 | 0.00 | 7.50 | large | Yes | Indus |
| Naxi-Dongba | model_4 | Proto-Cuneiform_vs_Proto-Elamite | 0.28 | 0.08 | 0.20 | 1789.15 | 0.00 | 5.06 | large | Yes | Proto-Cuneiform |
| Old-Naxi | model_0 | Indus_vs_Proto-Cuneiform | 0.62 | 0.07 | 0.55 | 3605.37 | 0.00 | 10.20 | large | Yes | Indus |
| Old-Naxi | model_0 | Indus_vs_Proto-Elamite | 0.62 | 0.10 | 0.52 | 3376.88 | 0.00 | 9.55 | large | Yes | Indus |
| Old-Naxi | model_0 | Proto-Cuneiform_vs_Proto-Elamite | 0.07 | 0.10 | -0.03 | -400.45 | 0.00 | -1.13 | large | Yes | Proto-Elamite |
| Old-Naxi | model_1 | Indus_vs_Proto-Cuneiform | 0.58 | 0.08 | 0.50 | 2096.98 | 0.00 | 5.93 | large | Yes | Indus |
| Old-Naxi | model_1 | Indus_vs_Proto-Elamite | 0.58 | 0.05 | 0.53 | 2358.66 | 0.00 | 6.67 | large | Yes | Indus |
| Old-Naxi | model_1 | Proto-Cuneiform_vs_Proto-Elamite | 0.07 | 0.05 | 0.02 | 263.14 | 0.00 | 0.74 | medium | Yes | Proto-Cuneiform |
| Old-Naxi | model_2 | Indus_vs_Proto-Cuneiform | 0.66 | 0.08 | 0.58 | 3075.61 | 0.00 | 8.70 | large | Yes | Indus |
| Old-Naxi | model_2 | Indus_vs_Proto-Elamite | 0.66 | 0.09 | 0.57 | 2944.56 | 0.00 | 8.33 | large | Yes | Indus |
| Old-Naxi | model_2 | Proto-Cuneiform_vs_Proto-Elamite | 0.07 | 0.08 | -0.01 | -118.30 | 0.00 | -0.33 | small | Yes | Proto-Elamite |
| Old-Naxi | model_3 | Indus_vs_Proto-Cuneiform | 0.61 | 0.08 | 0.52 | 2145.70 | 0.00 | 6.07 | large | Yes | Indus |
| Old-Naxi | model_3 | Indus_vs_Proto-Elamite | 0.59 | 0.07 | 0.52 | 2082.42 | 0.00 | 5.89 | large | Yes | Indus |

Continued on next page

**Table A.1 Continued from previous page**

| Comparison Script | Model | Test | Mean 1 | Mean 2 | Difference | T-statistic | P-value | Cohen's d | Effect Size | Significant | Better Match |
|---|---|---|---|---|---|---|---|---|---|---|---|
| Old-Naxi | model_3 | Proto-Cuneiform_vs_Proto-Elamite | 0.08 | 0.07 | 0.01 | 202.07 | 0.00 | 0.57 | medium | Yes | Proto-Cuneiform |
| Old-Naxi | model_4 | Indus_vs_Proto-Cuneiform | 0.65 | 0.22 | 0.42 | 1745.20 | 0.00 | 4.94 | large | Yes | Indus |
| Old-Naxi | model_4 | Indus_vs_Proto-Elamite | 0.65 | 0.07 | 0.58 | 2550.84 | 0.00 | 7.21 | large | Yes | Indus |
| Old-Naxi | model_4 | Proto-Cuneiform_vs_Proto-Elamite | 0.22 | 0.07 | 0.16 | 1430.05 | 0.00 | 4.04 | large | Yes | Proto-Cuneiform |

## A.11 Similarity Summary

Table A.2: Script Similarity Analysis Summary Output

| *Script Similarity Analysis Summary* | | | | |
|---|---|---|---|---|
| Analysis Date: 2025-03-6 17:34:46 | | | | |
| | | | | |
| **Indus Similarities** | | | | |
| | | | | |
| Script | Mean Similarity | Std Dev | 95% CI Lower | 95% CI Upper |
| ——- | ————- | ——- | ———— | ———: |
| Indus | 1.0000 | 0.0000 | 1.0000 | 1.0000 |
| TYC | 0.9297 | 0.0145 | 0.9170 | 0.9424 |
| Proto-Cuneiform | 0.8873 | 0.0272 | 0.8634 | 0.9111 |
| Old-Naxi | 0.8777 | 0.0560 | 0.8287 | 0.9268 |
| Proto-Elamite | 0.8550 | 0.0420 | 0.8182 | 0.8918 |
| Naxi-Dongba | 0.8518 | 0.0179 | 0.8361 | 0.8675 |
| | | | | |
| **Proto-Cuneiform Similarities** | | | | |
| | | | | |
| Script | Mean Similarity | Std Dev | 95% CI Lower | 95% CI Upper |
| ——- | ————- | ——- | ———— | ———: |
| Proto-Cuneiform | 1.0000 | 0.0000 | 1.0000 | 1.0000 |
| Proto-Elamite | 0.9860 | 0.0067 | 0.9801 | 0.9919 |
| Naxi-Dongba | 0.9500 | 0.0177 | 0.9345 | 0.9655 |
| TYC | 0.9463 | 0.0300 | 0.9199 | 0.9726 |
| Indus | 0.9115 | 0.0201 | 0.8939 | 0.9292 |
| Old-Naxi | 0.8828 | 0.0821 | 0.8109 | 0.9548 |
| | | | | |
| **Proto-Elamite Similarities** | | | | |
| | | | | |
| Script | Mean Similarity | Std Dev | 95% CI Lower | 95% CI Upper |
| ——- | ————- | ——- | ———— | ———: |
| Proto-Elamite | 1.0000 | 0.0000 | 1.0000 | 1.0000 |
| Proto-Cuneiform | 0.9726 | 0.0144 | 0.9600 | 0.9852 |
| TYC | 0.9450 | 0.0173 | 0.9298 | 0.9601 |
| Old-Naxi | 0.9130 | 0.0276 | 0.8888 | 0.9372 |
| Naxi-Dongba | 0.9127 | 0.0252 | 0.8906 | 0.9347 |
| Indus | 0.9088 | 0.0182 | 0.8928 | 0.9247 |

## A.12 Model Training Data

Table A.3: Indus Training Summary Output

| model_idx | seed | val_loss | epoch | model_path |
|---|---|---|---|---|
| 1 | 42 | 9.65621223449707 | 3 | Indus_ensemble_1_hybrid_extractor_best.pth |
| 2 | 43 | 9.49264907836914 | 1 | Indus_ensemble_2_hybrid_extractor_best.pth |
| 3 | 44 | 9.3032808303833 | 11 | Indus_ensemble_3_hybrid_extractor_best.pth |
| 4 | 45 | 9.921699905395508 | 1 | Indus_ensemble_4_hybrid_extractor_best.pth |
| 5 | 46 | 10.057604217529297 | 1 | Indus_ensemble_5_hybrid_extractor_best.pth |

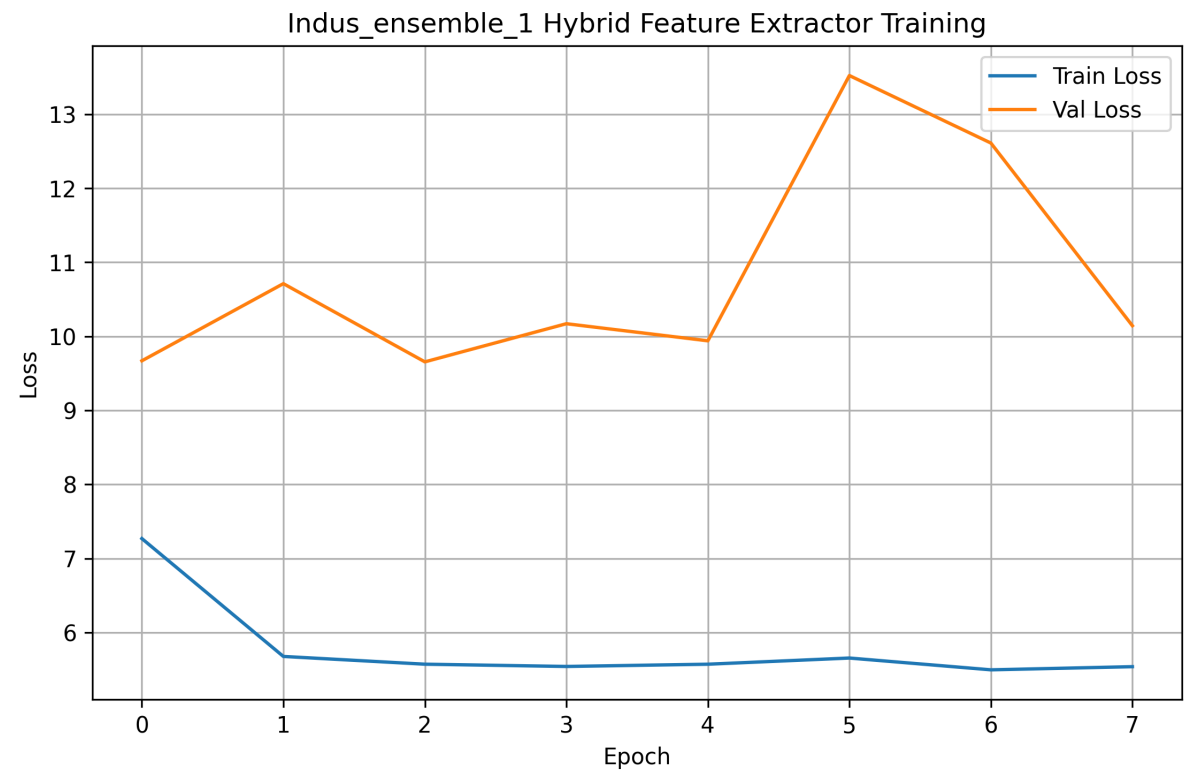

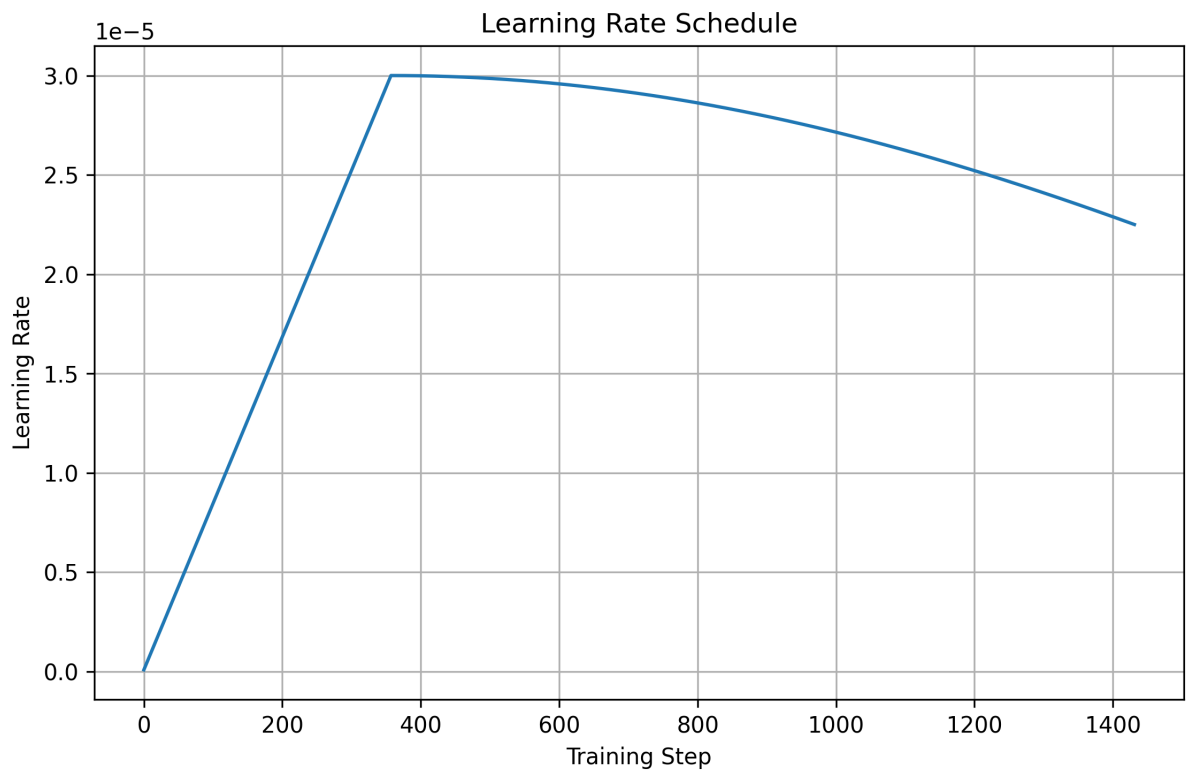

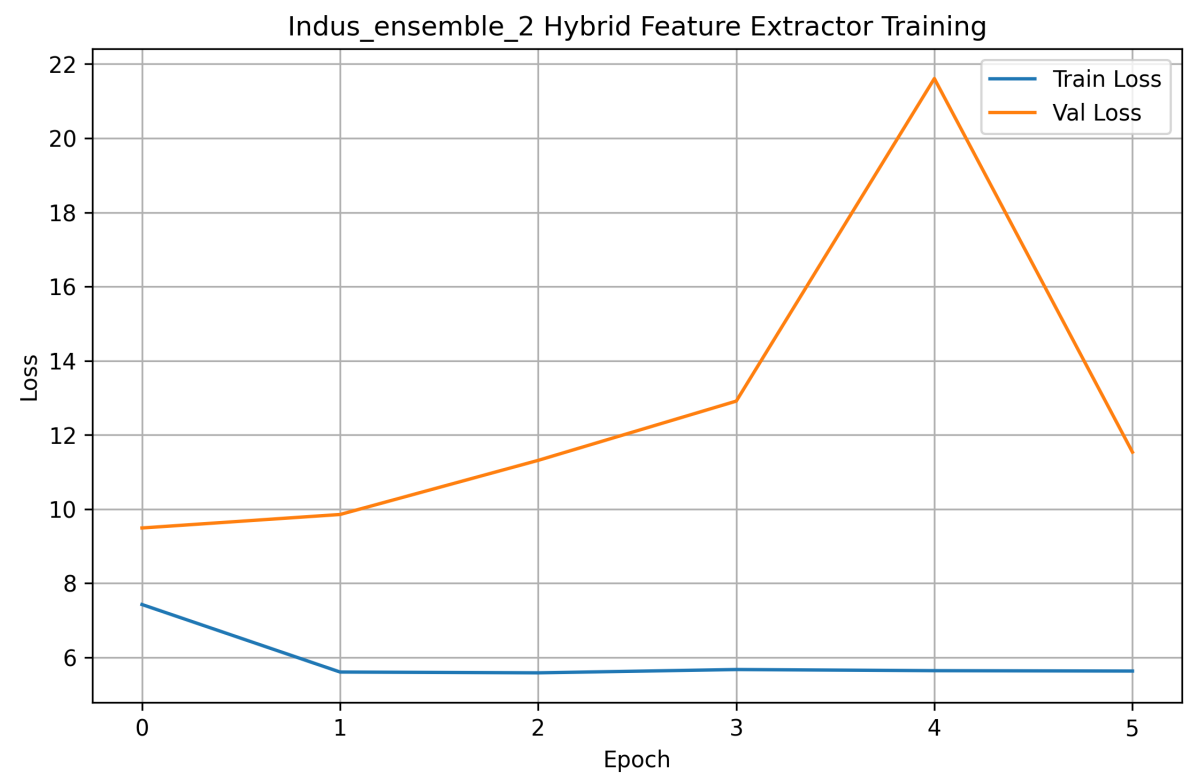

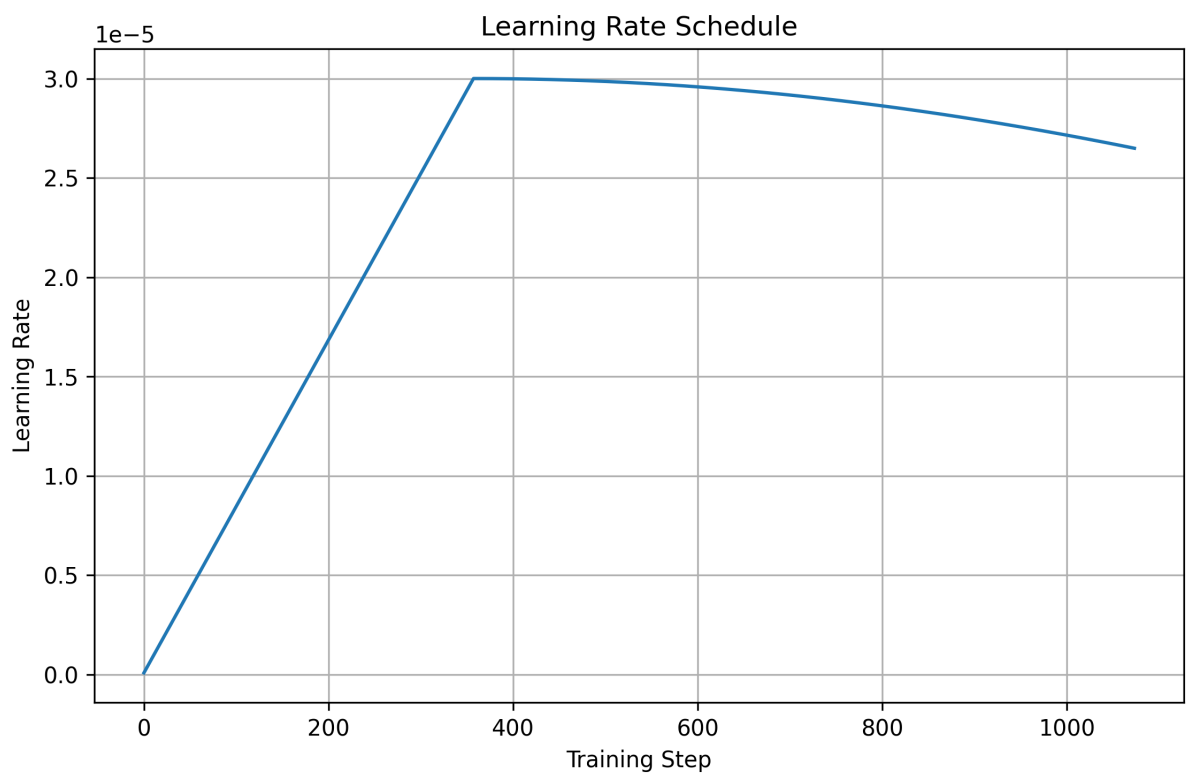

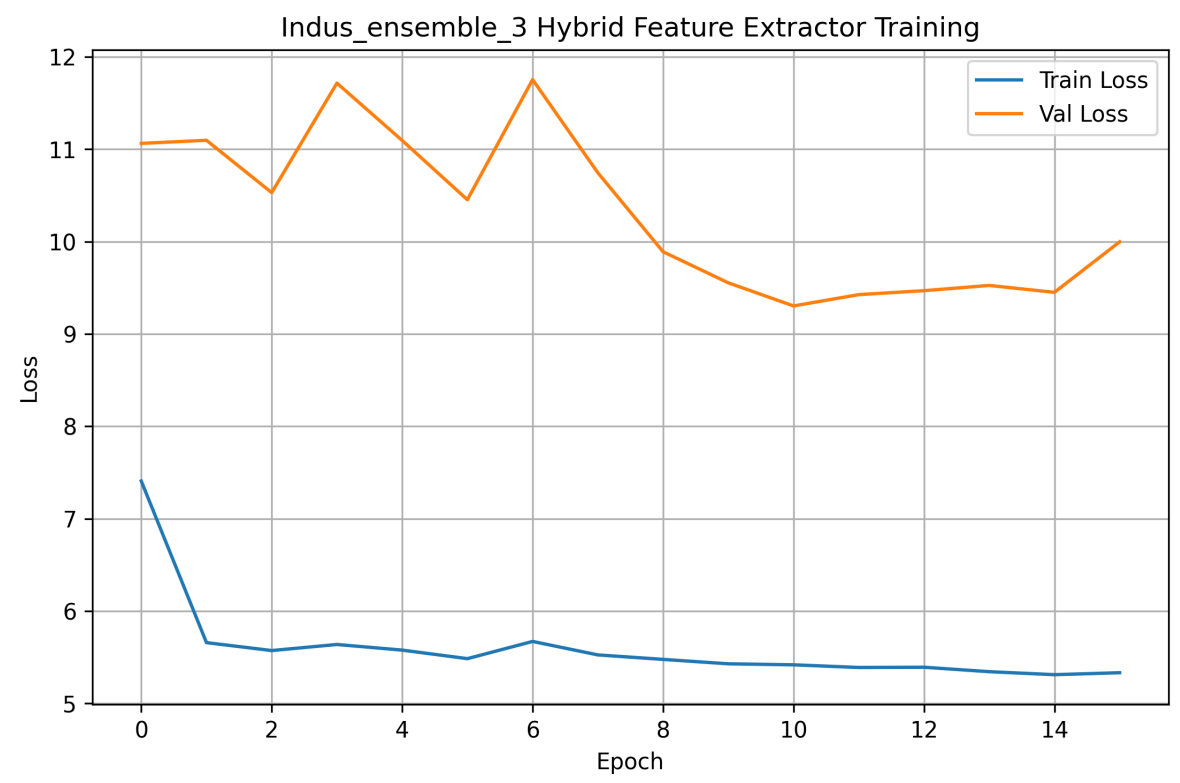

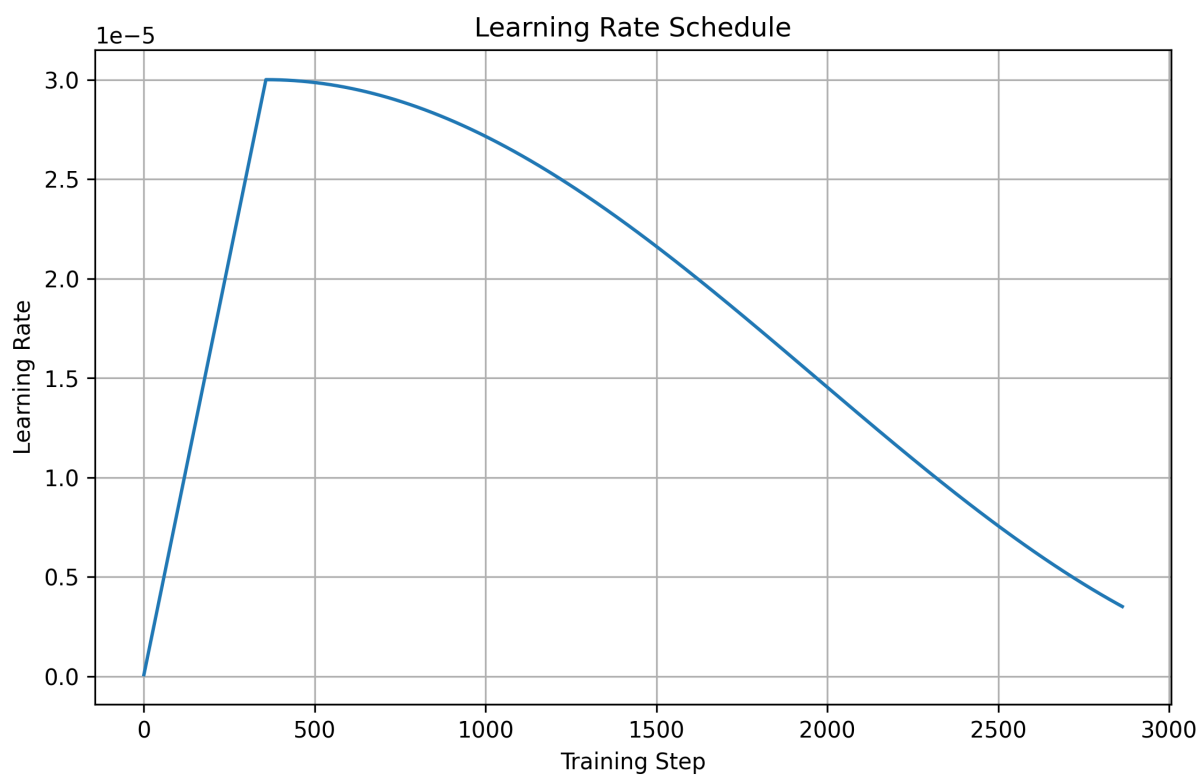

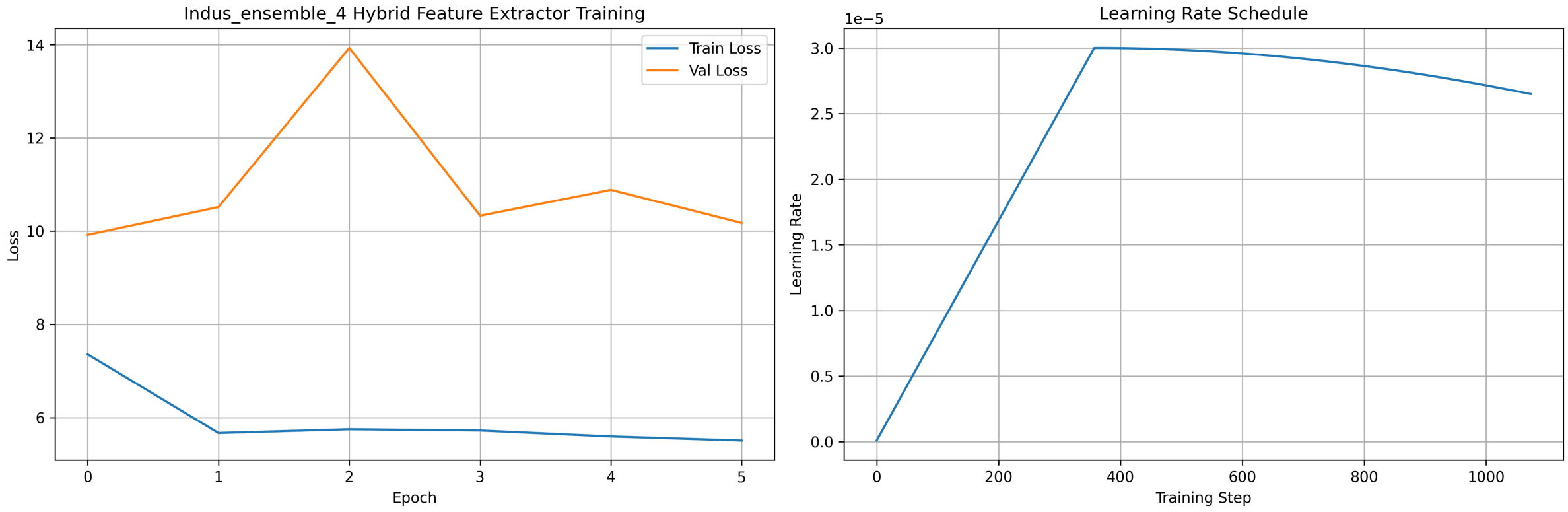
Indus_ensemble_4 Hybrid Feature Extractor Training
Train Loss
Val Loss
Loss
Epoch
Learning Rate Schedule
1e−5
Learning Rate
Training Step

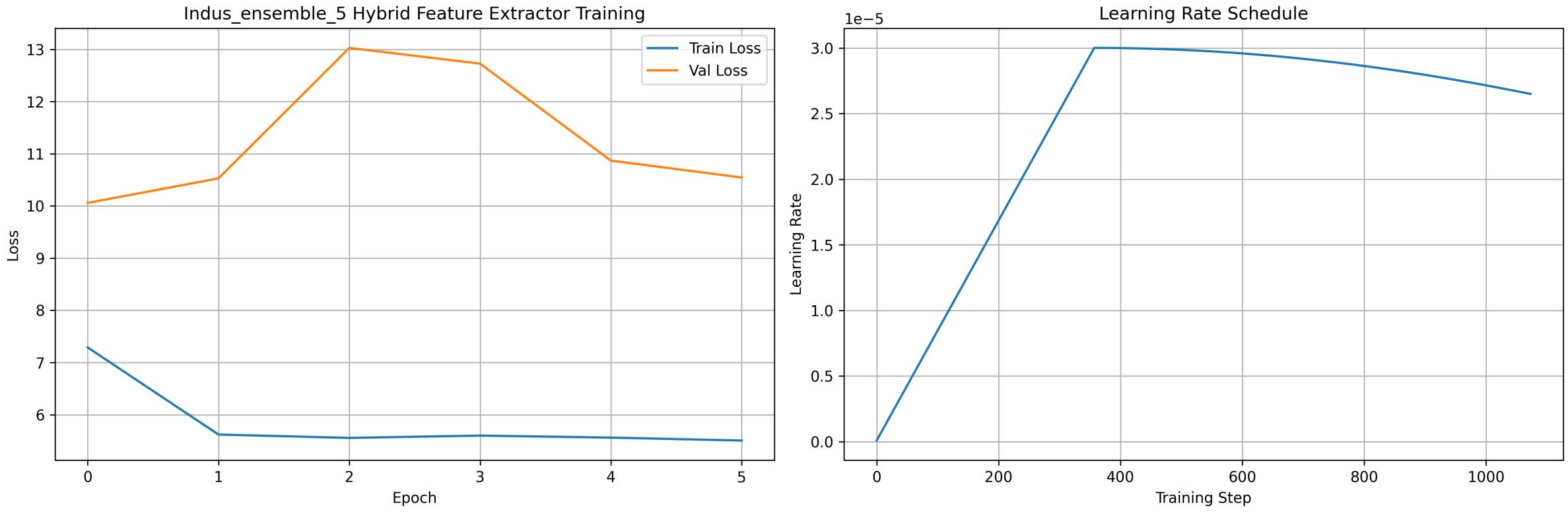
Indus_ensemble_5 Hybrid Feature Extractor Training
Train Loss
Val Loss
Loss
Epoch
Learning Rate Schedule
1e−5
Learning Rate
Training Step

Proto-Cuneiform training summary.

| model_idx | seed | val_loss | epoch | model_path |
|---|---|---|---|---|
| 1 | 42 | 8.801242896488734 | 11 | Proto-Cuneiform_ensemble_1_hybrid_extractor_best.pth |
| 2 | 43 | 8.71736410685948 | 11 | Proto-Cuneiform_ensemble_2_hybrid_extractor_best.pth |
| 3 | 44 | 8.819251537322998 | 7 | Proto-Cuneiform_ensemble_3_hybrid_extractor_best.pth |
| 4 | 45 | 8.773778438568115 | 9 | Proto-Cuneiform_ensemble_4_hybrid_extractor_best.pth |
| 5 | 46 | 9.27264860698155 | 1 | Proto-Cuneiform_ensemble_5_hybrid_extractor_best.pth |

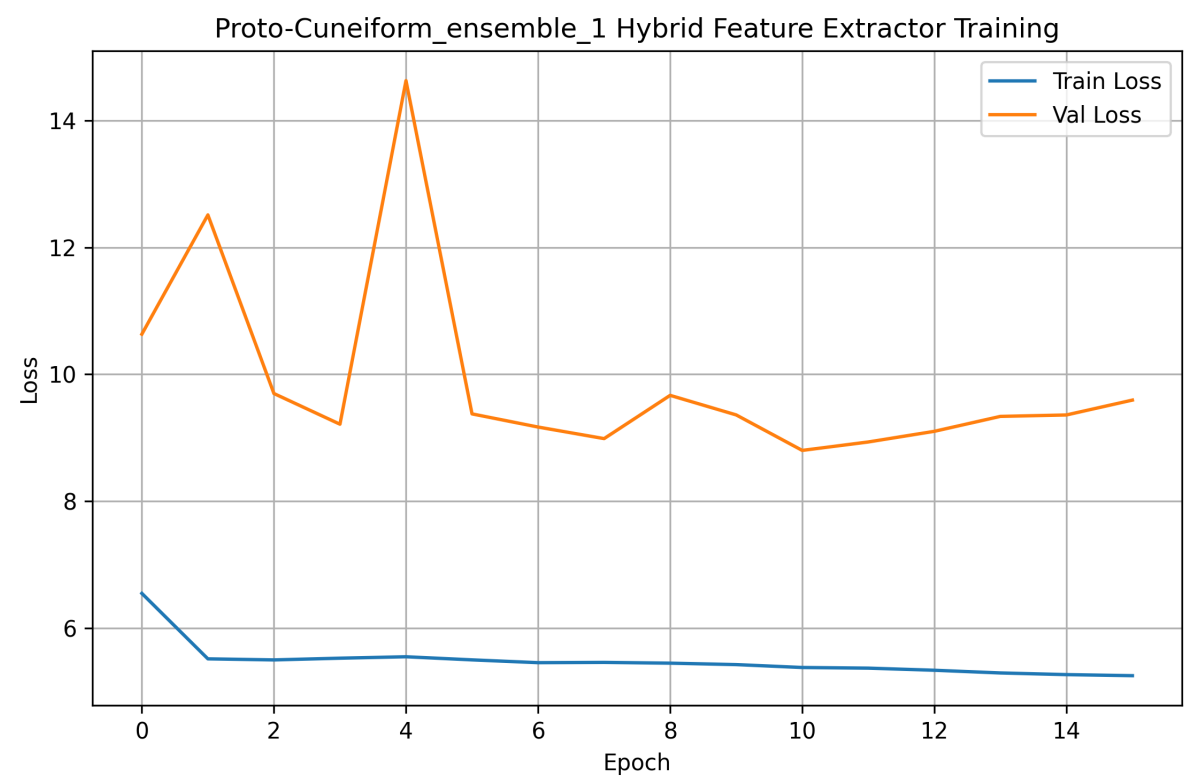

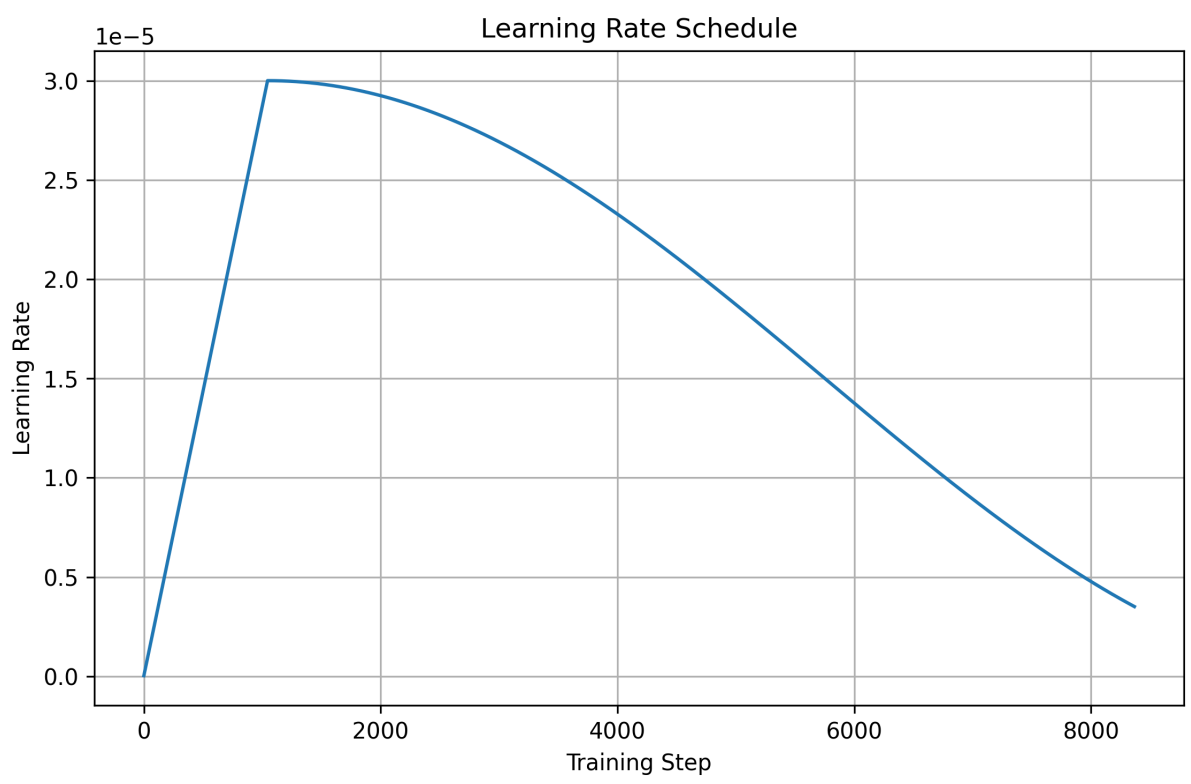

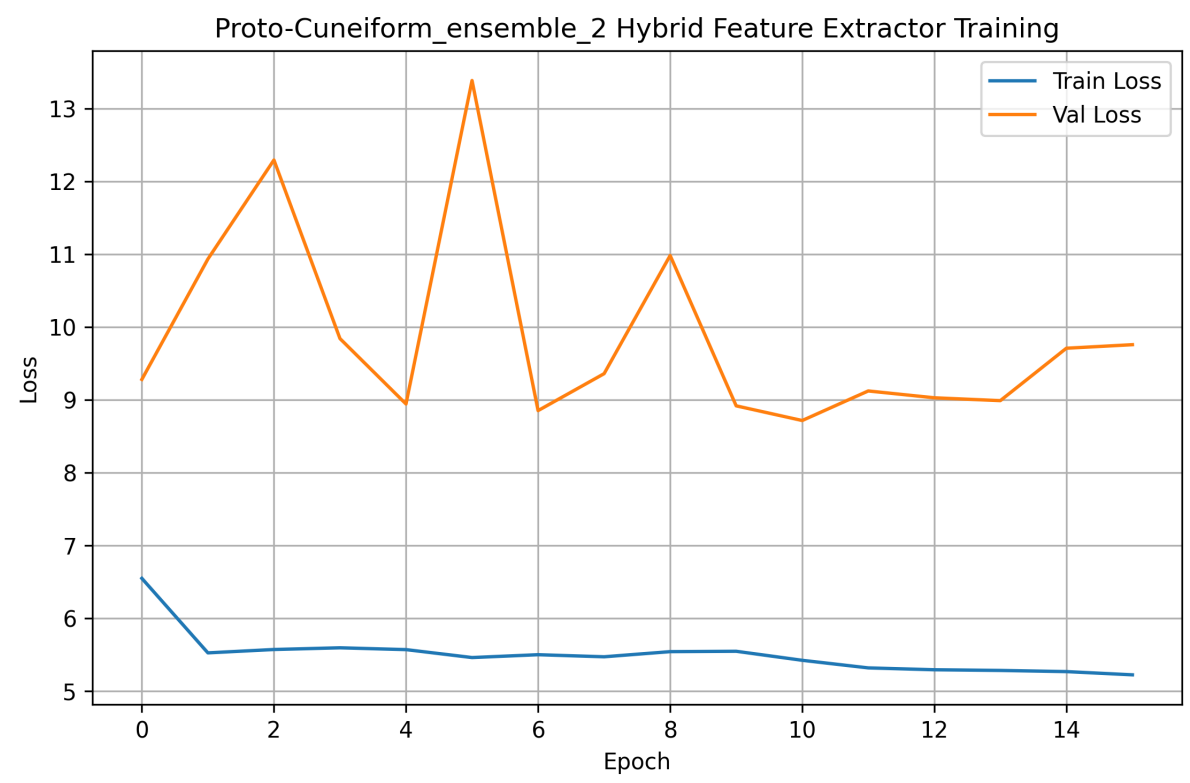

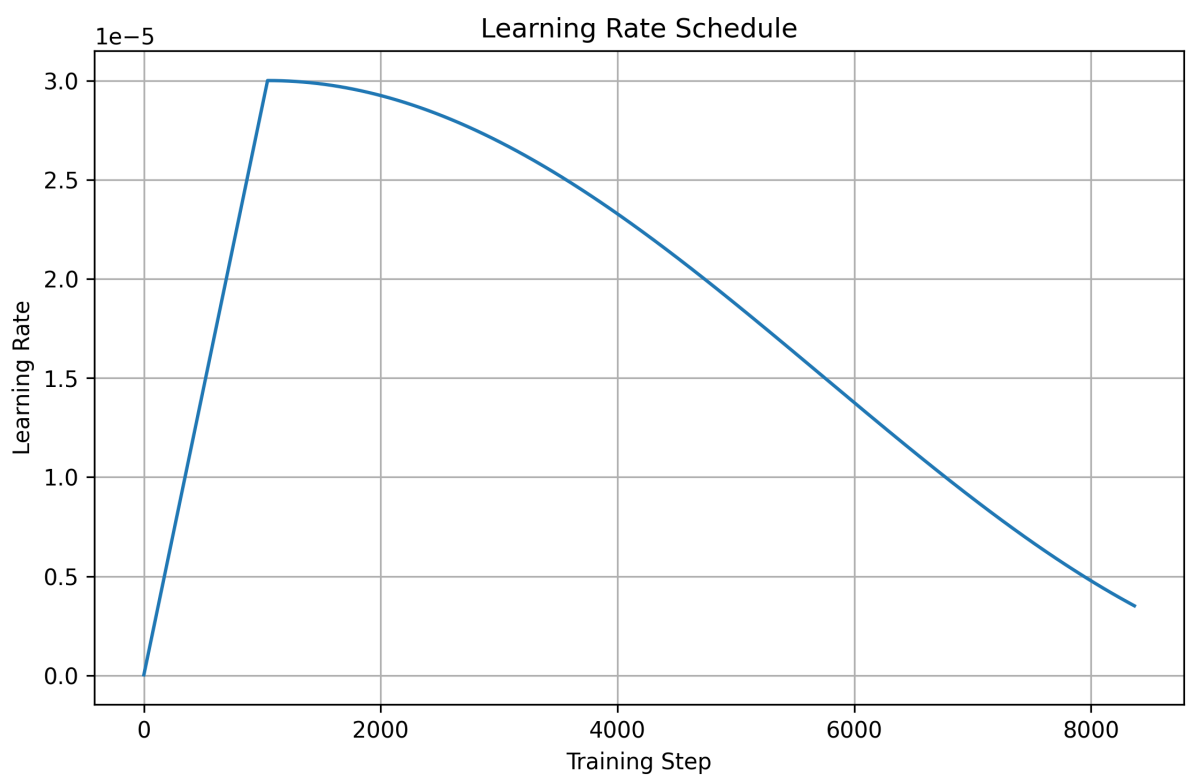

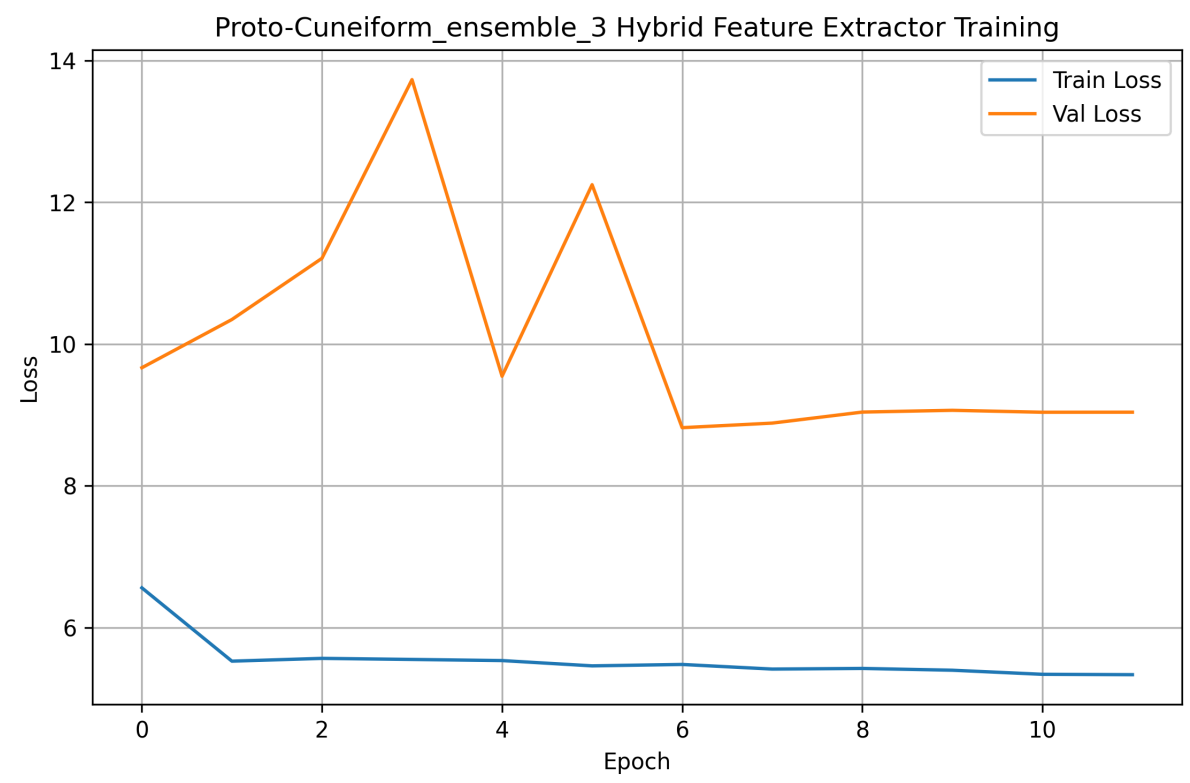

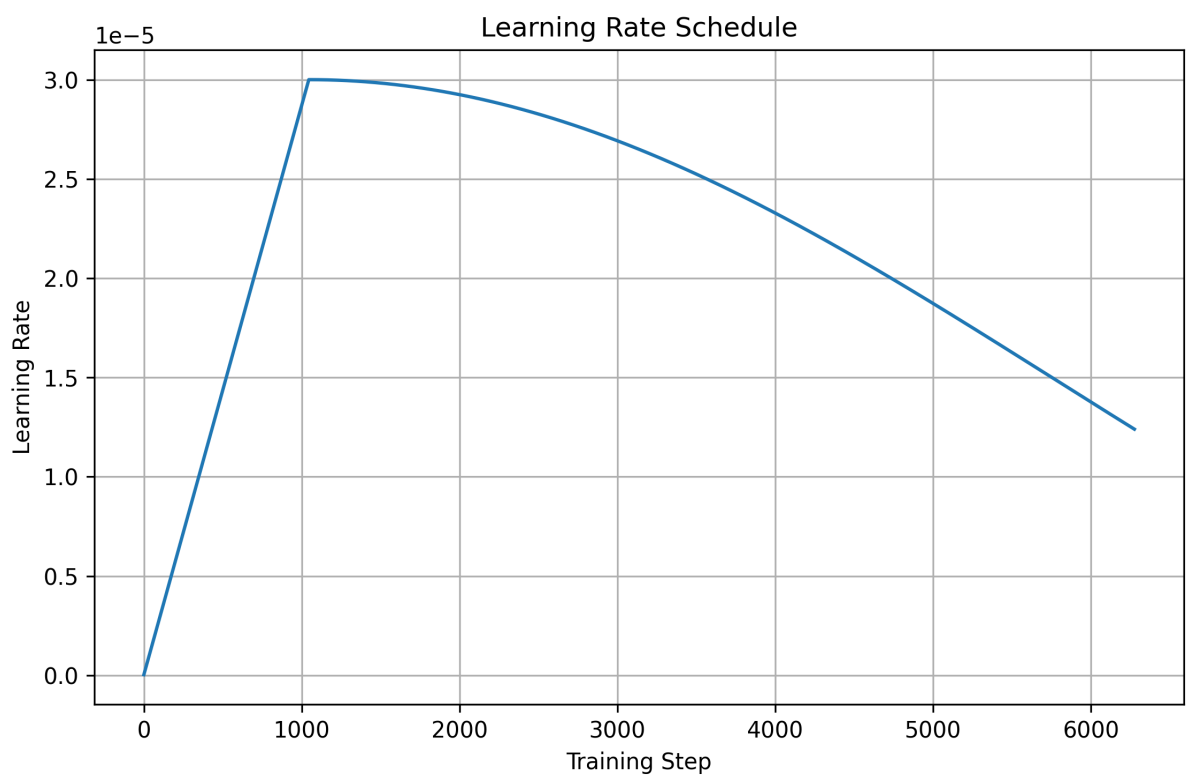

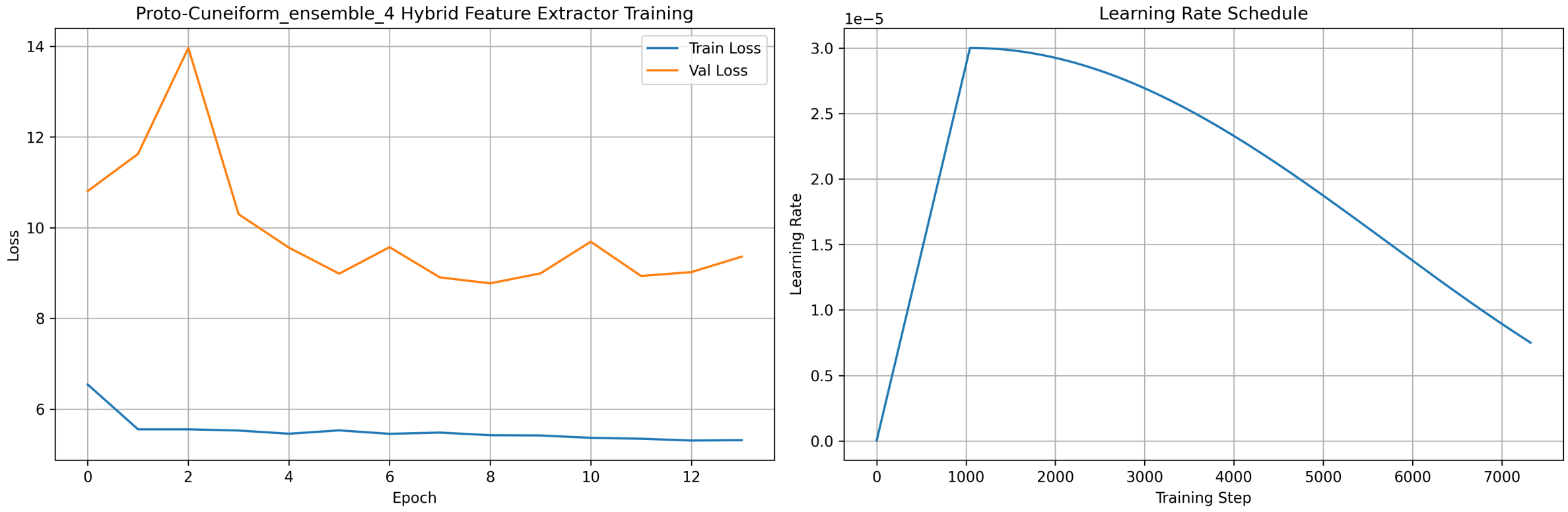
Proto-Cuneiform_ensemble_4 Hybrid Feature Extractor Training
Train Loss
Val Loss
Loss
Epoch
Learning Rate Schedule
1e−5
Learning Rate
Training Step

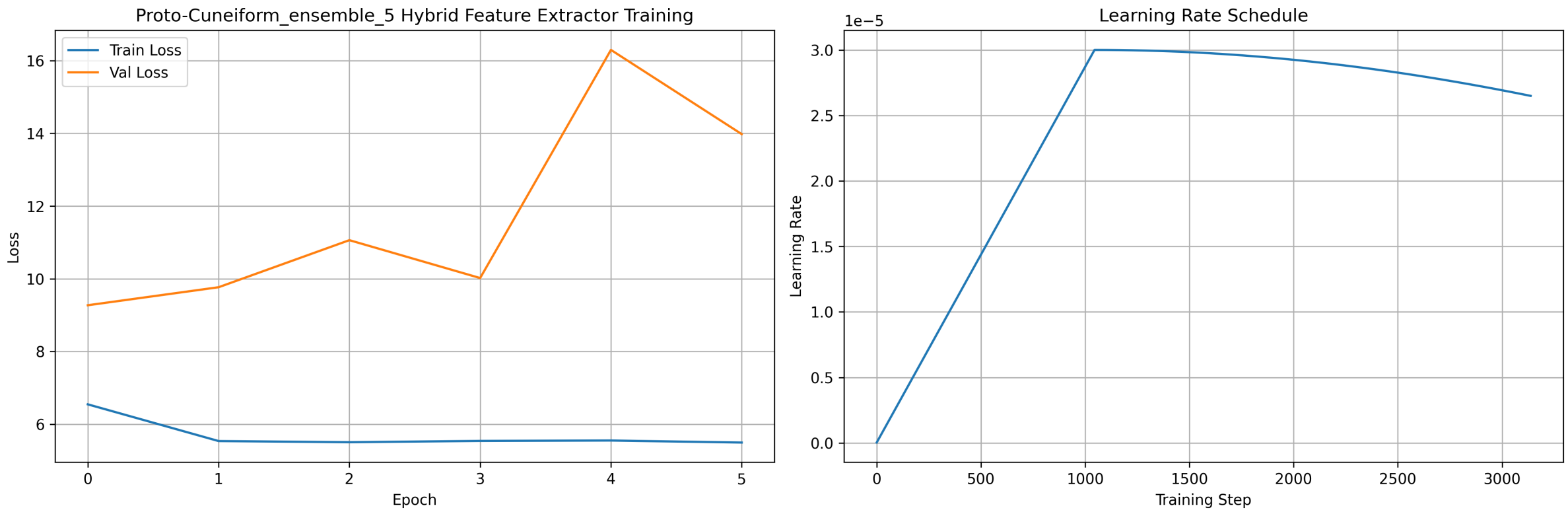
Proto-Cuneiform_ensemble_5 Hybrid Feature Extractor Training
Train Loss
Val Loss
Loss
Epoch
Learning Rate Schedule
1e−5
Learning Rate
Training Step

Proto-Elamite training summary.

| model_idx | seed | val_loss | epoch | model_path |
|---|---|---|---|---|
| 1 | 42 | 9.465515518188477 | 13 | Proto-Elamite_ensemble_1_hybrid_extractor_best.pth |
| 2 | 43 | 10.229036331176758 | 5 | Proto-Elamite_ensemble_2_hybrid_extractor_best.pth |
| 3 | 44 | 9.9594895362854 | 1 | Proto-Elamite_ensemble_3_hybrid_extractor_best.pth |
| 4 | 45 | 9.304905700683594 | 12 | Proto-Elamite_ensemble_4_hybrid_extractor_best.pth |
| 5 | 46 | 9.408125591278075 | 11 | Proto-Elamite_ensemble_5_hybrid_extractor_best.pth |

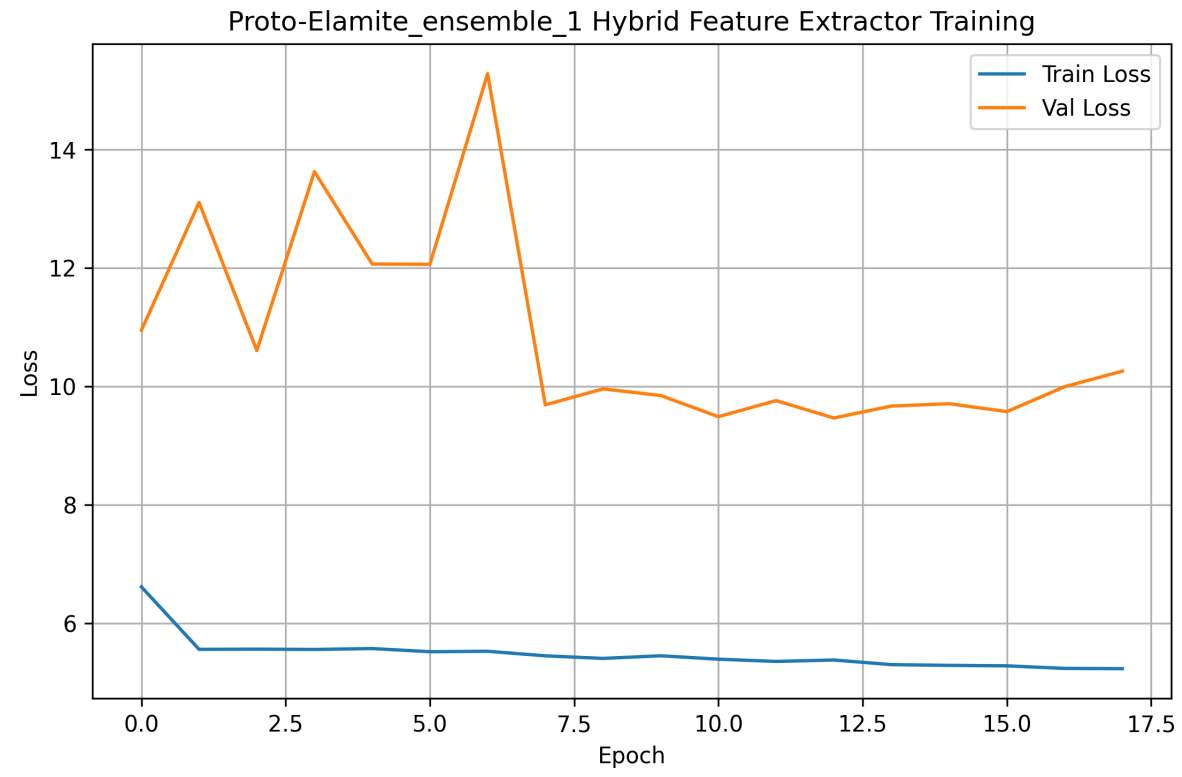

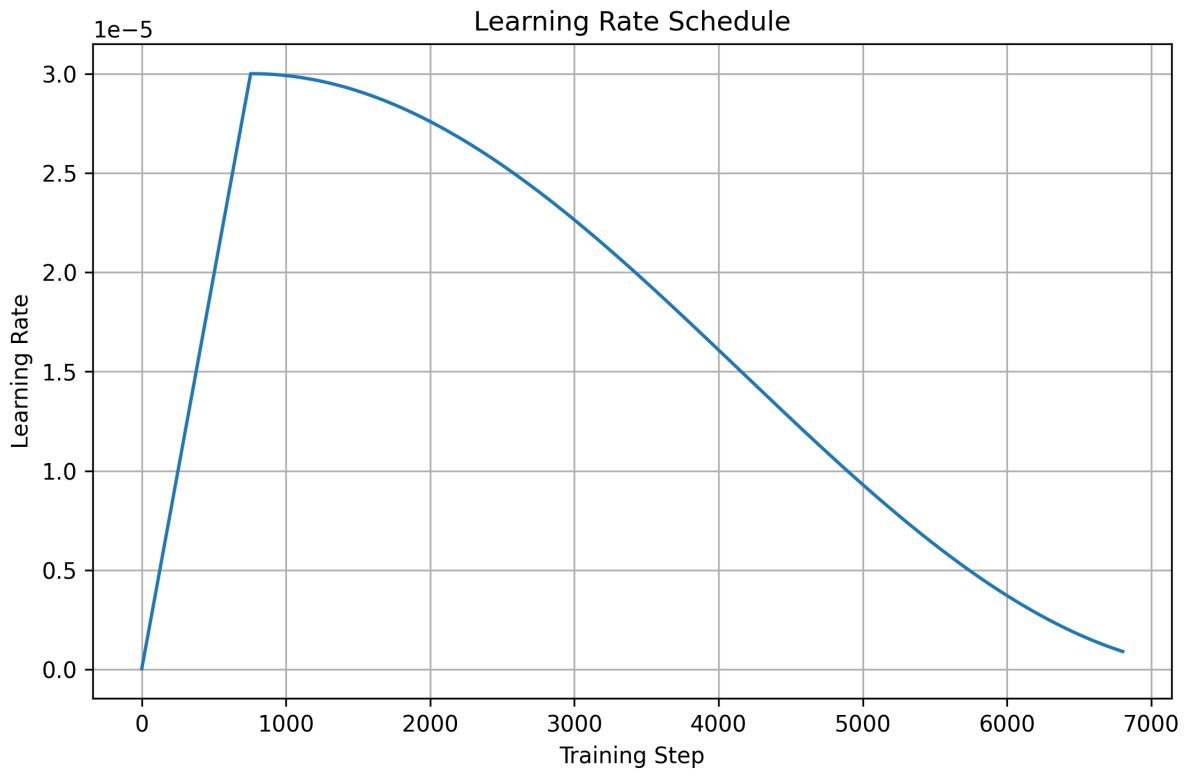

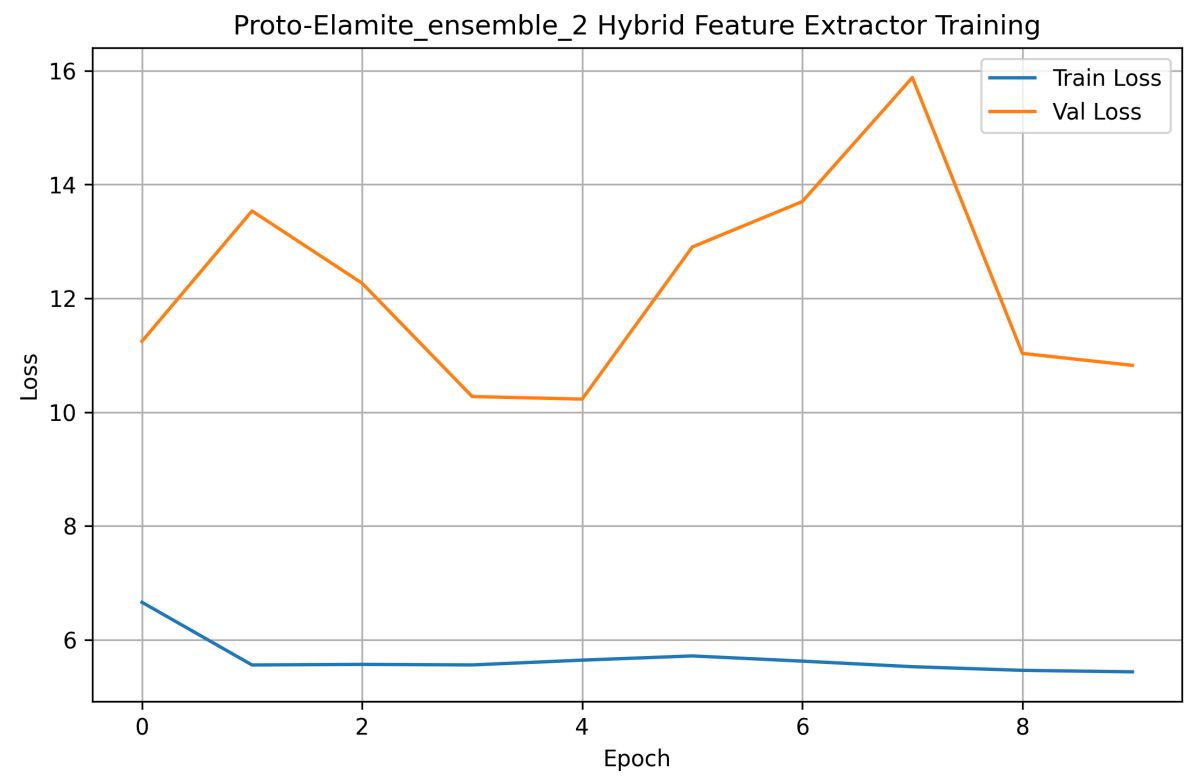

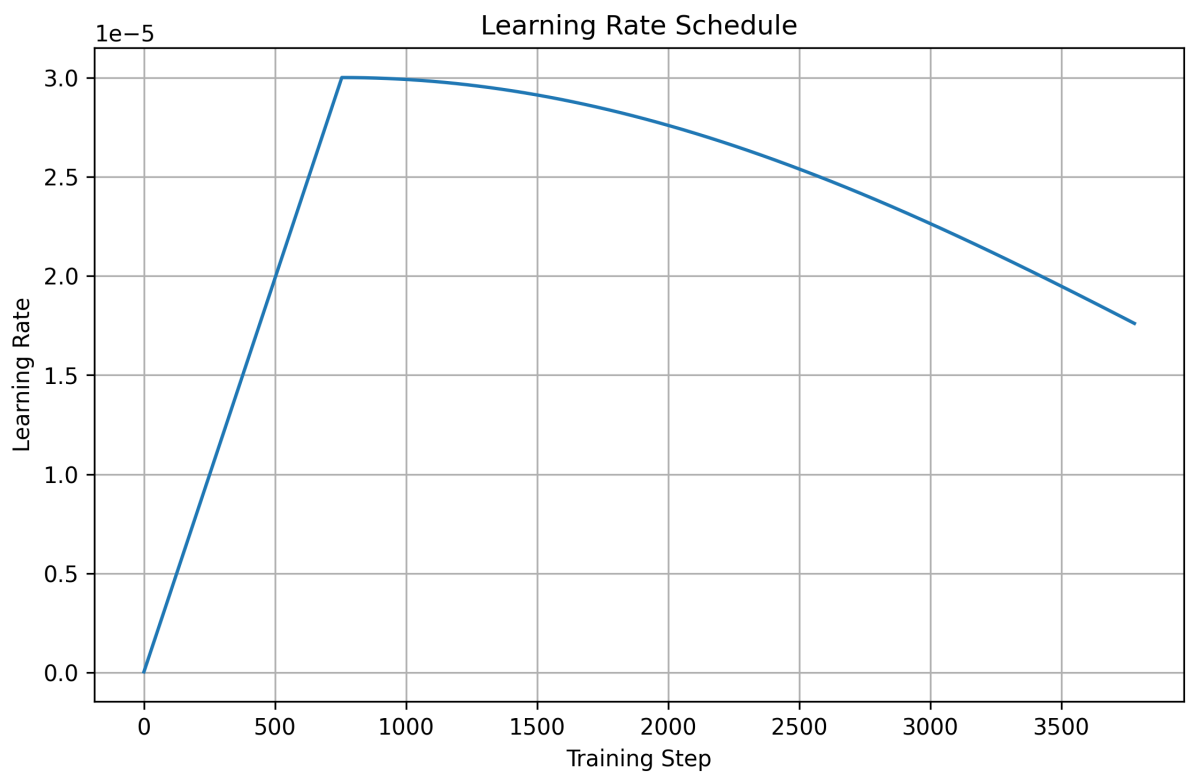

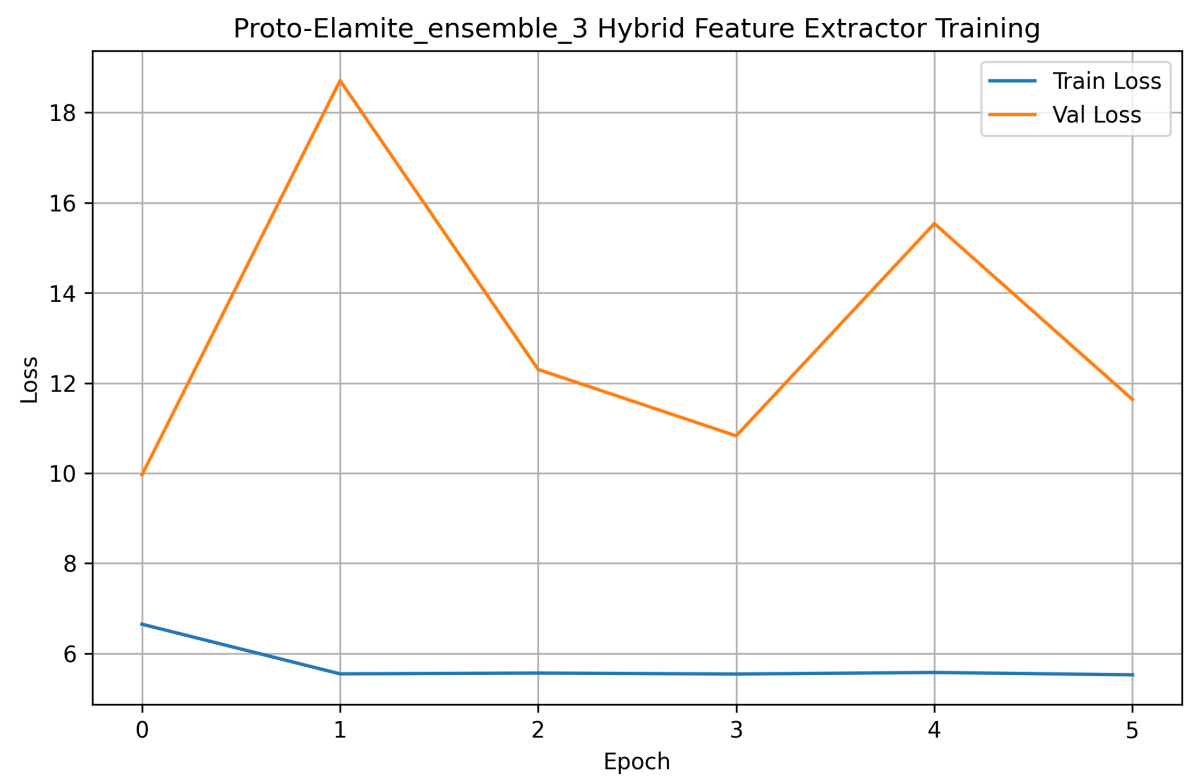

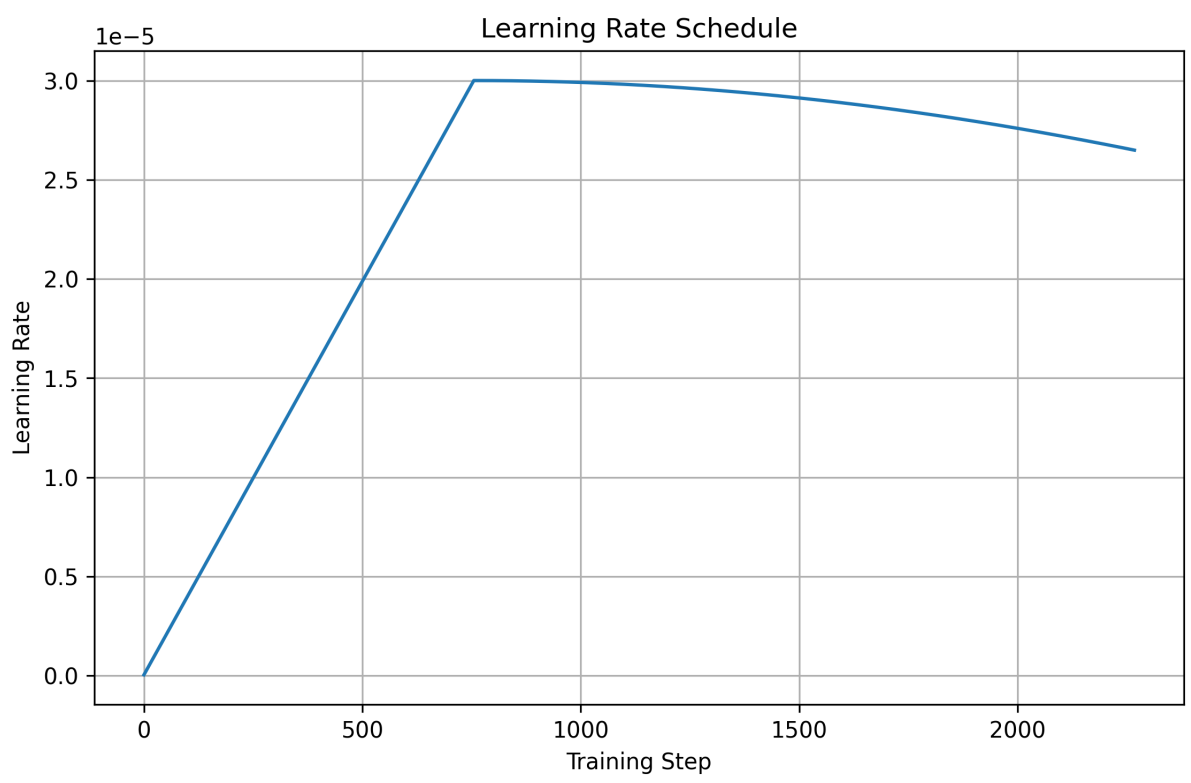

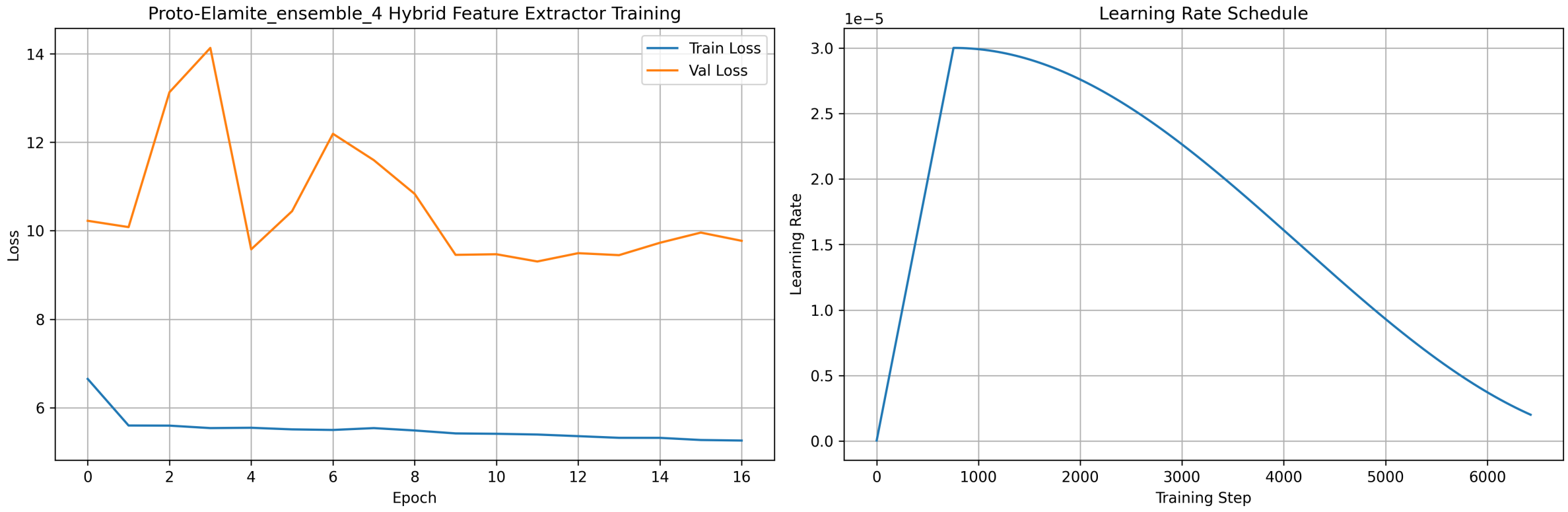
Proto-Elamite_ensemble_4 Hybrid Feature Extractor Training
Train Loss
Val Loss
Loss
Epoch
Learning Rate Schedule
1e−5
Learning Rate
Training Step

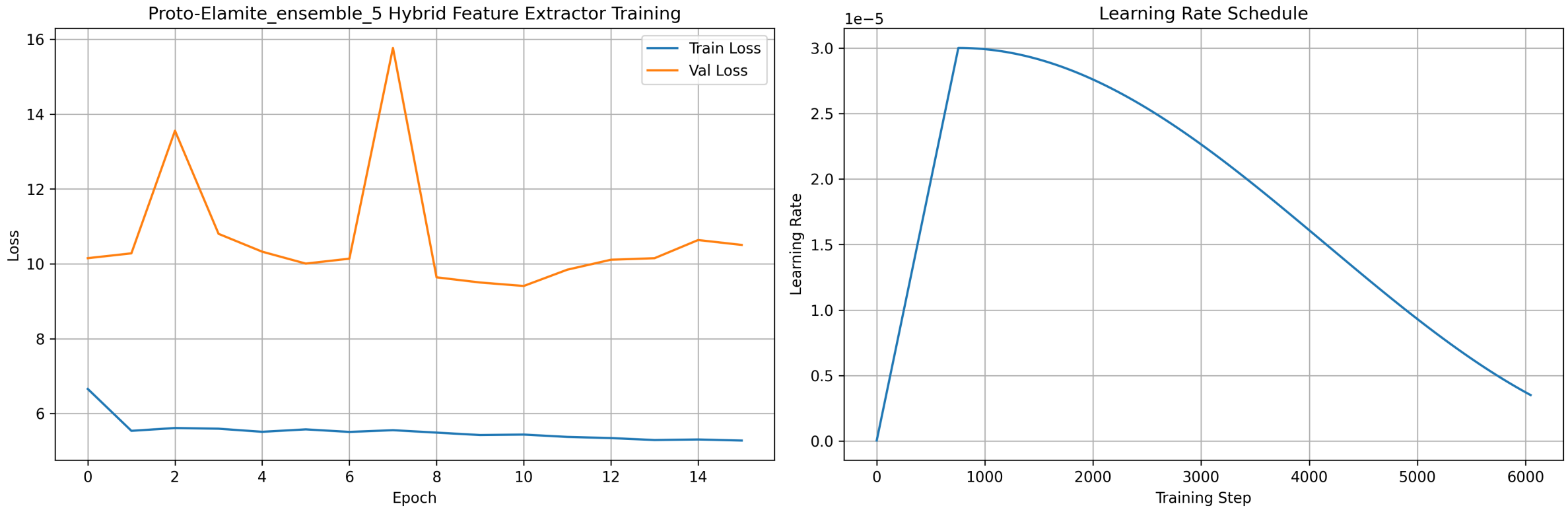
Proto-Elamite_ensemble_5 Hybrid Feature Extractor Training
Train Loss
Val Loss
Loss
Epoch
Learning Rate Schedule
1e−5
Learning Rate
Training Step